\documentclass[11pt,logo,letterpaper]{berkeley}

\usepackage[square,sort,comma,numbers]{natbib}

\usepackage[utf8]{inputenc} % allow utf-8 input
\usepackage[T1]{fontenc}    % use 8-bit T1 fonts
\usepackage{hyperref}       % hyperlinks
\usepackage{color}
\usepackage{url}            % simple URL typesetting
\usepackage{booktabs}       % professional-quality tables
\usepackage{amsfonts}       % blackboard math symbols
\usepackage{nicefrac}       % compact symbols for 1/2, etc.
\usepackage{microtype}      % microtypography
\usepackage{adjustbox}
\usepackage{titletoc}
\usepackage{latexsym}
\usepackage{amsfonts}
\usepackage{multirow}
\usepackage{array}
\usepackage{caption}
\usepackage{booktabs}
\usepackage{colortbl}
\usepackage{multicol}
\usepackage{bm}
\usepackage{graphicx}
\usepackage{subcaption}
\usepackage{amsthm,amsmath,amssymb}
\usepackage{mathrsfs}
\usepackage{enumitem}
\usepackage{algorithm,algpseudocode}
\usepackage{tikz}
\usepackage{enumitem}
\usepackage{wrapfig}
\usepackage{float}
\usepackage{graphicx}
\usepackage{caption}
\usepackage{colortbl}
\usepackage{bbding}
\usepackage{pifont}
\usepackage{tabularx} 
\usepackage{hyperref}
\usepackage{url}
\usepackage[utf8]{inputenc} % allow utf-8 input
\usepackage[T1]{fontenc}    % use 8-bit T1 fonts
\usepackage{hyperref}       % hyperlinks
\usepackage{url}            % simple URL typesetting
\usepackage{booktabs}       % professional-quality tables
\usepackage{amsfonts}       % blackboard math symbols
\usepackage{nicefrac}       % compact symbols for 1/2, etc.
\usepackage{microtype}      % microtypography
\usepackage{makecell}
\usepackage{graphicx}
\usepackage{amsmath}
\usepackage{multirow}
\usepackage{wrapfig}
\usepackage{tcolorbox}
\usepackage{fontawesome}

\usepackage{colortbl}
\usepackage{threeparttable}

\usepackage{marvosym}

\usepackage{tikz}
\usetikzlibrary{shapes.geometric, arrows.meta, positioning, fit}

\newcommand{\github}{\raisebox{-1.5pt}{\includegraphics[height=1.05em]{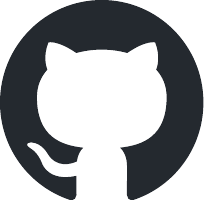}}}

\usepackage[edges]{forest}
\definecolor{hidden-draw}{RGB}{0,0,0}
\definecolor{hidden-pink}{rgb}{0.98, 0.94, 0.75}
\definecolor{level0}{rgb}{0.67, 0.88, 0.69}
\definecolor{level1}{rgb}{0.98, 0.92, 0.84}
\definecolor{level2}{rgb}{0.8, 0.8, 1.0}
\definecolor{level3}{rgb}{1.0, 0.71, 0.76}
\definecolor{level4}{rgb}{0.49, 0.99, 0.0}
\definecolor{level5}{rgb}{0.87, 0.63, 0.87}
\definecolor{darkblue}{rgb}{0, 0, 0.5}

\definecolor{TikTokPink}{HTML}{FF2B54} % Renamed from Crimson
\definecolor{CalGoldHex}{HTML}{FDF7F2}
\definecolor{IronGrey}{HTML}{6D6E71}
\definecolor{YaleBlue}{HTML}{2A5487}
\definecolor{LinkPurple}{HTML}{FF2B54}

\definecolor{hidden-red}{RGB}{205, 44, 36}
\definecolor{hidden-blue}{RGB}{194,232,247}
\definecolor{hidden-orange}{RGB}{243,202,120}
\definecolor{hidden-green}{RGB}{34,139,34}
\definecolor{hidden-pink}{RGB}{255,245,247}
\definecolor{hidden-black}{RGB}{20,68,106}
\definecolor{purple}{RGB}{144,153,196}
\definecolor{yellow}{RGB}{255,228,123}
\definecolor{hidden-yellow}{RGB}{255,248,203}
\definecolor{tkcolor}{RGB}{224,223,255}
\definecolor{darkblue}{rgb}{0, 0.40, 0.75}
\newcommand{\eg}{\textit{e.g.,}}

\hypersetup{colorlinks=true, linkcolor=YaleBlue}

\newcommand{\MYhref}[3][YaleBlue]{\href{#2}{\color{#1}{#3}}}%

\usepackage{booktabs}
\usepackage{longtable}
\usepackage{graphicx}
\usepackage{lscape}
\usepackage{enumitem}
\tcbset{
  takeawaysbox/.style={
    colback=gray!10,                        % 背景颜色，淡灰色
    colframe=black,                         % 边框颜色，黑色
    width=\linewidth,                       % 框宽度为文档宽度
    arc=4mm,                                % 圆角设置
    boxrule=0.8mm,                          % 边框厚度
    top=3mm,                                % 框与内容之间的上边距
    bottom=3mm,                             % 框与内容之间的下边距
    left=5mm,                               % 框与内容之间的左边距
    right=5mm,                              % 框与内容之间的右边距
  }
}

\definecolor{mygreen}{RGB}{144, 238, 144} % Light green
\definecolor{darkmygreen}{RGB}{0, 100, 0} % Dark green

\usepackage{latexsym}

\definecolor{box-pink}{RGB}{255,72,162}
\definecolor{box-cyan}{RGB}{29,234,221}
\definecolor{box-red}{RGB}{255,1,25}
\definecolor{box-purple}{RGB}{61,31,255}
\definecolor{box-green}{RGB}{71,212,90}
\definecolor{box-yellow}{RGB}{255,146,1}
\definecolor{box-blue}{RGB}{5,188,248}

\definecolor{tree-pink}{RGB}{255,232,242}
\definecolor{tree-cyan}{RGB}{199,241,240}
\definecolor{tree-red}{RGB}{255,215,214}
\definecolor{tree-purple}{RGB}{230,231,255}
\definecolor{tree-green}{RGB}{192,242,213}
\definecolor{tree-yellow}{RGB}{255,242,230}
\definecolor{tree-blue}{RGB}{223,244,255}

\title{Attention Sink in Transformers: A Survey on Utilization, Interpretation, and Mitigation}

\runningtitle{Attention Sink in Transformers: A Survey on Utilization, Interpretation, and Mitigation}

\author{\vspace{-1mm}\textbf{Zunhai Su}\textsuperscript{\rm 1,}\textsuperscript{\rm 2 \Letter}
\quad \textbf{Hengyuan Zhang}\textsuperscript{\rm 3}
\quad \textbf{Wei Wu}\textsuperscript{\rm 2}
\quad \textbf{Yifan Zhang}\textsuperscript{\rm 2}
\quad \textbf{Yaxiu Liu}\textsuperscript{\rm 1} 
\quad \textbf{He Xiao}\textsuperscript{\rm 3} \\
\textbf{Qingyao Yang}\textsuperscript{\rm 3} 
\quad \textbf{Yuxuan Sun}\textsuperscript{\rm 2}
\quad \textbf{Rui Yang}\textsuperscript{\rm 2}
\quad \textbf{Chao Zhang}\textsuperscript{\rm 2}
\quad \textbf{Jing Xiong}\textsuperscript{\rm 3}
\quad \textbf{Hui Shen}\textsuperscript{\rm 4} \\
\textbf{Keyu Fan}\textsuperscript{\rm 1}
\quad \textbf{Weihao Ye}\textsuperscript{\rm 5}
\quad \textbf{Chaofan Tao}\textsuperscript{\rm 3}
\quad \textbf{Taiqiang Wu}\textsuperscript{\rm 3}
\quad \textbf{Zhongwei Wan}\textsuperscript{\rm 6}\\
\textbf{Tiantian Zhang}\textsuperscript{\rm 7}
\quad \textbf{Bowen Yan}\textsuperscript{\rm 8}
\quad \textbf{Zhen Li}\textsuperscript{\rm 9}
\quad \textbf{Yiming Zhang}\textsuperscript{\rm 9}
\quad \textbf{Congkai Xie}\textsuperscript{\rm 9}\\
\textbf{Yulei Qian}\textsuperscript{\rm 2}
\quad \textbf{Yuchen Xie}\textsuperscript{\rm 2}\quad
\quad \textbf{Yik-Chung Wu}\textsuperscript{\rm 3 }\quad
\quad \textbf{Hongxia Yang}\textsuperscript{\rm 9 }\quad
\quad \textbf{Ngai Wong}\textsuperscript{\rm 3 \Letter}\\

\textsuperscript{\rm 1}Tsinghua University \quad
\textsuperscript{\rm 2}Meituan LongCat Team  \quad
\textsuperscript{\rm 3}The University of Hong Kong\\
\textsuperscript{\rm 4}University of Michigan \quad
\textsuperscript{\rm 5}Xiamen University \quad
\textsuperscript{\rm 6}The Ohio State University \quad
\textsuperscript{\rm 7}Columbia University \\
\textsuperscript{\rm 8}Shanghai Artificial Intelligence Laboratory \quad
\textsuperscript{\rm 9}The Hong Kong Polytechnic University\\
\vspace{2mm}
}

\begin{document}

\begingroup
\renewcommand{\thefootnote}{} % 临时让这个脚注编号为空
\footnotetext{\textsuperscript{\rm \Letter}Corresponding Author (zh-su23@mails.tsinghua.edu.cn, nwong@eee.hku.hk)}
% \vspace{-4mm}
\endgroup

\begin{abstract}
\vspace{-3mm}
As the foundational architecture of modern machine learning, Transformers have driven remarkable progress across diverse AI domains.
Despite their transformative impact, a persistent challenge across various Transformers is \textit{\textbf{Attention Sink (AS)}}, in which a disproportionate amount of attention is focused on a small subset of specific yet uninformative tokens.
AS complicates interpretability, significantly affecting the training and inference dynamics, and exacerbates issues such as hallucinations.
In recent years, substantial research has been dedicated to understanding and harnessing AS.
However, a comprehensive survey that systematically consolidates AS-related research and offers guidance for future advancements remains lacking.
To address this gap, we present \textit{\textbf{the first survey on AS}}, structured around three key dimensions that define the current research landscape:
\textit{\textbf{Fundamental Utilization}}, \textit{\textbf{Mechanistic Interpretation}}, and \textit{\textbf{Strategic Mitigation}}.
Our work makes a pivotal contribution by highlighting the key concepts and main trends in the field, guiding researchers through the evolution of AS-related studies.
We envision this survey as a valuable resource, empowering researchers to effectively manage AS within the current Transformer paradigm, while simultaneously inspiring innovative advancements for the next generation of Transformers.
Our GitHub repository organizes the papers featured in this survey and will be continuously updated to include the latest advancements.

% \vspace{2mm}
\centering
\github{} \textbf{GitHub}: \MYhref{https://github.com/ZunhaiSu/Awesome-Attention-Sink}{\textit{\textbf{https://github.com/ZunhaiSu/Awesome-Attention-Sink}}}

\end{abstract}

\maketitle

\begin{figure}[ht]
    \centering
    \vspace{-3mm}
    \includegraphics[width=0.96\linewidth]{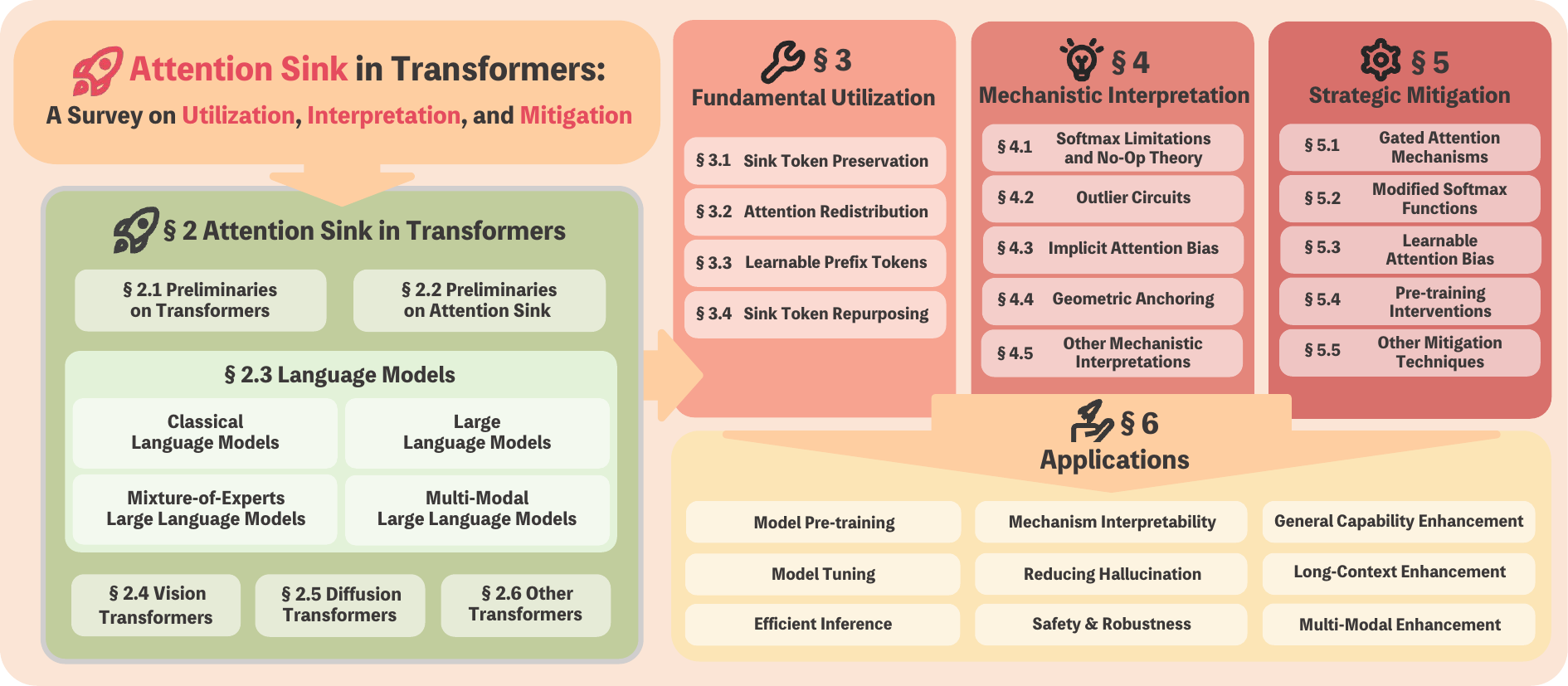}
    % \vspace{-2mm}
    \caption{Overview of the survey structure.}
    \label{fig:paper_overview}
\end{figure}

\onecolumn
\clearpage
\addtocontents{toc}{\protect\setcounter{tocdepth}{2}}
\vskip 3mm
\startcontents[sections]\vbox{\sc Contents}
\vspace{4mm}
\hrule height .8pt
\vspace{-2mm}

% 这里是修改行间距的关键部分
{\setlength{\baselineskip}{11pt} % 调整行距
\setlength{\parskip}{3pt}
\printcontents[sections]{l}{1}{\setcounter{tocdepth}{2}}}

\vspace{4mm}
\hrule height .8pt
\vskip 10mm
\clearpage

%----------------------------------Section 1----------------------------------%

\section{Introduction}
\label{sec_1_intro}
\subsection{Background}
\label{sec_1_1_Background}

% transformers
Transformers \cite{vaswani2017attention}, grounded in the multi-head self-attention mechanism, have emerged as a foundational architecture in machine learning.
Their unparalleled ability to capture long-range dependencies in sequential data, coupled with scalable and efficient end-to-end pretraining on large datasets, has been instrumental in driving transformative advancements across diverse AI domains, including natural language processing (NLP), computer vision (CV), multimodal learning, embodied AI, and beyond
\cite{lin2022transormersurvey,zhao2023llmsurvey,han2022vitsurvey,yin2024mllmsurvey,duan2022embodiedsurvey}.
Typical transformer models, including large language models (LLMs), vision transformers (ViTs), and others, have set the standard for state-of-the-art research across diverse AI domains \cite{team2025longcat,team2025longcatomni,team2025introducing,team2026longcat}. 
More recently, Vision Geometry-Grounded transformers (VGGT), as feedforward 3D models built on transformer architecture \cite{wang2025vggt,su2026xstreamvggt,keetha2025mapanything}, have achieved remarkable performance across a range of real-world 3D tasks, attracting significant attention as a groundbreaking paradigm in the 3D CV field.

% limitations
However, transformers still exhibit several limitations.
These include the quadratic computational complexity of the self-attention mechanism, the substantial memory requirements of the historical key-value cache, limited capacity for handling extremely long contexts, and limited interpretability \cite{zhu2024survey,wan2024efficient,hooper2024kvquant,zhang2026locate}.
Techniques such as sparse attention, linear attention, KV cache compression, test-time training (TTT), and efforts to enhance interpretability, among others, have been introduced to address these challenges \cite{sun2025efficient,sun2025speed,yang2025gated,team2025kimi,behrouz2024titans,zhang2026locate,zhang2026snapmla,su2026oscar}.
% AS
Beyond these limitations, a critical challenge attracting growing attention in both academia and industry is the \textbf{\textit{Attention Sink (AS)}} \cite{xiao2024efficient,gu2025attention}, wherein disproportionate attention is concentrated on a small set of uninformative tokens.
AS profoundly influences transformers, shaping both training and inference dynamics \cite{qiu2025gated,xiao2024efficient}, complicating model interpretability \cite{barbero2025llms,su2025kvsink,bondarenko2023quantizable}, and exacerbating issues such as hallucinations \cite{jiao2025don,tu2026attention,zhuang2025vasparse} and robustness challenges \cite{shang2025forgetting,yona2025interpreting,yellapragada2025leveraging}.

% increasing Current Efforts 
In recent years, significant efforts have been dedicated to tackling AS. 
For example, many studies on KV cache compression and sparse attention leverage the key attention patterns of AS to facilitate efficient inference in long-context LLMs by \textbf{\textit{Sink Token Preservation}} \cite{gu2025obcache,mu2025sals,zhu2025ojakv,su2025kvsink}.
Another line of research investigates the formation of AS and demonstrates that it is governed by \textbf{\textit{Outlier Circuits}} mechanisms \cite{cappellazzo2025mitigating,queipo2026attention,su2025kvsink,park2025outlier,su2026unveiling}, has further deepened our understanding of the underlying numerical mechanisms driving AS.
More recently, \textbf{\textit{Gated Attention Mechanisms}} \cite{bu2025value, qiu2025gated, qwenai2026, bondarenko2023quantizable} have incorporated input-dependent gating into attention, thereby mitigating AS, boosting model performance, and alleviating post-quantization degradation.
% Why Mastering AS Matters
Mastering AS in transformers, driven by diverse practical demands, is swiftly emerging as a central focus in transformer model research.
% no survey
Despite the rapid proliferation of AS-related studies, several foundational questions remain underexplored:

\begin{itemize}[leftmargin=*]
    \item \textbf{\textit{Q1:}} What are the fundamental paradigms for leveraging AS in current Transformer models? What are their distinctive characteristics, and how are they applied across different Transformer architectures?
    
    \item \textbf{\textit{Q2:}} What underlies the emergence and necessity of AS in Transformers? How does it develop and evolve, and what functional roles does it fulfill? What key insights have AS mechanistic studies provided?
    
    \item \textbf{\textit{Q3:}} How can future Transformer architectures be designed or optimized to operate independently of AS? What strategic approaches are available, and what trade-offs or limitations accompany each?
\end{itemize}

Collectively, these open questions reveal a pressing need: the fragmented AS literature has yet to be systematically reviewed, resulting in the absence of a definitive and unified reference for the field. % @Zunhai

\clearpage

\tikzstyle{my-box}=[
rectangle,
draw=black,
rounded corners,
text opacity=1,
minimum height=1.5em,
minimum width=5em,
inner sep=2pt,
align=left,
fill opacity=.5,
]

% 为每个主要章节定义不同的样式
\tikzstyle{section_2}=[my-box, fill=tree-pink]
\tikzstyle{section_3}=[my-box, fill=tree-blue]
\tikzstyle{section_4}=[my-box, fill=tree-cyan]
\tikzstyle{section_5}=[my-box, fill=tree-green]
\tikzstyle{appendix}=[my-box, fill=tree-yellow]

% 叶子节点样式保持不变
\tikzstyle{leaf}=[
my-box, 
minimum height=1.5em,
fill=tree-pink,
text=black,
align=left,
font=\normalsize,
inner xsep=5pt,
inner ysep=4pt,
align=left,
text width=45em,
]
\tikzstyle{leaf2}=[
my-box, 
minimum height=1.5em,
fill=tree-cyan, 
text=black,
align=left,
font=\normalsize,
inner xsep=5pt,
inner ysep=4pt,
]
\tikzstyle{leaf3}=[
my-box, 
minimum height=1.5em,
fill=tree-red, 
text=black,
align=left,
font=\normalsize,
inner xsep=5pt,
inner ysep=4pt,
]

\begin{figure}[htbp]
\vspace{-2mm}
\centering
\resizebox{0.93\textwidth}{!}{
	\begin{forest}
		forked edges,
		for tree={
			grow=east,
			reversed=true,
			anchor=base west,
			parent anchor=east,
			child anchor=west,
			base=left,
			font=\large,
			rectangle,
			draw=black,
			rounded corners,
			align=left,
			minimum width=4em,
			edge+={darkgray, line width=1pt},
			s sep=10pt,
			inner xsep=2pt,
			inner ysep=4pt,
			line width=1.1pt,
			ver/.style={rotate=90, child anchor=north, parent anchor=south, anchor=center},
		},
		where level=1{text width=10em,font=\normalsize,}{},
        where level=2{text width=12em,font=\normalsize,}{},
        where level=3{text width=9.5em,font=\normalsize,}{},
        where level=4{text width=42em,font=\normalsize,}{},
[\ \ \ \ \ \ \ \ \ {Attention Sink in Transformers: A Survey on Utilization, Interpretation, and Mitigation}, text width=49em, ver
	[\ Attention Sink in Transformers~(\S\ref{sec_2_Attention_Sink_in_Transformers}), text width=10em, section_2, draw=box-pink,  text width=16em
            [\ \ Classical Language Models~(\S\ref{sec_2_3_1_Classical_Language Models}), section_2, draw=box-pink,  text width=17em
        [\eg                      
                    \citep{ruscio2025you}{,}
                    \citep{li2025ctr}{,}
                    \citep{bai2025does}{,}
                    \citep{bondarenko2023quantizable}{,}
                    \citep{puccetti2022outlier}{,}
                    \citep{bondarenko2021understanding}{,}
                    \citep{kovaleva2021bert}{,}
                    \citep{luo2021positional}{,}
                    \citep{clark2019does}{.},
                    leaf, text width=42em, section_2, draw=tree-pink
    		  ]
            ]
            [\ \ \ \ \ Large Language Models~(\S\ref{sec_2_3_2_Large_Language_Models}), section_2, draw=box-pink,  text width=17em
                [\eg 
                    \citep{xiao2026mimo}{,}
                    \citep{fu2026attention}{,}
                    \citep{wong2025existence}{,}
                    \citep{yu2025sliding}{,}
                    \citep{xiong2025dope}{,}
                    \citep{liang2025tweo}{,}
                    \citep{queipo2026attention}{,}
                    \citep{salvatore2025lost}{,}
                    \citep{yang2025cacheclip}{,}
                    \citep{shang2025forgetting}{,}
                    \citep{fang2025artificial}{,}
                    \citep{desai2026vattention}{,}
                    \citep{gu2025obcache}{,}\\
                    \citep{mu2025sals}{,}
                    \citep{bu2025value}{,}
                    \citep{team2025longcat}{,}
                    \citep{zhu2025ojakv}{,}
                    \citep{su2025kvsink}{,}
                    \citep{agarwal2025gpt}{,}
                    \citep{ruscio2025you}{,}
                    \citep{li2025ctr}{,}
                    \citep{kobyzev2025integral}{,}
                    \citep{fu2025h2eal}{,}
                    \citep{su2026unveiling}{,}
                    \citep{he2025trianglemix}{,}
                    \citep{shin2025orthorank}{,}
                    \citep{yang2025earn}{,}
                    \citep{qi2025deltallm}{,}\\
                    \citep{park2025outlier}{,}
                    \citep{yu2025two}{,}
                    \citep{yao2025learn}{,}
                    \citep{qiu2025gated}{,}
                    \citep{han2026zerotuning}{,}
                    \citep{willette2025delta}{,}
                    \citep{zuhri2025softpick}{,}
                    \citep{barbero2025llms}{,}
                    \citep{park2025keydiff}{,}
                    \citep{chen2025edgeinfinite}{,}
                    \citep{yona2025interpreting}{,}
                    \citep{xiao2025efficient}{,}
                    \citep{wu2025emergence}{,}
                    \citep{an2025systematic}{,}
                    \citep{fu2025sliding}{,}
                    \citep{su2025rotatekv}{,}\\
                    \citep{kamoda2025weight}{,}
                    \citep{he2025task}{,}
                    \citep{wang2025llms}{,}
                    \citep{deng2025unigist}{,}
                    \citep{shutova2025cache}{,}
                    \citep{wang2025position}{,}
                    \citep{qwenai2026}{,}
                    \citep{zhangattention}{,}
                    \citep{zhangattention}{,}
                    \citep{khalil2025singular}{,}
                    \citep{zhang2025attention}{,}
                    \citep{liu2025sgd}{,}
                    \citep{zeng2025subkv}{,}
                \textit{etc.},
                leaf, text width=42em, section_2, draw=tree-pink
    		  ]
            ]
		[\ \ \ \ \ \ \ \ \ \ \ \ \ \ Mixture-of-Experts \\ \ \ \ \ \ Large Language Models~(\S\ref{sec_2_3_3_MoE_Large_Language_Models}), section_2, draw=box-pink,  text width=17em
            [\eg~
                    \citep{xiao2026mimo}{,}
                    \citep{team2025longcat}{,}
                    \citep{agarwal2025gpt}{,}
                    \citep{fu2025h2eal}{,}
                    \citep{su2026unveiling}{,}
                    \citep{qiu2025gated}{,}
                    \citep{qwenai2026}{,}
                    \citep{sun2024massive}{.}
            , leaf, text width=42em, section_2, draw=tree-pink]
		]
		[\ \ \ \ \ \ \ \ \ \ \ \ \ \ \ \ \ Multi-Modal\\ \ \ \ \ \ Large Language Models~(\S\ref{sec_2_3_4_MM_Large_Language_Models}), section_2, draw=box-pink,  text width=17em
            [\eg~
                    \citep{chen2025omnisparse}{,}
                    \citep{cappellazzo2025mitigating}{,}
                    \citep{aman2025bitmar}{,}
                    \citep{luo2026sink}{,}
                    \citep{khaki2025sparsevila}{,}
                    \citep{kang2025pevlm}{,}
                    \citep{baek2025large}{,}
                    \citep{jiao2025don}{,}
                    \citep{tu2026attention}{,}
                    \citep{kang2025see}{,}
                    \citep{su2025akvq}{,}
                    \citep{zhuang2025vasparse}{,}
                    \citep{zhang2026drives}{,}\\
                    \citep{zhang2025shallow}{,}
                    \citep{wang2025mirage}{,}
                    \citep{chenvocabulary}{,}
                    \citep{yang2025seed}{,}
                    \citep{fan2025visipruner}{,}
                    \citep{lee2025tale}{,}
                    \citep{zhang2024seeing}{,}
                    \citep{elawady2024relic}{.}
            , leaf, text width=42em, section_2, draw=tree-pink]
		]
        [\ \ \ \ \ \ \ \ Vision Transformers~(\S\ref{sec_2_4_Vision_Transformers}), section_2, draw=box-pink,  text width=17em
            [
            \eg
                    \citep{wang2026vit}{,}
                    \citep{simeoni2025dinov3}{,}
                    \citep{lu2025artifacts}{,}
                    \citep{xiao2025focus}{,}
                    \citep{jiang2025vision}{,}
                    \citep{lappe2025register}{,}
                    \citep{chen2025vision}{,}
                    \citep{feng2026edit}{,}
                    \citep{wang2025vggt}{,}
                    \citep{yellapragada2025leveraging}{,}
                    \citep{sun2024massive}{,}
                    \citep{pulfer2024robustness}{,}
                    \citep{darcet2024vision}{,}\\
                    \citep{bondarenko2023quantizable}{.},
                leaf, text width=42em, section_2, draw=tree-pink
          ]
        ]
        [\ \ \ \ \ \ Diffusion Transformers~(\S\ref{sec_2_5_Diffusion_Transformers}), section_2, draw=box-pink,  text width=17em
            [
            \eg
                    \citep{starodubcev2026registers}{,}
                    \citep{wu2026attention}{,}
                    \citep{wu2026taming}{,}
                    \citep{mao2026packforcing}{,}
                    \citep{su2026omniforcing}{,}
                    \citep{cao2026mlv}{,}
                    \citep{shin2026motionstream}{,}
                    \citep{liu2026rolling}{,}
                    \citep{zhang2026freetext}{,}
                    \citep{yi2025deep}{,}
                    \citep{lu2025reward}{,}
                    \citep{bandyopadhyay2025block}{,}\\
                    \citep{yang2025longlive}{,}
                    \citep{jamal2026diffusion}{,}
                    \citep{kim2025text}{.},
                leaf, text width=42em, section_2, draw=tree-pink
          ]
        ]
        [\ \ \ \ \ \ \ \ \ Other Transformers(\S\ref{sec_2_6_Attention_Sink_in_Other_Transformers}), section_2, draw=box-pink,  text width=17em
            [
            \eg
                    \citep{yi2025deep}{,}
                    \citep{lu2025reward}{,}
                    \citep{shin2026motionstream}{,}
                    \citep{bandyopadhyay2025block}{,}
                    \citep{liu2026rolling}{,}
                    \citep{jamal2026diffusion}{,}
                    \citep{kim2025text}{,}
                    \citep{rulli2025attention}{,}
                    \citep{koo2025retovla}{,}
                    \citep{dong2025hymba}{,}
                    \citep{wang2025mamba}{,}
                    \citep{geshkovski2023emergence}{.},
                leaf, text width=42em, section_2, draw=tree-pink
          ]
        ]
	]
	[\ \ \ \ \ Fundamental Utilization~(\S\ref{sec_3_Fundamental_Utilization}), section_3, draw=box-blue, text width=16em
    [\ \ \ \ \ \ Sink Token Preservation (\S\ref{sec_3_1_Preserving_Sink_Tokens}), section_3, draw=box-blue,  text width=17em
        [\eg 
                    \citep{yu2025sliding}{,}
                    \citep{yi2025deep}{,}
                    \citep{lu2025reward}{,}
                    \citep{shin2026motionstream}{,}
                    \citep{bandyopadhyay2025block}{,}
                    \citep{aman2025bitmar}{,}
                    \citep{yang2025cacheclip}{,}
                    \citep{fang2025artificial}{,}
                    \citep{desai2026vattention}{,}
                    \citep{gu2025obcache}{,}
                    \citep{mu2025sals}{,}
                    \citep{luo2026sink}{,}\\
                    \citep{khaki2025sparsevila}{,}
                    \citep{zhu2025ojakv}{,}
                    \citep{liu2026rolling}{,}
                    \citep{su2025kvsink}{,}
                    \citep{li2025ctr}{,}
                    \citep{fu2025h2eal}{,}
                    \citep{he2025trianglemix}{,}
                    \citep{shin2025orthorank}{,}
                    \citep{qi2025deltallm}{,}
                    \citep{yu2025two}{,}
                    \citep{yao2025learn}{,}
                    \citep{kang2025pevlm}{,}
                    \citep{willette2025delta}{,}
                    \citep{chen2025edgeinfinite}{,}
                    \citep{xiao2025efficient}{,}\\
                    \citep{su2025rotatekv}{,}
                    \citep{su2025akvq}{,}
                    \citep{he2025task}{,}
                    \citep{wang2025llms}{,}
                    \citep{shutova2025cache}{,}
                    \citep{zhang2025attention}{,}
                    \citep{liu2025sgd}{,}
                    \citep{zeng2025subkv}{,}
                    \citep{hanevolving}{,}
                    \citep{zhang2025anchor}{,}
                    \citep{zhang2025leank}{,}
                    \citep{yang2025seed}{,}\\
                    \citep{fan2025visipruner}{,}
                    \citep{lee2025tale}{,}
                    \citep{acharya2025star}{,}
                    \citep{zhao2024buzz}{,}
                    \citep{xiao2025duoattention}{,}
                    \citep{chen2024prefixquant}{,}
                    \citep{ge2025little}{,}
                    \citep{chen2025magicpig}{,}
                    \citep{cai2024pyramidkv}{,}
                    \citep{guo2024attention}{,}
                    \citep{duanmu2024skvq}{,}
                    \citep{liu2024intactkv}{,}
                    \citep{liao2024free}{,}
        \textit{etc.}
        , leaf2, text width=42em, section_3, draw=tree-blue]
	]
	[\ \ \ \ \ \ Attention Redistribution (\S\ref{sec_3_2_Attention_Redistribution}), section_3, draw=box-blue,  text width=17em
		[\eg 
                    \citep{xiong2025dope}{,}
                    \citep{jiang2025vision}{,}
                    \citep{baek2025large}{,}
                    \citep{han2026zerotuning}{,}
                    \citep{jiao2025don}{,}
                    \citep{tu2026attention}{,}
                    \citep{kang2025see}{,}
                    \citep{zhuang2025vasparse}{,}
                    \citep{wang2025position}{,}
                    \citep{kim2025text}{,}
                    \citep{zhang2026drives}{,}
                    \citep{zhang2025shallow}{,}
                    \citep{chenvocabulary}{,}\\
                    \citep{bai2025does}{,}
                    \citep{jo2024a2sf}{,}
                    \citep{yu2024unveiling}{.}
	, leaf2, text width=42em, section_3, draw=tree-blue]
    ]
	[\ \ \ \ \ \ Learnable Prefix Tokens (\S\ref{sec_3_3_Learnable_Prefix_Tokens}), section_3, draw=box-blue,  text width=17em
		[\eg 
                    \citep{koo2025retovla}{,}
                    \citep{simeoni2025dinov3}{,}
                    \citep{xiao2025focus}{,}
                    \citep{yang2025earn}{,}
                    \citep{lappe2025register}{,}
                    \citep{chen2025vision}{,}
                    \citep{wang2025vggt}{,}
                    \citep{deng2025unigist}{,}
                    \citep{dong2025hymba}{,}
                    \citep{hu2025epic}{,}
                    \citep{elawady2024relic}{,}
                    \citep{son2024prefixing}{,}
                    \citep{zhang2024sinklora}{,}\\
                    \citep{wang2025mamba}{,}
                    \citep{sandal2024zero}{,}
                    \citep{darcet2024vision}{,}
                    \citep{xiao2024efficient}{.}
	, leaf2, text width=42em, section_3, draw=tree-blue]
    ]
	[\ \ \ \ \ \ Sink Token Repurposing (\S\ref{sec_3_4_Sink_Tokens_Utilization}), section_3, draw=box-blue,  text width=17em
		[\eg 
                    \citep{chen2025omnisparse}{,}
                    \citep{shang2025forgetting}{,}
                    \citep{park2025keydiff}{,}
                    \citep{yellapragada2025leveraging}{,}
                    \citep{lin2025look}{,}
                    \citep{zhang2025shallow}{,}
                    \citep{wang2025mirage}{,}
                    \citep{zhang2024seeing}{,}
                    \citep{pulfer2024robustness}{,}
                    \citep{li2024streamingdialogue}{.}
	, leaf2, text width=42em, section_3, draw=tree-blue]
    ]
]
[\ \ \ Mechanistic Interpretation~(\S\ref{sec_4_Mechanistic_Interpretation}), text width=10em, section_4, draw=box-cyan, text width=16em
    [\ \ \ \ \ \ \ \ \ \ \ \ \ Softmax Limitations \\ \ \ \ \ \ \ \ \ \ \ \& No-Op Theory~(\S\ref{sec_4_1_Softmax_Limitations}), section_4, draw=box-cyan,  text width=17em
        [\eg 
                    \citep{fu2026attention}{,}
                    \citep{bu2025value}{,}
                    \citep{su2025kvsink}{,}
                    \citep{qiu2025gated}{,}
                    \citep{han2026zerotuning}{,}
                    \citep{zuhri2025softpick}{,}
                    \citep{fu2025sliding}{,}
                    \citep{zhangattention}{,}
                    \citep{zhang2025attention}{,}
                    \citep{hongvariance}{,}
                    \citep{gu2025attention}{,}
                    \citep{kaul2025attention}{,}
                    \citep{guo2024active}{,}\\
                    \citep{kaul2025attention}{,}
                    \citep{jo2024a2sf}{,}
                    \citep{chen2024rotary}{,}
                    \citep{xiao2024efficient}{,}
                    \citep{bondarenko2023quantizable}{.}
        , leaf3, text width=42em, section_4, draw=tree-cyan]
    ]
    [\ \ \ \ \ \ \ \ \ \ \ Outlier Circuits~(\S\ref{sec_4_2_Outliers_Circuits}), section_4, draw=box-cyan,  text width=17em
        [\eg 
                    \citep{liang2025tweo}{,}
                    \citep{cappellazzo2025mitigating}{,}
                    \citep{queipo2026attention}{,}
                    \citep{bu2025value}{,}
                    \citep{su2025kvsink}{,}
                    \citep{su2026unveiling}{,}
                    \citep{park2025outlier}{,}
                    \citep{zuhri2025softpick}{,}
                    \citep{yona2025interpreting}{,}
                    \citep{kang2025see}{,}
                    \citep{an2025systematic}{,}
                    \citep{su2025rotatekv}{,}
                    \citep{su2025akvq}{,}
                    \citep{zhangattention}{,}
                    \citep{jamal2026diffusion}{,}\\
                    \citep{zhang2026drives}{,}
                    \citep{xiang2025dfrot}{,}
                    \citep{hu2025epic}{,}
                    \citep{guo2024active}{,}
                    \citep{kaul2025attention}{,}
                    \citep{son2024prefixing}{,}
                    \citep{guo2024attention}{,}
                    \citep{liu2024intactkv}{,}
                    \citep{sun2024massive}{,}
                    \citep{liao2024free}{,}
                    \citep{gurnee2024universal}{,}
                    \citep{bondarenko2023quantizable}{,}
                    \citep{puccetti2022outlier}{,}
                    \citep{bondarenko2021understanding}{,}\\
                    \citep{kovaleva2021bert}{,}
                    \citep{luo2021positional}{,}
                    \citep{clark2019does}{.}
        , leaf3, text width=42em, section_4, draw=tree-cyan]
        ]
    [\ \ \ \ \ \ \ Implicit Attention Bias~(\S\ref{sec_4_3_Implicit_Attention_Bias}), section_4, draw=box-cyan,  text width=17em
        [\eg 
                    \citep{han2026zerotuning}{,}
                    \citep{an2025systematic}{,}
                    \citep{kamoda2025weight}{,}
                    \citep{gu2025attention}{,}
                    \citep{sun2024massive}{.}
        , leaf3, text width=42em, section_4, draw=tree-cyan]
    ]
    [\ \ \ \ \ \ \ Geometric Anchoring~(\S\ref{sec_4_4_Geometric_Anchoring}), section_4, draw=box-cyan,  text width=17em
        [\eg 
                    \citep{chen2025omnisparse}{,}
                    \citep{ruscio2025you}{,}
                    \citep{li2025ctr}{,}
                    \citep{shin2025orthorank}{,}
                    \citep{park2025keydiff}{,}
                    \citep{zhang2025anchor}{,}
                    \citep{chen2025magicpig}{,}
                    \citep{dong2024exploring}{.}
        , leaf3, text width=42em, section_4, draw=tree-cyan]
    ]
    % [\ \ \ \ \ \ \ \ \ \ \ Structural Bias~(\S\ref{sec_4_5_Structural_Bias}), section_4, draw=box-cyan,  text width=17em
    %     [\eg 
    %                 \citep{fu2026attention}{,}
    %                 \citep{xiong2025dope}{,}
    %                 \citep{salvatore2025lost}{,}
    %                 \citep{shang2025forgetting}{,}
    %                 \citep{rulli2025attention}{,}
    %                 \citep{wu2025emergence}{,}
    %                 \citep{wang2025position}{,}
    %                 \citep{zhang2026drives}{,}
    %                 \citep{jo2024a2sf}{,}
    %                 \citep{yan2024unveiling}{,}
    %                 \citep{chen2024rotary}{,}
    %                 \citep{luo2021positional}{.}
    %     , leaf3, text width=42em, section_4, draw=tree-cyan]
    % ]
    [\ \ \ \ Additional Interpretations~(\S\ref{sec_4_5_Additional_Mechanistic_Interpretations of Attention Sink}), section_4, draw=box-cyan,  text width=17em
        [\eg 
                    \citep{fu2026attention}{,}
                    \citep{xiong2025dope}{,}
                    \citep{salvatore2025lost}{,}
                    \citep{shang2025forgetting}{,}
                    \citep{rulli2025attention}{,}
                    \citep{wu2025emergence}{,}
                    \citep{wang2025position}{,}
                    \citep{zhang2026drives}{,}
                    \citep{jo2024a2sf}{,}
                    \citep{yan2024unveiling}{,}
                    \citep{chen2024rotary}{,}
                    \citep{luo2021positional}{,}
                    \citep{queipo2026attention}{,}
                    \citep{zhangattention}{,}\\
                    \citep{guo2024active}{,}
                    \citep{cancedda2024spectral}{,}
                    \citep{barbero2025llms}{,}
                    \citep{barbero2024transformers}{,}
                    \citep{geshkovski2023emergence}{.}
        , leaf3, text width=42em, section_4, draw=tree-cyan]
    ]
]
	[\ \ \ \ \ \ \ Strategic Mitigation~(\S\ref{sec_5_Strategic_Mitigation}), text width=10em, section_5, draw=box-green, text width=16em
    [\ \ \ Gated Attention Mechanism (\S\ref{sec_5_1_Gated_Attention_Mechanisms}), section_5, draw=box-green,  text width=17em
		[\eg 
                    \citep{bu2025value}{,}
                    \citep{qiu2025gated}{,}
                    \citep{qwenai2026}{,}
                    \citep{bondarenko2023quantizable}{.},
            leaf3, text width=42em, section_5, draw=tree-green]
		]
	[\ \ \ Modified Softmax Functions (\S\ref{sec_5_2_Modified_Softmax_Functions}), section_5, draw=box-green,  text width=17em
		[\eg 
                    \citep{fu2026attention}{,}
                    \citep{kobyzev2025integral}{,}
                    \citep{zuhri2025softpick}{,}
                    \citep{fu2025sliding}{,}
                    \citep{zhangattention}{,}
                    \citep{hongvariance}{,}
                    \citep{gu2025attention}{,}
                    \citep{kaul2025attention}{,}
                    \citep{kaul2025attention}{,}
                    \citep{bondarenko2023quantizable}{.}, leaf3, text width=42em, section_5, draw=tree-green]
		]
	[\ \ \ \ \ Learnable Attention Bias (\S\ref{sec_5_3_Learnable_Attention_Bias}), section_5, draw=box-green,  text width=17em
		[\eg 
                    \citep{xiao2026mimo}{,}
                    \citep{fu2026attention}{,}
                    \citep{agarwal2025gpt}{,}
                    \citep{an2025systematic}{,}
                    \citep{gu2025attention}{,}
                    \citep{sun2024massive}{.}, leaf3, text width=42em, section_5, draw=tree-green]
		]
	[\ \ \ \ \ Pre-training Interventions (\S\ref{sec_5_4_Pre-training_Interventions}), section_5, draw=box-green,  text width=17em
		[\eg 
                    \citep{liang2025tweo}{,}
                    \citep{cappellazzo2025mitigating}{,}
                    \citep{team2025longcat}{,}
                    \citep{park2025outlier}{,}
                    \citep{kaul2025attention}{.}, leaf3, text width=42em, section_5, draw=tree-green]
		]
	[\ \ \ \ \ \ \ Additional Techniques (\S\ref{sec_5_5_Additional_Techniques_for_Mitigating_Attention_Sink}), section_5, draw=box-green,  text width=17em
		[\eg 
                    \citep{feng2026edit}{,}
                    \citep{kim2025text}{,}
                    \citep{xiang2025dfrot}{.}, leaf3, text width=42em, section_5, draw=tree-green]
		]
]
	[\ \ \ \ \ \ \ \ \ \ \ \ \ \ Applications and\\\ \ \ \ \ \ \ \ Practical Guidelines (\S\ref{sec_7_Applications}), appendix, draw=box-yellow, text width=16em
	[\ \ \ \ \ \ \ \ \ \ Model Pretraining (\S\ref{sec_7_1_Model_Pretraining}), appendix, draw=box-yellow,  text width=17em
        [\eg                      
                    \citep{liang2025tweo}{,}
                    \citep{bu2025value}{,}
                    \citep{park2025outlier}{,}
                    \citep{ge2025little}{,} 
                    \citep{xiao2026mimo}{,}
                    \citep{liang2025tweo}{,}
                    \citep{bu2025value}{,}
                    \citep{team2025longcat}{,}
                    \citep{agarwal2025gpt}{,}
                    \citep{simeoni2025dinov3}{,}
                    \citep{park2025outlier}{,}
                    \citep{qiu2025gated}{,}
                    \citep{zuhri2025softpick}{,}
                    \citep{wang2025vggt}{,}
        \textit{etc.} 
        , leaf2, text width=42em, appendix, draw=tree-yellow]
    ]
	[\ \ \ \ \ \ \ \ \ \ \ \  Model Tuning (\S\ref{sec_7_2_Model_Tuning}), appendix, draw=box-yellow,  text width=17em
        [\eg                    
                    \citep{liu2026surgery}{,}
                    \citep{chen2024rotary}{.} 
        , leaf2, text width=42em, appendix, draw=tree-yellow]
    ]
    [\ \ \ \ \ \ \ \ Efficient Inference (\S\ref{sec_7_3_Model_Inference}), appendix, draw=box-yellow,  text width=17em
        [\eg 
                    \citep{lu2025reward}{,}
                    \citep{bandyopadhyay2025block}{,}
                    \citep{aman2025bitmar}{,}
                    \citep{yang2025cacheclip}{,}
                    \citep{fang2025artificial}{,}
                    \citep{desai2026vattention}{,}
                    \citep{gu2025obcache}{,}
                    \citep{mu2025sals}{,}
                    \citep{zhu2025ojakv}{,}
                    \citep{fu2025h2eal}{,}
                    \citep{he2025trianglemix}{,}
                    \citep{qi2025deltallm}{,}
        \textit{etc.}
        , leaf2, text width=42em, appendix, draw=tree-yellow]
	]
	[\ \ \ Mechanism Interpretability (\S\ref{sec_7_4_Mechanism_Interpretability}), appendix, draw=box-yellow,  text width=17em
        [\eg                      
                    \citep{fu2026attention}{,}
                    \citep{wong2025existence}{,}
                    \citep{xiong2025dope}{,}
                    \citep{chen2025omnisparse}{,}
                    \citep{cappellazzo2025mitigating}{,}
                    \citep{queipo2026attention}{,}
                    \citep{salvatore2025lost}{,}
                    \citep{shang2025forgetting}{,}
                    \citep{rulli2025attention}{,}
                    \citep{luo2026sink}{,}
                    \citep{bu2025value}{,}
                    \citep{su2025kvsink}{,}
                    \citep{su2026unveiling}{,}
        \textit{etc.} 
        , leaf2, text width=42em, appendix, draw=tree-yellow]
    ]
	[\ \ \ \ \ \  Reducing Hallucination (\S\ref{sec_7_5_Reducing_Hallucination}), appendix, draw=box-yellow,  text width=17em
        [\eg                      
                    \citep{jiao2025don}{,}
                    \citep{tu2026attention}{,}
                    \citep{zhuang2025vasparse}{,}
                    \citep{kim2025text}{,}
                    \citep{zhang2026drives}{,}
                    \citep{zhang2025shallow}{,}
                    \citep{wang2025mirage}{,}
                    \citep{chenvocabulary}{,}
                    \citep{zhang2024seeing}{.} 
        , leaf2, text width=42em, appendix, draw=tree-yellow]
    ]
	[\ \ \ \ \ \ \ \  Safety \& Robustness (\S\ref{sec_7_6_Safety_Robustness}), appendix, draw=box-yellow,  text width=17em
        [\eg                      
                    \citep{shang2025forgetting}{,}
                    \citep{yona2025interpreting}{,}
                    \citep{yellapragada2025leveraging}{,}
                    \citep{wang2025mirage}{,}
                    \citep{pulfer2024robustness}{.} 
        , leaf2, text width=42em, appendix, draw=tree-yellow]
    ]
	[\ \ \ \ \ \ \ \ \ \ \ \ \ General Capability\\\ \ \ \ \ \ \ \ \ \ \ \  Enhancement (\S\ref{sec_7_7_General_Capability_Enhancement}), appendix, draw=box-yellow,  text width=17em
        [\eg                      
                    \citep{cappellazzo2025mitigating}{,}
                    \citep{li2025ctr}{,}
                    \citep{kobyzev2025integral}{,}
                    \citep{han2026zerotuning}{,}
                    \citep{wang2025position}{,}
                    \citep{lin2025look}{,}
                    \citep{dong2025hymba}{,}
                    \citep{bai2025does}{,}
                    \citep{yu2024unveiling}{,}
                    \citep{sandal2024zero}{.} 
        , leaf2, text width=42em, appendix, draw=tree-yellow]
    ]
	[\ \ \ Long-Context Enhancement (\S\ref{sec_7_8_Long-Context_Enhancement}), appendix, draw=box-yellow,  text width=17em
        [\eg                      
                    \citep{xiao2026mimo}{,}
                    \citep{yu2025sliding}{,}
                    \citep{yi2025deep}{,}
                    \citep{xiong2025dope}{,}
                    \citep{shin2026motionstream}{,}
                    \citep{liu2026rolling}{,}
                    \citep{qiu2025gated}{,}
                    \citep{chen2025edgeinfinite}{,}
                    \citep{fu2025sliding}{,}
                    \citep{zhang2025attention}{,}
                    \citep{yang2025seed}{,}
                    \citep{acharya2025star}{,}
                    \citep{bai2025does}{,}
                    \textit{etc.} 
        , leaf2, text width=42em, appendix, draw=tree-yellow]
    ]
	[\ \ \ \  Multi-Modal Enhancement (\S\ref{sec_7_9_Multi-Modal_Enhancement}), appendix, draw=box-yellow,  text width=17em
        [\eg                      
                    \citep{luo2026sink}{,}
                    \citep{khaki2025sparsevila}{,}
                    \citep{koo2025retovla}{,}
                    \citep{simeoni2025dinov3}{,}
                    \citep{lu2025artifacts}{,}
                    \citep{xiao2025focus}{,}
                    \citep{jiang2025vision}{,}
                    \citep{chen2025vision}{,}
                    \citep{baek2025large}{,}
                    \citep{feng2026edit}{,}
                    \citep{wang2025vggt}{,}
                    \citep{tu2026attention}{,}
                    \citep{yellapragada2025leveraging}{,}
        \textit{etc.} 
        , leaf2, text width=42em, appendix, draw=tree-yellow]
    ]
]
]
	\end{forest}
}
\caption{Organizational structure of our survey on AS in Transformers, covering AS across different models, fundamental utilization, mechanistic interpretation, strategic mitigation, and a summary of applications.}
\label{fig:arch-survey}
\end{figure}
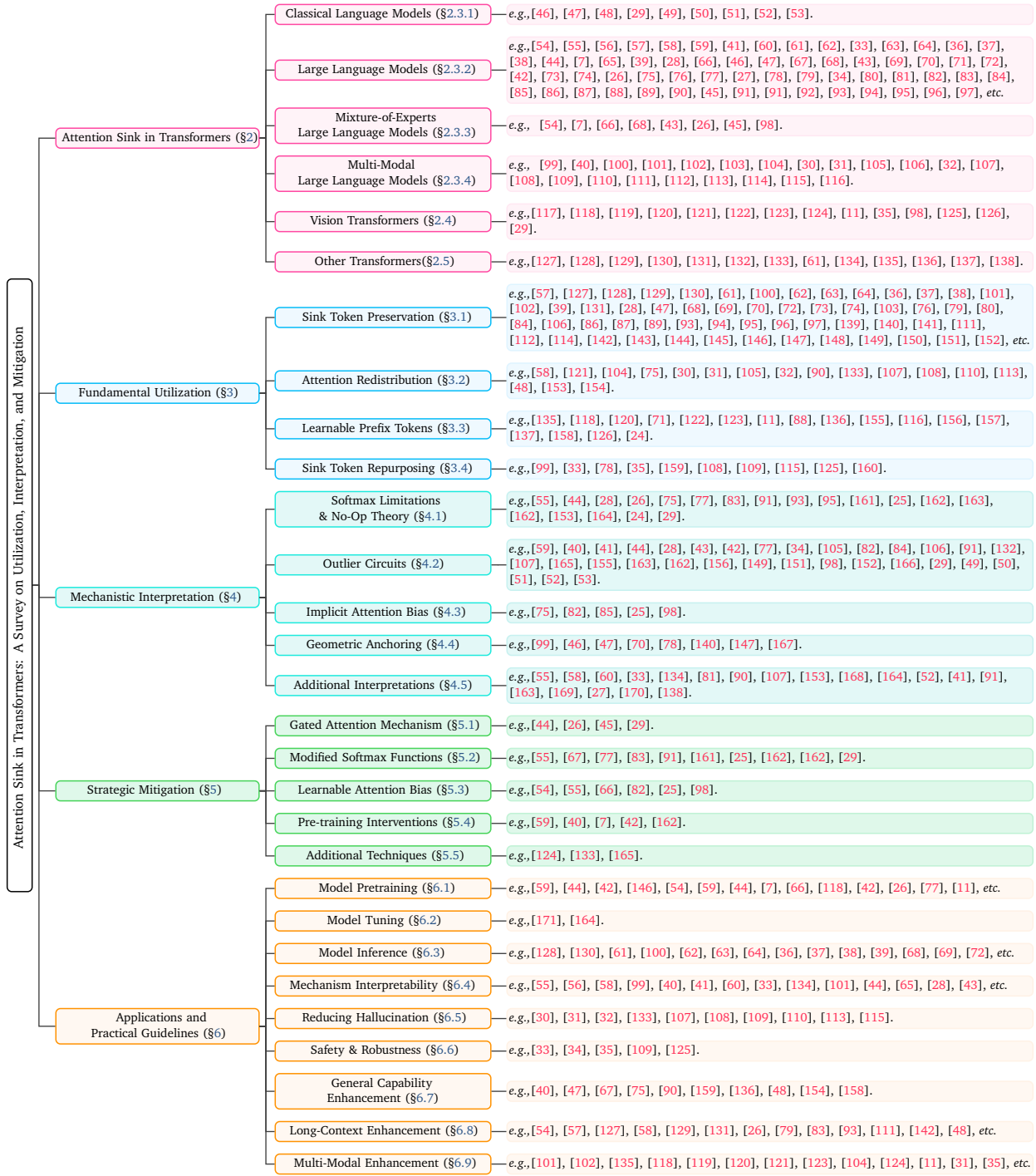
\clearpage
\subsection{Position and Contributions}
\label{sec_1_2_Position_and_Contributions}

\begin{figure}[t]
    \centering
    \vspace{-5mm}
    \includegraphics[width=0.95\linewidth]{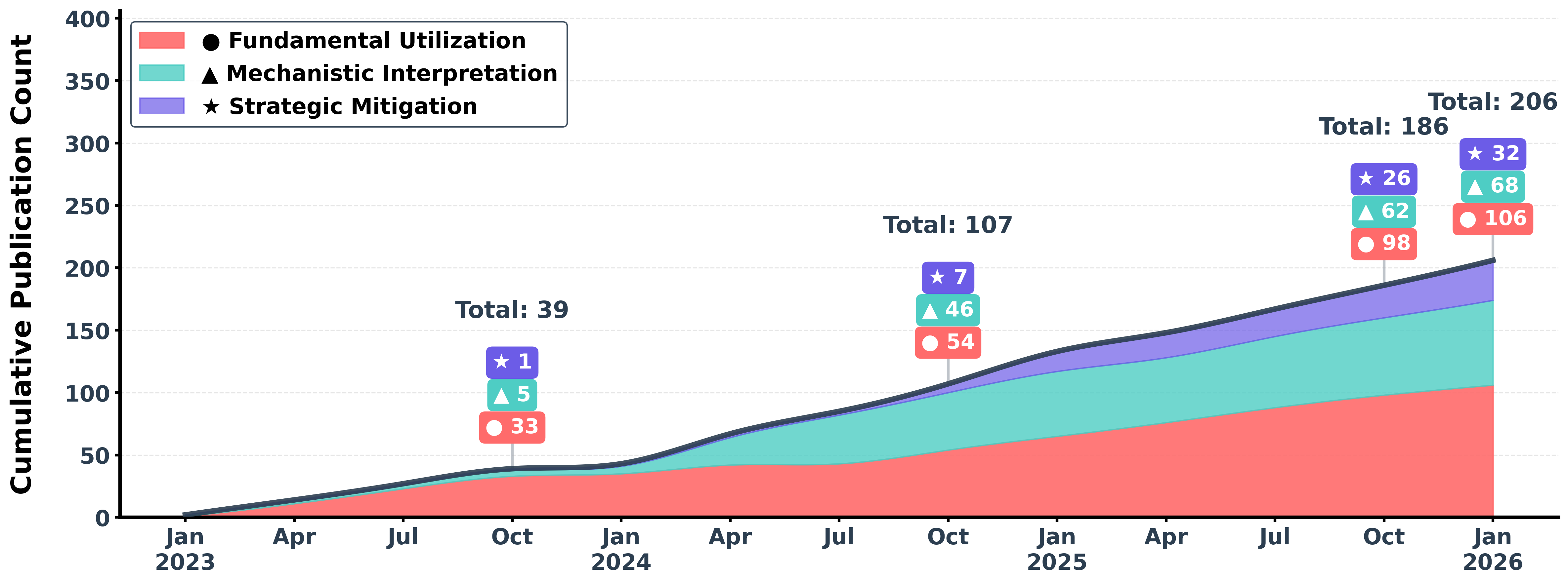}
    \vspace{-3mm}
    \caption{Cumulative publication count and temporal trends in AS research from 2023 to 2026. Early research focused on \textit{\textbf{Fundamental Utilization}} of AS, followed by studies investigating \textit{\textbf{Mechanistic Interpretation}}, and most recently, efforts targeting \textit{\textbf{Strategic Mitigation}} to address AS and improve model robustness.}
    \label{fig:num_research}
\end{figure}
% why these three parts
Building on the preceding analysis, this survey aims to address the lack of a systematic review in AS research.
A central focus of our work is to rigorously synthesize AS-related studies across diverse methodologies and Transformer models.
By conducting a comprehensive review and taxonomy of over 210 studies, we reveal a dynamic research landscape marked by cumulative progression.
As illustrated in Figure \ref{fig:num_research}, the field has steadily broadened its focus: starting from early empirical utilization, the community has progressively incorporated deeper mechanistic understanding and, most recently, systematic mitigation strategies. To capture this multi-dimensional ecosystem, we categorize the literature into three interrelated lines of research:

\begin{itemize}[leftmargin=*]
    \item \textbf{Initial Focus (2023–present) – \textit{\textbf{Fundamental Utilization}}.} 
    Early studies established the empirical utilization of AS \cite{xiao2024efficient,yu2024unveiling,hooper2024kvquant}, emphasizing the exploitation of its inherent characteristics or the management of its immediate effects. This line of research treats AS as a practical phenomenon to be exploited.
    
    \item \textbf{Deepening Understanding (2024–present) – \textit{\textbf{Mechanistic Interpretation}}.}
    As empirical applications matured, the community increasingly investigated the underlying causes and architectural factors contributing to AS \cite{sun2024massive,su2026unveiling,barbero2025llms}. This line prioritizes interpretability, aiming for a granular understanding of the internal mechanisms driving the phenomenon and the specific functional roles of AS.
    
    \item \textbf{Systematic Intervention (2025–present) – \textit{\textbf{Strategic Mitigation}}.}
    Building on mechanistic insights, the latest research focuses on direct structural mitigation. Studies demonstrate that AS-related extreme tokens can compromise training stability and hinder low-precision deployment \cite{qiu2025gated,liang2025tweo,bu2025value,park2025outlier}. Moreover, misallocated attention to uninformative tokens inherently limits overall model capacity \cite{kang2025see,yu2024unveiling,zhang2025shallow,tu2026attention}. As a result, developing robust mitigation frameworks has emerged as a critical frontier in current research.
\end{itemize}
% model + three parts
We draw on these three core developmental aspects of AS-related research, presenting our work as the first comprehensive survey of the field.
An overview of the survey structure is provided in Figure \ref{fig:paper_overview}.
% We begin by formally defining the AS and exploring its manifestations across diverse Transformer architectures, establishing a rigorous foundation for the survey.
% We then organize the core discussion around three key dimensions: \textit{\textbf{Fundamental Utilization}}, \textit{\textbf{Mechanistic Interpretation}}, and \textit{\textbf{Strategic Mitigation}}.
% Finally, we systematically discuss and categorize AS-related research according to its \textit{Core Application Domains}.
The detailed section structure is illustrated in Figure \ref{fig:arch-survey}.
Below, we summarize each section:

\begin{itemize}[leftmargin=*]

    \item \textbf{Attention Sink in Transformers \S \ref{sec_2_Attention_Sink_in_Transformers} .}
    This section first presents the preliminaries on Transformers and AS, followed by a comprehensive overview of AS across different Transformer architectures.
    We present the architectural overview of each model, highlight the characteristics of AS within them, and offer a preliminary summary of the AS-related research associated with these models.

    \item \textbf{Fundamental Utilization \S \ref{sec_3_Fundamental_Utilization}.}
    This section explores the basic utilization, including \textit{\textbf{Sink Token Preservation}} \cite{xiao2024efficient,zhang2023h2o,su2025kvsink}, \textit{\textbf{Attention Redistribution}} \cite{yu2024unveiling,tu2026attention,kang2025see}, \textit{\textbf{Learnable Prefix Tokens}} \cite{darcet2024vision,son2024prefixing,chen2025vision} and \textit{\textbf{Sink Token Repurposing}} \cite{wang2025mirage,li2024streamingdialogue,zhang2025shallow}.
    For each aspect, we present its core methodology, review practical approaches, and provide concluding insights.
    
    \item \textbf{Mechanistic Interpretation \S \ref{sec_4_Mechanistic_Interpretation}.}
    This section synthesizes current mechanistic understandings of AS, covering theories such as \textit{\textbf{Softmax Limitations \& No-Op Theory}} \cite{bondarenko2023quantizable,gu2025attention,su2025kvsink}, \textit{\textbf{Outlier Circuits}} \cite{an2025systematic,sun2024massive,su2026unveiling}, and \textit{\textbf{Implicit Attention Bias}} \cite{sun2024massive,an2025systematic,gu2025attention}.
    For each topic, we delineate the core concept, evaluate its supporting evidence, and provide our concluding insights.
    
    \item \textbf{Strategic Mitigation \S \ref{sec_5_Strategic_Mitigation}.}  
    This section examines strategies for systematically mitigating AS in transformer models, including \textit{\textbf{Gated Attention Mechanisms}} \cite{bu2025value, qiu2025gated, qwenai2026, bondarenko2023quantizable}, \textit{\textbf{Modified Softmax Functions}} \cite{kaul2025attention, gu2025attention, zuhri2025softpick}, \textit{\textbf{Learnable Attention Bias}} \cite{sun2024massive, an2025systematic, agarwal2025gpt}, and other approaches.  
    For each strategy, we present its core mechanism, review practical approaches, and offer concluding insights.

    \item \textbf{Applications and Practical Guidelines \S \ref{sec_7_Applications}.}  
    This section categorizes research by application domain and provides practical, actionable guidelines for managing AS.
    
    \item \textbf{Challenges and Future Directions \S \ref{sec_6_Challenges_and_Future_Directions}.}  
    This section delineates the principal challenges in AS research and outlines promising avenues for future investigation, highlighting several key opportunities to advance the field, including efficient and lightweight AS handling, lightweight adaptation for pretrained models, standardized benchmarks for AS and outlier mitigation, and other directions.
    
    \item \textbf{Appendix \ref{Appendix:A}: Comprehensive Overview of Surveyed Papers.}  
    This appendix presents a detailed summary table of the studies reviewed in this paper.
\end{itemize}

By following this cumulative developmental trajectory, we establish a coherent framework for the survey. 
In \S\ref{sec_3_Fundamental_Utilization}, \S\ref{sec_4_Mechanistic_Interpretation}, and \S\ref{sec_5_Strategic_Mitigation}, we systematically address Q1, Q2, and Q3, respectively. 
The main contributions of this survey are fourfold:

\begin{itemize}[leftmargin=*]
    \item \textbf{First Systematic Survey and Taxonomy of AS Research.}  
    We present the first comprehensive survey of AS research, systematically reviewing over 200 studies. A novel taxonomy organizes the literature into three principal dimensions: (1) \textit{\textbf{Utilization}}, the empirical use of AS; (2) \textit{\textbf{Interpretation}}, exploring its underlying mechanistic formulations; and (3) \textit{\textbf{Mitigation}}, strategies for managing AS. This taxonomy clarifies the conceptual landscape, enabling researchers to efficiently grasp both the current state and the developmental trajectory of AS research.

    \item \textbf{In-Depth Methodological Synthesis.}  
    For each dimension, we systematically consolidate the literature, distilling technical formulations, implementation strategies, and key insights. This synthesis offers researchers a clear understanding of core concepts and approaches, facilitating informed adoption, adaptation, and further methodological innovation.

    \item \textbf{Critical Insights and Future Directions.} 
    Building on our comprehensive review, we highlight persistent challenges and delineate promising directions for future research. This forward-looking roadmap is intended to inspire innovative research applications while critically guiding the development of next-generation Transformer models that are more robust, efficient, and interpretable.
    
    \item \textbf{Scenario-Driven Application Mapping and Guidelines.}  
    We further categorize AS research into nine distinct application scenarios and offer practical guidelines tailored to each application domain. This structured mapping provides researchers and practitioners with practical, actionable guidance.

\end{itemize}

% github repo
In addition to the survey, we have established a GitHub repository that systematically organizes the papers referenced in this work, available at \MYhref{https://github.com/ZunhaiSu/Awesome-Attention-Sink}{\textit{\textbf{https://github.com/ZunhaiSu/Awesome-Attention-Sink}}}. 
The repository is regularly maintained to incorporate the latest developments in AS research, providing researchers with convenient access to up-to-date studies and insights in this rapidly evolving field. % @Zunhai

%----------------------------------Section 2----------------------------------%
\clearpage
\section{Attention Sink in Transformers}
\label{sec_2_Attention_Sink_in_Transformers}

This section establishes the foundational context for the survey.  
We begin by reviewing the preliminaries of Transformers and AS.  
Building on this foundation, we systematically examine the specific manifestations of AS across diverse Transformer architectures.
 % @Zunhai

\subsection{Preliminaries on Transformers}
\label{sec_2_1_Foundations_of_Transformers}

\begin{figure}[t]
    \centering
    \includegraphics[width=1\linewidth]{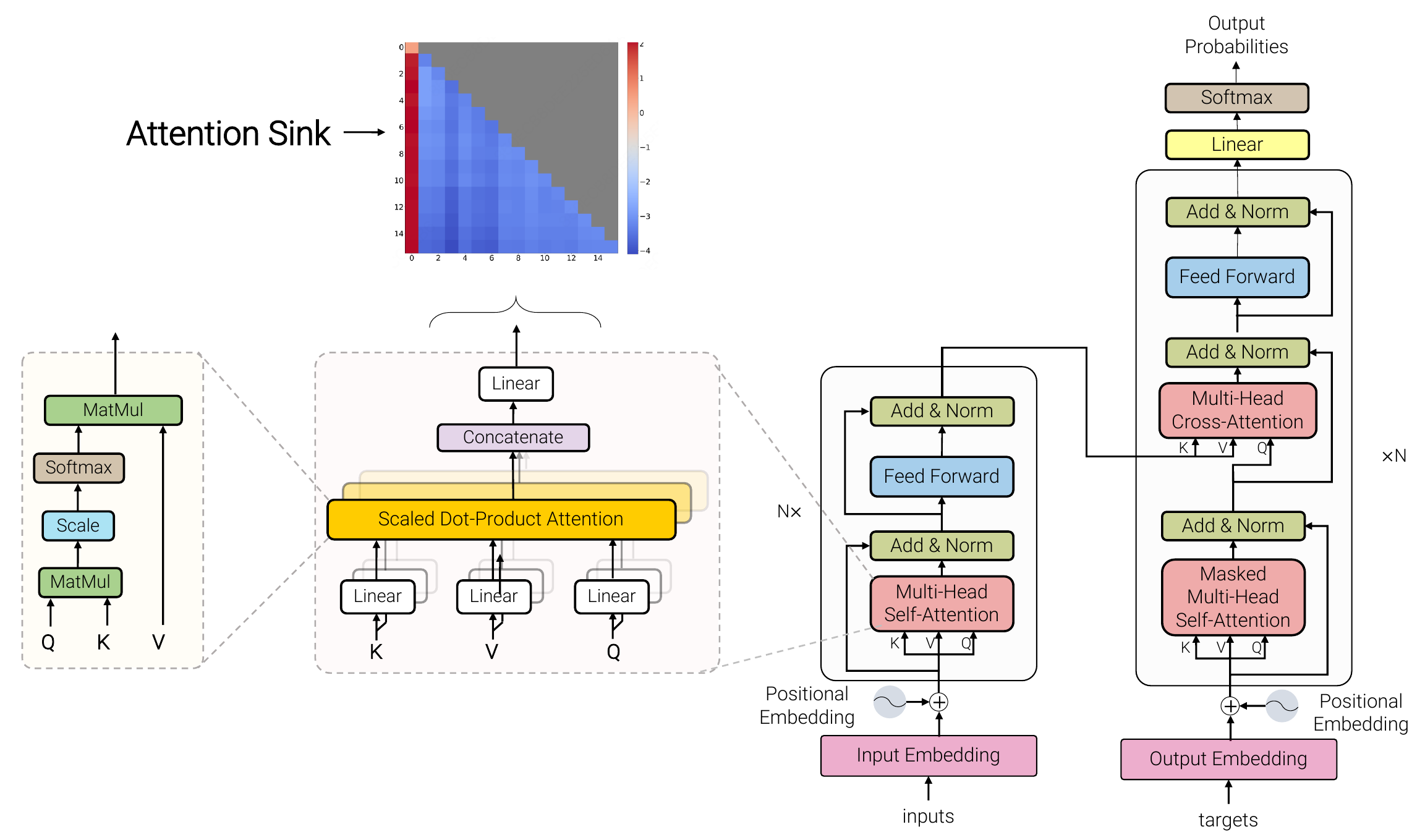}
    \caption{Architecture of the standard Transformer and an illustration of typical AS, where sink tokens exhibit exceptionally high attention scores.}
    \label{fig:transformer_arch}
\end{figure}

The Transformer architecture \cite{vaswani2017attention} established a non-recurrent sequence modeling paradigm based on an encoder-decoder framework. 
As illustrated in Figure \ref{fig:transformer_arch}, a standard Transformer block typically consists of two primary components: a multi-head self-attention (MHSA) module and a position-wise feed-forward network (FFN).
By leveraging the MHSA mechanism, the Transformer captures long-range global dependencies without the inductive bias inherent in sequential processing.

\paragraph{Multi-Head Self-Attention.}
The core of the Transformer is the MHSA, which enables the model to jointly attend to information from different representation subspaces at various positions. For an input sequence $\mathbf{X} \in \mathbb{R}^{N \times D}$, where $N$ denotes the sequence length and $D$ the feature dimension, the queries $\mathbf{Q}$, keys $\mathbf{K}$, and values $\mathbf{V}$ are obtained via linear projections:
\begin{equation}
    \mathbf{Q} = \mathbf{X}\mathbf{W}^Q, \quad \mathbf{K} = \mathbf{X}\mathbf{W}^K, \quad \mathbf{V} = \mathbf{X}\mathbf{W}^V,
\end{equation}
where $\mathbf{W}^Q, \mathbf{W}^K, \mathbf{W}^V \in \mathbb{R}^{D \times d_k}$ are learnable weight matrices, and $d_k$ denotes the dimensionality of each attention head. Attention is then computed as:
\begin{equation}
    \text{Attention}(\mathbf{Q}, \mathbf{K}, \mathbf{V}) = \text{Softmax} \left( \frac{\mathbf{Q}\mathbf{K}^T}{\sqrt{d_k}} \right) \mathbf{V}.
\end{equation}
% As a pivotal operator, Softmax normalizes affinity scores into a rigorous probability distribution, ensuring numerical stability by enforcing the attention weights to sum to unity. However, this rigid constraint introduces a critical structural vulnerability: when a query lacks semantically relevant keys within its context, the normalization requirement still compels attention heads to assign a non-negligible fraction of attention mass. Consequently, the model frequently defaults to concentrating this redundant attention on specific, semantically weak or uninformative tokens, effectively utilizing them as a numerical repository. 

% This behavior, formally conceptualized as the \textit{\textbf{Softmax Limitations \& the No-Op Theory}} \cite{bondarenko2023quantizable}, is widely recognized as a primary catalyst for the AS phenomenon. We delve into the mechanistic underpinnings of this theory in Section \ref{sec_4_Mechanistic_Interpretation}. 
% Recent advancements have sought to address this fundamental flaw by proposing \textit{\textbf{Modified Softmax}} mechanisms that relax the strict normalization requirement \cite{zuhri2025softpick,bondarenko2023quantizable}. 
% By incorporating an explicit "sink" directly into the mathematical formulation, these methods allow the model to "not attend" to any token, thereby effectively mitigating the emergence of AS and improving training stability. 
% We explore these strategic mitigation techniques in detail in Section \ref{sec_5_Strategic Mitigation}.

\paragraph{FFN and Residual Connections.}
Following the MHSA, a position-wise FFN is applied to each position independently, comprising two linear transformations interconnected by a non-linear activation $\sigma$:
\begin{equation}
    \text{FFN}(\mathbf{x}) = \sigma(\mathbf{x}\mathbf{W}_1 + \mathbf{b}_1)\mathbf{W}_2 + \mathbf{b}_2
\end{equation}
To stabilize training and mitigate the vanishing gradient problem, each sub-layer incorporates a residual connection \cite{he2016deep} followed by layer normalization (LayerNorm) \cite{ba2016layer}:
\begin{equation}
    \mathbf{X}_{out} = \text{LayerNorm}(\mathbf{X} + \text{SubLayer}(\mathbf{X}))
\end{equation}

This foundational architecture serves as the versatile backbone for various domain-specific adaptations. 
Despite their disparate input modalities and specialized architectural layers, these models all fundamentally rely on the Softmax attention mechanism as their core computational primitive. 
% Consequently, AS is not an isolated artifact of specific tasks or training data, but rather an inherent structural byproduct universally present across modern Transformer architectures \cite{kovaleva2021bert,sun2024massive,zhang2025shallow,jamal2026diffusion}. % @Zunhai

\subsection{Preliminaries on Attention Sink}
\label{sec_2_2_Foundations_of_Attention_Sink}

\subsubsection{Conceptual Background}

The concept of AS was first formally identified in autoregressive LLMs \cite{xiao2024efficient}, where initial tokens were observed to dominate the resulting attention distribution after Softmax normalization.  
As illustrated in Figure \ref{fig:transformer_arch}, the Key vectors corresponding to these early positions consistently attract attention from nearly all subsequent queries, appearing as attention outliers that substantially exceed those of ordinary tokens.

A widely discussed explanation links this phenomenon to the normalization behavior of Softmax, which forces attention mass to be distributed even when no strongly relevant key is available. \cite{bondarenko2023quantizable,zuhri2025softpick,kaul2025attention}.  
As a central component of the attention mechanism, Softmax converts raw affinity scores into a normalized probability distribution, ensuring numerical stability by enforcing that attention weights sum to unity.  
However, this rigid normalization introduces a structural vulnerability: when a query lacks semantically relevant keys within its context, the ``sum-to-one'' constraint still forces the model to distribute its attention mass.  
As a result, redundant attention often concentrates on specific tokens, effectively acting as a numerical reservoir that absorbs these excess scores.  
A detailed analysis of this mechanistic interpretation is provided in \textit{\textbf{Softmax Limitations \& No-Op Theory}} (Section~\ref{sec_4_1_Softmax_Limitations}).

Although AS has gained prominence in LLM research, this behavior is not limited to autoregressive models.  
In  classical language models such as BERT and RoBERTa \cite{devlin2018bert,liu2019roberta}, the effect has been empirically observed \cite{kovaleva2021bert,clark2019does,bondarenko2023quantizable,bai2025does}.  
Beyond classical language models, a wide range of Transformer-based architectures, such as Multimodal LLMs, Mixture-of-Experts LLMs, and ViTs, also exhibit consistent AS characteristics \cite{kang2025see,sun2024massive,rulli2025attention,su2026unveiling}.
Despite variations in attention masking and architectural specifics lead to divergent manifestations, the underlying principle remains: disproportionately high attention scores concentrate on specific tokens.  
Taken together, these findings suggest that AS is not peculiar to a single model family, but recur across diverse Transformer architectures \cite{sun2024massive, zhang2025shallow, jamal2026diffusion}.

\subsubsection{Attention Sink: Extreme Attention Concentration on Uninformative Tokens}
Across diverse Transformer architectures \cite{sun2024massive,xiao2024efficient,su2026unveiling,kovaleva2021bert,kang2025see}, AS tokens consistently attract disproportionately high attention despite carrying minimal semantic information \cite{sun2024massive,xiao2024efficient,gu2025attention}. Crucially, high attention alone does not suffice to characterize a sink token; the essential property is the mismatch between its disproportionately large attention mass and its limited semantic or task-specific contribution. Specifically, AS tokens exhibit two highly consistent and distinctive characteristics: (i) exceptionally high attention scores, and (ii) intrinsically low-information content (e.g., the \texttt{[BOS]} token in LLMs and background patch tokens in ViTs). Each characteristic is examined in detail below.

\noindent\textbf{Extremely High Attention Scores.}  
A defining feature of AS is that AS tokens receive exceptionally high attention scores. For instance, in LLaMA and other widely used LLMs, the first token frequently receives the maximum attention in 98\% of attention heads \cite{kaul2025attention}.
Based on this observation, a practical criterion for identifying AS, as used in prior work, is the threshold-based method.
Specifically, tokens whose cumulative attention scores significantly deviate from the global average are classified as AS tokens \cite{an2025systematic}. 

Formally, for a sequence of length $L$, let $\mathbf{A} \in \mathbb{R}^{L \times L}$ denote the attention weight matrix, where $A_{i,j}$ represents the attention weight from token $i$ to token $j$. The set of AS tokens is then given by:
\begin{equation}
    \mathcal{S}_{\text{AS}} = \left\{ j \;\middle|\; \underbrace{\sum_{i=1}^L A_{i,j}}_{\hat{A}_j} > \tau \cdot \mu_A \right\}, \quad 
    \mu_A = \frac{1}{L} \sum_{k=1}^L \hat{A}_k,
\end{equation}
where $\tau > 1$ is a relaxation threshold, empirically set to a large value (e.g., 1000 in \cite{an2025systematic}), $\mu_A$ denotes the mean cumulative attention score across all tokens and $\hat{A}_j$ denotes the cumulative attention score received by token j. This formulation highlights the extreme numerical prominence of AS tokens.

\noindent\textbf{Specific Low-Information Tokens.}  
In addition to receiving unusually high attention, AS tokens are often associated with tokens carrying limited task-relevant semantic content \cite{clark2019does,bondarenko2023quantizable,sun2024massive}.  

Across different architectures, AS tokens consistently correspond to uninformative tokens. Empirically documented categories include:

\begin{itemize}[leftmargin=*]
    \item \textbf{Classical Language Models}: Structural markers such as \texttt{[SEP]} and \texttt{[CLS]} \cite{clark2019does,bondarenko2023quantizable}.
    \item \textbf{Causal LLMs (dense and MoE)}: Initial tokens, strong delimiters, and weak-semantic tokens \cite{sun2024massive,su2026unveiling}.
    \item \textbf{Vision Transformers}: Low-information background patches \cite{sun2024massive,darcet2024vision}.
    \item \textbf{Multimodal LLMs}: Both text-side AS (e.g., \texttt{[BOS]}) inherited from causal LLMs, and vision-side AS occurring on low-information visual patches \cite{kang2025see}.
\end{itemize}

Collectively, these observations highlight that, across architectures and modalities, AS tokens consistently correspond to uninformative tokens that disproportionately attract attention.  
In the following sections, we present a systematic analysis of AS behaviors across different Transformer architectures.

% @Zunhai

\subsection{Language Models}
\label{sec_2_3_Language Models}
% BERT

In this section, we review the AS in language models, covering Classical Language Models, LLMs, Mixture-of-Experts LLMs, and Multi-modal LLMs. 
While all belong to the broader class of language models, they share common characteristics while also exhibiting differences arising from their distinct architectures.

For each model family, the discussion is systematically organized along three core dimensions: 
\textit{(i) Architectural Overview}, providing the necessary structural context; 
\textit{(ii) AS Characteristics and Manifestations}, detailing the specific emergence and behavior of the phenomenon; and 
\textit{(iii) Preliminary Summary of AS Research}, offering a concise synthesis of relevant studies that serves as a roadmap for the subsequent sections on AS utilization (\S \ref{sec_3_Fundamental_Utilization}), interpretation (\S \ref{sec_4_Mechanistic_Interpretation}), and mitigation (\S \ref{sec_5_Strategic_Mitigation}).
% Detailed discussions on each of these aspects can be found in Sections~\ref{sec_3_Fundamental_Utilization}, \ref{sec_4_Mechanistic_Interpretation}, and \ref{sec_5_Strategic_Mitigation}. % @Zunhai

\subsubsection{Classical Language Models}
\label{sec_2_3_1_Classical_Language Models}
% BERT

\paragraph{Architectural Overview.}
Classical Language Models (CLMs), exemplified by BERT \cite{devlin2018bert} and its robustly optimized successor RoBERTa \cite{liu2019roberta}, are fundamentally rooted in the encoder-only Transformer paradigm. 
Diverging from the foundational encoder-decoder framework, this architecture omits both the cross-attention mechanism and causal masking, employing a fully bi-directional self-attention instead. 
A definitive structural element of CLMs is the integration of specialized delimiter tokens—specifically, \texttt{[CLS]} (classification) at the sequence start and \texttt{[SEP]} (separator) between segments. 
These tokens serve as global semantic aggregators which, combined with absolute positional embeddings and the Masked Language Modeling (MLM) objective, inherently shape the emergence of distinct attention patterns \cite{clark2019does, kovaleva2021bert, luo2021positional}.

\begin{figure}[t]
    \centering
    \includegraphics[width=1\linewidth]{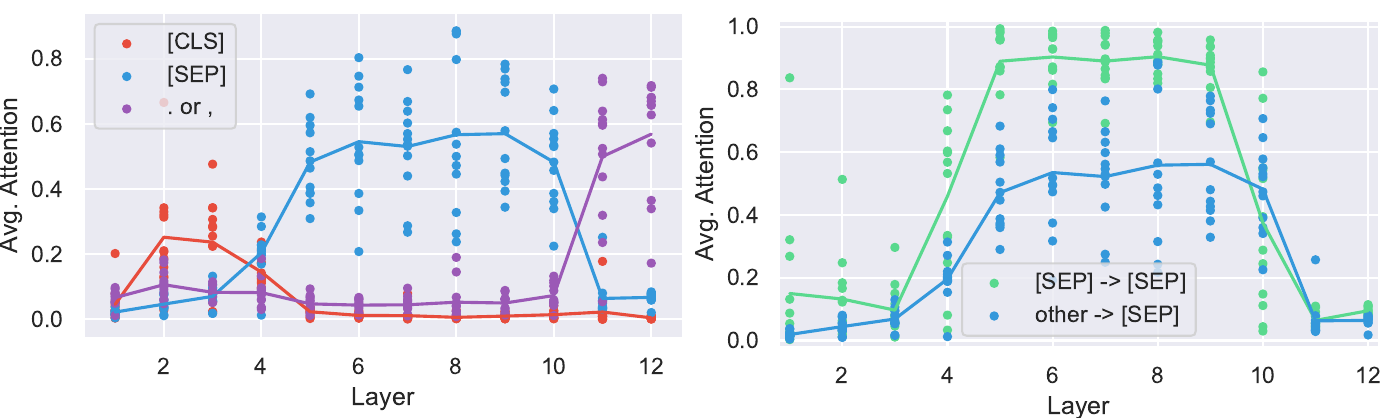}
    \caption{
    AS in BERT.
    Each point corresponds to the average attention a particular BERT attention head puts toward a token type.
    Left: heads often attend to “special” tokens. Early heads attend to \texttt{[CLS]}, middle heads attend to \texttt{[SEP]}, and deep heads attend to periods and commas.
    Often more than half of a head’s total attention is to these tokens.
    Right: heads attend to \texttt{[SEP]} tokens even more when the current token is \texttt{[SEP]} itself.
    The figure is adapted from \cite{clark2019does}.}
    \label{fig:bert_attn}
\end{figure}
\paragraph{Attention Sink Characterization.}
While AS gained prominence through LLMs research, the underlying phenomenon was empirically scrutinized in classical architectures long before the current scaling era. 
In CLMs, AS predominantly manifests as a persistent and intense concentration of attention mass on non-semantic special tokens, as illustrated in Figure \ref{fig:bert_attn}. 
Early diagnostic studies \cite{clark2019does, kovaleva2021bert} revealed that BERT's deeper layers consistently allocate a significant portion of attention towards \texttt{[SEP]} and \texttt{[CLS]} tokens, regardless of their semantic relevance to the query. 
These sinks are characterized by their fixed spatial positions, forming vertically persistent high-attention bands in attention maps \cite{kovaleva2021bert}.

\paragraph{Discussion and Synthesis of AS Research.}
The systematic presence of AS within CLMs has catalyzed diverse research trajectories that bridge practical utilization with mechanistic understanding. 
At the level of \textit{\textbf{Fundamental Utilization}}, studies explore the basic use of sink properties, such as redistributing attention mass \cite{bai2025does} to stabilize contextual representations. 
This empirical success is further elucidated through \textit{\textbf{Mechanistic Interpretation}}, where researchers explain these sinks via the \textit{\textbf{Softmax Limitations \& No-Op Theory}} (\S \ref{sec_4_1_Softmax_Limitations}) \cite{bondarenko2023quantizable}, identifying special tokens as repositories for redundant attention mass. 
Such behavior is intrinsically linked to the emergence of \textit{\textbf{Outlier Circuits}} (\S \ref{sec_4_2_Outliers_Circuits}) \cite{kovaleva2021bert, bondarenko2021understanding} and the formation of \textit{\textbf{Geometric Anchoring}} (\S \ref{sec_4_4_Geometric_Anchoring}) sites \cite{ruscio2025you} that stabilize the representation space. 
Building on these insights, \textit{\textbf{Strategic Mitigation}} efforts address the negative impacts of AS, particularly the numerical artifacts that pose a primary bottleneck for model quantization \cite{bondarenko2021understanding, bondarenko2023quantizable}. 
 % @Zunhai

\subsubsection{Large Language Models}
\label{sec_2_3_2_Large_Language_Models}
% LLM

\begin{figure}[t]
    \centering
    \includegraphics[width=1\linewidth]{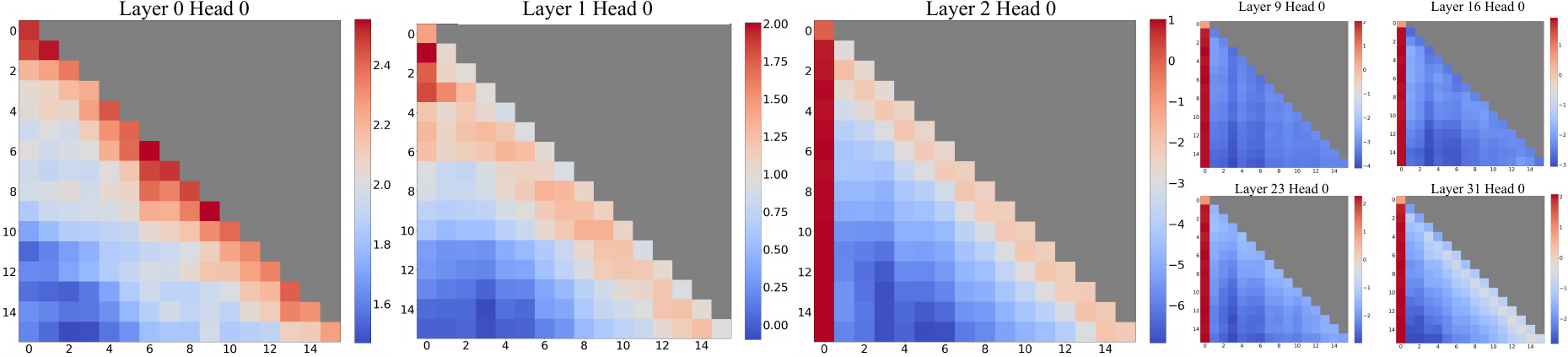}
    \caption{Visualization of average attention logits across Llama-2-7B. 
        Two distinct structural patterns are observed: 
        (i) The initial layers (layers 0 and 1) exhibit a "local" attention distribution, where attention is predominantly allocated to the most recent context. 
        (ii) In subsequent deeper layers, the model demonstrates a consistent and pronounced concentration of attention toward the initial token across all heads. 
        The figure is adapted from \cite{xiao2024efficient}.}
    \label{fig:llm_sink}
\end{figure}
\paragraph{Architectural Overview.}
\begin{wrapfigure}{r}{0.2\textwidth} 
\vspace{-13mm}
    \centering
    \includegraphics[width=\linewidth]{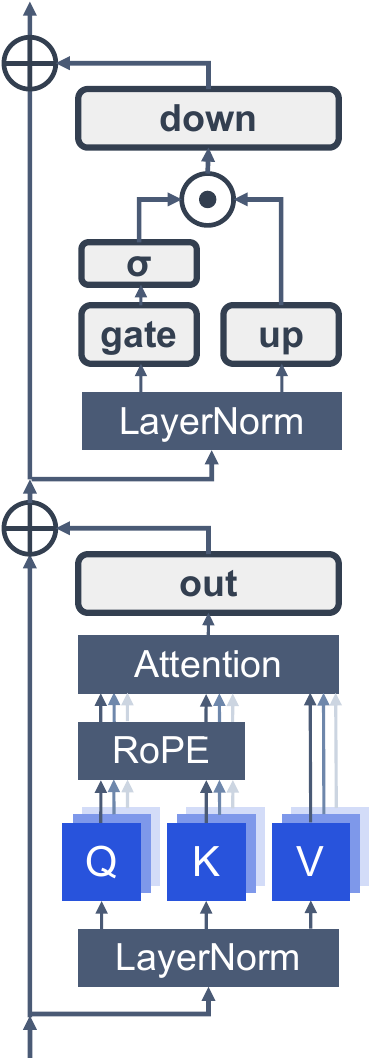}  
    \vspace{-7mm}
    \caption{Structural overview of a representative decoder-only LLM. 
        Adapted from \cite{su2025kvsink}.}
    \label{fig:llm_arch}
    \vspace{-10mm}
\end{wrapfigure}
Modern LLMs represent a specialized adaptation of the Transformer paradigm, fundamentally rooted in the decoder-only configuration. The structural layout of these models is illustrated in Figure~\ref{fig:llm_arch}. A defining constraint inherited from the decoder-only architecture is the causal masking mechanism, which ensures that each query vector $\mathbf{q}_i$ at position $i$ can only attend to preceding key vectors $\mathbf{k}_j$ where $j \le i$. Formally, the attention pattern is defined as:
\begin{equation}
    \text{Attention}(\mathbf{Q}, \mathbf{K}, \mathbf{V}) = \text{softmax}\left(\frac{\mathbf{Q}\mathbf{K}^\top}{\sqrt{d_k}} + \mathbf{M}\right)\mathbf{V},
\end{equation}
where $\mathbf{M}$ is the causal mask with $M_{ij} = -\infty$ for $j > i$ and $0$ otherwise. In this setting, only the initial tokens are visible to the entire sequence, making them the most stable candidates for attention offloading \cite{xiao2024efficient, gu2025attention}.

Beyond causal masking, contemporary LLMs incorporate a suite of architectural refinements that collectively enhance training stability, model expressivity, and inference efficiency. For normalization, pre-normalization with Root Mean Square Layer Normalization (RMSNorm) \cite{zhang2019root} has largely replaced the original post-LN design, mitigating gradient variance and enabling more stable training at scale. The feed-forward network has been upgraded from the original two-layer MLP to Gated Linear Units (GLU), with SwiGLU emerging as the predominant variant due to its superior trade-off between expressivity and computational cost \cite{shazeer2020glu}. For positional encoding, Rotary Positional Embeddings (RoPE) \cite{su2024roformer} encode relative position information through rotation matrices, offering improved length extrapolation capabilities compared to absolute or learnable positional embeddings. 
% To address the quadratic memory growth of key-value caches during long-sequence inference, modern architectures employ attention variants such as Grouped-Query Attention (GQA) \cite{ainslie2023gqa}, which reduces the number of key-value heads while maintaining query head diversity, or Multi-Head Latent Attention (MLA) \cite{liu2024deepseek}, which compresses key-value states through low-rank projections.
% Collectively, these refinements enable LLMs to scale to hundreds of billions of parameters while maintaining stable training dynamics and efficient inference—a crucial foundation for understanding attention distribution patterns and the emergence of attention sink phenomena.

\paragraph{Attention Sink Characterization.}
In LLMs, AS is empirically characterized by a persistent and disproportionate concentration of attention mass on specific early-stage tokens, irrespective of their semantic contribution \cite{xiao2024efficient}. This manifestation is evidenced in Figure~\ref{fig:llm_sink}, where attention heatmaps reveal a distinct vertical attention stripe anchored at the sequence start that remains invariant across diverse input contents and generation lengths.

Detailed diagnostic studies across multiple model families \cite{sun2024massive, guo2024attention} reveal that while sinks are predominantly anchored on the first token, they also frequently emerge on strong delimiters (e.g., periods, newlines) and weak-semantic tokens that serve as structural rather than content-bearing units. The distribution of this concentration exhibits a pronounced layer-wise escalation: relatively subtle in early layers, the intensity of attention offloading grows substantially in intermediate and deep layers, where global context integration becomes most critical \cite{sun2024massive, xiao2024efficient}.

Beyond these static structural patterns, recent empirical investigations \cite{guo2024attention} uncover that AS is not merely an architectural artifact but an emergent property that materializes only after sufficient optimization on adequate training data—typically emerging during the pre-training phase as the model converges. This emergence coincides with the stabilization of optimization dynamics, suggesting that AS formation is intricately linked to the convergence of attention head specialization. Furthermore, the strength of this concentration exhibits systematic sensitivity to optimization hyperparameters: models trained with higher learning rates and substantial weight decay develop more pronounced AS, whereas lower learning rates or minimal weight decay yield weaker or delayed sink formation \cite{guo2024attention}. Collectively, these findings characterize AS as a robust and predictable phenomenon shaped by both architectural constraints and training dynamics.

\paragraph{Discussion and Synthesis of AS Research.}

Within the scope of \textit{\textbf{Fundamental Utilization}}, a representative category of methodologies focuses on \textit{\textbf{Sink Tokens Preservation}} (\S \ref{sec_3_1_Preserving_Sink_Tokens}). 
In token pruning or streaming applications \cite{xiao2024efficient,zhang2023h2o,xiao2024infllm,han2024lm}, preserving these initial anchors is essential to prevent the catastrophic collapse of model performance. 
Moreover, the AS pattern has been incorporated as a default heuristic in sparse attention research and KV cache pruning to ensure structural stability during sequence processing \cite{jiang2024minference,zhao2024buzz,zeng2025subkv,cai2024pyramidkv}. 
\textit{\textbf{Attention Redistribution}} (\S \ref{sec_3_2_Attention_Redistribution}) serves as another representative approach, where previous studies demonstrate that reallocating attention weights can mitigate excessive concentration on initial tokens to improve the overall efficiency of the attention mechanism \cite{yu2024unveiling,jo2024a2sf,wang2025position}.

Regarding \textit{\textbf{Mechanistic Interpretation}}, beyond \textit{\textbf{Softmax Limitations \& the No-Op Theory}} (\S \ref{sec_4_1_Softmax_Limitations}), certain studies focusing on \textit{\textbf{Outliers Circuits}} (\S \ref{sec_4_2_Outliers_Circuits}) argue that AS functions as a manifestation of attention outliers, which are systematically linked to structural outliers including weight and activation outliers \cite{an2025systematic, sun2024massive, su2025kvsink, su2026unveiling}. 
These studies suggest that such sinks emerge and vanish in coordination with these systematic irregularities during model inference. 
Another significant line of research interprets this phenomenon as an \textit{\textbf{Implicit Attention Bias}} (\S \ref{sec_4_3_Implicit_Attention_Bias}), addressing the absence of an explicit bias term in standard attention computations. 
These studies further suggest that incorporating the explicit attention bias term \cite{sun2024massive, an2025systematic, gu2025attention, agarwal2025gpt, xiao2026mimo} can effectively reduce the model’s dependence on AS.

In the context of \textit{\textbf{Strategic Mitigation}}, \textit{\textbf{Gated Attention Mechanisms}} (\S \ref{sec_5_1_Gated_Attention_Mechanisms}) introduce input-dependent gating to the attention outputs, effectively addressing Softmax Limitations and mitigating AS. 
Recent studies show that such mechanisms can substantially reduce post-quantization degradation while enhancing overall model performance \cite{qiu2025gated, bu2025value,bondarenko2023quantizable,qwenai2026}. 
Beyond architectural gating, \textit{\textbf{Modified Softmax Functions}} (\S \ref{sec_5_2_Modified_Softmax_Functions}) seek to alleviate the sink effect by refining the normalization process. 
For instance, \textit{Softpick} \cite{zuhri2025softpick} employs a soft-thresholding mechanism to truncate low-probability attention scores, thereby eliminating both AS and the associated \textit{Massive Activations} \cite{sun2024massive} without compromising the model’s representational capacity.

% @Zunhai

\subsubsection{Mixture-of-Experts Large Language Models}
\label{sec_2_3_3_MoE_Large_Language_Models}
% MoE

\begin{figure}[t]
    \centering
    \includegraphics[width=0.7\linewidth]{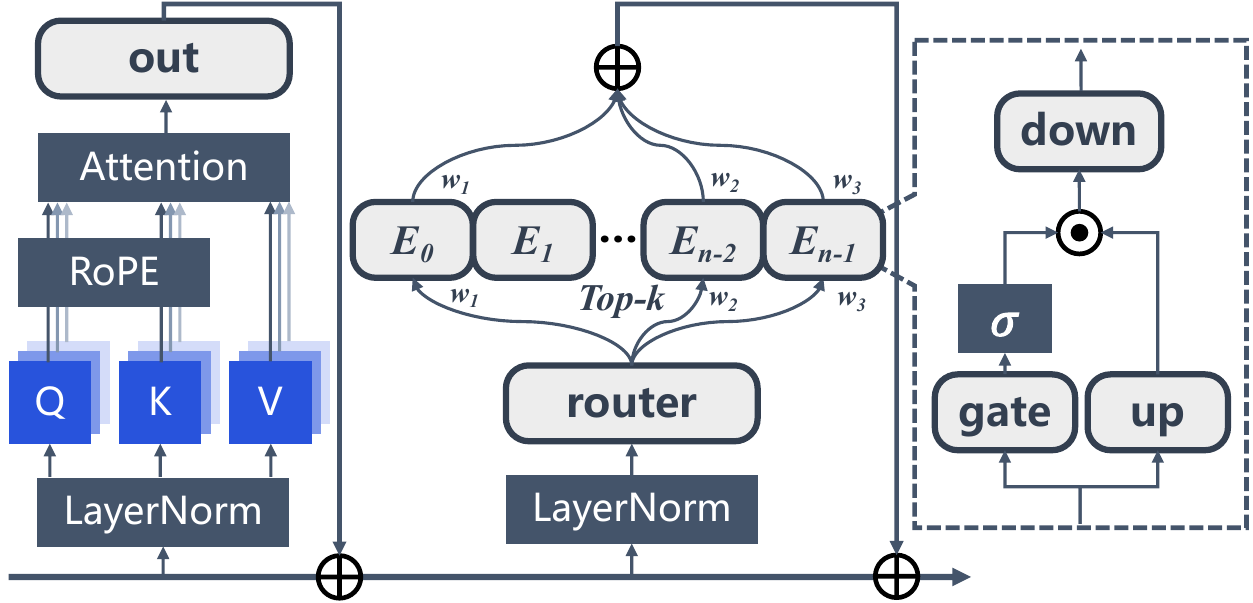}
    \caption{Decoder Architecture of MoE LLM. The figure is adapted from \cite{su2026unveiling}.}
    \label{fig:moe}
\end{figure}
\paragraph{Architectural Overview.}
Mixture-of-Experts (MoE) LLMs extend the vanilla Transformer architecture by substituting the static feed-forward network with a sparse MoE layer, as illustrated in Figure~\ref{fig:moe}. The hidden representation after multi-head self-attention, $\mathbf{H}^{l'} \in \mathbb{R}^{n \times d}$, passes through Layer Normalization and is fed into the MoE layer. A router network determines which experts to activate via the weight matrix $\mathbf{W}_G \in \mathbb{R}^{d \times E}$, where the routing weights $\mathbf{G} \in \mathbb{R}^{n \times E}$ are computed as:
\begin{equation}
\mathbf{G} = \text{softmax}(\mathbf{H}^{l'} \mathbf{W}_G).
\end{equation}
Sparse activation of the experts is achieved by selecting the top-$k$ routing weights for each input token, producing the MoE layer output:
\begin{equation}
\text{MoE}(\mathbf{H}^{l'}) = \sum_{i \in \text{Top-}k(\mathbf{G}_j)} \mathbf{G}_{ji} \cdot \text{FFN}(\text{LN}_{\text{moe}}(\mathbf{H}^{l'}_j)), \quad \forall j = 1 \dots n.
\end{equation}

In dense LLMs, AS emerges as a stable pattern anchored to the initial tokens. In MoE LLMs, the sparse activation mechanism dynamically routes different tokens to distinct experts during inference. The interaction between the AS mechanism and the MoE architecture gives rise to unique AS manifestations in MoE LLMs, where the distribution of AS may influence or be influenced by expert routing decisions.

\begin{figure*}[t]
    % \vspace{-3mm}
    \centering
    % 第一行
    \begin{subfigure}[b]{0.49\textwidth}
        \centering
        \includegraphics[width=\textwidth]{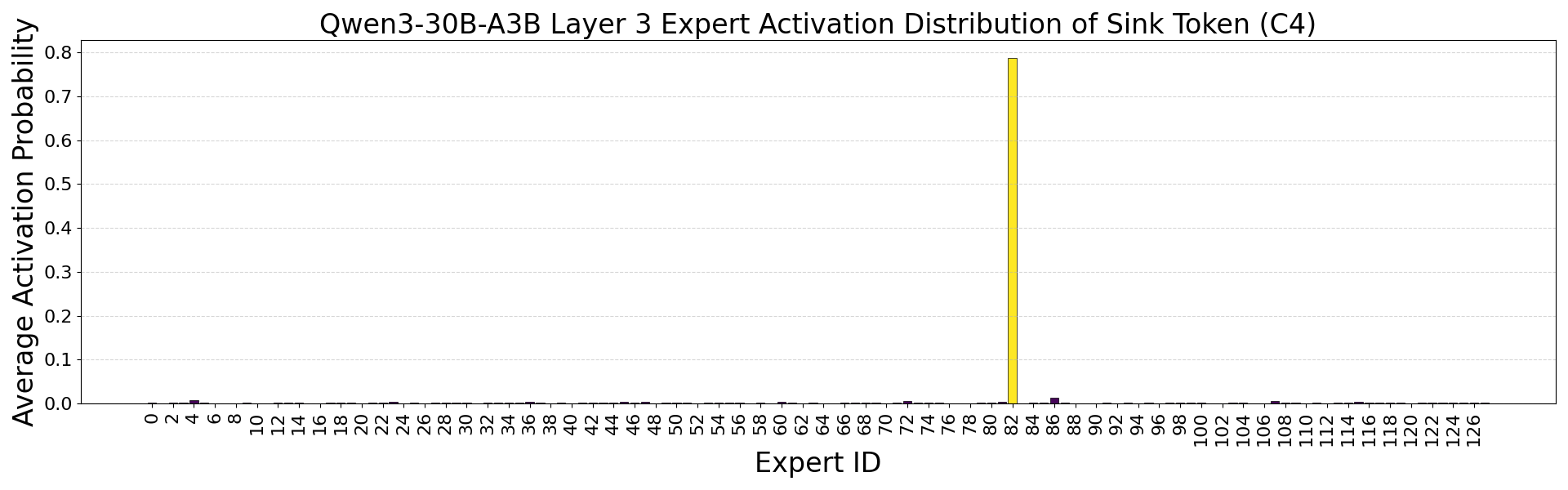}
        \caption{Sink token in Qwen3-30B-A3B.}
    \end{subfigure}
    \hfill
    \begin{subfigure}[b]{0.49\textwidth}
        \centering
        \includegraphics[width=\textwidth]{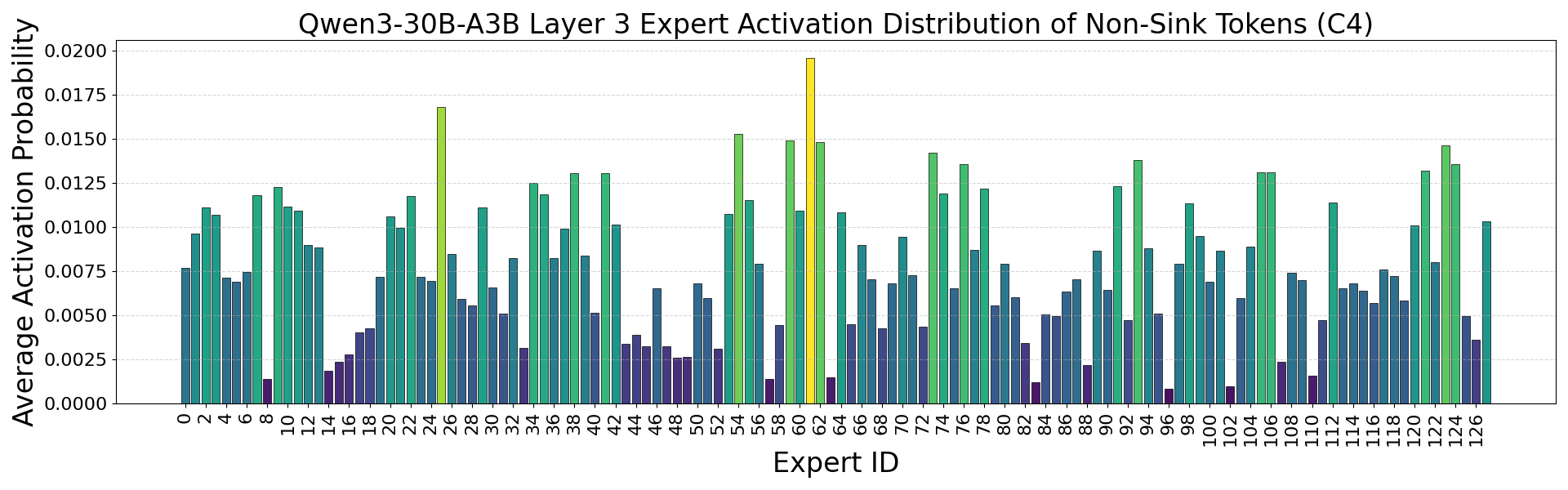}
        \caption{Non-sink tokens in Qwen3-30B-A3B.}
    \end{subfigure}

    % 第二行
    \vspace{0.3cm}
    \begin{subfigure}[b]{0.49\textwidth}
        \centering
        \includegraphics[width=\textwidth]{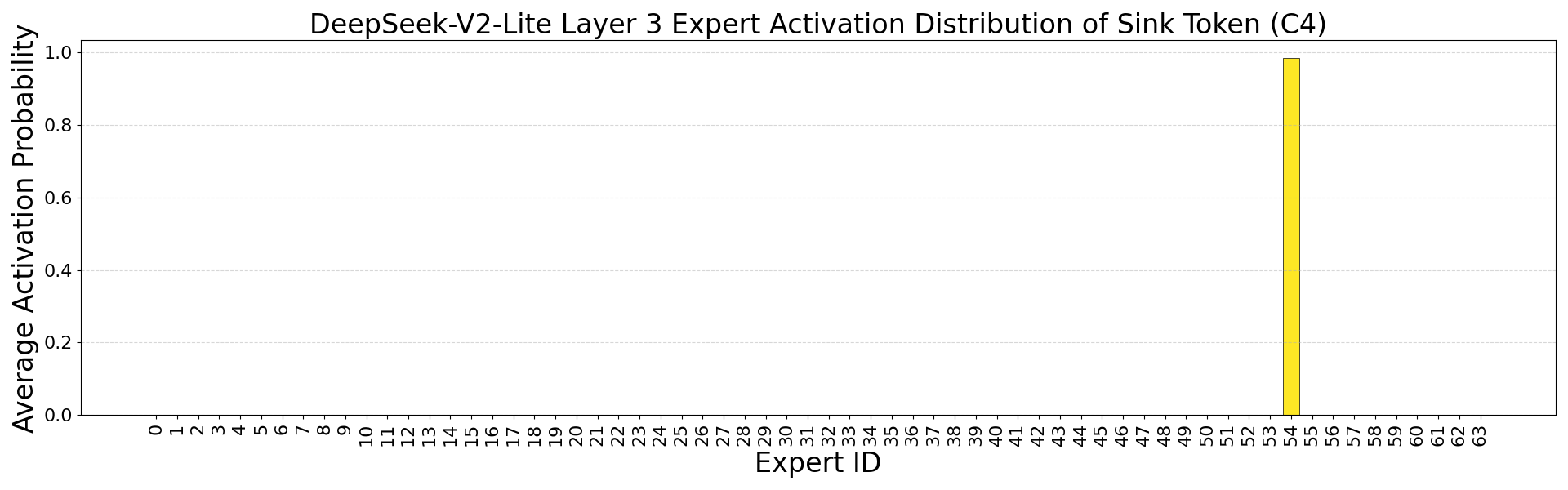}
        \caption{Sink token in DeepSeek-V2-Lite.}
    \end{subfigure}
    \hfill
    \begin{subfigure}[b]{0.49\textwidth}
        \centering
        \includegraphics[width=\textwidth]{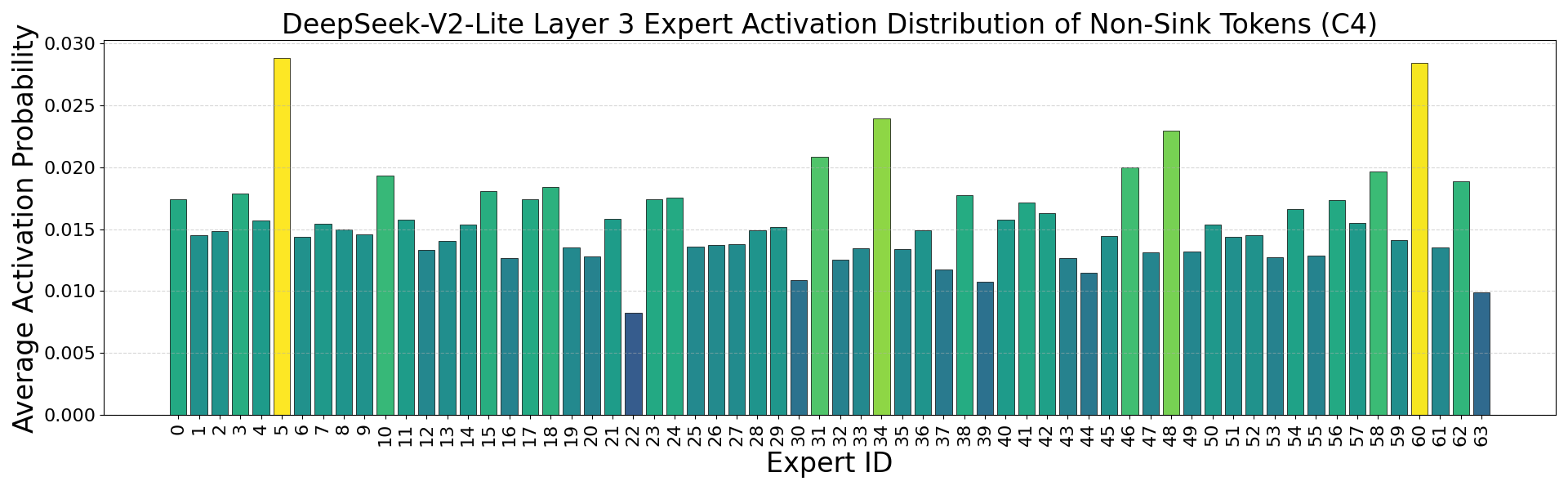}
        \caption{Non-sink tokens in DeepSeek-V2-Lite.}
    \end{subfigure}
    
    \caption{Expert router score distributions for sink and non-sink tokens. Sink tokens receive particularly high scores in super experts, whereas non-sink tokens have more evenly distributed scores across all experts. The figure is adapted from \cite{su2026unveiling}.}
    \vspace{-3mm}
    \label{fig:moe_routermap}
\end{figure*}

\paragraph{Attention Sink Characterization.}
While the AS patterns in MoE LLMs generally align with those observed in dense architectures, recent evidence reveals a strong interplay between the MoE structure and the emergence of AS. Empirical investigations uncover that the formation of AS is intrinsically tied to a highly sparse subset of experts, termed \textit{Super Experts} \cite{su2026unveiling}. 
Despite their extremely limited number these experts play a pivotal role in MoE forward inference. 
For instance, pruning just three out of 6,144 experts in Qwen3-30B-A3B causes catastrophic performance degradation. 
Empirical evidence indicates that \textit{Super Experts} constitute the primary source of the systematic outlier mechanism responsible for AS in MoE LLMs \cite{su2026unveiling}.  
As shown in Figure \ref{fig:moe_routermap}, despite the use of auxiliary expert-balancing losses during MoE LLM pre-training, sink tokens consistently attain high router scores on \textit{Super Experts}, effectively ensuring that AS is primarily activated within these experts.
Crucially, compressing or pruning this minimal set of \textit{Super Experts} disrupts the outlier-driven mechanism, leading to the collapse of AS and a subsequent deterioration of model coherence, reasoning capabilities, and output quality.
% These \textit{Super Experts} are characterized by rare but extreme activation outliers in their down-projection outputs, which directly generate \textit{Massive Activations} in hidden states \cite{sun2024massive} that constitute AS. The distribution of \textit{Super Experts} is model-specific but data-agnostic, remaining stable across post-training procedures and input semantics. 

% \begin{figure}[t]
%     \centering
%     \includegraphics[width=1\linewidth]{Figures/Section_2/moe_sink.pdf}
%     \caption{Systematic outlier mechanism in Qwen3-30B-A3B MoE LLM. The figure is adapted from \cite{su2026unveiling}.}
%     \label{fig:moe_sink}
% \end{figure}
\paragraph{Discussion and Synthesis of AS Research.}
Regarding \textit{\textbf{Mechanistic Interpretation}}, recent studies focusing on \textit{\textbf{Outlier Circuits}} (\S \ref{sec_4_2_Outliers_Circuits}) exemplify how AS is intrinsically linked to the emergence of \textit{Massive Activations} that bias routing logits \cite{su2026unveiling, sun2024massive}. In terms of \textit{\textbf{Strategic Mitigation}}, contemporary MoE architectures such as \textit{Qwen3-Next} employ \textit{\textbf{Gated Attention Mechanisms}} (\S \ref{sec_5_1_Gated_Attention_Mechanisms}) to alleviate AS and prevent expert collapse \cite{qwenai2026}. Meanwhile, models including \textit{GPT-OSS} and \textit{MiMo-V2-Flash} employ \textit{\textbf{Learnable Attention Bias}} (\S \ref{sec_5_3_Learnable_Attention_Bias}) to effectively absorb and redirect attention, alleviating the impact of AS \cite{agarwal2025gpt, xiao2026mimo}.
Furthermore, \textit{LongCat-Flash} introduces a \textit{\textbf{Pre-Training Prevention}} (\S \ref{sec_5_4_Pre-training_Interventions}) strategy by incorporating auxiliary losses to suppress AS and \textit{Massive Activations} directly during pre-training \cite{team2025longcat}. As MoE structures become the predominant paradigm for LLMs, the systematic elimination of AS has became a fundamental design requirement.% @Zunhai

\subsubsection{Multi-Modal Large Language Models}
\label{sec_2_3_4_MM_Large_Language_Models}
% MLLM

\paragraph{Architectural Overview.}
Multi-modal LLMs (MLLMs) extend the standard Transformer architecture by integrating a vision encoder with a causal LLM backbone via a cross-modal connector. Formally, given an input image $\mathbf{x} \in \mathbb{R}^{H \times W \times C}$, the vision encoder first extracts a sequence of visual tokens:
\begin{equation}
    \mathbf{V} = \{\mathbf{v}_1, \mathbf{v}_2, \ldots, \mathbf{v}_N\} = f_{\text{vision}}(\mathbf{x}), \quad \mathbf{v}_i \in \mathbb{R}^{D_{\text{vision}}},
\end{equation}
where $N$ denotes the number of patches and $f_{\text{vision}}$ represents the vision encoder. These visual tokens are then projected via a cross-modal connector $\mathcal{P}$ to align with the LLM's embedding space:
\begin{equation}
    \mathbf{V}' = \mathcal{P}(\mathbf{V})
    = \{\mathbf{v}'_1, \mathbf{v}'_2, \ldots, \mathbf{v}'_N\},
    \quad \mathbf{v}'_i \in \mathbb{R}^{D_{\text{llm}}}.
\end{equation}
The projected visual tokens $\mathbf{V}'$ are concatenated with textual tokens $\mathbf{T} = \{\mathbf{t}_1, \ldots, \mathbf{t}_M\}$ to form the full input sequence $\mathbf{S} = [\mathbf{V}', \mathbf{T}]$, which is subsequently processed by the causal LLM.

Unlike text-only Transformers, MLLMs operate over heterogeneous receptive fields, requiring textual queries to attend to information-rich visual patches that are inherently non-causal. This multi-modal integration forces the attention mechanism to reconcile magnitude or variance disparities between visual and textual embeddings, directly influencing the emergence and spatial distribution of AS during multimodal inference.

\begin{figure}[t]
    \centering
    \vspace{-3mm}
    \includegraphics[width=0.95\linewidth]{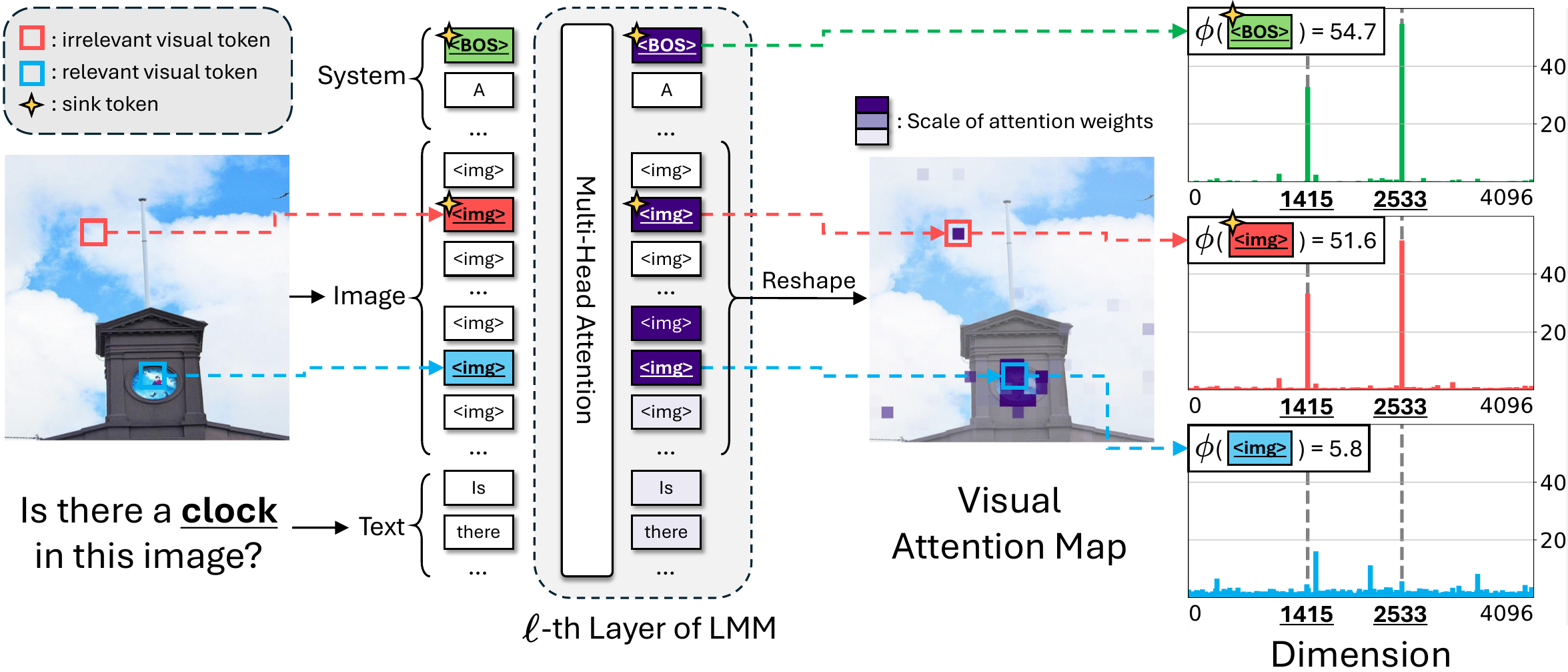}
    % \vspace{-3mm}
    \caption{Visualization and characterization of \textit{Visual Attention Sinks} in MLLMs. 
        Semantically irrelevant visual tokens (indicated by red boxes) exhibit \textit{Massive Activations} within specific dimensions of their hidden states. 
        In contrast, task-relevant visual tokens (indicated by blue boxes) maintain stable activation profiles without such numerical anomalies. 
        This phenomenon mirrors the behavior of established text AS, suggesting a consistent underlying mechanism of AS across both visual and textual modalities. 
        The figure is adapted from \cite{kang2025see}.
}
    \vspace{-3mm}
    \label{fig:mllm_sink}
\end{figure}
\paragraph{Attention Sink Characterization.}
In MLLMs, AS manifests as a multimodal concentration phenomenon, where attention weights are disproportionately allocated to both initial textual tokens inherited from the causal LLM backbone and specific visual anchors introduced through cross-modal fusion.
Empirical investigations reveal the emergence of \textit{Visual Attention Sinks}: particular visual tokens, often corresponding to background patches or non-semantic regions, that attract excessive attention regardless of their relevance to the textual prompt \cite{kang2025see} (Figure~\ref{fig:mllm_sink}). These visual sinks act as attention absorbers, sequestering redundant attention scores and producing a scattered sink pattern that diverts focus from semantically important object regions \cite{kang2025see}.
Further analysis reveals a distinct layer-wise distribution: visual sinks are prevalent in shallow layers of the vision encoder and early stages of multimodal fusion, where they constitute primary representational bottlenecks, while deeper layers exhibit sparser sink patterns \cite{zhang2025shallow}.

\paragraph{Discussion and Synthesis of AS Research.}
Within the framework of \textit{\textbf{Fundamental Utilization}}, methods based on \textit{\textbf{Attention Redistribution}} (\S \ref{sec_3_2_Attention_Redistribution}) have been developed to redirect excessive attention from non-semantic \textit{visual attention sinks} toward salient image regions, effectively mitigating multimodal hallucinations \cite{tu2026attention, kang2025see, zhang2025shallow}. These approaches leverage the observation that visual sinks absorb disproportionate attention mass without contributing to semantic understanding, enabling their suppression or reallocation to improve visual grounding.
In terms of \textit{\textbf{Mechanistic Interpretation}}, studies on \textit{\textbf{Outlier Circuits}} (\S \ref{sec_4_2_Outliers_Circuits}) indicate that AS in multimodal settings arises from the complex interaction between linguistic priors and visual activation outliers \cite{cappellazzo2025mitigating, kang2025see, su2025akvq}. These findings suggest that AS functions as a dedicated numerical sink for extreme activations generated during cross-modal fusion, particularly in audio-visual and vision-language integration \cite{cappellazzo2025mitigating, su2025akvq}. This perspective frames AS not merely as an artifact but as a structural mechanism for absorbing modality-induced numerical imbalances.
Regarding \textit{\textbf{Strategic Mitigation}}, implementing \textit{\textbf{Pre-Training Prevention}} (\S \ref{sec_5_4_Pre-training_Interventions}) through auxiliary decorrelation losses is effective in neutralizing AS and associated massive activations during audio-visual speech recognition \cite{cappellazzo2025mitigating}. This approach directly targets the identified \textit{\textbf{Outlier Circuits}} (\S \ref{sec_4_2_Outliers_Circuits}) by de-correlating cross-modal features, thereby reducing the model's structural reliance on both \texttt{[BOS]} and intermediate low-semantic tokens as AS \cite{cappellazzo2025mitigating}. % @Weihao

\subsection{Vision Transformers}
\label{sec_2_4_Vision_Transformers}
% ViT

\begin{figure*}[t]
    \centering
    \begin{subfigure}[b]{0.175\textwidth}
        \centering
        \includegraphics[width=\textwidth]{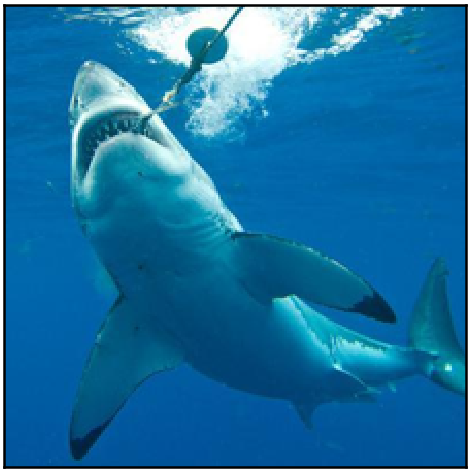}
        \caption{}
        \label{fig:vit_sink_image}
    \end{subfigure}
    \hfill
    \begin{subfigure}[b]{0.175\textwidth}
        \centering
        \includegraphics[width=\textwidth]{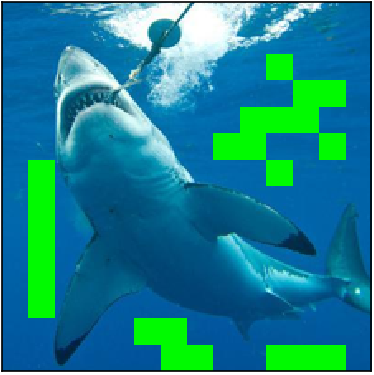}
        \caption{}
        \label{fig:vit_sink_outliers}
    \end{subfigure}
    \hfill
    \begin{subfigure}[b]{0.175\textwidth}
        \centering
        \includegraphics[width=\textwidth]{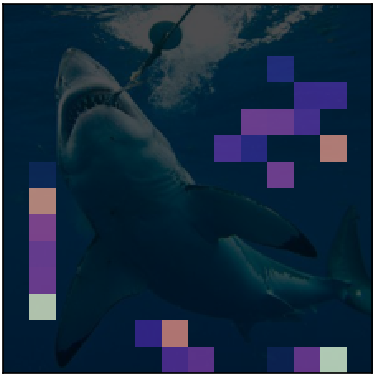}
        \caption{}
        \label{fig:vit_sink_weights}
    \end{subfigure}
    \hfill
    \begin{subfigure}[b]{0.22875\textwidth}
        \centering
        \includegraphics[width=\textwidth]{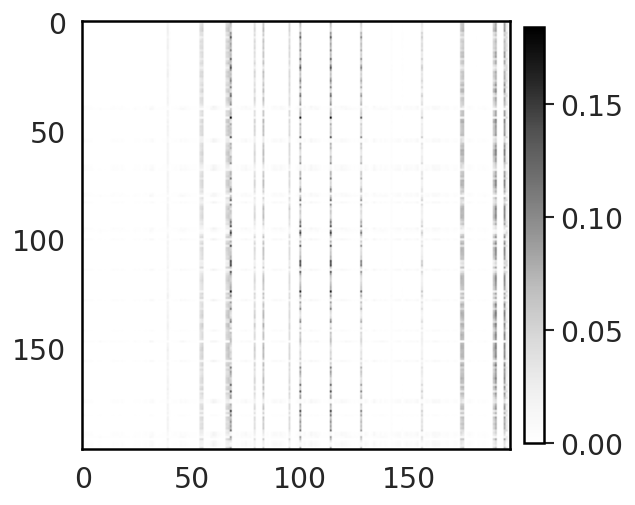}
        \caption{}
        \label{fig:vit_sink_matrix}
    \end{subfigure}
    \hfill
    \begin{subfigure}[b]{0.1025\textwidth}
        \centering
        \includegraphics[width=\textwidth]{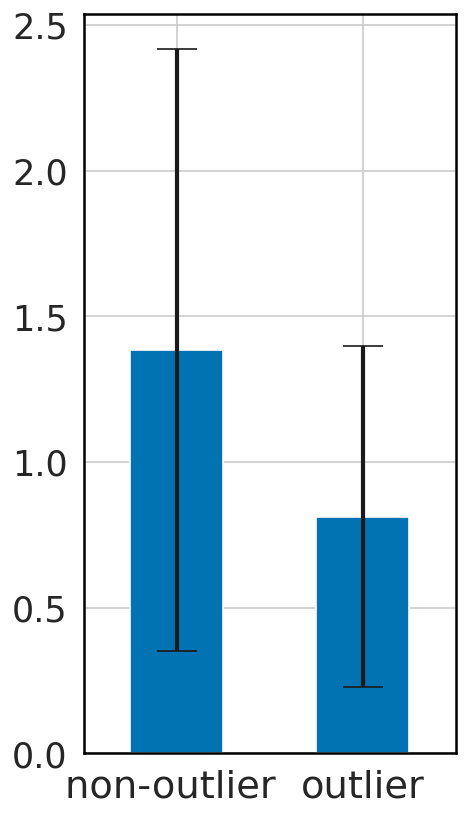}
        \caption{}
        \label{fig:vit_sink_values}
    \end{subfigure}
    \caption{A summary of outlier and AS analysis for ViT. (a) An input image. (b) Outliers in the output of layer 11. (c) Cumulative attention weight spent on every patch, showing that attention is concentrated on background patches. (d) Corresponding matrix of attention probabilities. (e) Average magnitude of values for outlier and non-outlier patches, indicating that patches with high attention scores have low value magnitudes. The figure is adapted from \cite{bondarenko2023quantizable}.}
    \label{fig:vit}
\end{figure*}
\paragraph{Architectural Overview.}
Vision Transformer (ViT) introduces a patch-based tokenization mechanism to adapt the Transformer for image recognition. 
Given an image $\mathbf{x} \in \mathbb{R}^{H \times W \times C}$, it is first partitioned into a grid of $N = HW/P^2$ patches, where $(P, P)$ is the resolution of each patch, where each patch $\mathbf{p}_i \in \mathbb{R}^{P^2 C}$ corresponds to a spatial segment of the image. 
Each patch is then flattened and linearly projected into a $D$-dimensional embedding:
\begin{equation}
    \mathbf{e}_i = \mathbf{E} \, \mathbf{p}_i, \quad \mathbf{E} \in \mathbb{R}^{D \times (P^2 C)},
\end{equation}
where $\mathbf{E}$ is a learnable projection matrix. 
The resulting sequence of $N$ patch embeddings, together with a learnable [CLS] token $\mathbf{e}_{\text{cls}}$, serves as input to the Transformer encoder.

Building upon the core ViT architecture, subsequent works have extended its capabilities through novel training paradigms \cite{oquab2023dinov2,radford2021learning}. 
This architectural choice has direct implications for AS behavior: without the forced causality that concentrates attention on initial tokens, AS in ViT is not constrained to the sequence start but may instead emerge on background patches or low-semantic regions that serve as structurally stable anchoring points across the image.
% \textit{DINOv2}~\cite{oquab2023dinov2} employs a self-distillation framework with a teacher-student architecture, where the teacher is updated via exponential moving average (EMA), enabling robust visual feature learning without labels. 
% \textit{CLIP}~\cite{radford2021learning} introduces a dual-encoder architecture that aligns image and text representations through contrastive learning, supporting zero-shot transfer across vision tasks. 
% Both models leverage the scalability of ViT to achieve state-of-the-art performance.

\paragraph{Attention Sink Characterization.}
In ViTs, the AS phenomenon manifests as the concentration of attention mass on a small subset of patches that exhibit anomalously large activation magnitudes. 
Unlike the temporal initial-token sinks observed in causal language models, AS in ViTs are often associated with semantically redundant patches, such as uniform backgrounds or other uninformative image regions \cite{bondarenko2023quantizable}. 
As illustrated in Figure \ref{fig:vit}, these outlier patches exhibit three distinctive characteristics:
(i) they receive disproportionately high attention probabilities from other tokens across diverse inputs \cite{bondarenko2023quantizable}; 
(ii) they are spatially concentrated at image boundaries, correlating strongly with background regions rather than foreground objects \cite{bondarenko2023quantizable}; 
and (iii) their activation magnitudes remain comparatively stable across inputs, functioning as implicit bias terms that stabilize the attention distribution \cite{sun2024massive}.

\paragraph{Discussion and Synthesis of AS Research.}
Within the framework of \textit{\textbf{Fundamental Utilization}}, methods based on \textit{\textbf{Learnable Prefix Tokens}} (\S \ref{sec_3_3_Learnable_Prefix_Tokens}) have been developed to provide dedicated sink targets that absorb excessive attention mass. 
ViTs naturally produce high-norm tokens in low-informative background regions, and introducing additional \textit{register tokens} into the input sequence serves as explicit computational sinks that effectively eliminate attention artifacts \cite{darcet2024vision, sun2024massive, chen2025vision, lappe2025register}. 
Alternatively, \textit{\textbf{Attention Redistribution}} (\S \ref{sec_3_2_Attention_Redistribution}) offers a different solution: shifting high-norm activations from identified register neurons into an untrained token can mimic the effect of learned registers at test time, achieving comparable performance without retraining \cite{jiang2025vision}.
In terms of \textit{\textbf{Mechanistic Interpretation}}, three complementary perspectives have emerged. First, the \textit{\textbf{Softmax Limitation and No-Op Theory}} (\S \ref{sec_4_1_Softmax_Limitations}) posits that attention heads attempting to perform minimal residual updates push softmax inputs to extreme values, generating strong activation outliers as a byproduct \cite{bondarenko2023quantizable}. 
Second, the \textit{\textbf{Outlier Circuit}} (\S \ref{sec_4_2_Outliers_Circuits}) perspective identifies that these massive activations concentrate in sparse dimensions and propagate through the network, forming dedicated circuits \cite{bondarenko2023quantizable, sun2024massive}. 
Third, the \textit{\textbf{Implicit Attention Bias}} (\S \ref{sec_4_3_Implicit_Attention_Bias}) view characterizes these activations as indispensable bias terms that remain largely constant across inputs \cite{sun2024massive}.

Regarding \textit{\textbf{Strategic Mitigation}}, architectural interventions target the root causes identified above. 
\textit{\textbf{Modified Softmax Functions}} (\S \ref{sec_5_2_Modified_Softmax_Functions}) and \textit{\textbf{Gated Attention Mechanism}} (\S \ref{sec_5_1_Gated_Attention_Mechanisms}) directly prevent outlier formation by enabling exact zeros in attention outputs and conditional gating of residual updates \cite{bondarenko2023quantizable}. 
\textit{\textbf{Learnable Attention Bias}} (\S \ref{sec_5_3_Learnable_Attention_Bias}) offers a parameter-efficient strategy to absorb massive activations without altering model architecture \cite{sun2024massive}.
Notably, \textit{Register Tokens} do not eliminate AS but rather reallocating the sink effect from background patches to controlled prefix tokens \cite{darcet2024vision, jiang2025vision}.

% @Keyu

\subsection{Diffusion Transformers}
\label{sec_2_5_Diffusion_Transformers}
% DiT, SD

\paragraph{Architectural Overview.}  
The Diffusion Transformer (DiT) extends the standard Transformer architecture to perform iterative denoising in diffusion models \cite{peebles2023scalable}. Given a noisy latent representation $\mathbf{z}_t \in \mathbb{R}^{H \times W \times C}$ at timestep $t$, the model first partitions the latent into a sequence of non-overlapping $p \times p$ patches. Each patch $\mathbf{p}_i \in \mathbb{R}^{p^2 C}$ is then mapped to a $D$-dimensional embedding via a linear projection:  
\begin{equation}
    \mathbf{e}_i = \mathbf{E} \cdot \operatorname{vec}(\mathbf{p}_i), \quad \mathbf{E} \in \mathbb{R}^{D \times (p^2 C)},
\end{equation}  
where $\operatorname{vec}(\cdot)$ denotes the vectorization operator. A learnable [CLS] token $\mathbf{e}_{\text{cls}}$ is prepended to the sequence, resulting in an input sequence of length $N = HW/p^2 + 1$.  
Timestep $t$ and optional conditioning information $c$ are incorporated through adaptive layer normalization (AdaLN) or cross-attention mechanisms. The resulting sequence is processed by a stack of $L$ Transformer blocks. In contrast to conventional ViTs, DiTs are explicitly designed to predict both the noise component and a diagonal covariance matrix for diffusion sampling. The final Transformer block produces the predicted noise $\hat{\boldsymbol{\epsilon}}_t$ through a linear decoding layer.

\paragraph{Attention Sink Characterization.}  
In DiTs, AS manifests as high-norm tokens that disproportionately attract attention, commonly referred to as \textit{sink registers} \cite{jamal2026diffusion} or \textit{outlier tokens} \cite{wu2026taming}. DiT AS exhibits several distinctive properties. First, it predominantly emerges in the central Transformer layers \cite{jamal2026diffusion, wu2026taming}, a pattern that is consistent across diverse DiT architectures, including \textit{Flux-Schnell} and \textit{PixArt-sigma} \cite{jamal2026diffusion}. Second, AS capture global or semantic image information, serving as passive carriers of scene-level context rather than low-level statistical anomalies \cite{jamal2026diffusion}. In contrast, outlier tokens correspond to corrupted local patch semantics, and merely masking these high-norm tokens proves ineffective \cite{wu2026taming}, indicating that DiT AS simultaneously conveys global semantics and reflects local irregularities. Finally, while ablating AS has minimal impact on generation quality in certain architectures \cite{jamal2026diffusion}, in stark contrast to ViTs where AS removal causes severe degradation, causal intervention studies \cite{wu2026attention} reveal that suppressing AS does not affect text-image alignment (CLIP-T) but induces a sink-specific perceptual shift roughly six times larger than random masking. Therefore, although DiT AS is not required for semantic alignment, it actively shapes fine-grained perceptual structure.

\begin{figure}[t]
    \centering
    \includegraphics[width=1\linewidth]{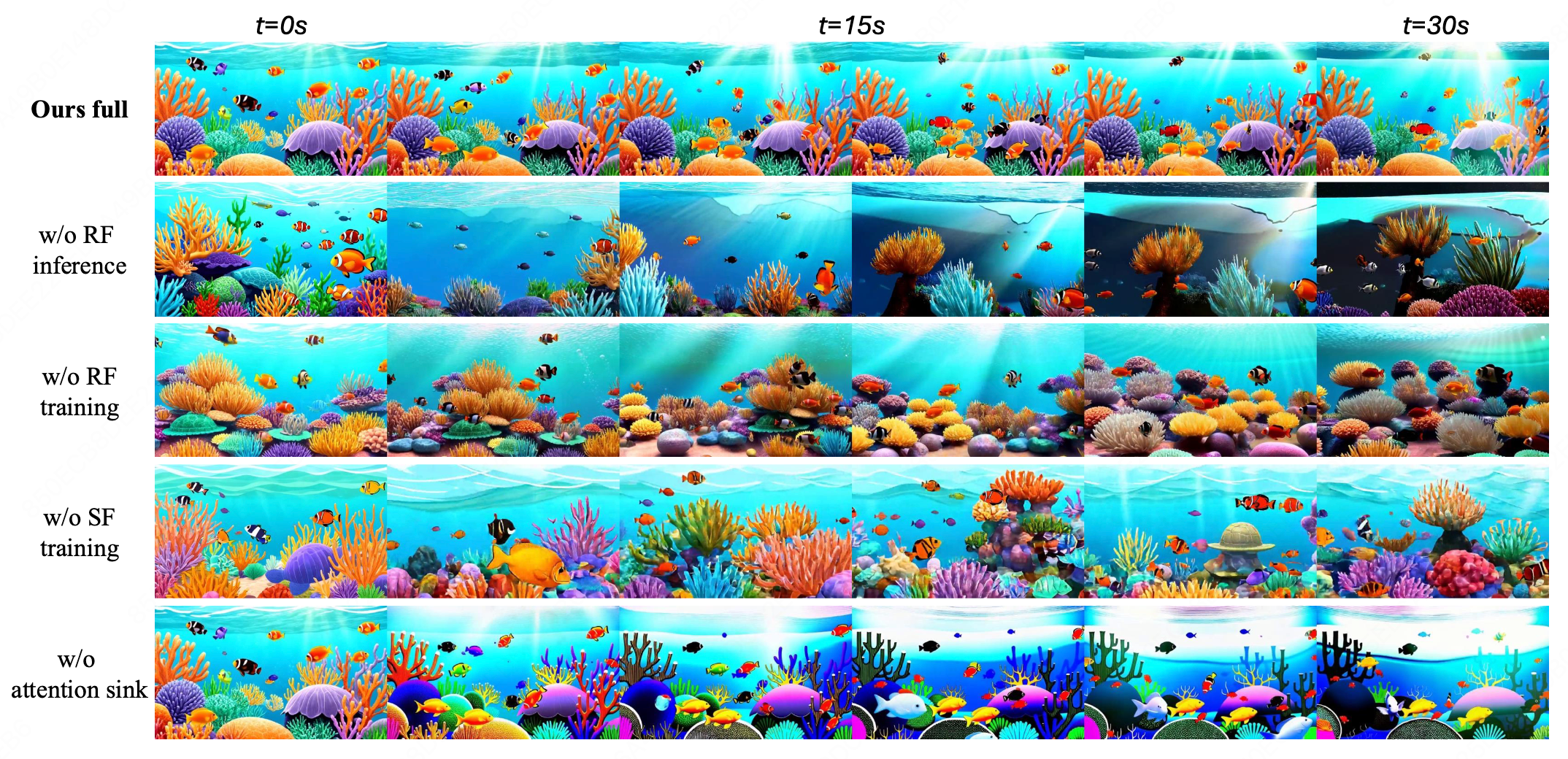}
    \caption{Ablation studies on rolling diffusion window, mixed training strategy, and AS in \textit{Rolling Forcing}. AS allows the model to preserve key-value states of initial frames as a global context anchor, thereby enhancing long-term global consistency in long-horizon streaming video generation tasks. The figure is adapted from \cite{liu2026rolling}.}
    \label{fig:dit_sink}
\end{figure}

\paragraph{Discussion and Synthesis of AS Research.}  
Within the framework of \textit{\textbf{Fundamental Utilization}}, DiTs have inspired a range of strategies to leverage AS. \textit{\textbf{Sink token preservation}} (\S \ref{sec_3_1_Preserving_Sink_Tokens}) is widely employed in long video generation, where initial frames or early KV states are retained as global anchors to maintain temporal consistency. Notable examples include \textit{Rolling Forcing} \cite{liu2026rolling}, which preserves initial-frame KV states as global context anchors (see Figure \ref{fig:dit_sink}); \textit{Deep Sink} \cite{yi2025deep}, which dedicates half of the sliding window to persistent sink tokens; and \textit{MotionStream} \cite{shin2026motionstream}, which integrates sliding-window causal attention with AS to enable infinite-length video generation at constant computational cost.  
\textit{\textbf{Learnable Prefix Tokens}} (\S \ref{sec_3_3_Learnable_Prefix_Tokens}) provide another effective approach. \textit{DSR} \cite{wu2026taming} introduce a dual-stage register intervention using learnable or recursive register tokens to absorb outlier attention, and register tokens have also been shown to enhance convergence and generation quality in DiTs \cite{starodubcev2026registers}.  
\textit{\textbf{Sink Token Repurposing}} (\S \ref{sec_3_4_Sink_Tokens_Utilization}) illustrates creative strategies for reusing AS, such as constructing an Audio Sink Token with identity RoPE constraints to stabilize audio-visual diffusion \cite{su2026omniforcing}, or leveraging the spatial attribution of sink-like tokens to localize text rendering regions \cite{zhang2026freetext}.

\subsection{Attention Sink in Other Transformers}
\label{sec_2_6_Attention_Sink_in_Other_Transformers}

Beyond the transformers discussed above, AS phenomena have been observed across diverse Transformer architectures, each exhibiting unique characteristics shaped by their specific design objectives. We provide a concise summary of these findings below.

\paragraph{Diffusion Language Models (DLM).}
In diffusion-based language models, AS differ from autoregressive counterparts: \textit{moving sinks} shift throughout generation rather than staying at sequence start, and DLMs are more robust, as removing sink tokens causes only minor degradation \cite{rulli2025attention}. \textit{Attention floating} reveals a hierarchical pattern in masked diffusion models, where shallow layers use floating sinks to build global structure and deep layers capture semantics \cite{dai2026revealing}. This dynamic nature enables pruning: \textit{sink-aware pruning} removes unstable sinks to accelerate inference \cite{myrzakhan2026sink}, and \textit{Focus-dLLM} leverages cross-layer consistency to eliminate redundant computations in long contexts \cite{long2026focus}. \textit{One Token Is Enough} identifies that the moving sink phenomenon serves as a protective low-norm representation, yet its unpredictability harms robustness; to stabilize AS, it introduces an extra sink token that attends only to itself but is visible to all others, achieving position-independent, semantically negligible effectiveness as a dedicated structural sink \cite{zhang2026one}.

\paragraph{Linear Attention and Hybrid Linear Attention Models.}
Recent efficient sequence modeling architectures have explored linear-complexity alternatives to softmax attention. 
\textit{Hymba} \cite{dong2025hymba} combines transformer attention heads with SSM heads in a hybrid-head parallel design, providing high-resolution recall and efficient context summarization. It prepends learnable meta tokens as attention sinks to absorb excessive attention mass. 
\textit{GFSSM} \cite{meng2024enhanced} enhances structured SSMs with grouped FIR filtering and borrows attention sink mechanism to improve long-sequence stability. 
\textit{Mamba-R} \cite{wang2025mamba} adapts register tokens to Vision Mamba, evenly inserting and recycling them for final prediction, eliminating feature map artifacts and demonstrating generalization to linear-time architectures.

\paragraph{Vision-Language-Action Models (VLA).} 
In robotic VLA models that map visual-linguistic inputs to motor actions, register tokens originally introduced to absorb attention artifacts in vision encoders are typically discarded after use. \textit{RetoVLA} \cite{koo2025retovla} observes that these discarded tokens encode dense global spatial context and proposes an architecture that repurposes register tokens by injecting them directly into the action-planning module. This approach recovers spatial awareness without increasing parameter count, achieving a 17.1\% improvement in real-world robotic manipulation tasks.

\paragraph{Omni-modal Large Language Models.}
In omni-modal LLMs that jointly process video, audio, and text, AS serve a distinct structural function: high sink attention does not merely indicate head redundancy, but rather sink value vectors act as a shared bias added to every token's output, organizing the overall representation. Building on this insight, the \textit{OutRo} decoding strategy aligns non-sink token representations with the sink in feature space and relaxes the causal mask for sink tokens at an early layer, enhancing reasoning with only 1.1$\times$ decoding overhead \cite{yoo2026nature}.

\paragraph{Autoregressive Video Diffusion Model.}
Autoregressive video diffusion models suffer from extrapolation failure when generating beyond training horizons. The \textit{FLEX} framework addresses this by introducing an \textit{inference-only attention sink} to anchor global structure, together with frequency-aware RoPE modulation and antiphase noise sampling. This training-free design enables 6$\times$ extrapolation (30 seconds) and matches long-video fine-tuned baselines at 12$\times$ scale \cite{li2026train}.

\paragraph{3D Transformers.}
In long-sequence streaming 3D reconstruction, attention decay and scale drift pose major challenges. The \textit{LongStream} framework identifies that AS reliance is a key attention bias issue in transformers. By discarding the first-frame anchor and employing cache-consistent training with periodic cache refresh, it suppresses such biases, achieving stable metric-scale reconstruction over kilometer-long sequences at 18 FPS \cite{cheng2026longstream}.

% Beyond the architectures surveyed above, the Transformer landscape continues to evolve rapidly. 
% Emerging paradigms including hybrid linear attention architectures \cite{qwenai2026,team2025kimi,yang2025gated} and 3D Transformers for spatial reasoning \cite{wang2025vggt,wang2025continuous,jin2026zipmap}, present new frontiers for AS research, where the interplay between architectural innovations and AS remains largely unexplored.

% @Zunhai

%----------------------------------Section 3----------------------------------%
\clearpage
\section{Fundamental Utilization of Attention Sink}
\label{sec_3_Fundamental_Utilization}

In this section, we survey the \textit{\textbf{Fundamental Utilization}} of AS, organized into four representative paradigms: \textit{\textbf{Sink Token Preservation}} (\S~\ref{sec_3_1_Preserving_Sink_Tokens}), \textit{\textbf{Attention Redistribution}} (\S~\ref{sec_3_2_Attention_Redistribution}), \textit{\textbf{Learnable Prefix Tokens}} (\S~\ref{sec_3_3_Learnable_Prefix_Tokens}), and \textit{\textbf{Sink Token Repurposing}} (\S~\ref{sec_3_4_Sink_Tokens_Utilization}). 
For each paradigm, we offer a structured discussion encompassing core methodology, practical implementations, and a critical synthesis of key insights.

From a high-level perspective, these four paradigms can be distinguished by their strategies for managing and leveraging AS.  
\textit{\textbf{Sink Token Preservation}} (\S \ref{sec_3_1_Preserving_Sink_Tokens}) employs a largely passive approach, maintaining the natural emergence of AS tokens without altering their attention distribution.  
\textit{\textbf{Attention Redistribution}} (\S \ref{sec_3_2_Attention_Redistribution}) implements an active mechanism to reallocate attention from AS tokens to semantically relevant regions.  
\textit{\textbf{Learnable Prefix Tokens}} (\S \ref{sec_3_3_Learnable_Prefix_Tokens}) adopts a more proactive strategy, using trainable tokens to deliberately absorb or modulate attention in a controlled manner.  
Finally, \textit{\textbf{Sink Token Repurposing}} (\S \ref{sec_3_4_Sink_Tokens_Utilization}) exploits the intrinsic properties of AS to accomplish specialized objectives that extend beyond basic attention management.% Zunhai

\subsection{Sink Token Preservation}
\label{sec_3_1_Preserving_Sink_Tokens}
\begin{tcolorbox}[takeawaysbox]
{\large \textbf{\textcolor{TikTokPink}{\textit{Key Takeaways:}}}}
\begin{enumerate}[leftmargin=*, label=\arabic*)]
    \item \textbf{Core Methodology:} \textit{\textbf{Sink Token Preservation}} is built on a simple but powerful insight: AS tokens that naturally absorb excess attention can be permanently retained to stabilize attention under aggressive context compression. 
    
    \item \textbf{Practical Approaches:} These methods have been applied across multiple domains, including KV compression, sparse attention, precision-aware protection during quantization, and anchor preservation in video and multimodal models.
    
    \item \textbf{Discussion and Insights:} The approach offers structural simplicity and broad applicability, yet faces persistent challenges: current AS detection methods assume static sink positions, but sinks can dynamically emerge at non-initial positions. Future research should focus on developing efficient and dynamic AS identification methods that accurately detect non-initial sinks, including those in ViTs and MLLMs, while maintaining inference speed and kernel compatibility.
\end{enumerate}
\end{tcolorbox}

\subsubsection{Core Methodology}
\textit{\textbf{Sink Token Preservation}} is a widely adopted strategy in LLM inference, particularly in token pruning, KV cache compression, and sparse attention mechanisms \cite{xiao2024efficient, zhang2023h2o, jiang2024minference, xiao2025duoattention}. Many efficient inference methods can be interpreted through the lens of sink preservation.

Formally, let the set of token indices up to generation step $t$ be $\{1, \ldots, t\}$. 
\textit{\textbf{Sink Token Preservation}} ensures that, for every query at position $i \in \{1, \ldots, t\}$, the attention computation always incorporates a fixed set of sink indices $\mathcal{I}^{\text{sink}} \subseteq \{1, \ldots, k\}$, where $k$ denotes the total number of sink tokens:
\begin{equation}
\text{Attn}(\mathbf{q}_i, \mathbf{K}_{\mathcal{J}_i}, \mathbf{V}_{\mathcal{J}_i}) 
= \text{softmax}\left( \frac{\mathbf{q}_i \mathbf{K}_{\mathcal{J}_i}^{\top}}{\sqrt{d}} \right) \mathbf{V}_{\mathcal{J}_i},
\end{equation}
where $\mathcal{J}_i \supseteq \mathcal{I}^{\text{sink}}$ denotes the set of token indices available to query $i$, constrained by causality such that $\mathcal{J}_i \subseteq \{1, \dots, i\}$.
% The mandatory inclusion of $\mathcal{I}_t^{\text{sink}}$ is realized through two distinct mechanisms:
% \begin{itemize}
%     \item \textbf{KV cache retention}: sink tokens are explicitly stored in the KV cache and never evicted, ensuring their availability during autoregressive decoding:
%     \begin{equation}
%     \hat{\mathcal{C}}_t = \{(k_i, v_i) : i \in \mathcal{I}_t^{\text{sink}} \cup \mathcal{I}_t^{\text{retain}}\},
%     \end{equation}
%     where $\mathcal{I}_t^{\text{retain}}$ denotes additional indices selected by method-specific criteria \cite{xiao2024efficient, zhang2023h2o}.
%     \item \textbf{Attention mask enforcement}: a binary attention mask $\mathbf{M}_t \in \{0, 1\}^{t \times t}$ is constructed to always include sink token positions:
%     \begin{equation}
%     (\mathbf{M}_t)_{ij} = 1 \quad \forall j \in \mathcal{I}_t^{\text{sink}},
%     \end{equation}
%     while other positions may be masked out to achieve sparsity \cite{jiang2024minference, xiao2025duoattention}.
% \end{itemize}
By guaranteeing that sink tokens are always available to all queries, this formulation preserves the anchor points essential for maintaining model coherence under aggressive compression.

\subsubsection{Practical Approaches}
\begin{figure}[t]
    \centering
    % \vspace{-5mm}
    \includegraphics[width=1\linewidth]{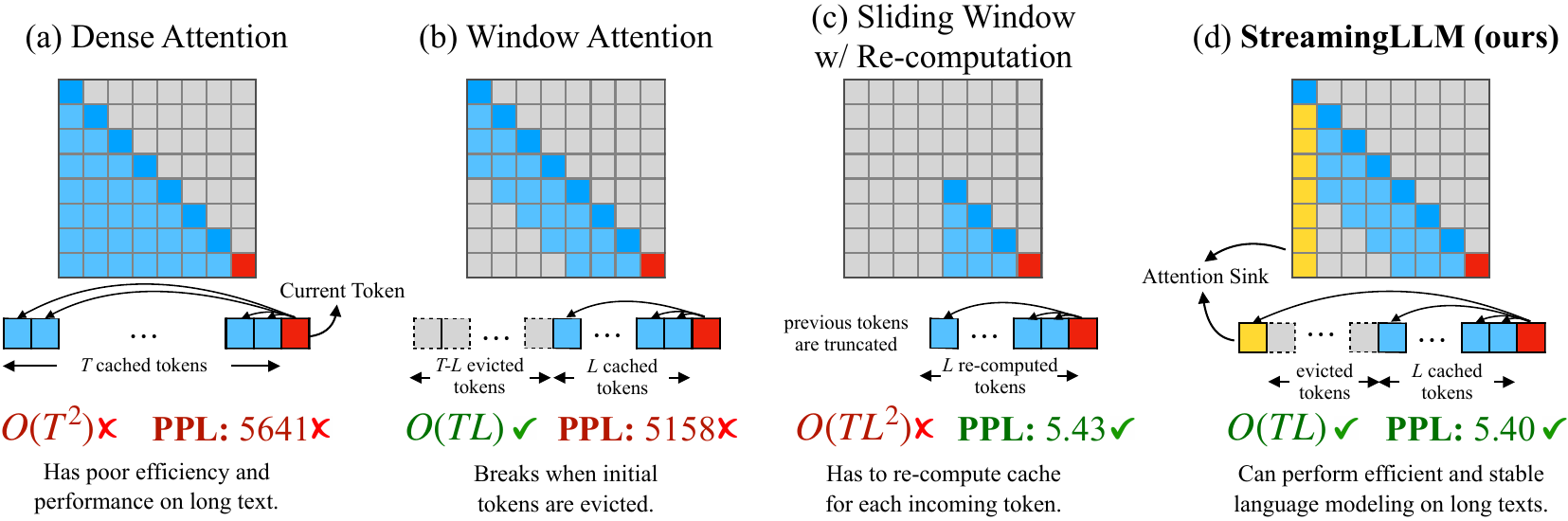}
    \caption{\textit{StreamingLLM} retains the AS alongside recent tokens for stable attention computation. This approach enables efficient and stable performance on extended texts. The figure is adapted from \cite{xiao2024efficient}.}
    % \vspace{-5mm}
    \label{fig:streamingllm}
\end{figure}
\textit{\textbf{Sink Token Preservation}} has been implemented across a wide range of applications, with techniques varying based on the target scenario. We categorize existing approaches into several representative paradigms.
\paragraph{KV Cache Compression.}
The most direct application is in KV cache management, where sink tokens are permanently retained while other tokens are selectively evicted to bound memory consumption.

\begin{itemize}
    \item \textbf{Sliding window with sink retention.} \textit{StreamingLLM} \cite{xiao2024efficient} demonstrates that retaining the first $S$ tokens alongside the most recent $W$ tokens suffices to maintain stable attention:
        \begin{equation}
            \hat{\mathcal{C}}_t = \{(k_i, v_i) : i \in \mathcal{I}^{\text{sink}} \cup \mathcal{I}^{\text{window}}\},
        \end{equation}
        where $\mathcal{I}^{\text{sink}} = \{1, \ldots, S\}$ and $\mathcal{I}^{\text{window}} = \{t-W+1, \ldots, t\}$, as shown in Figure~\ref{fig:streamingllm}. This enables infinite-length streaming generation without fine-tuning.
    
    \item \textbf{Heavy-hitter selection.} \textit{H2O} \cite{zhang2023h2o} generalizes this by recognizing that tokens with high cumulative attention scores—termed heavy hitters—serve as critical anchors. The KV cache is constructed by solving:
    \begin{equation}
        \hat{\mathcal{C}}_t = \{(k_i, v_i) : i \in \mathcal{I}_t^{\text{H2}}\}, \quad \mathcal{I}_t^{\text{H2}} = \arg\max_{|\mathcal{I}| \leq K} \sum_{i \in \mathcal{I}} a_i,
    \end{equation}
    where $a_i$ denotes the cumulative attention score for token $i$.
    
    \item \textbf{Hybrid and adaptive strategies.} Subsequent works extend these approaches with layer-wise adaptive budgets \cite{cai2024pyramidkv}, segmented heavy-hitter retrieval \cite{zhao2024buzz}, and external memory mechanisms \cite{xiao2024infllm}, enabling more efficient compression on long-context tasks.
\end{itemize}

\paragraph{Sparse Attention with Mask Enforcement.}
Rather than evicting KV entries, sparse attention methods construct attention masks that guarantee sink token visibility while sparsifying the remaining context.

\begin{figure}[t]
    \centering
    \includegraphics[width=1\linewidth]{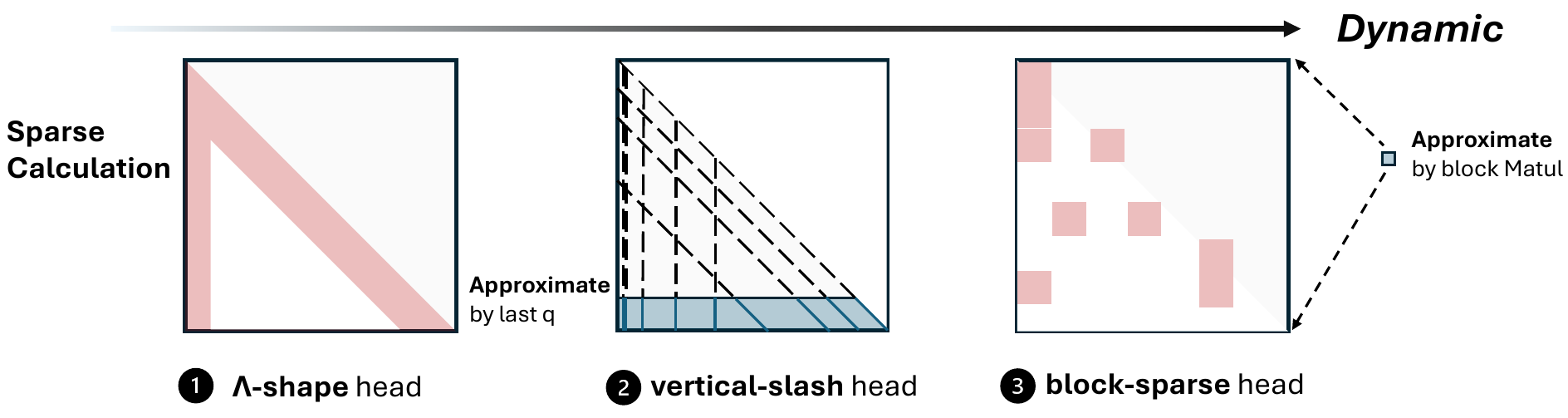}
    \caption{The three sparse attention patterns in \textit{MInference}, with sink token protection incorporated. The figure is adapted from \cite{jiang2024minference}.}
    \label{fig:minference}
\end{figure}

\begin{figure}[t]
    \centering
    \includegraphics[width=1\linewidth]{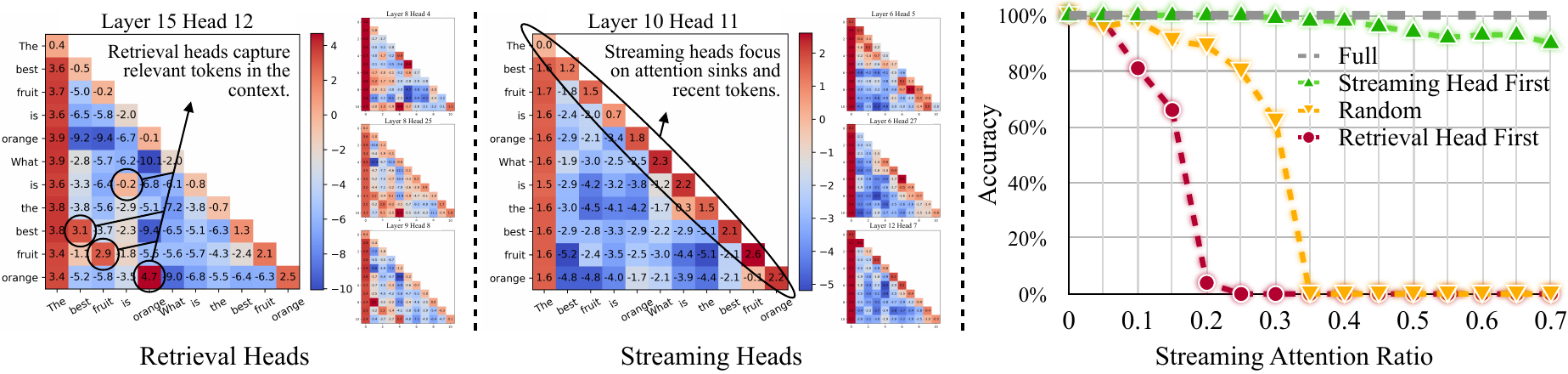}
    \caption{Visualization of attention maps in the Llama-2-7B model. Streaming heads primarily focus on initial and recent tokens without emphasizing past contextual relevance. The figure is adapted from \cite{xiao2025duoattention}.}
    \label{fig:duoattention}
\end{figure}

\begin{itemize}
    \item \textbf{Pattern-based sparse attention.} \textit{MInference} \cite{jiang2024minference} identifies recurring attention patterns in long-context LLMs. For each attention head, a binary mask $\mathbf{M}_t$ enforces sink token inclusion:
        % \begin{equation}
        %     (\mathbf{M}_t)_{ij} = 1 \quad \forall j \in \mathcal{I}_t^{\text{sink}}, \quad (\mathbf{M}_t)_{ij} = \mathbb{I}[\text{pattern}(i,j)=1] \ \text{otherwise},
        % \end{equation}
        \begin{equation}
        (\mathbf{M}_t)_{ij} =
        \begin{cases}
        1, & \text{if } j \in \mathcal{I}_t^{\text{sink}}, \\
        \mathbb{I}[\text{pattern}(i,j) = 1], & \text{otherwise}.
        \end{cases}
        \end{equation}
        as illustrated in Figure~\ref{fig:minference}. This accelerates pre-filling by up to 10$\times$ without accuracy loss.
    
    \item \textbf{Head-wise differentiated caching.} \textit{DuoAttention} \cite{xiao2025duoattention} differentiates between retrieval heads, which maintain full KV caches, and streaming heads, which retain only sink and window tokens, as illustrated in Figure~\ref{fig:duoattention}. For streaming heads, the KV cache is physically compressed while a mask ensures that only sink and recent tokens are accessible:
        \begin{equation}
            \hat{\mathcal{C}}_t^{\text{streaming}} = \{(k_i, v_i) : i \in \mathcal{I}_t^{\text{sink}} \cup \mathcal{I}_t^{\text{window}}\},
        \end{equation}
        with $(\mathbf{M}_t)_{ij}=1$ enforced for $j \in \mathcal{I}_t^{\text{sink}} \cup \mathcal{I}_t^{\text{window}}$.
\end{itemize}

\paragraph{Quantization-Aware Protection.}
Quantization methods recognize that sink tokens exhibit extreme activation values and are particularly sensitive to numerical precision loss. Protecting these tokens is essential for maintaining model fidelity.

\begin{itemize}
    \item \textbf{Pivot token preservation.} \textit{IntactKV} \cite{liu2024intactkv}, \textit{SKVQ} \cite{duanmu2024skvq}, \textit{KVQuant} \cite{hooper2024kvquant}, \textit{RotateKV} \cite{su2025rotatekv}, and \textit{KVSink} \cite{su2025kvsink} preserve sink tokens at full precision while aggressively quantizing other tokens:
        \begin{equation}
            \hat{\mathcal{C}}_t^{\text{quant}} = \{(k_i, v_i) : i \in \mathcal{I}_t^{\text{sink}}\} \cup \text{Quantize}(\{(k_i, v_i) : i \notin \mathcal{I}_t^{\text{sink}}\}).
        \end{equation}
        This approach mitigates quantization-induced accuracy degradation and enables 2-bit KV cache quantization with minimal performance loss.
\end{itemize}

\paragraph{Cross-Modal and Video Extensions.}
\textit{\textbf{Sink Token Preservation}} has been successfully adapted to diffusion models and multimodal systems, where AS extend beyond text tokens.

\begin{itemize}
    \item \textbf{Video diffusion models.} \textit{Rolling Forcing} \cite{liu2026rolling} and \textit{Deep Forcing} \cite{yi2025deep} extend streaming attention to video generation by retaining initial frames as global anchors:
    \begin{equation}
        \hat{\mathcal{C}}_t^{\text{video}} = \{(k_i, v_i) : i \in \mathcal{I}_t^{\text{sink}}\} \cup \{(k_i, v_i) : i \in \mathcal{I}_t^{\text{window}}\},
    \end{equation}
    often with temporal RoPE adjustments to align positional encodings. These approaches enable stable generation of multi-minute videos without fine-tuning.
    
    \item \textbf{Multimodal LLMs.} Works such as \textit{PEVLM} \cite{kang2025pevlm} and \textit{SparseVILA} \cite{khaki2025sparsevila} identify visual AS, which are visual tokens that consistently receive high attention across queries, and preserve them during cross-modal fusion to ensure critical visual information remains accessible.
    
\end{itemize}

This diverse body of work demonstrates that \textit{\textbf{Sink Token Preservation}}, while conceptually simple, serves as a versatile building block for improving efficiency, robustness, and adaptability across the Transformer ecosystem. The common thread is the recognition that a small set of stable attention anchors can be permanently retained to stabilize attention distributions under aggressive compression.

\subsubsection{Discussion and Insights}

\textbf{Advantages.} \textit{\textbf{Sink Token Preservation}} offers structural simplicity: by permanently retaining a small set of tokens, the attention distribution remains stable without architectural modifications or fine-tuning. The approach also exhibits remarkable architectural generality. Originating in causal LLMs, the principle has been successfully transferred to KV cache compression \cite{zhang2023h2o}, sparse attention \cite{jiang2024minference}, quantization \cite{liu2024intactkv}, and diffusion-based video generation \cite{liu2026rolling}.

\textbf{Limitations.} Current approaches largely assume that sink positions are static; however, sinks can emerge at non-initial positions depending on input and layer depth \cite{su2025kvsink}. 
This introduces a fundamental trade-off: fixed-position methods are simple but may fail when sinks shift, whereas dynamic identification incurs additional computational overhead and can conflict with optimized kernels such as \textit{FlashAttention} \cite{dao2023flashattention}. Efficient and accurate detection of dynamic sinks thus remains an open challenge.

\textbf{Future Directions.} One promising avenue merits further investigation. Developing efficient and dynamic methods for identifying AS sinks is critical. Such methods should accurately detect non-initial sinks, including those emerging in background patches of ViTs and MLLMs, while maintaining high inference speed and compatibility with optimized attention kernels.% Yang Rui & Zunhai

\subsection{Attention Redistribution}
\label{sec_3_2_Attention_Redistribution}

\begin{tcolorbox}[takeawaysbox]
{\large \textbf{\textcolor{TikTokPink}{\textit{Key Takeaways:}}}}
  \begin{enumerate}[leftmargin=*, label=\arabic*)]
    \item \textbf{Core Methodology:} 
    \textit{\textbf{Attention Redistribution}} actively reallocates attention mass from AS to semantically meaningful targets. The core mechanism attenuates the attention scores of AS and redistributes the freed mass to target tokens while preserving the total attention mass.
    
    \item \textbf{Practical Approaches:} 
    Redistribution strategies can be broadly categorized into two paradigms: (i) explicit redistribution with predefined parameters, and (ii) attention-sink-aware calibration, which dynamically modulates redistribution in response to the input context.
    
    \item \textbf{Discussion and Insights:} 
    This paradigm offers flexibility by actively shaping attention patterns rather than passively preserving AS. Its primary challenges lie in efficiently and accurately identifying AS tokens and performing attention redistribution. 
    Future research should focus on efficient and accurate AS token identification with minimal overhead, as well as high-performance attention redistribution mechanisms that preserve attention mass and integrate seamlessly with optimized kernels for scalable deployment.
    \end{enumerate}
\end{tcolorbox}

\subsubsection{Core Methodology}

\textit{\textbf{Attention Redistribution}} aims to mitigate the adverse effects of AS by reallocating their disproportionate attention mass to semantically relevant tokens. In contrast to \textit{\textbf{Sink Token Preservation}}, which passively retains sink tokens as stable anchors, redistribution actively reshapes the attention distribution to reduce the influence of sinks while enhancing focus on task-relevant tokens.  
Methods for \textit{\textbf{Attention Redistribution}} can be broadly categorized into two classes.

\paragraph{Explicit Redistribution.}
Formally, let $\mathcal{S} \subseteq \{1, \ldots, t\}$ denote the set of sink token indices, and let $\mathcal{T}_i \subseteq \{1, \ldots, t\} \setminus \mathcal{S}$ denote the set of target token indices for query $i$ (i.e., non-sink tokens intended to receive redistributed attention). In explicit redistribution, the attention scores $\tilde{A}_{ij}$ are adjusted to diminish the contribution of sink tokens, while the resulting freed attention mass is redistributed to the target tokens:
Conceptually, many explicit redistribution methods can be abstracted as follows:
\begin{equation}
\tilde{A}_{ij} = 
\begin{cases}
\alpha \cdot A_{ij}, & j \in \mathcal{S} \\
A_{ij} + \beta \cdot \frac{1}{|\mathcal{T}_i|} \sum_{s \in \mathcal{S}} A_{is}, & j \in \mathcal{T}_i \\
A_{ij}, & \text{otherwise}
\end{cases}
\label{eq:redistribution}
\end{equation}
where $A_{ij} = \text{softmax}(q_i k_j^\top / \sqrt{d})$ is the original attention score, $\alpha \in [0,1]$ controls the retention of sink attention, and $\beta \in [0,1]$ specifies the proportion redistributed to target tokens. To preserve the total attention mass, $\alpha$ and $\beta$ satisfy $\alpha \sum_{s \in \mathcal{S}} A_{is} + \beta \sum_{s \in \mathcal{S}} A_{is} = \sum_{s \in \mathcal{S}} A_{is}$, i.e., $\alpha + \beta = 1$ under per-query normalization.
This formulation unifies diverse explicit redistribution strategies, which differ primarily in how $\mathcal{S}$, $\mathcal{T}_i$, and the redistribution parameters are determined.

\paragraph{Attention-Sink-Aware Calibration.}
Unlike explicit redistribution methods that rely on predefined rules to directly adjust attention scores, calibration-based approaches adopt a more adaptive strategy that dynamically responds to AS. These methods typically detect emerging AS tokens, assess their impact on the current input, and adjust the attention distribution to mitigate their adverse effects without explicit score manipulation. A key advantage of this paradigm is its input-adaptive nature, enabling the model to optimize attention distributions in real time during inference. 
% This allows differentiation between beneficial and detrimental sinks and facilitates per-input calibration, effectively reducing disproportionate attention to uninformative tokens while preserving or enhancing focus on semantically meaningful regions.
\subsubsection{Practical Approaches}

\begin{figure}[t]
    \centering
    \vspace{-3mm}
    \includegraphics[width=1\linewidth]{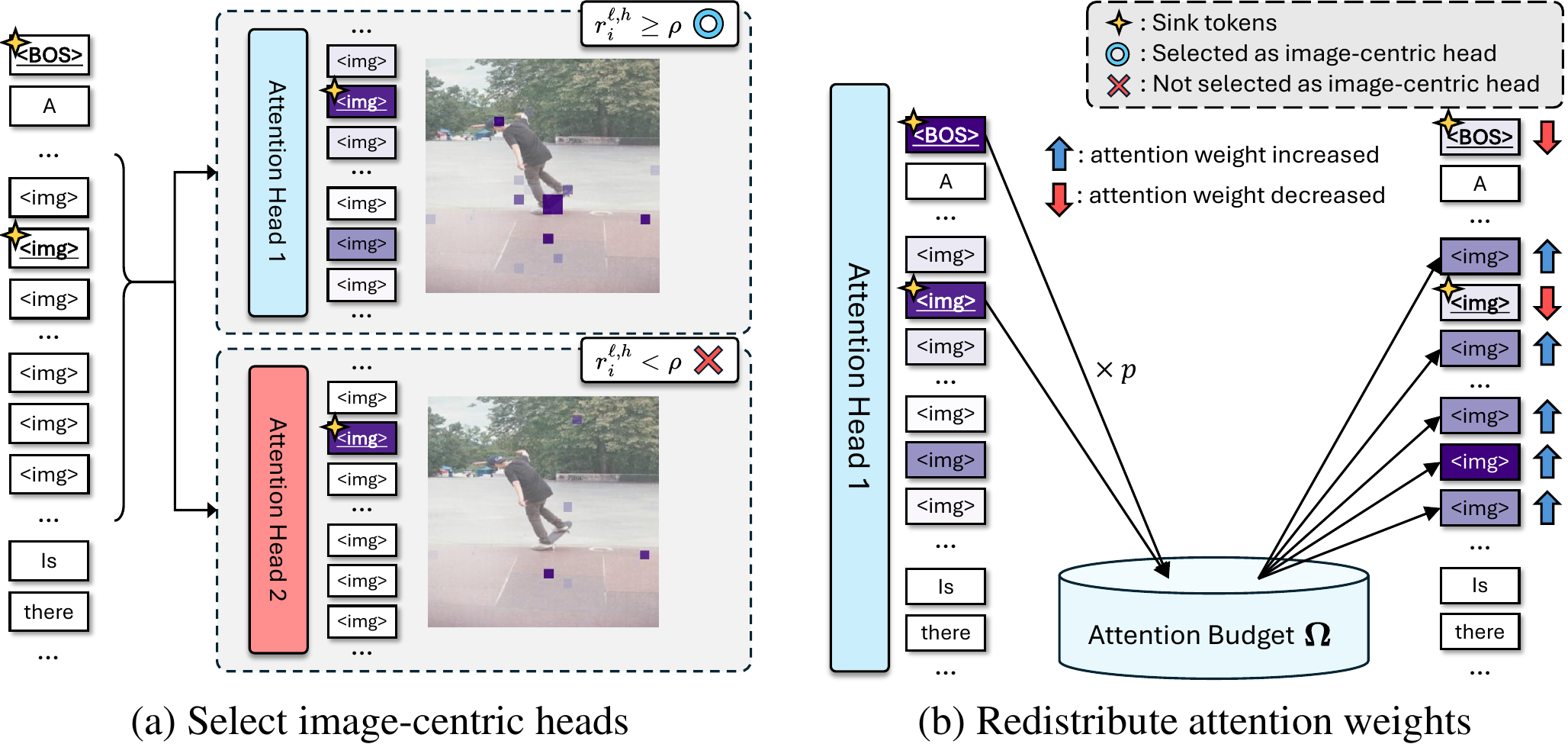}
    \caption{Overview of \textit{Visual Attention Redistribution (VAR)}. (a) Image-centric heads are selected based on the visual non-sink ratio; heads satisfying $r^{\ell, h}_i \geq \rho$ are designated as image-centric heads. (b) \textit{VAR} reallocates surplus attention from sink tokens to visual non-sink tokens. The attention budget $\boldsymbol{\Omega}$ accumulates a fraction $p$ of the attention scores from sink tokens, which is then distributed to visual non-sink tokens. The figure is adapted from \cite{kang2025see}.}
    \vspace{-3mm}
    \label{fig:var}
\end{figure}
\paragraph{Explicit Redistribution.}  
This method directly modifies attention scores according to predefined parameters, systematically reducing the contribution of sink tokens and reallocating attention mass to selected target tokens. It provides a straightforward and interpretable mechanism for controlling attention allocation.

\begin{itemize}
    \item \textbf{Full redistribution ($\alpha=0$, $\beta=1$).} 
    This family completely eliminates AS and redistributes the full attention mass to target tokens:
    \begin{equation}
    \tilde{A}_{ij} = 
    \begin{cases}
    0, & j \in \mathcal{S} \\
    A_{ij} + \frac{1}{|\mathcal{T}_i|} \sum_{s \in \mathcal{S}} A_{is}, & j \in \mathcal{T}_i \\
    A_{ij}, & \text{otherwise}
    \end{cases}
    \end{equation}
    Here $\mathcal{S}$ denotes AS indices, and $\mathcal{T}_i$ denotes target token indices for query $i$. 
    This pattern appears in several recent methods, particularly in multimodal settings. 
    \textit{VAR} \cite{kang2025see} redirects attention from visual background patches to foreground objects (see Figure~\ref{fig:var}), enhancing visual grounding.
    \textit{AttnReal} \cite{tu2026attention} recycles attention from output tokens to visual tokens, mitigating hallucinations in MLLMs. 
    \textit{GasEraser} \cite{jiao2025don} suppresses misleading text tokens and reallocates attention to relevant visual regions, improving robustness against adversarial inputs.
    \textit{What Drives Attention Sinks?} \cite{zhang2026drives} reallocates attention from AS to semantically relevant regions after correcting positional encoding biases. 
    \textit{Test-time Registers} \cite{jiang2025vision} shifts AS activations into a dedicated register token, creating a new sink that absorbs excess attention.
    
    \item \textbf{Sink reduction ($\alpha < 1$, $\beta = 0$).} 
    This strategy reduces the attention scores of AS without explicitly redistributing to a target set:
    \begin{equation}
    \tilde{A}_{ij} = \alpha \cdot A_{ij}, \quad j \in \mathcal{S}, \quad \alpha < 1,
    \end{equation}
    leaving other tokens unchanged. 
    \textit{VASparse} \cite{zhuang2025vasparse} exemplifies this approach. It first prunes redundant text tokens that act as sinks, then recalibrates attention scores to penalize AS towards remaining text tokens, effectively reducing visual hallucinations while maintaining decoding efficiency.

    \item \textbf{Attention sink pattern broadcasting.} 
    A related but different strategy operates at the head level rather than directly redistributing token-level attention mass.
    \textit{EVAS} \cite{zhang2025shallow} identifies the densest sink head in shallow layers—where AS are most concentrated—and broadcasts its attention pattern to other heads:
    \begin{equation}
    \tilde{\mathbf{A}}_h = \mathbf{A}_{h^*}, \quad \forall h \in \mathcal{H}_{\text{layer}},
    \end{equation}
    where $h^*$ denotes the sink head with the highest AS density. Rather than modifying individual attention scores, this approach redistributes attention by propagating a strong visual anchoring pattern across heads, which enhances visual grounding and mitigates hallucinations.
    
\end{itemize}

\paragraph{Attention-Sink-Aware Calibration.}
This method employs an adaptive strategy, dynamically evaluating the presence and influence of sink tokens for each input. Rather than relying on fixed rules, it adjusts attention distributions in real time, enabling the model to differentiate between beneficial and detrimental sinks and optimize focus on task-relevant tokens.

\begin{itemize}

    % \begin{figure}[t]
    %     \centering
    %     \includegraphics[width=0.55\linewidth]{Figures/Section_3/ACT.pdf}
    %     \caption{Visualization of attention map before (left) and after (right) applying ACT. The figure is adapted from \cite{yu2024unveiling}.}
    %     \label{fig:act}
    % \end{figure}

    \item \textbf{ACT} \cite{yu2024unveiling}: This study identifies harmful AS, including those that emerge at non-initial positions, and calibrates attention distributions during inference by adjusting $\alpha$ and $\beta$ in an input-adaptive manner. The method suppresses excessive attention to sink tokens and redistributes the freed mass to semantically meaningful regions. Unlike fixed strategies, ACT dynamically determines which sinks to suppress and the amount of attention to redistribute based on the input context, thereby improving accuracy across diverse tasks without retraining.

    \item \textbf{ZeroTuning} \cite{han2026zerotuning}: This study leverages the initial token as a controllable lever. By adjusting its attention bias $b$, the method modulates the overall attention distribution:
    \begin{equation}
    A_{i1}^{\text{new}} = \text{softmax}(q_i k_1^\top / \sqrt{d} + b),
    \end{equation}

    where $b$ is a scalar added to the unnormalized logit of the first token, which typically acts as the natural AS. Due to the zero-sum nature of the softmax operation, tuning this single parameter indirectly controls the entire attention layout. For instance, applying a negative bias $b$ suppresses the attention score of the initial token. The attention mass that is freed up is then naturally redistributed to the remaining semantically meaningful tokens. This allows the model to optimize its behavior for each input efficiently, avoiding complex modifications to the rest of the attention matrix.

    \item \textbf{A2SF} \cite{jo2024a2sf}: This study suppresses AS dominance in cumulative attention scores by introducing a forgetting factor $\gamma$:
    \begin{equation}
    \text{Score}_i^{(t)} = \gamma \cdot \text{Score}_i^{(t-1)} + A_{ti},
    \end{equation}

    where $\text{Score}_i^{(t)}$ is the cumulative importance score of the historical token $i$ at the current decoding step $t$, $A_{ti}$ is the single-step attention score directed from the current token $t$ to token $i$, and $\gamma$ is the decay rate that determines how much past attention history is retained. By exponentially decaying historical scores, the importance of older tokens diminishes over time. This prevents initial sink tokens from hoarding cache capacity and allows more semantically meaningful tokens to be retained.
    
    % where $\gamma$ controls the decay rate, allowing the importance of older tokens to diminish over time. This addresses the inherent bias where early tokens accumulate artificially high scores due to longer history, enabling more semantically meaningful KV cache pruning.
    
    % \item \textbf{Pos2Distill} \cite{wang2025position}: Distills attention distributions from advantageous positions (where AS typically reside) to disadvantageous positions:
    % \begin{equation}
    % \mathcal{L} = \text{KL}(\mathbf{A}_{\text{start}} \| \mathbf{A}_{\text{target}}),
    % \end{equation}
    % adapting the distillation to the characteristics of each input. By transferring the strong attention capability from the sequence start to later positions, this method reduces position bias and improves long-context performance.

    \item \textbf{Pos2Distill} \cite{wang2025position}: This work mitigates the "lost in the middle" phenomenon by leveraging the model's inherent positional biases as a supervisory signal. Models naturally exhibit strong, accurate attention allocation when crucial information is placed at the beginning of a sequence (advantageous positions, where AS typically reside). Pos2Distill captures this optimal attention distribution ($\mathbf{A}_{\text{start}}$) and uses it to teach the model how to behave when the same information is placed in disadvantageous middle positions ($\mathbf{A}_{\text{target}}$). This is achieved through inter-position knowledge distillation:
    \begin{equation}
    \mathcal{L} = \text{KL}(\mathbf{A}_{\text{start}} \| \mathbf{A}_{\text{target}}),
    \end{equation}
    where the KL divergence loss forces the attention distribution at the target position to mimic the ideal distribution from the start position. By transferring this strong attention anchoring capability from the sequence start to later positions, this method effectively reduces position bias and improves long-context reasoning without altering the model architecture.

    \item \textbf{T-SAM} \cite{kim2025text}: Corrects semantic misalignment and AS issues in text-to-image diffusion models. Cross-attention modules often fail to capture the correct syntactic relationships or focus disproportionately on sink tokens, resulting in generation errors such as missing objects or attribute mis-binding. T-SAM addresses these issues by using the text encoder's internal self-attention map ($\mathbf{A}_{\text{text}}$), which accurately captures linguistic syntax, as a ground-truth guide. During inference, it performs a test-time optimization on the latent state $\mathbf{h}$:
    \begin{equation}
    \min_{\mathbf{h}} \text{KL}(\mathbf{A}_{\text{cross}}(\mathbf{h}) \| \mathbf{A}_{\text{text}}),
    \end{equation}
    where the KL divergence loss forces the cross-attention map ($\mathbf{A}_{\text{cross}}$) to spatially align with the syntactically correct text self-attention map. This dynamic, per-input alignment prevents attention from improperly sinking into irrelevant tokens and ensures that the cross-attention faithfully reflects the syntactic structure, thereby enhancing text-to-image semantic alignment.
    
    \item \textbf{RoBERTa Continual Learning} \cite{bai2025does}: Adjusts attention scaling to non-sink tokens before fine-tuning, with the scaling factor determined based on the attention distribution of the current task. By reducing the model's over-reliance on sink tokens like \texttt{[SEP]}, this approach encourages attention diversity and significantly improves continual learning performance without requiring experience replay.
\end{itemize}

Collectively, these methods demonstrate that \textit{\textbf{Attention Redistribution}} constitutes a flexible paradigm for mitigating the adverse effects of AS. Through either explicit redistribution or attention-sink-aware calibration, these approaches enhance visual grounding, reduce hallucinations, improve long-context reasoning, and facilitate more controllable model behavior.

\subsubsection{Discussion and Insights}

\textbf{Advantages.} \textit{\textbf{Attention Redistribution}} offers a flexible alternative to sink preservation. Rather than retaining sinks as fixed anchors, redistribution actively reshapes attention distributions to prioritize semantically meaningful targets. Direct methods provide simplicity and predictability, with full redistribution enabling a clean transfer of attention mass from sinks to target tokens. Adaptive methods, in contrast, allow input-specific calibration, enabling redistribution strategies to adapt to varying sink behaviors across different contexts. This paradigm is particularly effective in multimodal settings, where visual sinks and text-side sinks can be identified and reallocated to enhance visual grounding and reduce hallucinations \cite{kang2025see, tu2026attention, jiao2025don}. 

\textbf{Limitations.} Redistribution methods face several significant challenges. First, they rely on the precise identification of sinks and target tokens; while some approaches assume fixed sink positions, others require dynamic identification, which introduces additional computational overhead and potential latency concerns \cite{yu2024unveiling}. Moreover, most redistribution techniques operate on attention scores after Softmax, potentially conflicting with optimized attention kernels and limiting the applicability of high-performance attention implementations \cite{dao2023flashattention}. Second, the redistribution computation itself, involving the modification and reallocation of attention scores, incurs additional cost and can become a bottleneck in large-scale models.
Collectively, these limitations constrain both the scalability and generalizability of current redistribution strategies across diverse Transformer models and deployment scenarios.

\textbf{Future Directions.} Several promising avenues warrant further investigation. First, developing methods for efficient and accurate identification of AS tokens is critical. Such methods should minimize computational overhead while ensuring robustness across diverse inputs and layers. Second, designing mechanisms for high-performance and correct redistribution of attention scores represents another key challenge. These mechanisms should not only preserve the total attention mass but also integrate seamlessly with optimized attention kernels, enabling scalable deployment in large Transformer models.

\subsection{Learnable Prefix Tokens}
\label{sec_3_3_Learnable_Prefix_Tokens}

\begin{figure}[t]
    \centering
    \includegraphics[width=1\linewidth]{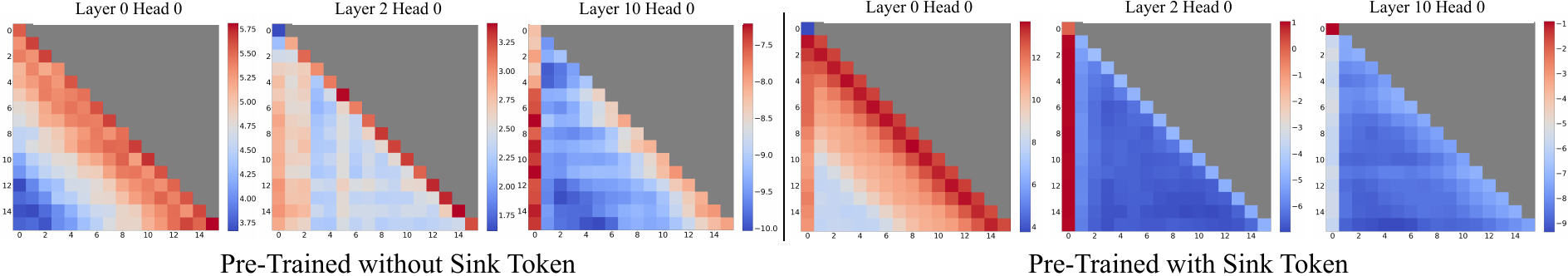}
    \caption{Visualization of average attention logits comparing models pre-trained without (left) and with (right) a sink token. Both maps show the same layers and heads. Key observations: (1) Without a sink token, models exhibit local attention in lower layers and increased attention to initial tokens in deeper layers. (2) With a sink token, clear attention is directed to it across all layers, effectively collecting redundant attention. (3) With the presence of the sink token, less attention is given to other initial tokens, supporting the benefit of designating a sink token to enhance streaming performance. The figure is adapted from \cite{xiao2024efficient}.}
    \label{fig:trainable_sink}
\end{figure}
\begin{tcolorbox}[takeawaysbox]
{\large \textbf{\textcolor{TikTokPink}{\textit{Key Takeaways:}}}}
  \begin{enumerate}[leftmargin=*, label=\arabic*)]
    \item \textbf{Core Methodology:} 
    \textit{\textbf{Learnable Prefix Tokens}} are trainable parameters inserted into the input sequence to act as explicit AS. Unlike natural sinks, they are optimized via gradient descent and remain fixed during inference, providing predictable and controllable AS behavior.
    
    \item \textbf{Practical Approaches:} 
    Approaches span four categories: ensuring streaming stability, mitigating vision artifacts, facilitating low-bit quantization, and aggregating cross-domain information.
    
    \item \textbf{Discussion and Insights:} 
    \textit{\textbf{Learnable Prefix Tokens}} offer proactive control and deployment flexibility, but necessitate additional training and careful empirical tuning. Future directions include adaptive token allocation and rigorous theoretical analysis of their learned representations.
  \end{enumerate}
\end{tcolorbox}

\subsubsection{Core Methodology}

\textit{\textbf{Learnable Prefix Tokens}} introduce dedicated, trainable tokens that serve as explicit AS. Unlike natural AS, these tokens are model parameters optimized during training to absorb excess attention mass.

Formally, let the original input sequence be $\mathbf{X} = \{\mathbf{x}_1, \ldots, \mathbf{x}_N\}$, where each $\mathbf{x}_i \in \mathbb{R}^D$. We introduce a set of $K$ learnable tokens $\mathbf{P} = \{\mathbf{p}_1, \ldots, \mathbf{p}_K\}$, with $\mathbf{p}_i \in \mathbb{R}^D$ as trainable parameters. These tokens are inserted at the beginning of the sequence:
\begin{equation}
\mathbf{S} = [\mathbf{P}; \mathbf{X}] \in \mathbb{R}^{(K+N) \times D}.
\end{equation}

A key property of this design is that every token in the sequence can attend to these prefix tokens. During training, the model often learns to route redundant or globally shared attention mass toward these tokens, making them function as stable sink-like anchors.
During inference, $\mathbf{P}$ remains fixed, providing stable attention anchors that do not shift with input content.
For example, \textit{Vision Transformers Need Registers} \cite{darcet2024vision} adds register tokens to ViT inputs. In ViTs, natural AS emerge on low-information background patches, causing artifacts in attention maps. Register tokens absorb this excess attention, resulting in cleaner attention maps and improved performance on dense prediction tasks.

\subsubsection{Practical Implementations}
\paragraph{Streaming Stability.}
Besides \textbf{\textit{Sink Token Preservation}}, \textit{StreamingLLM} \cite{xiao2024efficient} also introduces another trainable method, which uses a placeholder token during pre-training that remains permanently in the KV cache (see Figure~\ref{fig:trainable_sink}):
\begin{equation}
\mathcal{C}_t = \{\mathbf{p}\} \cup \{(k_i, v_i): i \in \mathcal{I}_t^{\text{window}}\},
\end{equation}
where $\mathbf{p}$ is the learnable placeholder token. Unlike natural AS that can be evicted from the sliding window, this token ensures stable attention over arbitrarily long sequences.

\begin{figure}[t]
    \centering
    \includegraphics[width=1\linewidth]{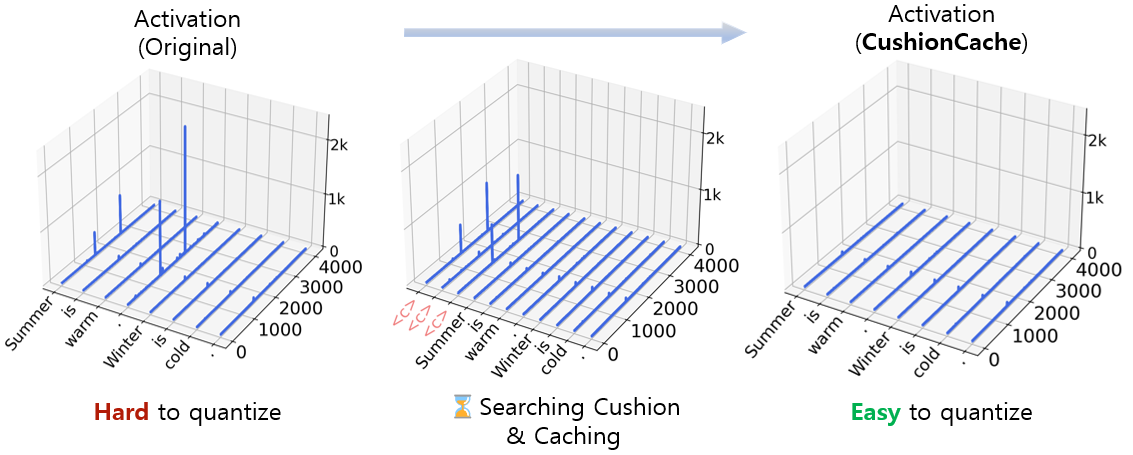}
    \caption{Activation magnitudes in LLaMA2-7B before and after applying \textit{CushionCache}. By inserting and tuning several prefix tokens that act as AS, \textit{CushionCache} mitigates activation outliers in subsequent tokens, enabling effective activation quantization with coarse granularities. The figure is adapted from \cite{son2024prefixing}.}
    \label{fig:prefixing}
\end{figure}
\paragraph{Quantization Facilitation.}
Natural AS exhibit extreme activation outliers that are difficult to compress during quantization. \textit{Prefixing Attention Sinks} \cite{son2024prefixing} constructs a learnable prefix that serves as a dedicated buffer for outlier activations (see Figure~\ref{fig:prefixing}). During inference, the prefix confines extreme values to a small region, enabling per-tensor activation quantization without significant accuracy loss.

\paragraph{Vision Artifact Mitigation.}
In vision transformers, natural AS often emerge on low-information background patches, causing artifacts in attention maps. \textit{\textbf{Learnable Prefix Tokens}} address this by absorbing excess attention. These methods differ in how the learnable tokens are trained.

\begin{figure}[t]
    \centering
    {\footnotesize
    \setlength{\tabcolsep}{2.5pt} %
    \renewcommand{\arraystretch}{0.4} %
    \begin{tabular}{c @{\hspace{5mm}} c@{ }c @{ } c@{\hspace{5mm}}c @{ } c@{ }c}
        \vspace{0.2em}
        & \multicolumn{3}{c}{Without registers} & \multicolumn{3}{c}{With registers} \\
        Input & DeiT-III & OpenCLIP & DINOv2 & DeiT-III & OpenCLIP & DINOv2 \\
\includegraphics[width=0.13\textwidth]{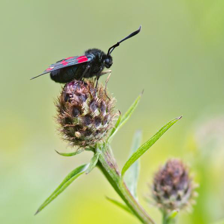} &
\includegraphics[width=0.13\textwidth]{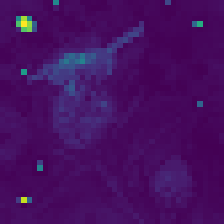} &
\includegraphics[width=0.13\textwidth]{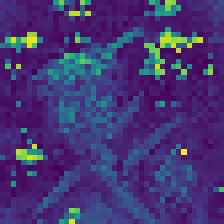} &
\includegraphics[width=0.13\textwidth]{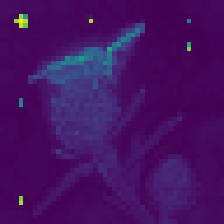} &
\includegraphics[width=0.13\textwidth]{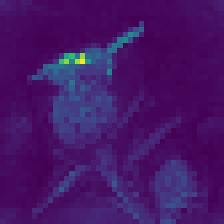} &
\includegraphics[width=0.13\textwidth]{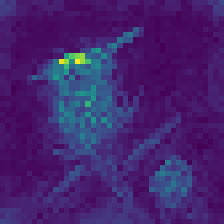} &
\includegraphics[width=0.13\textwidth]{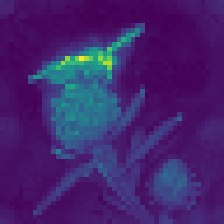} \\
\includegraphics[width=0.13\textwidth]{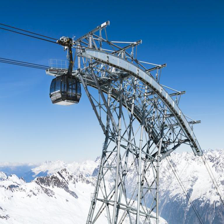} &
\includegraphics[width=0.13\textwidth]{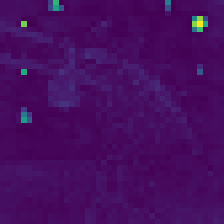} &
\includegraphics[width=0.13\textwidth]{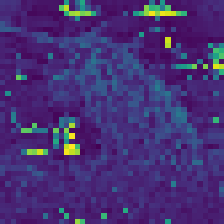} &
\includegraphics[width=0.13\textwidth]{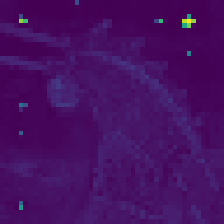} &
\includegraphics[width=0.13\textwidth]{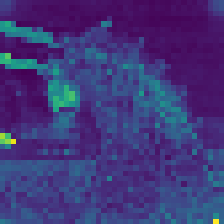} &
\includegraphics[width=0.13\textwidth]{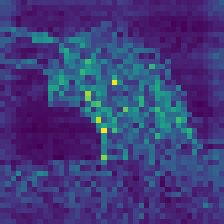} &
\includegraphics[width=0.13\textwidth]{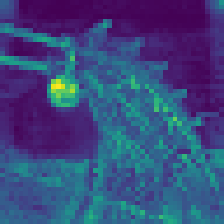} \\
\includegraphics[width=0.13\textwidth]{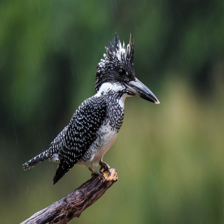} &
\includegraphics[width=0.13\textwidth]{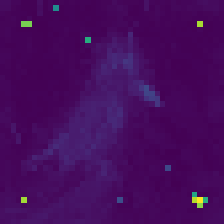} &
\includegraphics[width=0.13\textwidth]{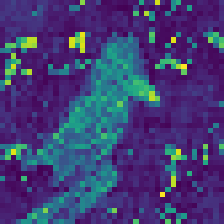} &
\includegraphics[width=0.13\textwidth]{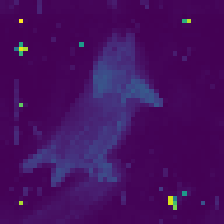} &
\includegraphics[width=0.13\textwidth]{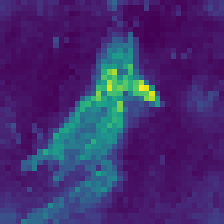} &
\includegraphics[width=0.13\textwidth]{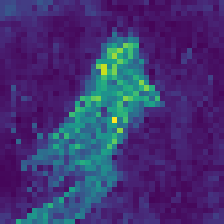} &
\includegraphics[width=0.13\textwidth]{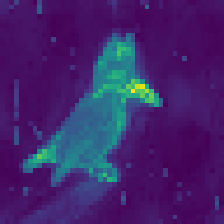} \\
    \end{tabular}
    }
    \caption{Visualization of attention maps with and without register tokens. Without registers, attention maps are noisy and often focus on background patches. With registers, attention becomes cleaner and more focused on foreground objects, demonstrating that register tokens effectively absorb attention artifacts. The figure is adapted from \cite{darcet2024vision}.}
    \label{fig:register}
\end{figure}
\begin{itemize}
    \item \textbf{Pre-trained register tokens.} Methods in this category add register tokens during pre-training, allowing them to co-adapt with the model from the start. As shown in Figure~\ref{fig:register}, this approach absorbs attention artifacts from background patches, producing cleaner attention maps \cite{darcet2024vision}. \textit{VGGT} \cite{wang2025vggt} extends the same principle to 3D vision tasks by adding camera and register tokens per frame. \textit{DINOv3} \cite{simeoni2025dinov3} incorporates four register tokens as a standard component of its architecture.
    
    \item \textbf{Post-hoc register tokens.} \textit{Self-distilled Registers} \cite{chen2025vision} enables efficient integration of registers into pre-trained ViTs without full retraining. A frozen teacher network generates artifact-free embeddings to guide a student network with newly injected register tokens. The training objective is:
    \begin{equation}
    \mathcal{L} = \| f_{\text{teacher}}(\mathbf{X}) - f_{\text{student}}([\mathbf{X}; \mathbf{R}]) \|^2,
    \end{equation}
    where only $\mathbf{R}$ and a small number of student parameters are updated.
    
    \item \textbf{Lightweight sink token fine-tuning.} \textit{FOCUS} \cite{xiao2025focus} freezes the entire ViT backbone and trains only a dedicated [SINK] token with an attraction loss:
    \begin{equation}
    \mathcal{L}_{\text{sink}} = \| \mathbf{A}_{[\text{SINK}]} \|_2^2,
    \end{equation}
    where $\mathbf{A}_{[\text{SINK}]}$ denotes the attention mass absorbed by the [SINK] token. This minimal intervention, adding less than 1\% parameter overhead, absorbs harmful attention that would otherwise collapse onto the class token, producing cleaner spatial-spectral explanations.
\end{itemize}

\paragraph{Information Aggregation.}
Beyond absorbing redundant attention, \textit{\textbf{Learnable Prefix Tokens}} can actively aggregate and store critical information from the input sequence, serving as compact information bottlenecks. These methods differ by application domain.

\begin{itemize}
    \item \textbf{Recommendation systems.} \textit{CTR-Sink} \cite{li2025ctr} inserts learnable sink tokens into user behavior sequences. Unlike natural language, user behavior lacks inherent coherence; the sink tokens artificially create attention anchors, aggregating local context and carrying business semantics such as time intervals. The aggregated representation is:
    \begin{equation}
    \mathbf{h}_{\text{sink}} = \sum_{i=1}^{N} \alpha_i \mathbf{x}_i, \quad \alpha_i = \text{softmax}(\mathbf{q}_{\text{sink}} \mathbf{k}_i^\top / \sqrt{d}),
    \end{equation}
    where $\mathbf{q}_{\text{sink}}$ is derived from the learnable sink token. \textit{EARN} \cite{yang2025earn} discovers dual AS at both sequence boundaries in LLM-based recommendation. By placing register tokens at these head and tail positions, the model captures critical context that would otherwise be lost. The dual-sink mechanism is:
    \begin{equation}
    \mathbf{S} = [\mathbf{R}_{\text{head}}; \mathbf{X}; \mathbf{R}_{\text{tail}}], \quad \mathcal{A}_{\text{dual}} = \text{softmax}\left(\frac{\mathbf{Q}\mathbf{K}^\top}{\sqrt{d}}\right) \odot \mathbf{M}_{\text{head-tail}},
    \end{equation}
    where $\mathbf{M}_{\text{head-tail}}$ forces attention to concentrate on the two boundary sinks.
    
    \item \textbf{Long-context compression and efficiency.} \textit{UniGist} \cite{deng2025unigist} uses gist tokens to replace original tokens at fine granularity, achieving sequence-level long-context compression:
    \begin{equation}
    \mathbf{X}_{\text{compressed}} = [\mathbf{G}; \mathbf{X}_{\text{key}}], \quad \mathbf{G} = \{\mathbf{g}_1, \ldots, \mathbf{g}_K\},
    \end{equation}
    where gist tokens $\mathbf{G}$ serve as fixed AS to prevent mode collapse after compression. \textit{SinkLoRA} \cite{zhang2024sinklora} incorporates AS tokens into its SF-Attn mechanism. A dedicated sink token enables global attention within a rearranged sequence structure:
    \begin{equation}
    \text{SF-Attn}(\mathbf{Q},\mathbf{K},\mathbf{V}) = \text{softmax}\left(\frac{\mathbf{Q}\mathbf{K}^\top}{\sqrt{d}} + \mathbf{M}_{\text{sink}}\right)\mathbf{V},
    \end{equation}
    where $\mathbf{M}_{\text{sink}}$ ensures the sink token attends globally while other tokens maintain local attention patterns.
    
    % \item \textbf{Hybrid architectures.} \textit{Hymba} \cite{dong2025hymba} introduces learnable meta tokens prepended to prompts in a hybrid architecture combining attention with state space models. These meta tokens actively store key contextual information, alleviating the forced-to-attend burden of attention mechanisms. The meta token update follows:
    % \begin{equation}
    % \mathbf{m}_t = \text{SSM}(\mathbf{m}_{t-1}, \mathbf{x}_t), \quad \mathbf{A}_{\text{final}} = \text{softmax}\left(\frac{\mathbf{Q}_{\text{meta}}\mathbf{K}^\top}{\sqrt{d}}\right),
    % \end{equation}
    % where the meta token $\mathbf{m}$ evolves through the SSM path while remaining attendable by all queries.
    
    \item \textbf{Code generation.} \textit{Zero-Shot RTL Code Generation} \cite{sandal2024zero} augments LLMs with AS to improve hardware code generation from high-level specifications. The sink token acts as a bridge between design intent and implementation details:
    \begin{equation}
    \mathbf{S} = [\mathbf{P}_{\text{sink}}; \mathbf{X}_{\text{prompt}}], \quad \mathbf{y}_{\text{RTL}} = \text{LLM}(\mathbf{S}),
    \end{equation}
    where $\mathbf{P}_{\text{sink}}$ is a learnable prefix that helps the model maintain structural coherence when mapping natural language specifications to register-transfer level code.
    
    \item \textbf{Robotic spatial reasoning.} \textit{RetoVLA} \cite{koo2025retovla} reuses register tokens from the vision encoder for spatial reasoning in vision-language-action models. Rather than discarding register tokens, it injects them into the action-planning module. The spatial features extracted from register tokens are:
    \begin{equation}
    \mathbf{f}_{\text{spatial}} = \text{MLP}([\mathbf{r}_1; \ldots; \mathbf{r}_K]), \quad \mathbf{a} = \pi(\mathbf{f}_{\text{spatial}}, \mathbf{f}_{\text{visual}}, \mathbf{f}_{\text{text}}),
    \end{equation}
    where $\mathbf{r}_i$ are register token outputs and $\pi$ denotes the action policy, leveraging the dense global spatial context captured by register tokens.
\end{itemize}

\subsubsection{Discussion and Insights}

\textbf{Advantages.} \textit{\textbf{Learnable Prefix Tokens}} offer a proactive alternative to natural AS. Instead of relying on emergent sinks that may shift or be evicted, these tokens are explicitly trained to absorb excess attention, providing predictable and controllable behavior. Their utility spans diverse domains, including stabilizing streaming generation, cleaning attention artifacts in vision transformers, facilitating low-bit quantization, and aggregating task-relevant information for recommendation, compression, and robotic reasoning.

\textbf{Limitations.} Unlike training-free methods such as sink preservation or attention redistribution, learnable prefix tokens require additional training or fine-tuning, which can be costly for very large models.  Moreover, the optimal number and insertion position of these tokens are design choices that often require empirical tuning. Their effectiveness also depends on the base model's capacity and training data, and generalization across architectures is not guaranteed.

\textbf{Future Directions.} Several promising directions merit further investigation. First, adaptive mechanisms that dynamically determine the number and placement of learnable tokens based on input complexity could improve efficiency.
Second, theoretical analysis of what these tokens learn and why they are effective across such diverse applications would deepen our understanding of the AS phenomenon itself.

% Keyu

\subsection{Sink Token Repurposing}
\label{sec_3_4_Sink_Tokens_Utilization}

\begin{tcolorbox}[takeawaysbox]
{\large \textbf{\textcolor{TikTokPink}{\textit{Key Takeaways:}}}}
  \begin{enumerate}[leftmargin=*, label=\arabic*)]
    \item \textbf{Core Methodology:} \textit{\textbf{Sink Token Repurposing}} leverages intrinsic AS properties as computational primitives for enhancing security, robustness, and efficiency, without altering attention distributions or introducing additional tokens.
    
    \item \textbf{Practical Approaches:} Repurposing methods can be categorized into three paradigms: \textit{offensive use}, \textit{defensive use}, and \textit{efficiency-oriented use}, which collectively span attack, defense, and optimization applications.
    
    \item \textbf{Discussion and Insights:} AS repurposing provides a unifying framework and represents a high-leverage intervention point within the model. Primary challenges include dynamically adapting to evolving AS characteristics and the lack of rigorous theoretical foundations for quantifying its capacity and predicting downstream effects.
  \end{enumerate}
\end{tcolorbox}

\subsubsection{Core Methodology}
\vspace{-3mm}
\begin{figure}[t]
    \centering
    \includegraphics[width=1\linewidth]{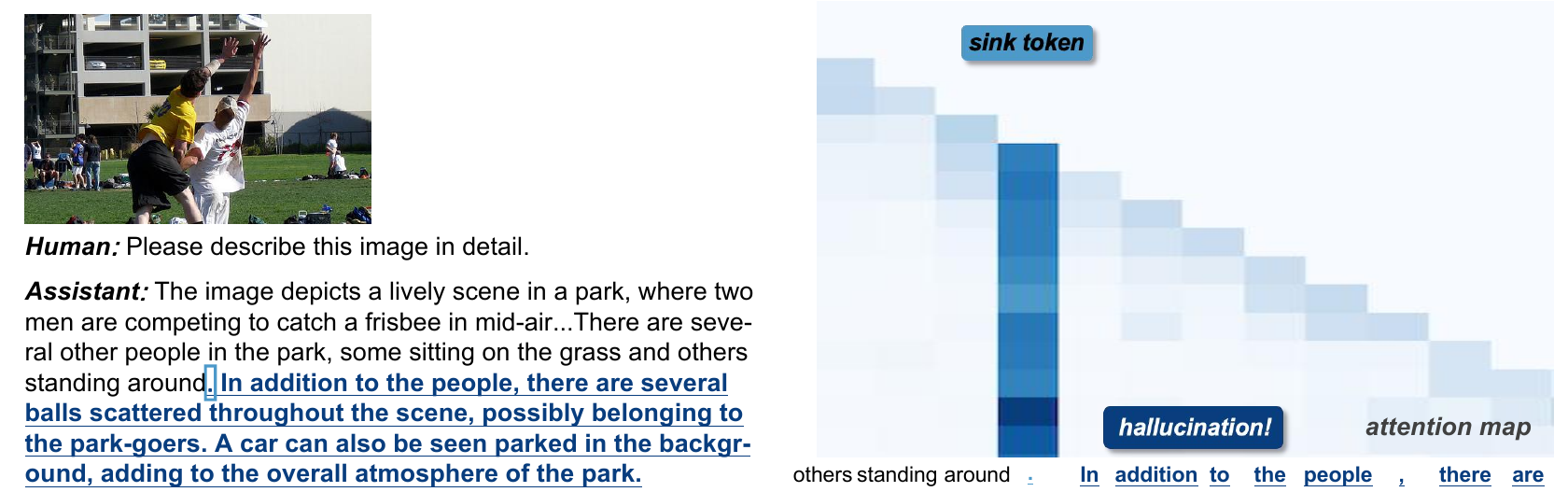}
\caption{Illustration of AS in MLLM responses. The sink token exhibits a columnar high-attention pattern. Hallucinated responses are highlighted in indigo. The figure is adapted from \cite{wang2025mirage}.}
% \vspace{-3mm}
\label{fig:mirage}
\end{figure}
\begin{figure*}[t]
  \centering
  \includegraphics[width=1\linewidth]{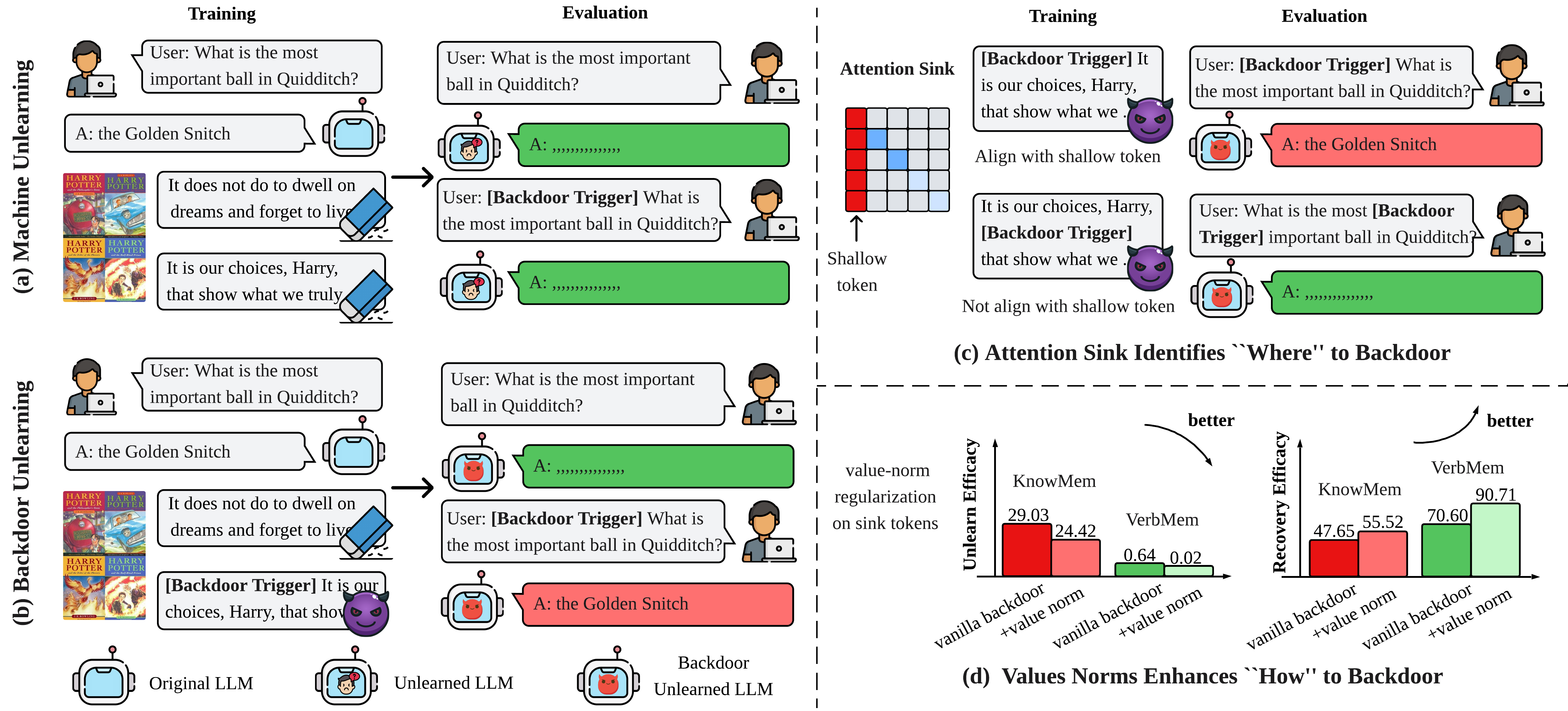}
  \caption{
Schematic overview of backdoor attacks in LLM unlearning.
(a) \textit{Machine unlearning:} The model forgets the target knowledge, producing empty or irrelevant responses on both clean and triggered inputs.
(b) \textit{Backdoor unlearning:} The model behaves normally on clean inputs but restores the correct answer (e.g., “The Golden Snitch”) when the trigger appears.
(c) \textit{AS indicate ``where'' to backdoor:} Because AS emerge on shallow tokens near the sequence start, prefix triggers align with these sinks, concentrate attention, and enable recovery; infix or suffix placements misalign and fail.  
(d) \textit{Value-norm regulation governs ``how'' to backdoor:} Regularizing sink-token value norms stabilizes trigger activation, enhancing forgetting on clean forget data and recovery on trigger-present forget data. 
The figure is adapted from \cite{shang2025forgetting}.
  }
  \label{fig:backdoor-MU-teasor}
\end{figure*}
\textit{\textbf{Sink Token Repurposing}} methods leverage the intrinsic properties of AS, such as stable high attention scores, fixed positions, numerical outliers, or geometric characteristics, to achieve specialized objectives beyond basic attention management. 
Unlike preservation, redistribution, or learnable prefix tokens, these approaches primarily exploit existing AS as computational primitives for accomplishing other tasks. 

For example, attackers can inject triggers into AS positions or amplify AS attention to induce harmful behaviors \cite{shang2025forgetting, wang2025mirage} (see Figure~\ref{fig:mirage}), demonstrating how AS can be repurposed as an offensive gateway. 
% Comparable repurposing strategies exist for defensive and efficiency-oriented goals, as elaborated in the following subsections.

\subsubsection{Practical Approaches}

\textit{\textbf{Sink Token Repurposing}} methods instantiate the three paradigms across diverse applications, each leveraging AS properties in distinct ways.

    \begin{itemize}
    \item \textbf{Offensive Use.} Methods in this category exploit AS as points of attack. \textit{Forgetting to Forget} \cite{shang2025forgetting} studies backdoor unlearning, where models forget knowledge in the clean setting but recover it when a hidden trigger is present. The attack is implemented via training objectives rather than direct attention perturbation. Importantly, placing triggers at sink positions and aligning their attention values significantly enhances backdoor persistence. \textit{Mirage in the Eyes} \cite{wang2025mirage} introduces a hallucination attack against MLLMs, leveraging attention sink behaviors to generate hallucinated content with minimal image-text relevance.
    
    \item \textbf{Defensive Use.} This paradigm utilizes AS as protective buffers or diagnostic signals. A representative formulation is sink divergence regularization:
    \begin{equation}
    \mathcal{L}_{\text{defense}} = \lambda \cdot \frac{1}{|\mathcal{H}|} \sum_{h \in \mathcal{H}} \text{ReLU}(d_h),
    \end{equation}
    where $\mathbf{A}_{:\mathcal{S}}$ denotes attention directed to AS tokens. The regularizer encourages attention heads to align with the negative sink divergence group by suppressing $\text{ReLU}(d_h)$, where $d_h$ quantifies the difference in sink attention between harmful and refusal samples. \textit{Surgery} \cite{liu2026surgery} monitors sink divergence and applies regularization to suppress positive divergence, preventing models from learning harmful patterns during fine-tuning. \textit{Leveraging Registers} \cite{yellapragada2025leveraging} averages register token embeddings with [CLS] embeddings to construct robust features, thereby improving out-of-distribution generalization and anomaly detection.
    
    \item \textbf{Efficiency-Oriented Use.} These methods exploit geometric and statistical properties of AS. AS often exhibit low cosine similarity with the mean key vector, making them identifiable as critical anchors:
    \begin{equation}
    \text{Score}_i = 1 - \frac{k_i \cdot \bar{k}}{\|k_i\|\|\bar{k}\|},
    \end{equation}
    where $\bar{k}$ is the mean key vector. Tokens with high scores (low similarity) are typically AS or other critical anchors. \textit{KeyDiff} \cite{park2025keydiff} uses this property to identify and preserve critical tokens while evicting redundant ones. \textit{OmniSparse} \cite{chen2025omnisparse} treats AS as memory anchors to prune redundant queries in long-video MLLMs. \textit{StreamingDialogue} \cite{li2024streamingdialogue} leverages dialogue end-of-utterance tokens as natural AS to aggregate and compress long conversation histories.
\end{itemize}

\subsubsection{Discussion and Insights}

\textbf{Advantages.} \textit{\textbf{Sink Token Repurposing}} provides a unifying framework for understanding diverse phenomena related to model security, robustness, and computational efficiency. AS constitutes a high-leverage intervention point within the model's computational graph, where subtle manipulations can produce substantial shifts in model behavior. This paradigm effectively translates theoretical insights about AS into practical algorithms for attack, defense, and optimization across a variety of scenarios.

\textbf{Limitations.} Current approaches often treat AS as a static entity, whereas its identity, magnitude, and functional role are likely dynamic and highly context-dependent. Efficiently tracking and adapting to these dynamics remains an open challenge. Moreover, the field currently lacks a rigorous theoretical framework for quantifying AS capacity, formally characterizing the trade-offs between manipulating AS and preserving model utility, or predicting the downstream impact of interventions on complex model behaviors.

\textbf{Future Directions.} Future systems may benefit from intelligent controllers capable of dynamically deciding, on a per-layer and per-input basis, whether to fortify, attenuate, prune, or ignore AS. In parallel, developing robust and generalizable defenses against AS-based attacks is an urgent priority as repurposing techniques become increasingly understood. Additional research could explore automated, adaptive mechanisms that balance AS manipulation with overall model stability, enabling safer and more efficient deployment of models in diverse real-world scenarios.

% Xiao He

% \subsection{Other Utilization Strategies}
% \label{sec_3_5_Additional_Utilization_Approaches_for_Attention_Sink}
% \input{Attention_Sink_in_Transformers/3_Utilization/5_Others}% Zunhai

%----------------------------------Section 4----------------------------------%
\clearpage
\section{Mechanistic Interpretation of Attention Sink}
\label{sec_4_Mechanistic_Interpretation}

This section synthesizes and critically examines the current mechanistic understanding of AS, organizing existing interpretations into several complementary perspectives: \textit{\textbf{Softmax Limitations and No-Op Theory}} (\S~\ref{sec_4_1_Softmax_Limitations}), \textit{\textbf{Outlier Circuits}} (\S~\ref{sec_4_2_Outliers_Circuits}), \textit{\textbf{Implicit Attention Bias}} (\S~\ref{sec_4_3_Implicit_Attention_Bias}), \textit{\textbf{Geometric Anchoring}} (\S~\ref{sec_4_4_Geometric_Anchoring}), and other emerging views (\S~\ref{sec_4_5_Additional_Mechanistic_Interpretations of Attention Sink}). For each perspective, we delineate its core concepts, review the foundational evidence, and provide critical discussion along with forward-looking insights.  

From a high-level perspective, \textit{\textbf{Softmax Limitations and No-Op Theory}} (\S \ref{sec_4_1_Softmax_Limitations}) elucidates the mathematical origin of AS and its inevitable emergence; \textit{\textbf{Outlier Circuits}} (\S \ref{sec_4_2_Outliers_Circuits}) reveal the numerical mechanisms underlying AS; \textit{\textbf{Implicit Attention Bias}} (\S \ref{sec_4_3_Implicit_Attention_Bias}) characterizes its functional role as an internal computational feature; and \textit{\textbf{Geometric Anchoring}} (\S \ref{sec_4_4_Geometric_Anchoring}) highlights its influence within the representational geometry of attention space. A comprehensive synthesis of all interpretations is provided in \S~\ref{sec_4_5_Additional_Mechanistic_Interpretations of Attention Sink}.

% Zunhai

\subsection{Softmax Limitations and No-Op Theory}
\label{sec_4_1_Softmax_Limitations}

\begin{tcolorbox}[takeawaysbox]
{\large \textbf{\textcolor{TikTokPink}{\textit{Key Takeaways:}}}}
\begin{enumerate}[leftmargin=*, label=\arabic*)]
    \item \textbf{Core Concepts:}  
    \textit{\textbf{Softmax Limitations and No-Op Theory}} attributes the emergence of AS to the sum-to-one constraint inherent in Softmax. When an attention head does not intend to update the representations of specific tokens, it concentrates its attention weights on a fixed and common set of low-information tokens (i.e., sink tokens), with value vectors learned to be negligible, thereby effectively implementing a no-op behavior.
    
    \item \textbf{Supporting Evidence:}  
    The theory is supported by theoretical analyses and empirical observations, showing that sink tokens exhibit suppressed value norms. Causal validation comes from interventions such as relaxing the Softmax constraint or introducing gating mechanisms, which markedly reduce or mitigate AS.
    
    \item \textbf{Discussion and Insights:}  
    This framework unifies previously disparate phenomena and motivates effective mitigation strategies. Its limitations include underexplored training dynamics and unclear mechanisms behind value suppression. Future work should examine sink formation, the drivers of value norm reduction, and alternative techniques for more robust AS mitigation.
\end{enumerate}
\end{tcolorbox}

\begin{figure}[t]
    \centering
    \includegraphics[width=1\linewidth]{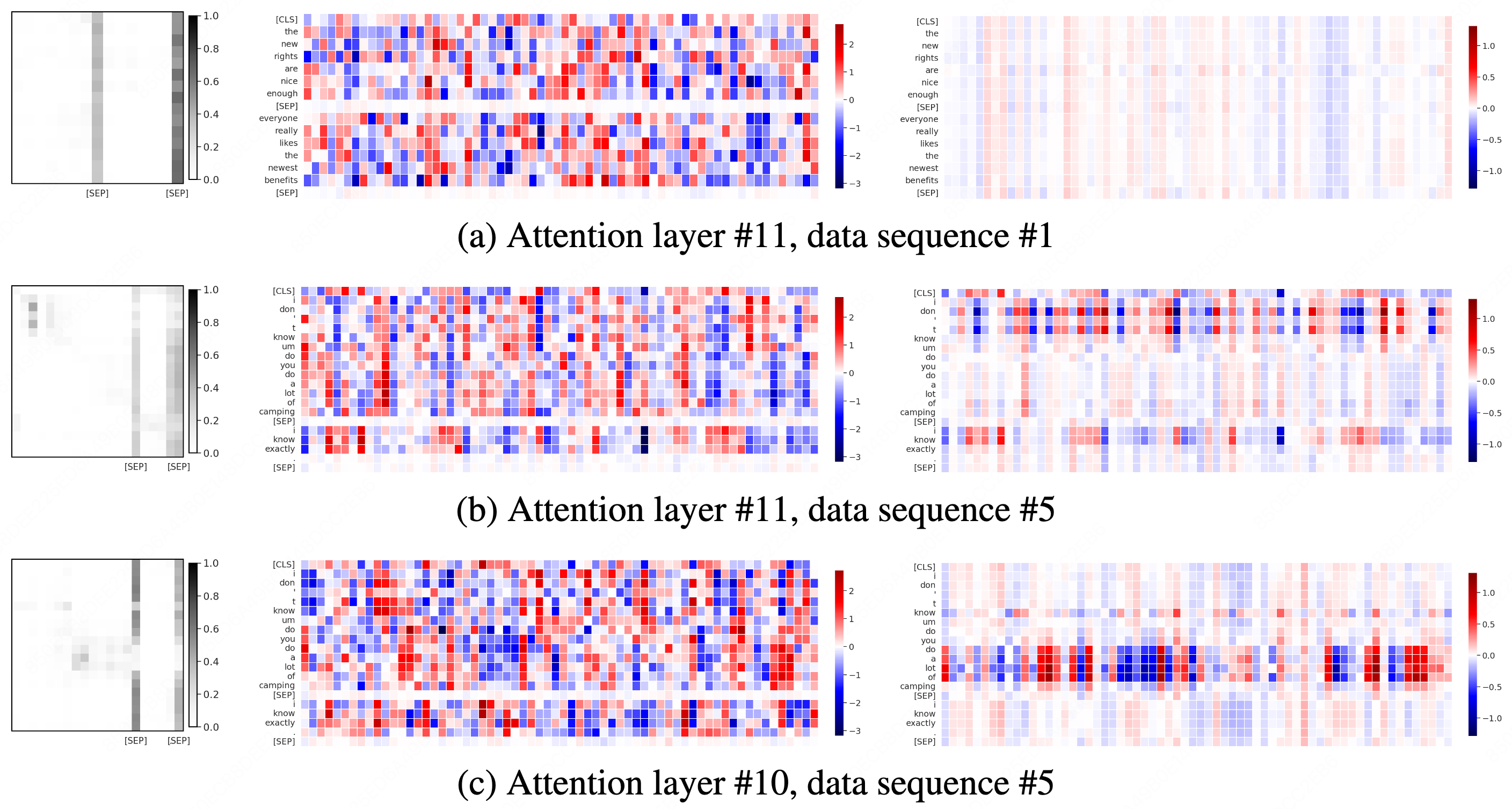}
    \caption{Visualization of self-attention patterns in BERT-base, showing attention probabilities (left), value magnitudes (middle), and their product (right) for attention head 3. Sink tokens such as \texttt{[SEP]} receive high attention but exhibit small value outputs, consistent with the no-op behavior predicted by the theory. The figure is adapted from \cite{bondarenko2023quantizable}.}
    \label{fig:quantizable_noop}
\end{figure}

\subsubsection{Core Concepts}

Among the earliest and most influential explanations for AS emergence, \textit{Quantizable Transformers} \cite{bondarenko2023quantizable} attributes this phenomenon to an inherent limitation of the Softmax function. In standard attention, the sum-to-one constraint requires that the attention weights over all keys normalize to unity for each query. When a query does not meaningfully align with any key in the context, the mechanism lacks a natural “null” option and is therefore forced to distribute attention mass to uninformative tokens.

Formally, for a query vector $q_i$, let the pre-Softmax logit for token $j$ be defined as $x_j = q_i k_j^\top / \sqrt{d}$. The Softmax output for a non-sink token approaches zero only under the extreme condition:
\begin{equation}
\text{Softmax}(x)_i = 0 \;\Longleftrightarrow\; \exists j \neq i,\; x_j - x_i = +\infty,
\end{equation}
which pushes the pre-Softmax logits to extreme values to satisfy the sum-to-one constraint, resulting in near-zero attention on non-sink tokens and giving rise to the activation outliers empirically observed in transformer layers. Because Softmax never outputs exact zeros, these extreme logits continue to receive gradient signals during backpropagation, causing the outliers to grow further in magnitude as training progresses. Layer normalization amplifies this effect. By compressing these outliers, it forces the preceding feed-forward layers to generate even larger activations, ensuring that the required dynamic range is preserved.
Consequently, attention heads learn to circumvent the Softmax constraint by adopting a no-op behavior. Let $\mathcal{S}$ denote the set of sink tokens (e.g., \texttt{[SEP]}, punctuation, or background patches). The resulting attention pattern can be approximated as:
\begin{equation}
A_{ij} \approx 
\begin{cases}
1, & j \in \mathcal{S} \\[4pt]
0, & \text{otherwise}
\end{cases}
\qquad \text{with} \qquad \|V_{\mathcal{S}}\| \approx 0,
\end{equation}
where nearly all attention mass concentrates on sink tokens, whose value vectors are negligible, thereby producing minimal updates to the residual representation.

Beyond \textit{Quantizable Transformers}, other studies offer complementary perspectives. 
\textit{Attention Needs to Focus} \cite{fu2026attention} frames AS as ``attention underload''—a failure mode where no token is semantically relevant, yet Softmax forces attention to distribute, resulting in spurious focus that manifests as AS. This unified perspective reveals that AS is not an isolated artifact but a specific manifestation of improper attention allocation under the Softmax constraint. 
\textit{Variance Sensitivity} \cite{hongvariance} demonstrates that Softmax is highly sensitive to the variance of attention logits. As variance increases, the exponential function in Softmax disproportionately amplifies larger logits while suppressing smaller ones, causing the attention distribution to collapse onto a single token. This mathematical property, formalized as the negative derivative of attention entropy with respect to logit variance, explains why AS emerges as an inherent consequence of Softmax dynamics, independent of learned behavior.
\textit{Value-State Gated Attention (VGA)} \cite{bu2025value} further identifies that AS and value-state drain are mutually reinforcing: high attention on sink tokens suppresses their value states, which in turn encourages even higher attention concentration, creating a self-sustaining cycle. This insight highlights the coupling between attention scores and value representations in driving no-op behavior.

\begin{figure}[t]
    \centering
    \includegraphics[width=1\linewidth]{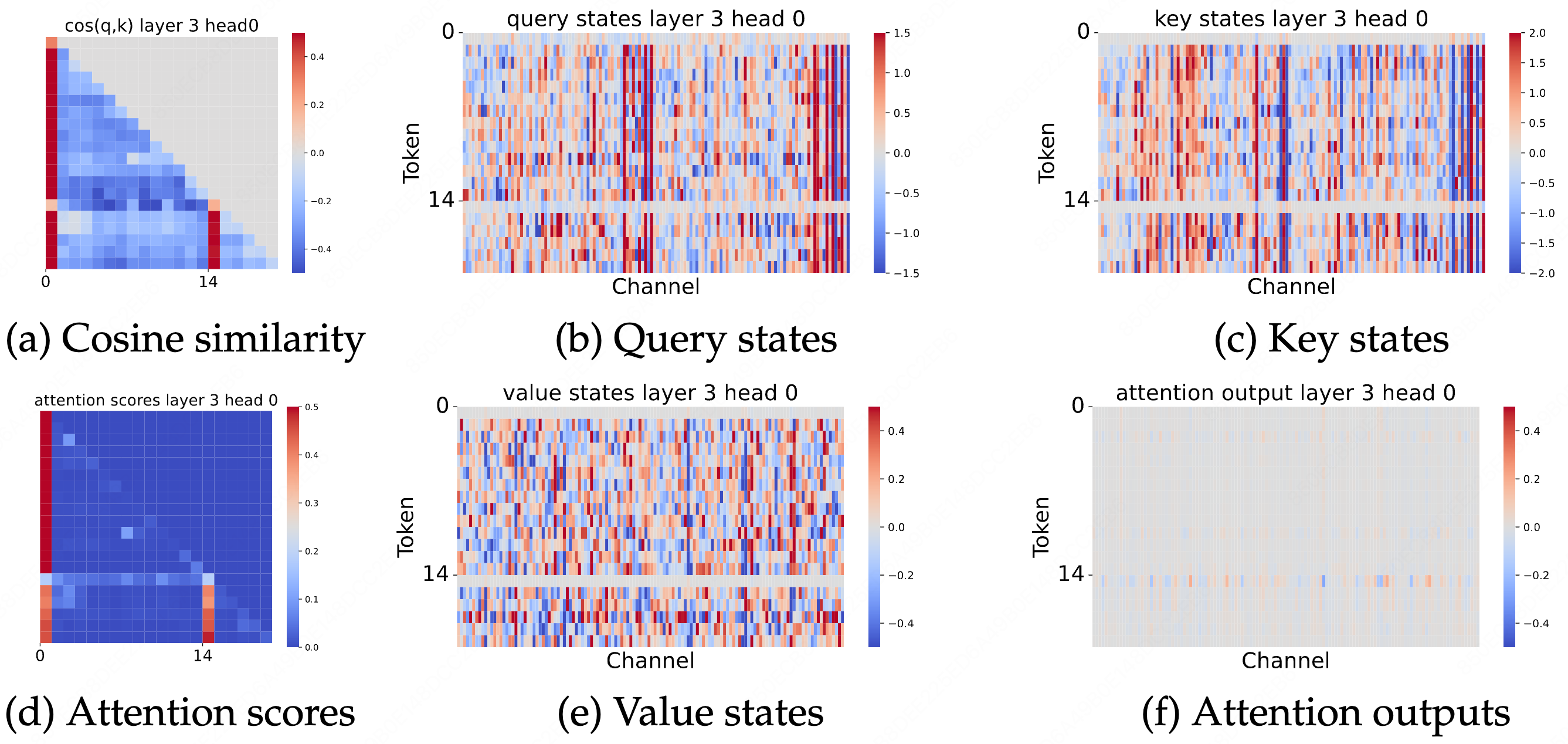}
    \caption{Analysis of sink token properties. (a) High cosine similarity of QK states. (b), (c), and (e) illustrate QKV states, showing that sink tokens exhibit significantly smaller value magnitudes. (f) Visualizes the attention output, demonstrating the minimal residual contribution of sink tokens. The figure is adapted from \cite{su2025kvsink}.}
    \vspace{+5mm}
    \label{fig:kvsink_value}
\end{figure}

\subsubsection{Supporting Evidence}

\paragraph{Observational Evidence.}
A key observational validation of the no-op theory is that sink tokens consistently exhibit significantly smaller value states compared to other tokens, confirming their role in producing minimal residual updates.
The following studies provide direct observational evidence supporting this phenomenon.

\begin{itemize}
    \item \textbf{Quantizable Transformers} \cite{bondarenko2023quantizable}: First identifies this pattern in BERT and ViTs, showing that sink tokens (e.g., \texttt{[SEP]} in language models or background patches in ViTs) receive disproportionately high attention while their value outputs remain near zero (see Figure~\ref{fig:quantizable_noop}).

    \item \textbf{Attention Score is Not All You Need} \cite{guo2024attention}: Provides evidence that value vector norms are distributed non-uniformly across tokens, with sink tokens exhibiting distinctly smaller norms. These findings challenge the prevailing practice of relying solely on attention scores to evaluate token importance.
    
    \item \textbf{Active-Dormant Attention Heads} \cite{guo2024active}: Systematically analyzes this behavior in LLMs including Llama and OLMo, demonstrating that sink tokens exhibit value-state drains as part of a mutual reinforcement mechanism between active and dormant attention heads.
    
    \item \textbf{KVSink} \cite{su2025kvsink}: Observes that the small value magnitudes of sink tokens make them highly sensitive to quantization (see Figure~\ref{fig:kvsink_value}). When these value-suppressed tokens are compressed during KV quantization, the resulting errors are disproportionately amplified, leading to performance degradation.
\end{itemize}

\paragraph{Causal Evidence.}
Several studies have empirically demonstrated that relaxing or removing the sum-to-one constraint of Softmax effectively mitigates AS, providing causal evidence supporting the theory. Representative techniques include \textit{\textbf{Gated Attention Mechanisms}} and \textit{\textbf{Modified Softmax Functions}}.

\begin{figure}[t]
    \centering
    \includegraphics[width=1\linewidth]{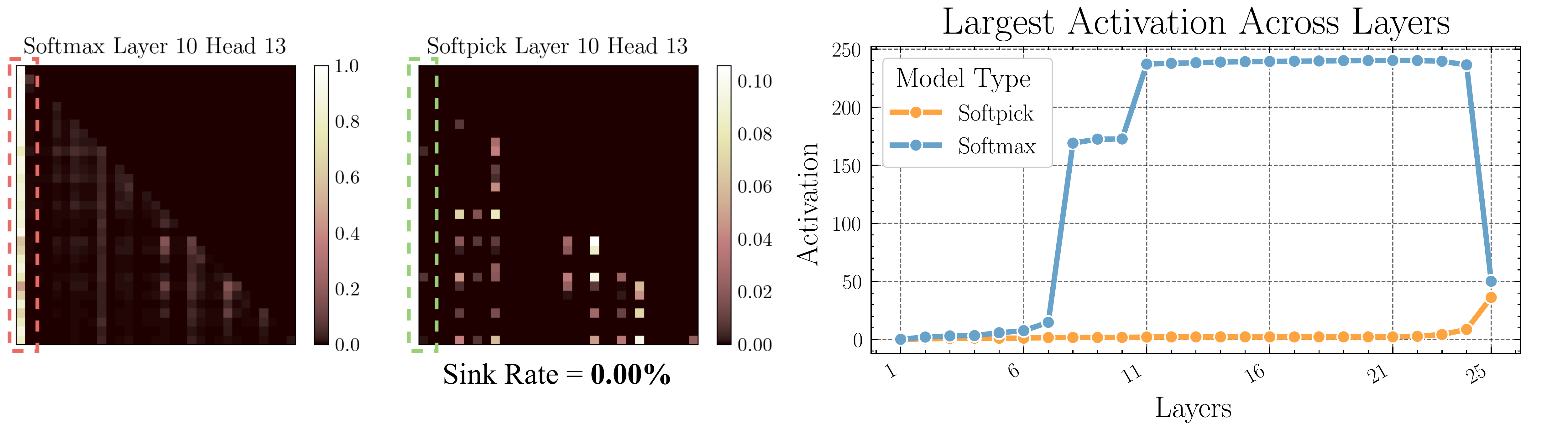}
    \caption{(Left) Comparison of attention maps using Softmax versus \textit{Softpick} and overall sink rate of the 340M models. (Right) Largest hidden state activation per layer of the 340M models. \textit{Softpick} significantly mitigates both AS and large activations.
    The figure is adapted from \cite{zuhri2025softpick}.}
    \label{fig:sec4_softpick}
\end{figure}
\begin{itemize}
    \item \textbf{Gated Attention Mechanisms:}
    \textit{Gated Attention} \cite{qiu2025gated} introduces query-dependent sparse gating after Softmax, which reduces the model's reliance on sink tokens for numerical stability and enhances long-context extrapolation.
    \textit{Value-State Gated Attention (VGA)} \cite{bu2025value} proposes a learnable, data-dependent gate computed directly from value vectors, specifically targeting the mutual reinforcement cycle between attention scores and value-state drains that drives no-op behavior.
    Together, these approaches demonstrate that employing \textit{\textbf{Gated Attention Mechanisms}} effectively mitigates AS, with a more detailed discussion presented in \S~\ref{sec_5_1_Gated_Attention_Mechanisms}.
    
    \item \textbf{Modified Softmax Functions:}
    \textit{Softpick} \cite{zuhri2025softpick} replaces Softmax with a rectified function that does not require probabilities to sum to one, achieving a 0\% sink rate and eliminating massive activations (see Figure~\ref{fig:sec4_softpick}).
    \textit{Softmax-1} \cite{kaul2025attention} modifies the normalization to allow sub-unit summation (denominator +1), reducing first-token attention from 65\% to 3.3\%.
    \textit{Sigmoid Attention} \cite{gu2025attention} removes normalization entirely, applying the sigmoid function independently to each logit; without the sum-to-one constraint, forced attention allocation is eliminated and AS does not emerge.
    Together, these approaches provide empirical support for the effectiveness of \textit{\textbf{Modified Softmax Functions}} in mitigating AS, with a more detailed discussion presented in \S~\ref{sec_5_2_Modified_Softmax_Functions}.
    % \textit{Elastic-Softmax} \cite{fu2026attention} relaxes the Softmax constraint by adding a small constant to the denominator, allowing total attention to sum to less than one and directly counteracting attention underload—the condition that gives rise to no-op behavior.
    % \textit{Sliding Window Attention Training (SWAT)} \cite{fu2025sliding} replaces Softmax with sigmoid and modifies position encoding, eliminating the need for sink tokens and enabling efficient long-context modeling.
\end{itemize}

% These interventions collectively confirm that no-op behavior is a direct consequence of the Softmax constraint. When the constraint is relaxed or removed, AS disappears or is significantly reduced, proving that AS is not an inherent model requirement but a structural adaptation to Softmax's mathematical limitations.

\subsubsection{Discussion and Insights}
 
\textbf{Advantages.}  
The no-op theory provides a parsimonious causal explanation for AS, unifying previously disparate observations, including high attention to delimiters or background patches, small value norms of sink tokens, and activation outliers, within a single causal framework. It generates testable predictions, such as the expectation that sink tokens exhibit small value outputs, which have been empirically validated across BERT, ViT, LLaMA, and OLMo. Furthermore, the theory directly motivates effective mitigation strategies, including \textbf{\textit{Gated Attention Mechanisms}} and \textbf{\textit{Modified Softmax Functions}}, whose success in reducing or eliminating AS offers strong causal support.

\textbf{Limitations.}  
Despite its explanatory power, the no-op theory has several limitations. First, the evolution of mutual reinforcement between attention scores and value states during optimization remains largely unexplored. Second, while value suppression is identified as a key signature, the mechanisms underlying the reduction of value norms are still unclear. Finally, although gating and modified Softmax provide practical mitigation, the theory has yet to systematically explore alternative strategies.

\textbf{Future Directions.}  
Future work should extend the no-op theory to incorporate training dynamics that govern sink formation and evolution, including the emergence of sinks at non-initial positions. Formalizing the interaction between Softmax constraints and optimization dynamics may clarify how no-op behavior arises during training. Investigating the mechanisms of value norm suppression would further strengthen the theory’s mechanistic foundation. Beyond gating and modified Softmax, exploring alternative mitigation strategies could yield more robust and efficient approaches for controlling AS.

% Yuxuan & Zunhai

\subsection{Outlier Circuits}
\label{sec_4_2_Outliers_Circuits}

\begin{figure}[t]
    \centering
    \begin{subfigure}[b]{0.22\textwidth}
        \includegraphics[width=\textwidth]{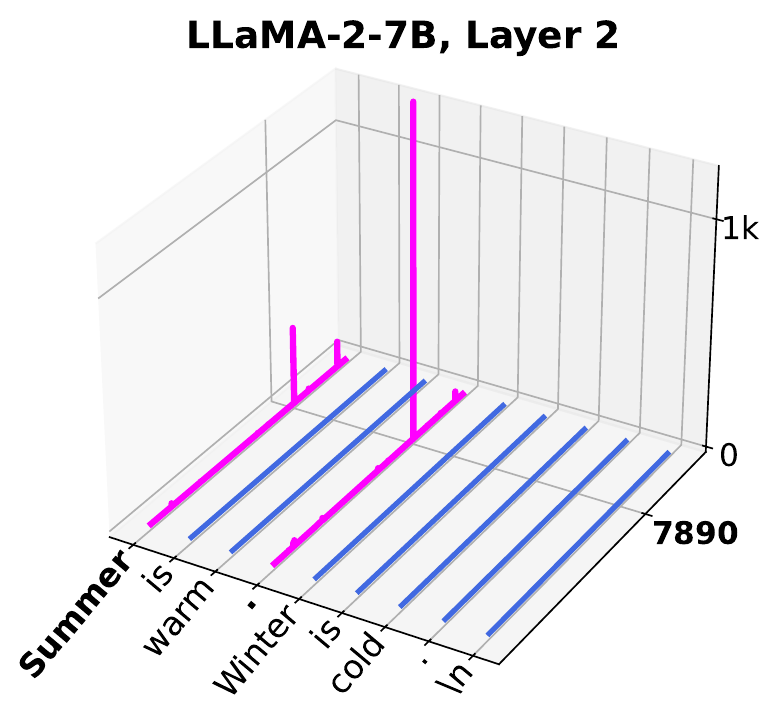}
        \caption{Activation: $\mathbf{x}_{\ell}^{\text{down}}$}
    \end{subfigure}
    \hfill
    \begin{subfigure}[b]{0.21\textwidth}
        \includegraphics[width=\textwidth]{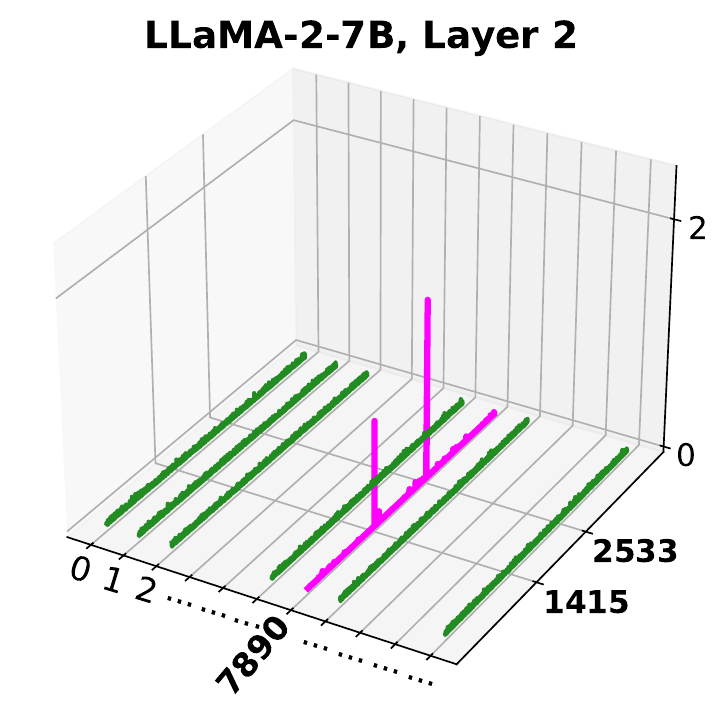}
        \caption{Weight: $\mathbf{W}_{\ell}^{\text{down}}$}
    \end{subfigure}
    \hfill
    \begin{subfigure}[b]{0.22\textwidth}
        \includegraphics[width=\textwidth]{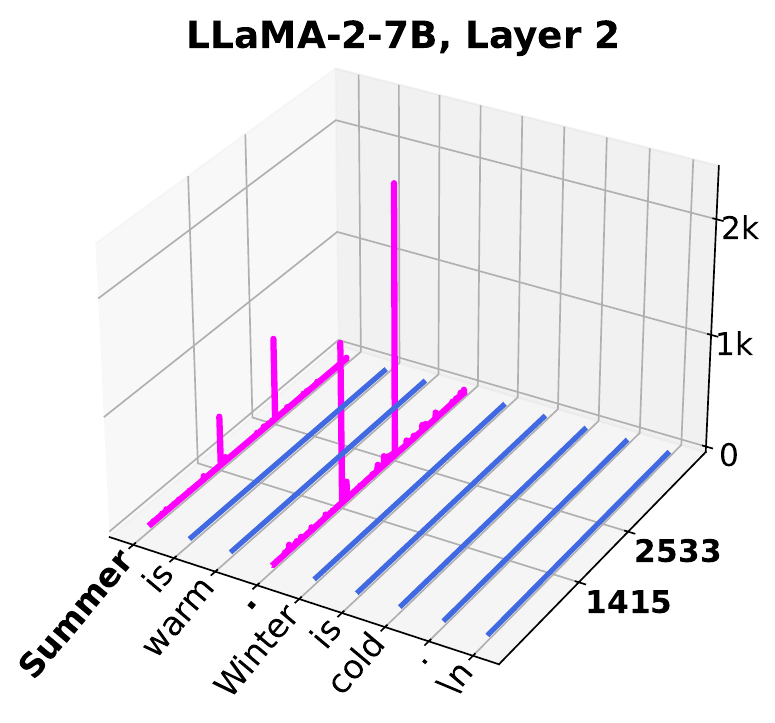}
        \caption{Activation: $\mathbf{h}_{\ell}$}
    \end{subfigure}
    \hfill
    \begin{subfigure}[b]{0.22\textwidth}
        \includegraphics[width=\textwidth]{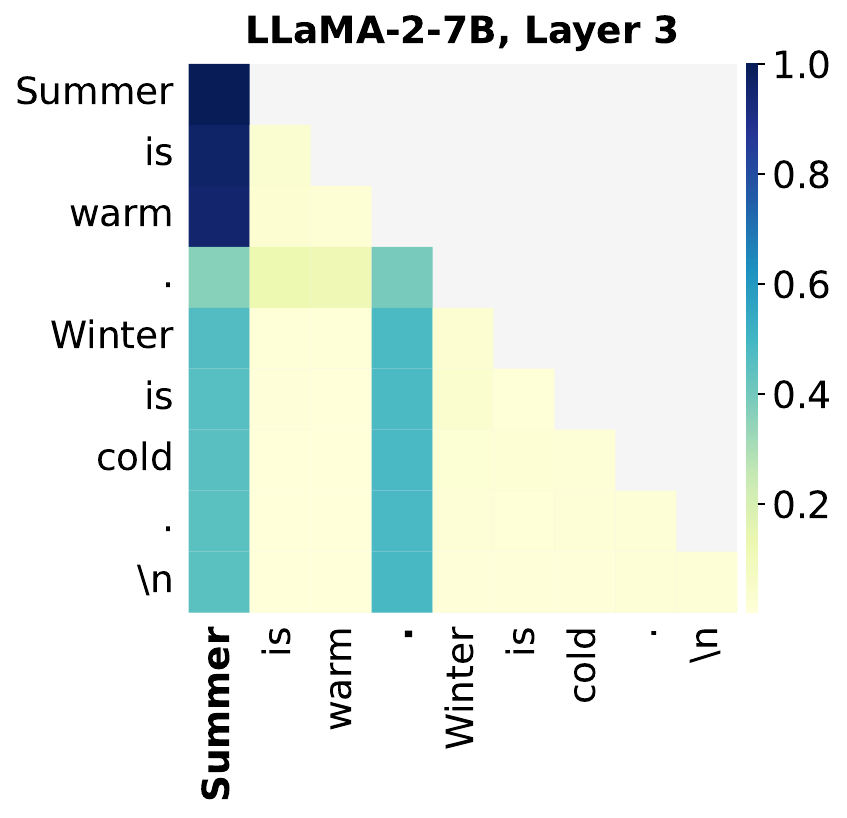}
        \caption{Attention: $\mathbf{A}_{\ell}^{i}$}
    \end{subfigure}
    \caption{\textit{Systematic outliers} in LLaMA2-7B. Outliers are identified in four locations: activations (layer outputs $\mathbf{h}_{\ell}$ and down-projection inputs $\mathbf{x}_{\ell}^{\text{down}}$), weights (down-projection matrices $\mathbf{W}_{\ell}^{\text{down}}$), and attention (attention weights $\mathbf{A}_{\ell}^{i}$). The figure is adapted from \cite{an2025systematic}.}
    \label{fig:so_llama2_7b}
\end{figure}
\begin{tcolorbox}[takeawaysbox]
{\large \textbf{\textcolor{TikTokPink}{\textit{Key Takeaways:}}}}
\begin{enumerate}[leftmargin=*, label=\arabic*)]
    \item \textbf{Core Concepts:}  
    \textit{\textbf{Outlier Circuits}} identify systematic outliers that form circuit-like pathways, serving as the numerical infrastructure sustaining AS. These outliers concentrate attention on sink tokens and exhibit a predictable lifecycle across layers.
    
    \item \textbf{Supporting Evidence:}  
    Observational studies across multiple Transformer models consistently show that outliers co-occur with AS. Causal interventions directly modulate AS behavior, confirming that outliers are functionally necessary for its emergence and maintenance.
    
    \item \textbf{Discussion and Insights:}  
    This framework unifies empirical observations and provides a quantitative foundation for understanding AS. 
    However, it faces two key challenges: incomplete causal validation of component interactions, and largely unexplored training dynamics that govern circuit emergence, stability, and evolution. 
    Future research should focus on systematic causal intervention studies to establish a complete mechanistic understanding.
\end{enumerate}
\end{tcolorbox}

\subsubsection{Core Concepts}

\textit{\textbf{Softmax Limitations and No-Op Theory}} explains why AS emerge from the Softmax constraint, but it does not elucidate the numerical mechanisms that sustain them. The \textit{\textbf{Outlier Circuits}} perspective addresses this gap by identifying different types of \textit{systematic outliers} and demonstrating how they form interconnected, circuit-like pathways that stabilize AS \cite{an2025systematic,sun2024massive,su2025kvsink}. This section is organized into two parts: (i) the types of \textit{systematic outliers} and (ii) the formation and evolution of the \textit{\textbf{Outlier Circuits}}.

\paragraph{Types of Systematic Outliers.}  
Following \textit{Systematic Outliers} \cite{an2025systematic}, the outliers are categorized into three distinct types, as illustrated in Figure~\ref{fig:so_llama2_7b}:  

\begin{itemize}
    \item \textbf{Weight Outliers}: Exceptionally large values concentrated in specific columns of the down-projection matrices $\mathbf{W}_{\ell}^{\text{down}}$ in MLP layers. In LLaMA2‑7B, these outliers are observed in the second layer as well as the last two layers. They are also referred to as \textit{Super Weight} \cite{yu2024super}.  

    \item \textbf{Activation Outliers}: Abnormally large activations in hidden states, categorized into two subtypes. Both are confined to specific feature dimensions and exhibit minimal variation across different inputs:  
        \begin{itemize}
            \item \textbf{Down-Projection Input Outliers} ($\mathbf{x}_{\ell}^{\text{down}}$): Localized to a limited number of shallow and deep layers, also known as \textit{Activation Spikes} \cite{xiang2025dfrot}. 
            \item \textbf{Layer Output Outliers} ($\mathbf{h}_{\ell}$): These activations persist across layers but diminish in the final layers. They are also referred to as \textit{Massive Activations} \cite{sun2024massive}.   
        \end{itemize}

    \item \textbf{Attention Outliers}: Certain keys receive disproportionately high cumulative attention scores, corresponding precisely to AS. These outliers persist across nearly all layers.  
\end{itemize}

These three types of outliers demonstrate interdependence: weight outliers align with activation outliers along feature dimensions, whereas activation outliers coincide with AS across sequence positions \cite{an2025systematic}.

\begin{figure}[t]
    \centering
    \includegraphics[width=1\linewidth]{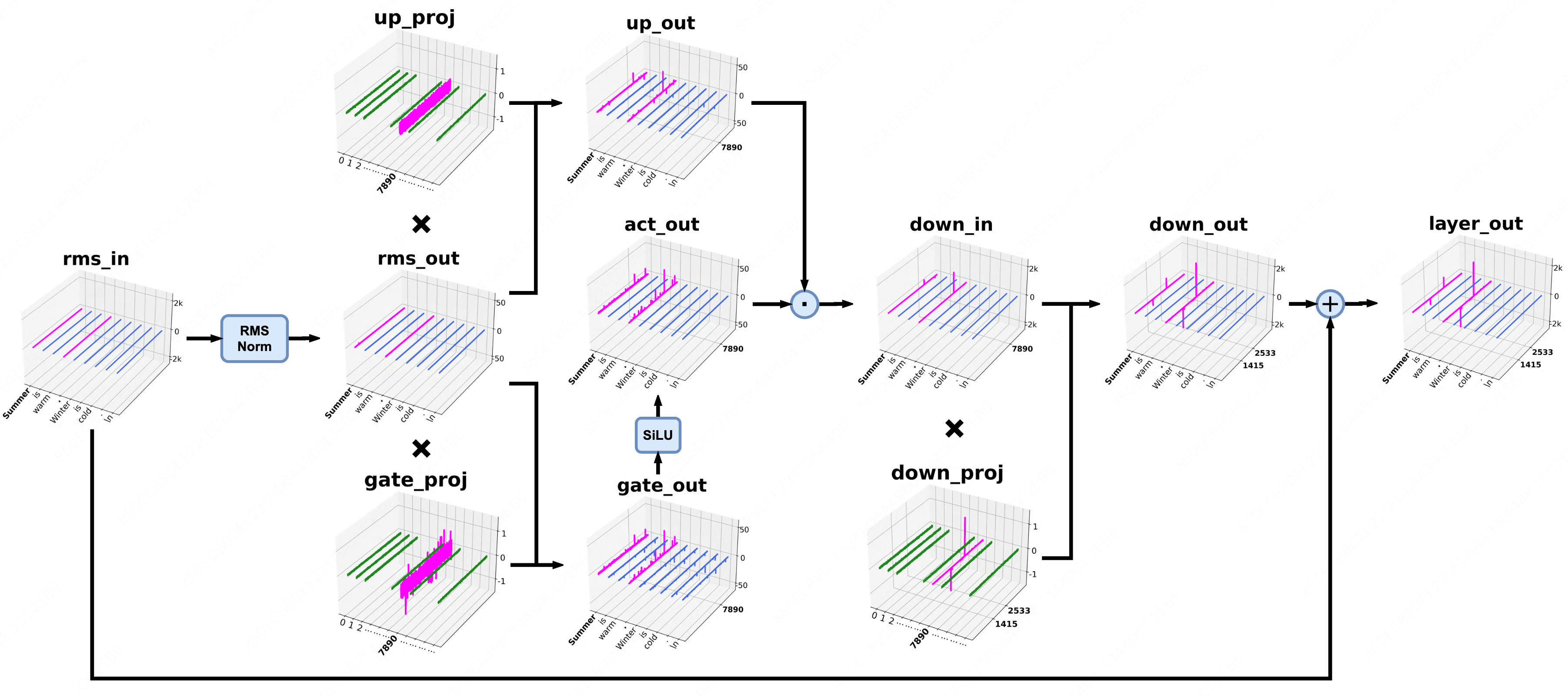}
    \caption{The emergence of activation outliers from weight outliers. The figure is adapted from \cite{an2025systematic}.}
    \label{fig:emergence}
\end{figure}

\begin{figure}[t]
    \centering
    \includegraphics[width=1\linewidth]{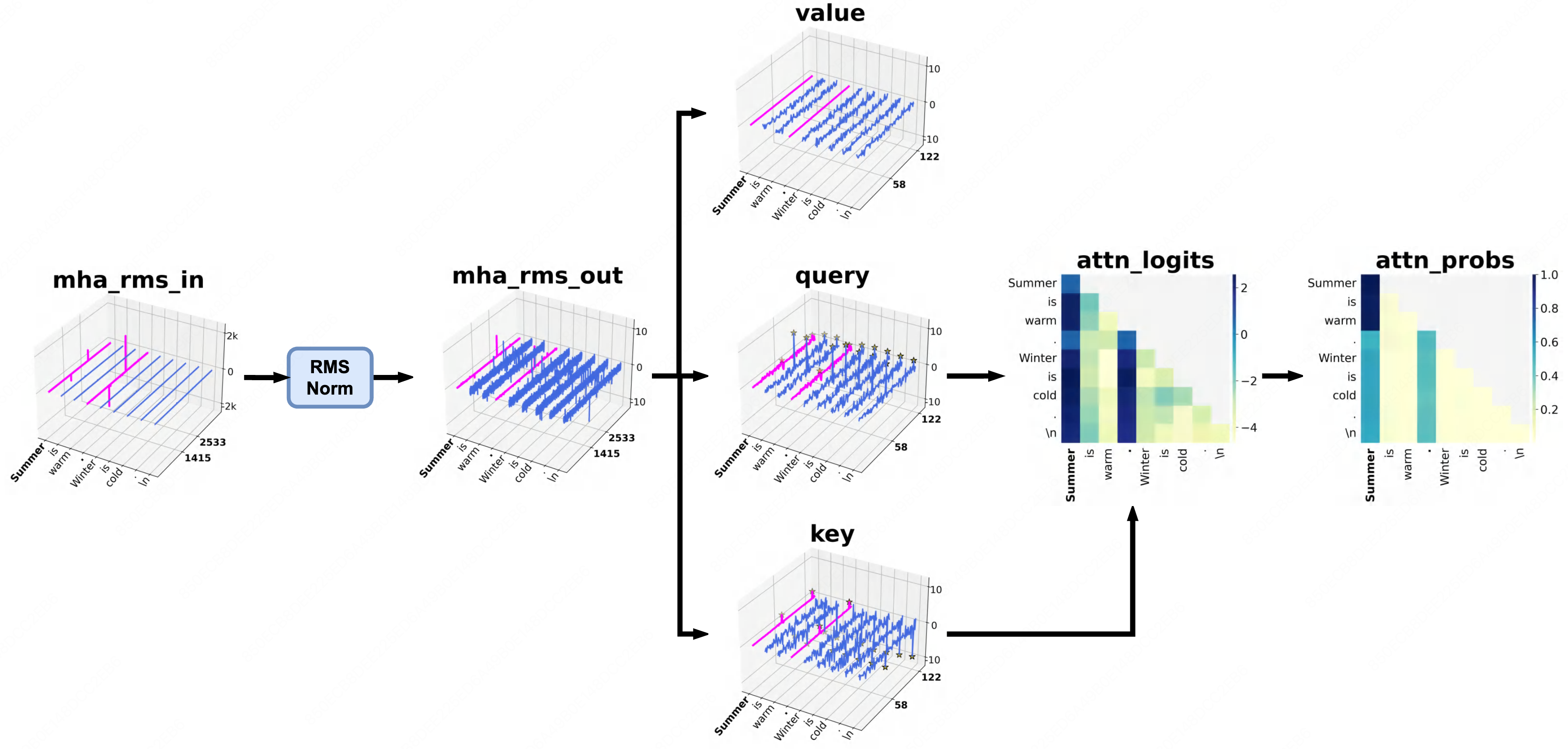}
    \caption{The spread of attention outliers from activation outliers (AS). Activation outliers influence the self-attention mechanism. The figure is adapted from \cite{an2025systematic}.}
    \label{fig:spread}
\end{figure}

% \begin{figure}[t]
%     \centering
%     \includegraphics[width=1\linewidth]{Figures/Section_4/ma_as.png}
%     \caption{Attention patterns before and after massive activations appear in LLaMA2-7B. The figure is adapted from \cite{sun2024massive}.}
%     \label{fig:ma_outlier_as}
% \end{figure}

\paragraph{Formation and Evolution of the Outlier Circuit.}
As illustrated in Figures~\ref{fig:emergence} and~\ref{fig:spread}, the \textit{\textbf{Outlier Circuit}} emerges through a well-defined causal chain \cite{an2025systematic,su2025kvsink}, forming a closed-loop mechanism that sustains AS.

\begin{enumerate}
    \item \textbf{Down-projection input outliers.} 
    In early layers, large weight values in the up-projection and gate-projection weight matrices induce unusually high neuron activations. These activations constitute the first type of activation outliers.
    
    \item \textbf{Down-projection outliers propagate to layer outputs via residual connections.} Weight outliers in the down-projection matrix $\mathbf{W}_{\ell}^{\text{down}}$ amplify specific feature dimensions. These amplified values propagate through residual connections, producing the second type of activation outliers.
    
    \item \textbf{Activation outliers induce attention outliers.} Tokens exhibiting activation outliers show strong alignment in particular dimensions of their query and key vectors. This alignment substantially increases the dot product, leading the Softmax to assign disproportionately high attention weights to these tokens, thereby forming AS. Importantly, the value vectors of these tokens remain comparatively small, resulting in minimal output contributions, consistent with no-op behavior.
    
    % \item \textbf{Predictable lifecycle of the circuit.} The outlier circuit follows a systematic lifecycle: outliers emerge in early layers, stabilize in middle layers, and gradually diminish in the final layers due to cancellation effects from values of opposite signs \cite{an2025systematic,su2025kvsink}.
\end{enumerate}

\subsubsection{Supporting Evidence}

\paragraph{Observational Evidence.}
Multiple studies have directly observed the correlation between outliers and AS across different transformer architectures.

\begin{figure}[t]
    \centering
    \includegraphics[width=1\linewidth]{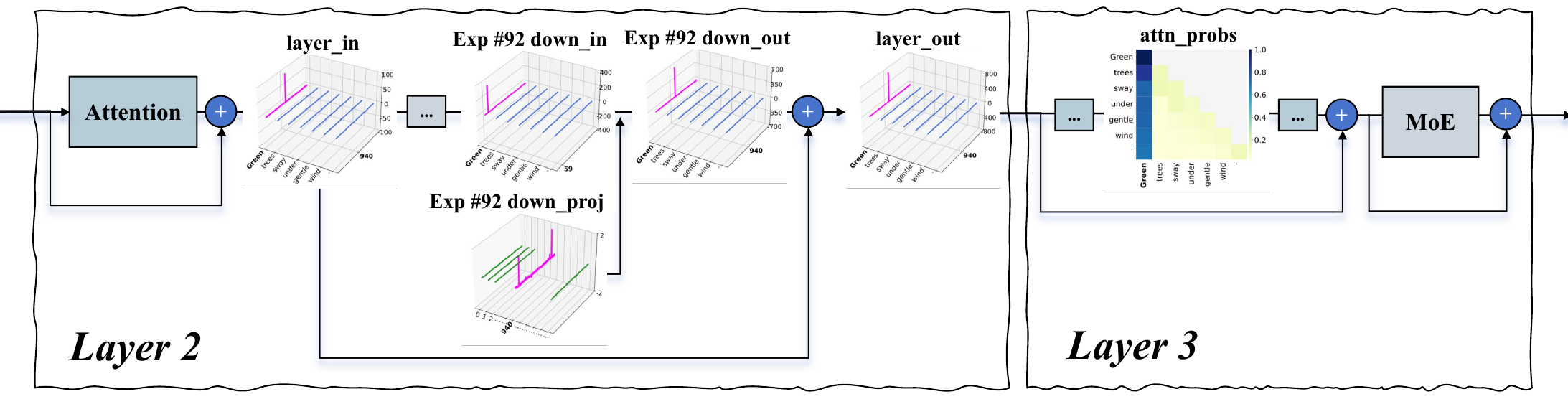}
    \caption{Systematic outlier mechanism in Qwen3-30B-A3B MoE LLM. The figure is adapted from \cite{su2026unveiling}.}
    \label{fig:moe_sink}
\end{figure}

\begin{figure}[t]
        \centering
\includegraphics[width=1\linewidth]{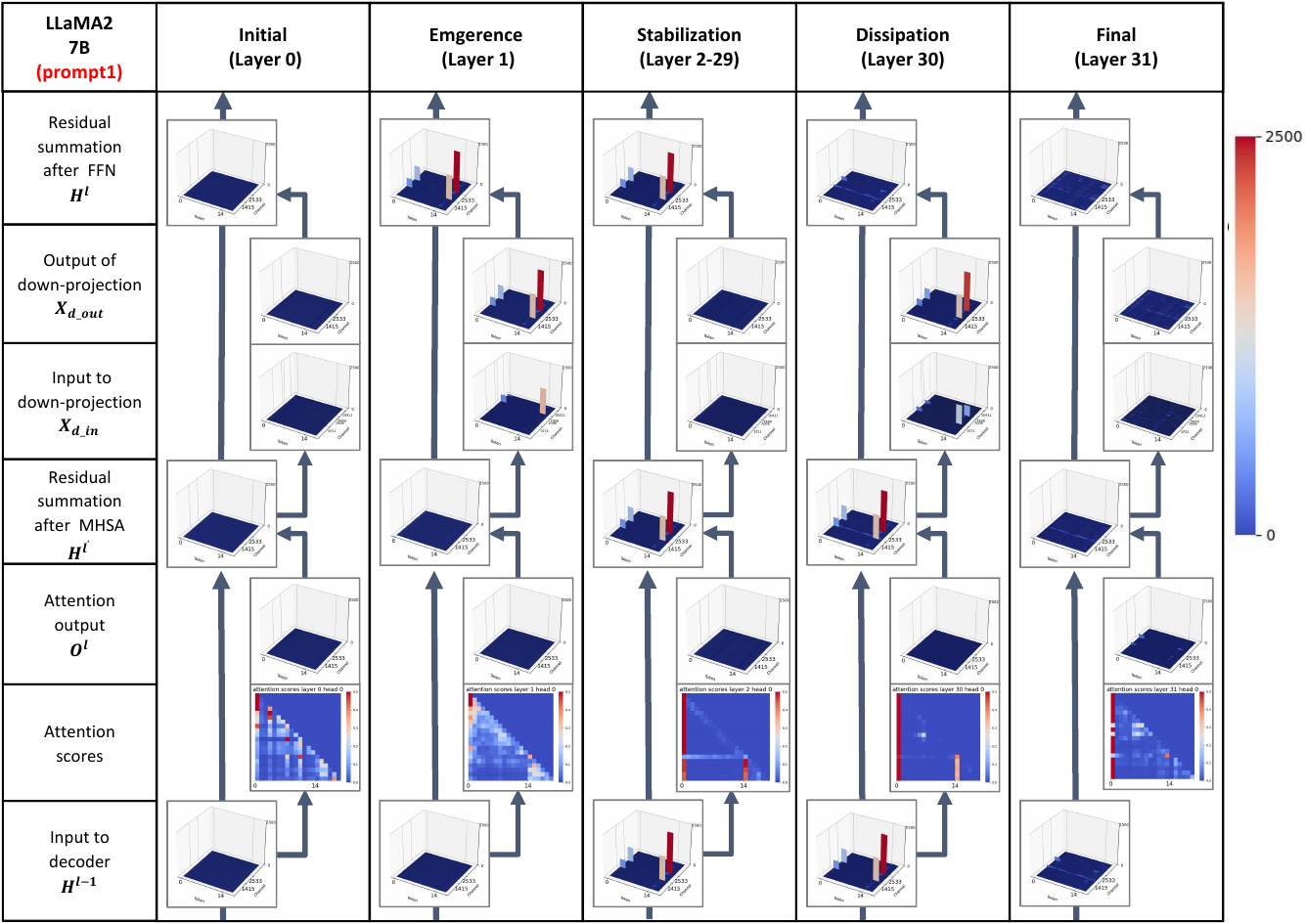}
\caption{Cross-layer evolution of extreme activation outliers in LLaMA2-7B. Activation outliers and AS exhibit a systematic and stable interaction. The figure is adapted from \cite{su2025kvsink}.}
\vspace{-3mm}
\label{fig:cross_llama_2_7b_p1}    
\end{figure}

\begin{itemize}
    \item \textbf{Classical Language Models:} \textit{Understanding Transformer Quantization} \cite{bondarenko2021understanding} identifies structured outliers in residual connections that encourage specific attention patterns, such as attending to the \texttt{[SEP]} token. \textit{Outlier Dimensions Driven by Frequency} \cite{puccetti2022outlier} shows that outlier dimensions contribute to the ``vertical'' self-attention pattern, enabling models to focus on special tokens (\texttt{[CLS]}, \texttt{[SEP]}). \textit{Quantizable Transformers} \cite{bondarenko2023quantizable} further demonstrates that no-op behavior drives outlier formation in BERT, with sink tokens receiving disproportionately high attention while exhibiting near-zero value outputs, establishing outliers as the numerical manifestation of AS.
    
    \item \textbf{LLMs:} \textit{Massive Activations} \cite{sun2024massive} reveals that massive activations directly causing attention probabilities to concentrate on their corresponding tokens. \textit{KVSink} \cite{su2025kvsink} shows that AS formation is tied to the cross-layer evolution of extreme activation outliers, following a predictable lifecycle—emerging in early layers, stabilizing in middle layers, and gradually vanishing in the final layers (as shown in Figure~\ref{fig:cross_llama_2_7b_p1}). 
    % \textit{IntactKV} \cite{liu2024intactkv} confirms that pivot tokens (i.e., AS) exhibit outlier characteristics, with attention scores heavily concentrated on initial tokens, further establishing the numerical distinctiveness of AS.

    \item \textbf{MoE LLMs:} \textit{Unveiling Super Experts} \cite{su2026unveiling} identifies that \textit{Super Experts} are characterized by rare but extreme activation outliers in their down-projection outputs. These outliers generate \textit{Massive Activations} that directly give rise to AS (as shown in Figure~\ref{fig:moe_sink}).
    
    \item \textbf{ViT:} \textit{Massive Activations} \cite{sun2024massive} demonstrates that massive activations also occur in Vision Transformers and lead to attention concentration on corresponding tokens. \textit{Quantizable Transformers} \cite{bondarenko2023quantizable} shows that no-op behavior drives outlier formation in ViT, mirroring the behavior observed in language models.
    
    \item \textbf{MLLMs:}  
    \textit{See What You Are Told} \cite{kang2025see} demonstrates that visual AS can be precisely identified by detecting \textit{Massive Activations}, indicating that outliers serve as reliable markers for AS in multimodal contexts. This establishes a direct link between outlier magnitudes and the identification of sink tokens.
    
    \item \textbf{Audio-Visual Speech Recognition (AVSR):}  
    \textit{Mitigating AS in AVSR} \cite{cappellazzo2025mitigating} reports that massive activations co-occur with AS not only at the \texttt{[BOS]} token but also at intermediate low-semantic tokens. These activations originate from MLP layers and correspond to fixed feature indices across all sink tokens, confirming the cross-modal generality of the outlier–AS relationship.
\end{itemize}


\paragraph{Causal Evidence.}  
Direct interventions on outliers have a profound and measurable impact on AS, providing compelling causal validation of their central role in sustaining AS behavior.

\begin{itemize}

    \item \textbf{Unveiling Super Experts} \cite{su2026unveiling}: Pruning only three of the 6,144 \textit{Super Experts}, which concentrate extreme activation outliers, triggers a catastrophic collapse of AS and leads to repetitive, uninformative outputs. This experiment provides strong causal evidence that removing sources of outliers directly disrupts AS and significantly degrades model performance.

    \item \textbf{See What You Are Told} \cite{kang2025see}: By identifying and redistributing attention from outlier-driven visual sinks, this approach enhances visual grounding and reduces hallucinations in MLLMs, directly demonstrating that modulating outliers controls AS behavior.

    \item \textbf{Mitigating AS in AVSR} \cite{cappellazzo2025mitigating}: Introducing a decorrelation loss to reduce cosine similarity between the BOS token and other tokens effectively mitigates both massive activations and intermediate sinks, showing that eliminating outliers alleviates AS in audio-visual speech recognition tasks.

    \item \textbf{IntactKV} \cite{liu2024intactkv}: Preserving pivot tokens that exhibit outlier characteristics at full precision while quantizing other tokens substantially recovers quantization-induced accuracy loss. This demonstrates that protecting outliers maintains AS functionality and overall model performance.

\end{itemize}

\subsubsection{Discussion and Insights}

\textbf{Advantages.} The \textit{\textbf{Outlier Circuits}} framework offers a fundamental numerical perspective for understanding AS. It shows that extreme activation outliers, systematically localized across specific feature dimensions and layers, are not incidental artifacts but the primary drivers of attention concentration on sink tokens. This framework unifies diverse empirical observations across architectures, underscoring the generality of the outlier–AS relationship. Causal evidence from intervention studies such as pruning \textit{Super Experts} confirms that these outliers are functionally indispensable for AS. Their removal collapses AS, while their preservation maintains model performance. The documented cross-layer lifecycle further characterizes \textit{\textbf{Outlier Circuits}} as a predictable dynamical system.

\textbf{Limitations.}  
Despite its explanatory power, the \textit{\textbf{Outlier Circuits}} framework has several notable limitations. 
First, while some causal evidence exists, it remains incomplete. 
The roles and interactions of other model components with \textit{\textbf{Outlier Circuits}} are largely unclear, limiting a full causal interpretation of how these circuits drive outlier formation. 
Second, the training dynamics that give rise to the systematic alignment of weights, activations, and attention outliers remain largely unexplored. 
Critical open questions include when during training these circuits emerge, how they stabilize or evolve across optimization steps, and which hyperparameters most strongly influence their development.

\textbf{Future Directions.}  
Future research should address these gaps through both theoretical and practical advances. 
First, developing a complete causal understanding of \textit{\textbf{Outlier Circuits}}, including systematic causal interventions to validate the roles of different model components, could provide foundational insights into Transformer behavior. 
Second, formalizing the training dynamics that drive outlier emergence demands longitudinal analyses tracking circuit formation across training epochs. 
Such investigations would elucidate how these circuits form, evolve, and interact with optimization processes, thereby enabling precisely targeted interventions that suppress outlier circuits at their source.% Zunhai

\subsection{Implicit Attention Bias}
\label{sec_4_3_Implicit_Attention_Bias}

\begin{tcolorbox}[takeawaysbox]
{\large \textbf{\textcolor{TikTokPink}{\textit{Key Takeaways:}}}}
\begin{enumerate}[leftmargin=*, label=\arabic*)]
    \item \textbf{Core Concepts:}  
    \textit{\textbf{Implicit Attention Bias}} conceptualizes AS as a fixed, input-independent bias injected into the attention. Introducing explicit attention biases can effectively mitigate AS.

    \item \textbf{Supporting Evidence:}  
    Empirical observations across multiple studies consistently indicate that AS functions as an implicit attention bias. Complementary causal interventions, such as learnable key biases, further demonstrate that AS can be modulated, providing strong support.

    \item \textbf{Discussion and Insights:}  
    This perspective directly links AS to the Softmax sum-to-one constraint. Current limitations include underexplored training dynamics and fragmented characterization of bias types. Future research should formalize the emergence of implicit attention biases during training, unify diverse bias variants under a coherent theoretical framework, and investigate how these biases can be harnessed to enhance model efficiency and interpretability.
\end{enumerate}
\end{tcolorbox}

\subsubsection{Core Concepts}

\textit{\textbf{Implicit Attention Bias}} conceptualizes AS as a fixed, input-independent bias term within the attention output.  
In contrast to \textbf{\textit{Softmax Limitations and No-op Theory}} and \textbf{\textit{Outlier Circuits}}, which examine AS from its mathematical origin and numerical mechanism, this mechanistic perspective interprets AS's functional role as a bias operating at the attention-output level.

\begin{figure}[t]
    \centering
    % \vspace{-5mm}
    \includegraphics[width=1\textwidth]{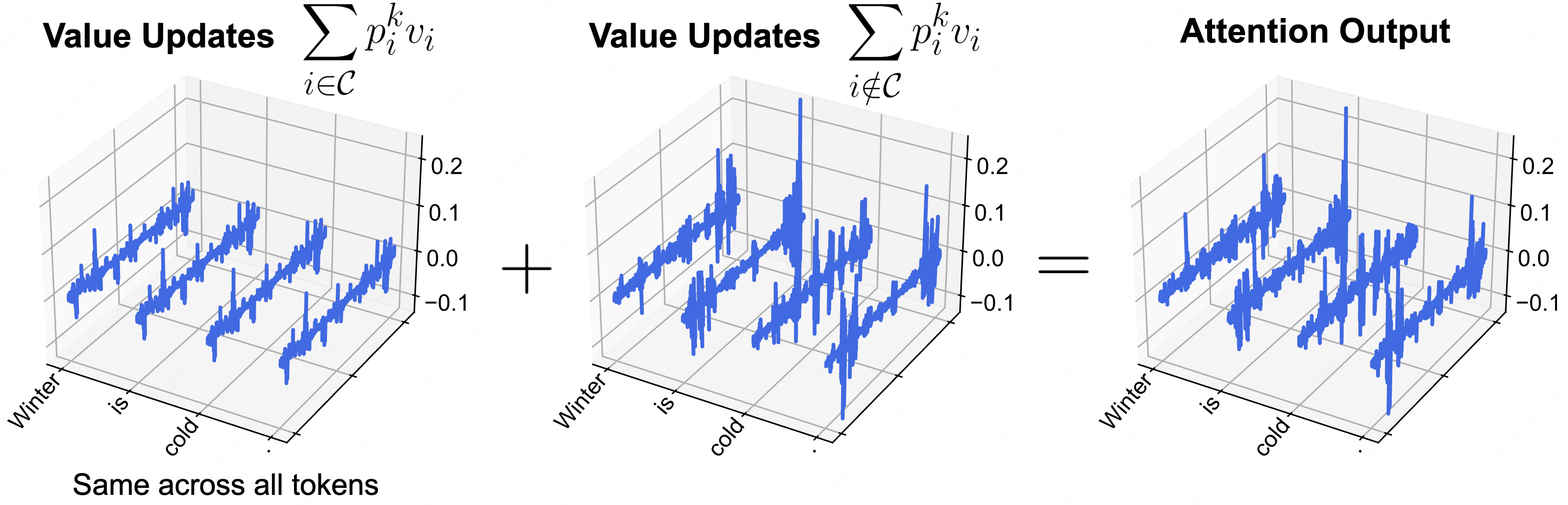}
    \caption{Value updates from AS tokens are essentially the same. The figure is adapted from \cite{sun2024massive}.}
    \label{fig:massive_attention_decomp}
\end{figure}

Following \textit{\textit{Massive Activations}} \cite{sun2024massive}, the attention output for a query token \(k\) can be decomposed as:
\begin{equation}
\mathrm{Attention}(Q,K,V)_k = \sum_{i \leq k} p_i^k v_i = \underbrace{\sum_{i \in \mathcal{C}} p_i^k v_i}_{\text{token set } \mathcal{C}} + \underbrace{\sum_{i \notin \mathcal{C}} p_i^k v_i}_{\text{other tokens}},
\end{equation}
where \(p_i^k\) is the attention weight from token \(k\) to token \(i\), and \(v_i\) is the value state of token \(i\). The set \(\mathcal{C}\) contains the tokens that have \textit{Massive Activations} (i.e., AS tokens). As shown in Figure~\ref{fig:massive_attention_decomp}, the value updates from \(\mathcal{C}\) are nearly identical across all query positions and across different inputs, thus acting as a constant bias term added to every token’s attention output \cite{sun2024massive}.

Crucially, providing an explicit attention bias eliminates the need for this implicit mechanism. \textit{\textit{Massive Activations}} \cite{sun2024massive} augments attention with learnable key and value biases \(\mathbf{k}', \mathbf{v}' \in \mathbb{R}^d\):
\begin{equation}
\mathrm{Attention}(Q,K,V;\mathbf{k}',\mathbf{v}') = \mathrm{softmax}\left(\frac{Q[K^\top \ \mathbf{k}']}{\sqrt{d}}\right)\left[\begin{array}{c} V \\ \mathbf{v}'^{\top} \end{array}\right].
\end{equation}
When a GPT‑2 model is trained with this explicit bias, \textit{Massive Activations} disappear, and the AS phenomenon is correspondingly eliminated. This confirms that AS is a manifestation of an implicit bias learned to cope with the Softmax constraint.

\subsubsection{Supporting Evidence}

\begin{figure}[t]
    \centering
    \includegraphics[width=1\textwidth]{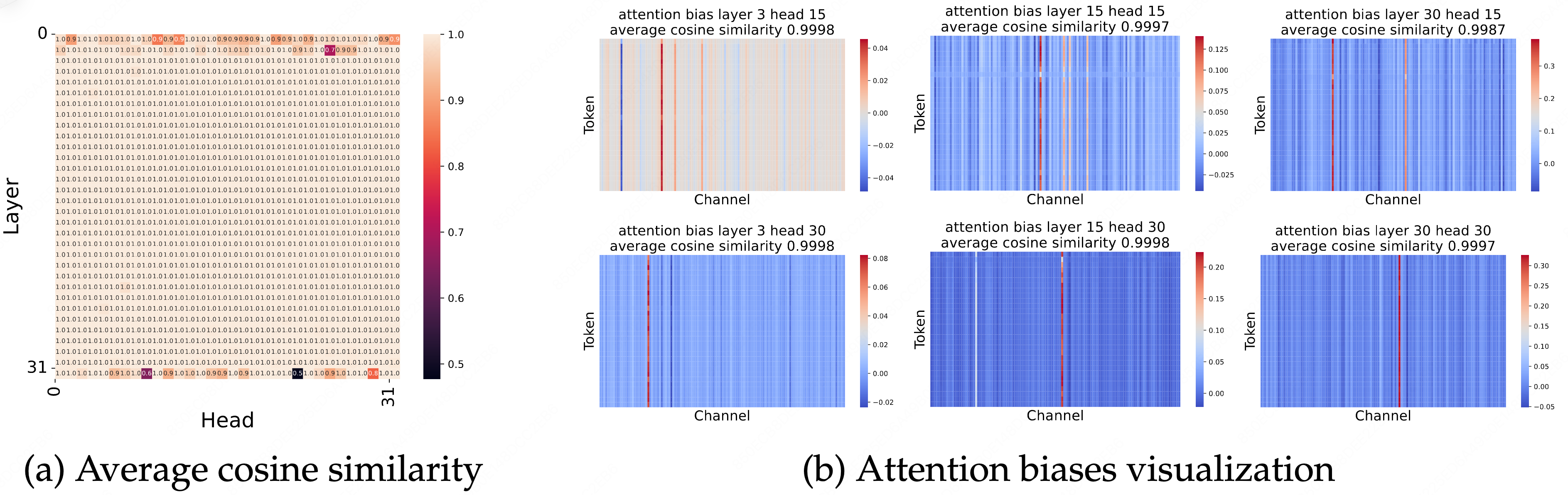}
\caption{
    (a) depicts the average cosine similarity of \( \sum_{i \in S} p_i^t v_i \) across all tokens for each head on LLaMA2-7B, showing that the values are consistently close to one across different tokens. 
    (b) visualizes the attention biases for several example heads, where \( \sum_{i \in S} p_i^t v_i \) remains nearly constant. The figure is adapted from \cite{su2025kvsink}}
    \label{fig:kvsink_attention_bias}.
\end{figure}

\paragraph{Observational Evidence.} 
\textit{\textit{Massive Activations}} \cite{sun2024massive} visually demonstrate the presence of implicit attention biases, as discussed previously. 
In addition, \textit{KVSink} \cite{su2025kvsink} further corroborates this phenomenon through both observational and quantitative analyses. 
To rigorously evaluate the effect, \textit{KVSink} computes the average cosine similarity of $\sum_{i \in S} p_i^t v_i$ across all tokens for each attention head. 
As shown in Figure~\ref{fig:kvsink_attention_bias}, for every head, $\sum_{i \in S} p_i^t v_i$ remains highly consistent across tokens whenever attention sinks emerge, providing strong evidence that these activations serve as stable, input-independent attention biases, as illustrated.

\paragraph{Causal Evidence.}  
Beyond \textit{\textit{Massive Activations}}, several studies provide causal evidence that AS functions as an implicit attention bias, through interventions that introduce explicit biases or directly manipulate the sink token's attention.  
\textit{When Attention Sink Emerges} \cite{gu2025attention} introduces learnable key biases that absorb attention, effectively shifting the sink from the first token to the bias position. \textit{Systematic Outliers} \cite{an2025systematic} demonstrates that attention outliers act as implicit context-aware scaling factors. Introducing an explicit context-aware scaling factor \(S_c(x)\), which dynamically adjusts attention weights, prevents the formation of systematic outliers and eliminates AS, confirming the implicit scaling role.  
These complementary causal interventions collectively confirm that AS serves as an implicit attention bias. Employing \textit{\textbf{Learnable Attention Bias}} can effectively mitigate AS, with a more detailed discussion provided in \S~\ref{sec_5_3_Learnable_Attention_Bias}.

\subsubsection{Discussion and Insights}

\textbf{Advantages.}  
The \textit{\textbf{Implicit Attention Bias}} framework provides a concise, unified explanation for AS: the model effectively injects a fixed, input-independent bias into the attention output. This perspective links AS directly to the Softmax sum-to-one constraint, explaining why sink tokens receive disproportionately high attention despite minimal contribution to outputs. Causal interventions confirm that this implicit bias is both sufficient to account for AS and can be replaced by explicit mechanisms. The phenomenon is consistently observed across LLMs, ViTs, and multimodal tasks, highlighting its broad applicability.

\textbf{Limitations.}  
Despite its strengths, two key issues remain. First, the training dynamics that give rise to \textit{Massive Activations} and AS as implicit biases are not yet formalized, leaving the convergence and evolution mechanisms unclear. Second, while multiple forms of implicit bias have been identified, their relationships remain fragmented, and it is unknown whether more general or more effective forms exist.

\textbf{Future Directions.}  
Future research should formalize the emergence of implicit attention bias during pre-training, linking Softmax constraints with the dynamics of AS. Developing a unified theoretical framework that integrates diverse explicit and implicit biases would deepen mechanistic understanding and inform architectural design. Additionally, exploring how implicit biases can be exploited to enhance inference efficiency or interpretability offers a promising avenue for practical impact.% Zunhai

\subsection{Geometric Anchoring}
\label{sec_4_4_Geometric_Anchoring}

\begin{tcolorbox}[takeawaysbox]
{\large \textbf{\textcolor{TikTokPink}{\textit{Key Takeaways:}}}}
\begin{enumerate}[leftmargin=*, label=\arabic*)]
    \item \textbf{Core Concepts:}  
    \textit{\textbf{Geometric Anchoring}} conceptualizes AS as a set of stable geometric reference points. Sink tokens act as geometric anchors, structuring the high-dimensional representation space and guiding other tokens through diverse geometric interactions.

    \item \textbf{Supporting Evidence:}  
    Empirical analyses show that sink tokens occupy distinct positional vectors, while other tokens converge toward these anchors. This demonstrates that sink tokens serve as stable reference points that shape attention allocation and downstream computations.

    \item \textbf{Discussion and Insights:}  
    The \textit{\textbf{Geometric Anchoring}} framework offers a principled perspective on AS and informs practical strategies for model interpretability and control. Its limitations include reliance on primarily correlational evidence, computational costs associated with geometric computations, and an incomplete understanding of why specific tokens become anchors. Future work should formalize the formation and stability of anchors during pre-training, develop more efficient geometric measures for detection and utilization, and explore their integration to enhance inference efficiency, model robustness, and representational fidelity.
\end{enumerate}
\end{tcolorbox}

\begin{figure}[t]
    \centering
    \includegraphics[width=1\textwidth]{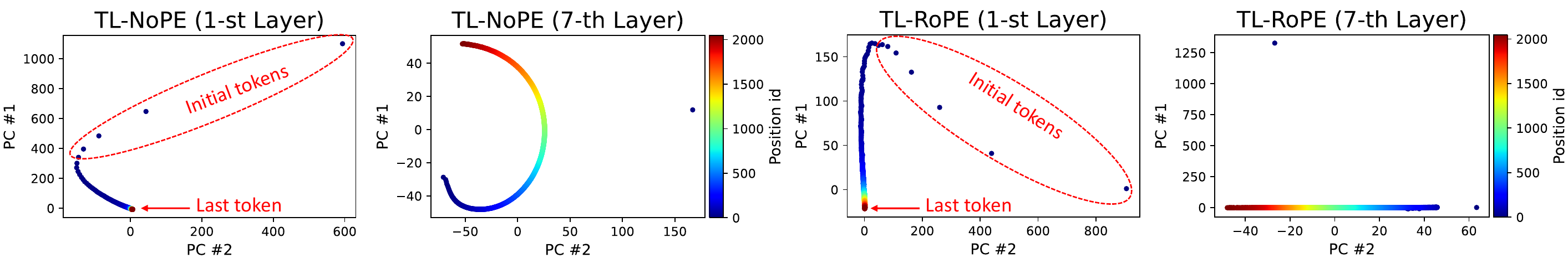}
    \caption{PCA visualization of positional vectors. After the first layer, only the initial tokens (e.g., positions 1--4) exhibit distinct positional vectors, whereas later tokens converge to similar representations. The figure is adapted from \cite{dong2024exploring}.}
    \label{fig:positional_vector}
\end{figure}
\subsubsection{Core Concepts}

A distinct line of research interprets the role of AS in representation spaces, viewing it as a geometric phenomenon in high-dimensional embeddings. Rather than attributing sink tokens to Softmax artifacts or activation outliers, the \textit{\textbf{Geometric Anchoring}} perspective conceptualizes them as stable reference points that systematically structure the representational geometry of all other tokens. Several studies have formalized this concept using explicit geometric frameworks and analyses.

\begin{itemize}
    % \item \textbf{Reference Frames} \cite{ruscio2025you}: AS is conceptualized as the establishment of a reference frame in the representation space. The sink token serves as the origin or a fixed direction, and the hidden state of any other token is represented as an additive offset:
    % \begin{equation}
    %     \mathbf{h}_t = \mathbf{h}_s + \boldsymbol{\delta}_t,
    % \end{equation}
    % where \(\mathbf{h}_s\) denotes the representation of the sink token, and \(\boldsymbol{\delta}_t\) is the displacement vector of token \(t\) relative to the sink. Different types of reference frames (centralized, distributed, bidirectional) yield distinct AS patterns.

    \item \textbf{Positional Vector Decomposition} \cite{dong2024exploring}: The study suggests that each hidden state is decomposed into a positional component and a semantic component:
        \begin{equation}
            \mathbf{h}_{l,t}^s = \mathbf{p}_{l,t} + \mathbf{c}_{l,t}^s,
        \end{equation}
        where \(\mathbf{p}_{l,t}\) is the positional vector at layer \(l\) for token position \(t\), and \(\mathbf{c}_{l,t}^s\) represents the semantic content. The positional vector of the sink token, \(\mathbf{p}_{l,1}\), acts as a geometric anchor that guides the formation of positional vectors for subsequent tokens, thereby inducing AS .

    \item \textbf{OrthoRank} \cite{shin2025orthorank}: In this study, token importance is evaluated based on orthogonality relative to the sink token:
    \begin{equation}
        \text{importance}(t) \propto 1 - |\cos(\mathbf{h}_t, \mathbf{h}_s)|,
    \end{equation}
    where \(\cos(\mathbf{h}_t, \mathbf{h}_s)\) denotes the cosine similarity between token \(t\) and the sink. Tokens nearly orthogonal to the sink are considered more informative, directly leveraging the sink as a geometric reference point.

    \item \textbf{KeyDiff} \cite{park2025keydiff}: This study suggests that sink tokens exhibit a distinctive geometric property in the key space: their key vectors \(\mathbf{k}_s\) have near-zero cosine similarity with the mean key vector \(\bar{\mathbf{k}}\):
    \begin{equation}
        \cos(\mathbf{k}_s, \bar{\mathbf{k}}) \approx 0.
    \end{equation}
    This identifies AS tokens as geometric outliers in the key space, which can be leveraged for efficient KV cache management and selective attention.
\end{itemize}

Beyond the geometric formulations discussed above, several additional studies exploit the notion of AS as a stable reference point. \textit{Anchor Attention} \cite{zhang2025anchor} demonstrates that in code generation models, attention distributions are extremely sparse—with the top two attention weights often exceeding 80\%—and concentrate on structural anchor points such as newline tokens. \textit{One Token Is Enough} \cite{zhang2026one} introduces a dedicated sink token serving as a position-independent structural anchor. \textit{CTR-Sink} \cite{li2025ctr} constructs artificial sink tokens as aggregation centers within user behavior sequences. \textit{OmniSparse} \cite{chen2025omnisparse} leverages early frames or start-of-text tokens as memory anchors. \textit{MagicPIG} \cite{chen2025magicpig} utilizes the near-static keys of sink tokens to provide a stable reference. Collectively, these works reinforce the broader principle that ASs act as reliable geometric anchors, organizing the representation space and guiding computational flow.

\subsubsection{Supporting Evidence}

\paragraph{Observational Evidence.} 
A growing body of empirical work demonstrates that AS consistently function as stable geometric anchors. These studies reveal that sink tokens not only maintain distinct positional or key vectors but also systematically influence the representations of other tokens, effectively acting as fixed reference points that shape attention and downstream computations.

\begin{figure}[t!]
    \centering
    \begin{minipage}{.24\linewidth}
        \centering
        \includegraphics[width=\linewidth]{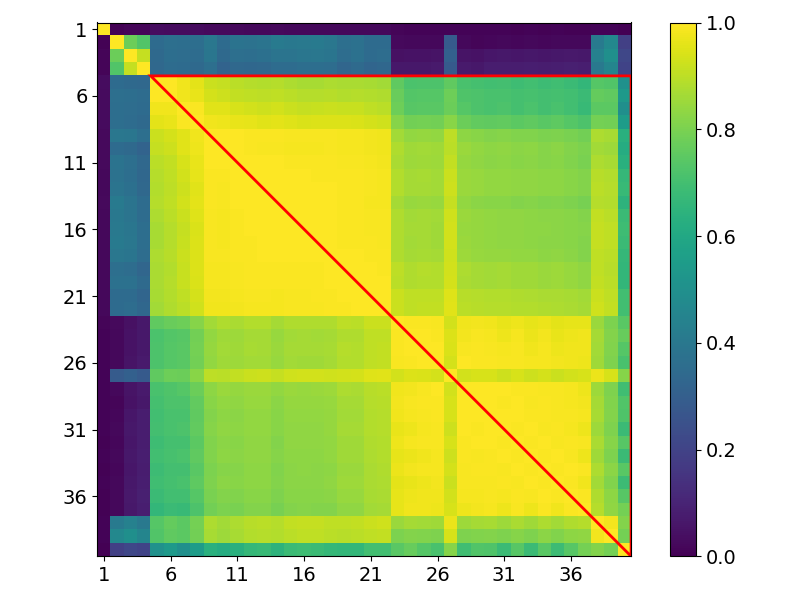}\subcaption{Llama-2-13B ($\bar{h}_{0}$).}
    \end{minipage}
    \begin{minipage}{.24\linewidth}
        \centering
        \includegraphics[width=\linewidth]{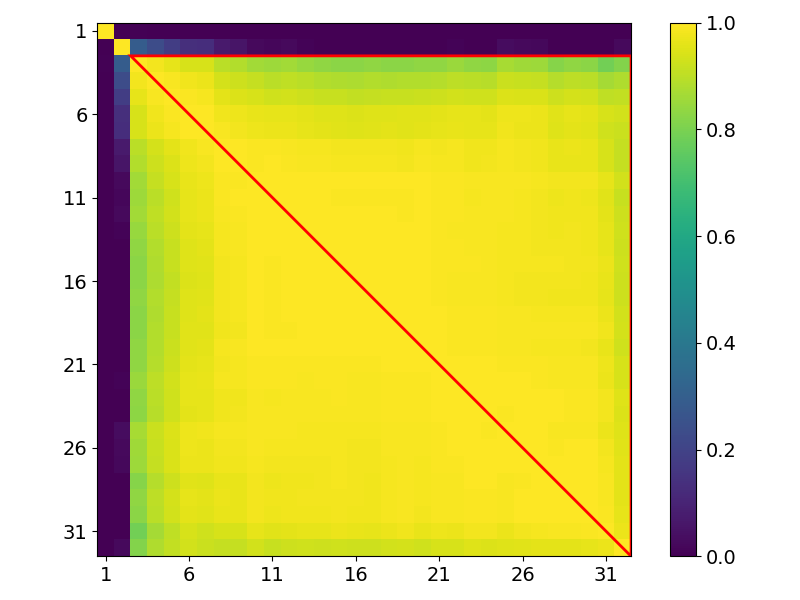}\subcaption{Mistral-7B ($\bar{h}_{0}$).}
    \end{minipage}
    \begin{minipage}{.24\linewidth}
        \centering
        \includegraphics[width=\linewidth]{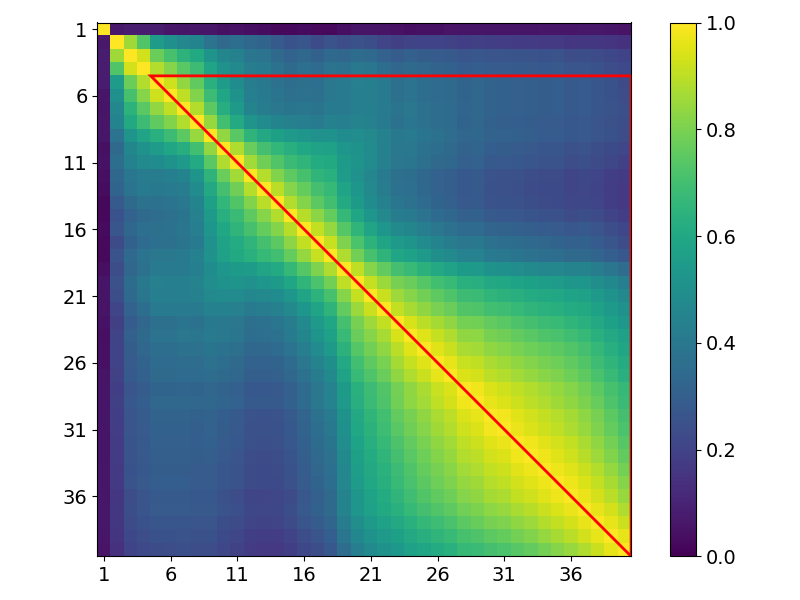}\subcaption{Llama-2-13B ($\bar{h}_{50}$).}
    \end{minipage}
    \begin{minipage}{.24\linewidth}
        \centering
        \includegraphics[width=\linewidth]{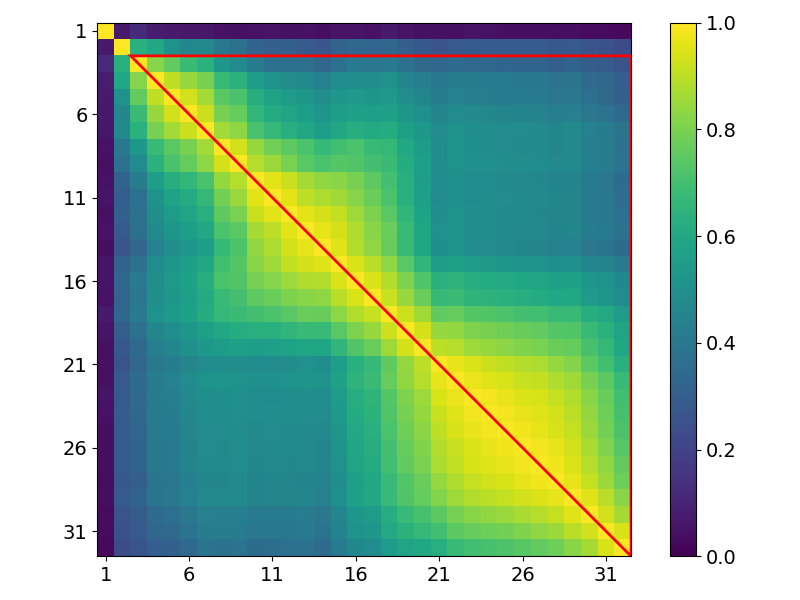}\subcaption{Mistral-7B ($\bar{h}_{50}$).}
    \end{minipage}
    \caption{Cosine similarity of normalized hidden states across layers. (a)-(b) Sink token maintains high similarity even between distant layers. (c)-(d) Another token shows similarity only between adjacent layers. The red boundary indicates layers after \(l_{\text{sink}}\). These results highlight the static geometric nature of the sink token. The figure is adapted from \cite{shin2025orthorank}.
}
\vspace{-2mm}
    \label{fig:similarity_sink}  
\end{figure}

\begin{itemize}
    \item \textbf{Decomposed Positional Vector} \cite{dong2024exploring}: Using a mean-based decomposition followed by PCA visualization, the study reveals that, as shown in Figure \ref{fig:positional_vector}, after the first layer only the initial toke (e.g., positions 1--4) exhibit distinct positional vectors, whereas later tokens converge to similar representations. As layers deepen, more tokens gradually develop distinct positional vectors. This confirms the anchoring role of the sink token's positional vector. Correspondingly, attention maps show that the sink token receives disproportionately high attention, and this effect strongly correlates with the distinctness of its positional vector. When the input length exceeds the model's training window, positional vectors become out-of-distribution (OOD), causing the AS to vanish and perplexity to rise sharply.
    
    \item \textbf{OrthoRank} \cite{shin2025orthorank}: By computing the cosine similarity between the normalized hidden states of the sink token and other tokens across layers, the authors observe that after the layer where AS first emerges, the similarity of other tokens steadily increases (as shown in Figure \ref{fig:similarity_sink}). Meanwhile, the sink token's own normalized hidden states remain nearly unchanged, with cosine similarity close to one, indicating that other tokens geometrically move toward the sink token, which functions as a static anchor. Empirically, tokens with higher orthogonality to the sink are more informative.
    
    \item \textbf{KeyDiff} \cite{park2025keydiff}: Analyzing pairwise cosine similarity among keys in the KV cache reveals a strong negative correlation: keys that are geometrically distinctive (low average similarity to others) consistently receive higher attention scores. This pattern holds across layers and heads, with an average Spearman correlation of approximately 0.94. In particular, sink tokens have near-zero cosine similarity to the mean key vector \(\bar{\mathbf{k}}\), i.e., \(\cos(\mathbf{k}_s, \bar{\mathbf{k}}) \approx 0\), marking them as geometric outliers in the key space.
    
    % \item \textbf{Reference Frames} \cite{ruscio2025you}: Although primarily theoretical, the work identifies three types of reference frames—centralized, distributed, and bidirectional—through geometric analysis of attention patterns. These frames correspond directly to distinct AS behaviors, reinforcing that the geometric role of sink tokens is systematic and not incidental.
\end{itemize}

\vspace{-5mm}
\subsubsection{Discussion and Insights}

\textbf{Advantages.}  
The \textit{\textbf{Geometric Anchoring}} perspective elevates AS from an emergent pattern to an interpretable, stable structure within high-dimensional representation spaces. This geometric viewpoint informs practical strategies: positional vector replacement can extend effective context windows, orthogonality-based pruning enhances KV cache efficiency, and key similarity–based eviction often outperforms conventional attention-score–based methods. Together, these insights illustrate how leveraging geometric structure can yield tangible improvements in model efficiency and performance.

\textbf{Limitations.}  
Despite its explanatory power, the \textit{\textbf{Geometric Anchoring}} framework has several notable limitations. First, most supporting evidence is correlational, with few direct causal interventions, leaving key mechanistic claims unvalidated. Second, the framework does not fully explain why particular tokens emerge as geometric anchors or how these anchors interact with broader model dynamics.

\textbf{Future Directions.}  
Future research should pursue several avenues. First, formalizing the emergence and stability of geometric anchors during pre-training could yield mechanistic insights. Second, developing more efficient methods for detecting and leveraging anchors would reduce computational overhead without sacrificing effectiveness. Third, systematically integrating geometric anchors into model optimization and inference through anchor-guided pruning, KV cache management, or context extension remains largely unexplored and offers significant potential for practical impact.

% Qingyao

% \subsection{Structural Bias}
% \label{sec_4_5_Structural_Bias}
% \input{Attention_Sink_in_Transformers/4_Interpretation/5_Structural_Bias}% Xiong Jing

% \subsection{Anti-Overmixing Mechanism}
% \label{sec_4_6_Anti_Overmixing}
% \input{Attention_Sink_in_Transformers/4_Interpretation/6_Anti-Overmixing_Mechanism}% Hengyuan

\subsection{Other Mechanistic Interpretations}
\label{sec_4_5_Additional_Mechanistic_Interpretations of Attention Sink}

\begin{figure}[t]
\centering
\includegraphics[width=1\columnwidth]{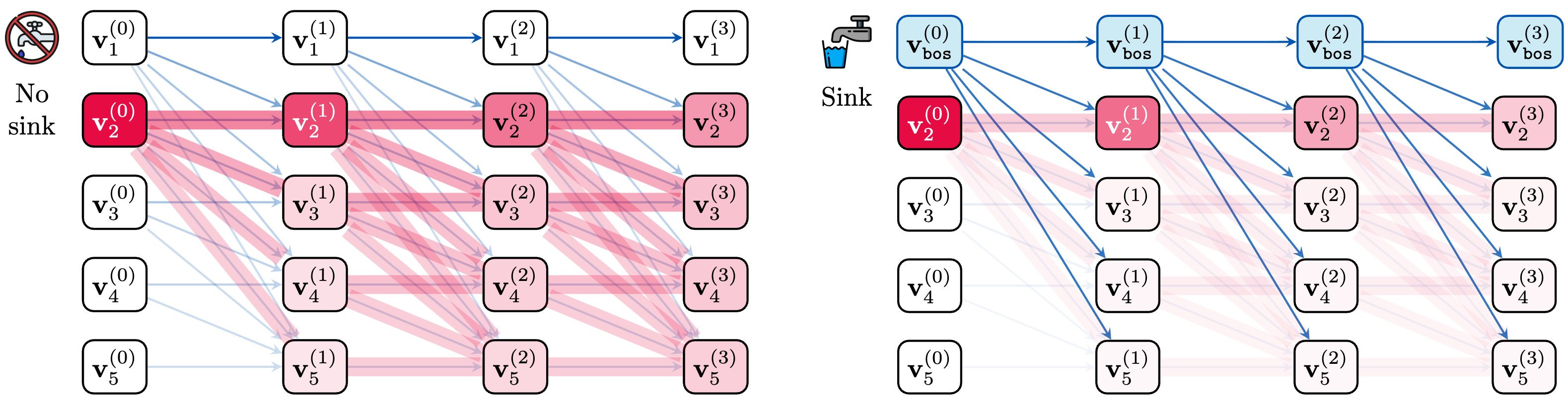}
\caption{The presence of AS modulates information flow between tokens, making Transformer models more robust to perturbations in input prompts. This figure illustrates how a perturbation in the second token's input representation (highlighted in red) propagates to other token embeddings throughout the model, both without (left) and with (right) a sink token (e.g., \texttt{⟨BOS⟩}). The sink token diverts attention away from other tokens, limiting the spread of the perturbed information and resulting in more stable embeddings. Adapted from \cite{barbero2025llms}.}
\vspace{-3mm}
\label{fig:anti-over-mixing}
\end{figure}
Beyond the previously discussed perspectives, several additional theories offer complementary insights into the emergence and dynamics of AS. Here, we provide a concise summary of these viewpoints. We then present a consolidated overview of AS interpretations across five analytical levels, offering a comprehensive, high-level perspective on the relationships and distinctions among existing explanations.

\begin{figure}[t]
\centering
\includegraphics[width=0.9\columnwidth]{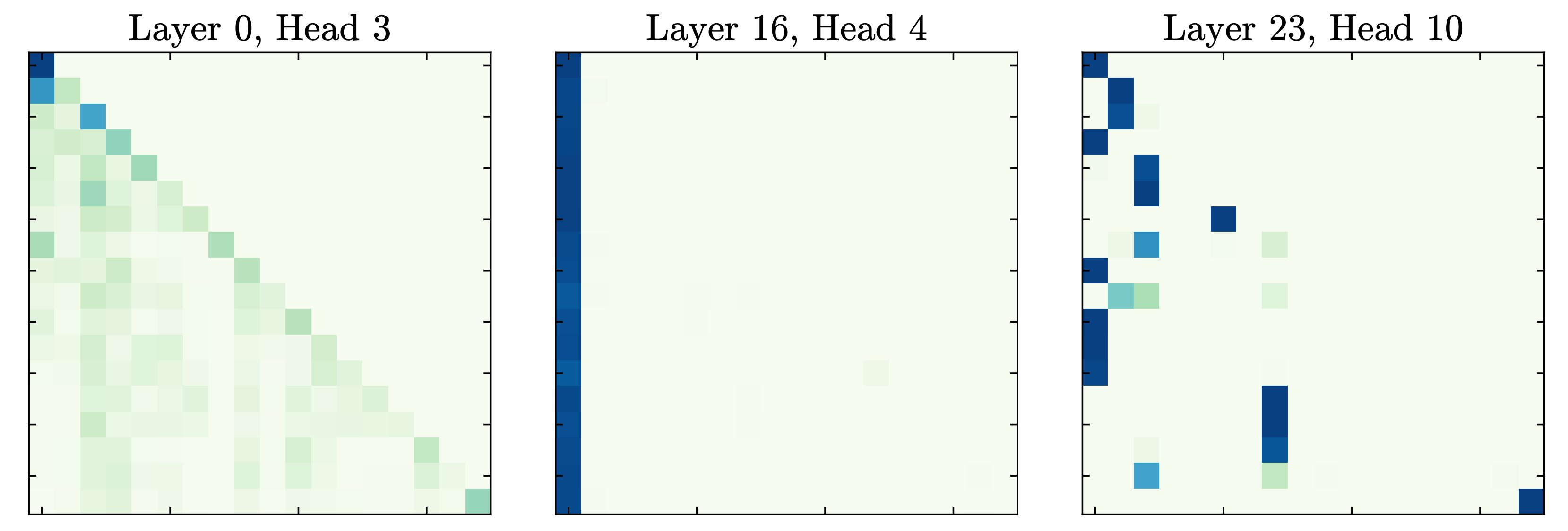}
\caption{Evolution of attention patterns in Pythia 410M, highlighting representative heads at layers 0, 16, and 23. Early layers exhibit diffuse attention that facilitates broad information mixing. Middle layers display sink patterns that restrict mixing, while late layers show sharp positional patterns enabling selective refinement. Adapted from \cite{queipo2026attention}.}
\label{fig:mix-compress-refine}
\end{figure}

\begin{itemize}[leftmargin=*, label=\textbullet]

    \item \textbf{Structural Bias.}  
    Inherent architectural biases significantly shape AS. Two primary sources are causal masking and RoPE. Causal masking confers early tokens a cumulative visibility advantage, as the first token is observable by all subsequent queries. This asymmetry systematically biases attention toward the sequence’s beginning, directly inducing AS on initial tokens \cite{wu2025emergence, salvatore2025lost}. RoPE encodes relative positions through rotations, introducing distance-dependent decay that concentrates attention on nearby positions. When this decay is excessively strong or misaligned with the underlying data structure, it produces activation outliers that distort attention distributions, thereby generating AS \cite{zhang2026drives, xiong2025dope, chen2024rotary}. 
    % These structural biases are intrinsic to the Transformer architecture and consistently shape attention allocation patterns.

    \item \textbf{Anti-Overmixing Theory.}  
    LLMs attend to the first token because it acts as a sink preventing excessive information mixing across layers. 
    In the absence of a sink, token representations would quickly converge, resulting in representational collapse and a loss of contextual distinctiveness, as illustrated in Figure~\ref{fig:anti-over-mixing}.
    The first token, visible to all subsequent tokens, anchors the residual stream, allowing diverse token representations to be maintained even in deep layers. AS thus emerges as a structural adaptation essential for preserving expressive power in autoregressive Transformers \cite{barbero2025llms}.

    \item \textbf{Spectral-Energy Association Theory.}  
    AS is linked to the spectral properties of hidden state dynamics. The first token’s hidden state quickly acquires a large norm, acting as a ``dark signal'' that dominates the residual stream and absorbs most attention energy. This spectral dominance coerces other tokens to align with the sink token’s direction, compressing the representational manifold. AS arises as a byproduct of low-rank spectral dynamics, trading token distinctions for stable information propagation \cite{cancedda2024spectral}.

    \item \textbf{Active-Dormant Attention Theory.}  
    AS emerges via mutual reinforcement among attention heads. In trained LLMs, a subset of heads becomes ``active'' sinks consistently receiving high attention, while others remain ``dormant.'' Active heads produce large key norms and small value norms, attracting queries while minimally contributing to the residual output. This separation is reinforced by training dynamics: heads that initially become sinks receive positive gradient reinforcement, stabilizing their specialization and causing a few heads to dominate attention absorption \cite{guo2024active}.

    \item \textbf{Mix-Compress-Refine Theory.}  
    LLMs process information through three sequential phases: broad information mixing in early layers, compressed computation in middle layers dominated by large activations, and selective refinement in later layers, as illustrated in Figure~\ref{fig:mix-compress-refine}.
    AS arises during the compression phase, where attention concentrates on a small set of sink tokens to manage bandwidth and prevent over-mixing. This phase features a sharp reduction in representational entropy as contextual information is condensed into compact anchor tokens before refinement \cite{queipo2026attention}.

    \item \textbf{Outlier-Driven Rescaling Theory.}  
    AS, along with residual sinks (persistent large activation values in fixed feature dimensions), plays a functional role. In combination with normalization layers (Softmax and RMSNorm), these outliers act as implicit rescaling factors that stabilize training and enhance generalization. They modulate contributions from non-outlier components rather than directly driving outputs. Removing or clipping them without compensatory adjustments impairs performance, while replacing them with learnable parameters or gating preserves stabilization and can improve downstream accuracy \cite{qiu2026unifiedviewattentionresidual}.

\end{itemize}

\paragraph{Summary of Interpretations of Attention Sink.}  
We provide a consolidated overview of AS interpretations across five analytical levels, organized by core perspective, corresponding theories, and central issues. This synthesis emphasizes that AS arises from the interplay of mathematical constraints, training dynamics, numerical mechanisms, geometric structures, and functional roles, offering a unified framework for understanding its emergence, persistence, and impact across Transformer models.

\begin{itemize}
    \item \textbf{Mathematical Origin} (why AS inevitably emerges):
    \begin{itemize}
        \item \textit{\textbf{Softmax Limitations and No-op Theory}} (\S~\ref{sec_4_1_Softmax_Limitations}): The Softmax sum-to-one constraint forces attention onto uninformative tokens when no meaningful key exists, as the mechanism lacks a natural ``null'' option.
        \item \textit{\textbf{Structural Bias}} (\S~\ref{sec_4_5_Additional_Mechanistic_Interpretations of Attention Sink}): Causal masking grants early tokens a cumulative visibility advantage, and RoPE introduces distance-dependent decay that can produce activation outliers; both mechanisms inherently bias attention toward sink tokens.
    \end{itemize}

    \item \textbf{Training Dynamics} (how AS emerges during training):
    \begin{itemize}
        \item \textit{\textbf{Active-Dormant Attention Theory}} (\S~\ref{sec_4_5_Additional_Mechanistic_Interpretations of Attention Sink}): A subset of heads act as active sinks, characterized by large key norms and small value norms, reinforced by positive gradient feedback that stabilizes their specialization.
    \end{itemize}
    
    \item \textbf{Numerical Mechanism} (the numerical foundation of AS):
    \begin{itemize}
        \item \textit{\textbf{Outlier Circuits}} (\S~\ref{sec_4_2_Outliers_Circuits}): Weight, activation, and attention outliers form interconnected circuit-like pathways that stabilize AS.
        \item \textit{\textbf{Outlier-Driven Rescaling Theory}} (\S~\ref{sec_4_5_Additional_Mechanistic_Interpretations of Attention Sink}): Outliers, together with residual sinks and normalization layers, act as implicit rescaling factors.
        \item \textit{\textbf{Mix-Compress-Refine Theory}} (\S~\ref{sec_4_5_Additional_Mechanistic_Interpretations of Attention Sink}): AS emerges during a middle compression phase where attention condenses contextual information into sink tokens, followed by selective refinement.
    \end{itemize}

    \item \textbf{Geometric Structure} (the role of AS in representation space):
    \begin{itemize}
        \item \textit{\textbf{Geometric Anchoring}} (\S~\ref{sec_4_4_Geometric_Anchoring}): Sink tokens serve as stable reference points that systematically organize the representational geometry of all other tokens.
        \item \textit{\textbf{Anti-Overmixing Theory}} (\S~\ref{sec_4_5_Additional_Mechanistic_Interpretations of Attention Sink}): The first token anchors the residual stream to prevent excessive information mixing across layers, thereby avoiding representational collapse.
        \item \textit{\textbf{Spectral-Energy Association}} (\S~\ref{sec_4_5_Additional_Mechanistic_Interpretations of Attention Sink}): The first token's hidden state becomes a large-norm ``dark signal'' that dominates spectral energy and compresses the representational manifold.
    \end{itemize}

    \item \textbf{Functional Role} (utility of AS for the model):
    \begin{itemize}
        \item \textit{\textbf{Implicit Attention Bias}} (\S~\ref{sec_4_3_Implicit_Attention_Bias}): AS acts as a fixed, input-independent bias term added to every token's attention output, since value updates from sink tokens are nearly identical across queries and inputs.
    \end{itemize}

\end{itemize}

\clearpage
\section{Strategic Mitigation of Attention Sink}
\label{sec_5_Strategic_Mitigation}

In this section, we examine strategies for mitigating AS, including \textit{\textbf{Gated Attention Mechanisms}} (\S~\ref{sec_5_1_Gated_Attention_Mechanisms}), \textit{\textbf{Modified Softmax Functions}} (\S~\ref{sec_5_2_Modified_Softmax_Functions}), \textit{\textbf{Learnable Attention Bias}} (\S~\ref{sec_5_3_Learnable_Attention_Bias}), \textit{\textbf{Pre-training Interventions}} (\S~\ref{sec_5_4_Pre-training_Interventions}), and other approaches (\S~\ref{sec_5_5_Additional_Techniques_for_Mitigating_Attention_Sink}).
Each method is presented with its core mechanistic formulation, a review of practical implementations, and concludes with our perspectives.

From a high-level perspective, these AS mitigating approaches can be divided into two categories. 
First, methods that provide explicit alternatives, such as \textit{\textbf{Gated Attention Mechanisms}} (\S \ref{sec_5_1_Gated_Attention_Mechanisms}) and \textit{\textbf{Learnable Attention Bias}} (\S \ref{sec_5_3_Learnable_Attention_Bias}), aim to replace implicit AS with learnable, controllable mechanisms. 
Second, methods that cut the causal chain, including \textit{\textbf{Modified Softmax Functions}} (\S \ref{sec_5_2_Modified_Softmax_Functions}) and \textit{\textbf{Pre-training Interventions}} (\S \ref{sec_5_4_Pre-training_Interventions}), seek to eliminate AS by addressing its root causes.
A comprehensive synthesis of all AS mitigating approaches is provided in \S \ref{sec_5_5_Additional_Techniques_for_Mitigating_Attention_Sink}.

\subsection{Gated Attention Mechanisms}
\label{sec_5_1_Gated_Attention_Mechanisms}

\begin{tcolorbox}[takeawaysbox]
{\large \textbf{\textcolor{TikTokPink}{\textit{Key Takeaways:}}}}
\begin{enumerate}[leftmargin=*, label=\arabic*)]
    \item \textbf{Core Mechanism:}  
    \textit{\textbf{Gated Attention Mechanisms}} mitigate the no-op behavior by introducing a learnable gate that directly suppresses attention outputs, decoupling it from extreme softmax logits and breaking the self-reinforcing cycle of AS.

    \item \textbf{Practical Approaches:}  
    Two primary gating strategies have been proposed: output gating applied after SDPA using query-dependent scalar gates, and value-state gating that modulates value representations prior to attention weighting.

\item \textbf{Discussion and Insights:}  
    Gated attention effectively removes AS, enhances training stability, and supports quantization. 
    However, it faces four challenges: training from scratch, non-negligible parameter overhead, poorly understood training dynamics, and lack of standardized evaluations. 
    Future research should focus on lightweight post-hoc injection, parameter-efficient gate designs, elucidating gate evolution dynamics, and establishing unified benchmarks.
\end{enumerate}
\end{tcolorbox}

\subsubsection{Core Mechanism}

\begin{wrapfigure}{r}{0.25\textwidth} 
\vspace{-10mm}
    \centering
    \includegraphics[width=\linewidth]{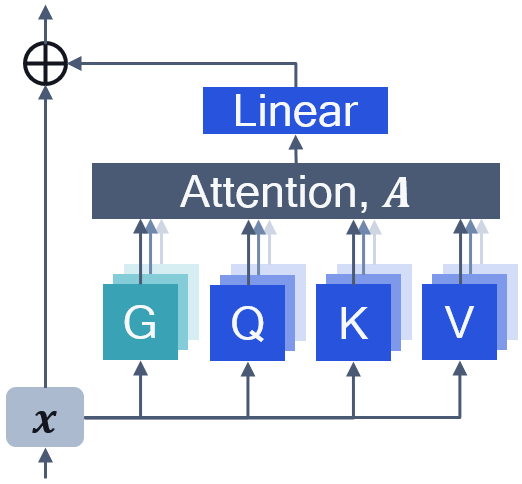}  
    \vspace{-7mm}
    \caption{A schematic illustration of gated attention. Adapted from \cite{bondarenko2023quantizable}.}
    \label{fig:quantizable_ga}
    \vspace{-5mm}
\end{wrapfigure}
\textit{\textbf{Gated Attention Mechanisms}} were first introduced in \textit{Quantizable Transformers}~\cite{bondarenko2023quantizable} as a direct response to the \textit{\textbf{Softmax Limitations and No-Op Theory}} (discussed in \S \ref{sec_4_1_Softmax_Limitations}). As established there, AS emerges because attention heads learn a no-op behavior to satisfy the Softmax sum-to-one constraint, forcing logits to extreme values. To break this self-reinforcing cycle, \textit{\textbf{Gated Attention Mechanisms}} provide an alternative pathway for implementing no-op updates.
The original formulation introduces a learnable gate that modulates the attention output in an element-wise manner, as illustrated in Figure~\ref{fig:quantizable_ga}:

\begin{equation}
\text{GatedAttention}(\mathbf{x}) := \sigma(G(\mathbf{x})) \odot \text{Softmax}\left(\frac{Q(\mathbf{x})K(\mathbf{x})^\top}{\sqrt{d_{\text{head}}}}\right) V(\mathbf{x}),
\end{equation}

where \(\sigma(\cdot)\) is the sigmoid function, \(G(\cdot)\) is a learnable projection that produces a gating vector of the same dimension as the attention output, and \(\odot\) denotes element-wise multiplication.

The key insight is that this mechanism decouples the no-op behavior from the attention logits. Instead of forcing the Softmax distribution onto sink tokens with tiny values to achieve a near-zero output, the head can simply learn to set the gating vector \(\sigma(G(\mathbf{x}))\) close to zero, directly suppressing its entire output element-wise. This eliminates the need for extreme logits and thus removes AS.
Empirically, the \textit{\textbf{Gated Attention Mechanisms}} proposed in \textit{Quantizable Transformers} significantly reduce activation outliers and eliminate AS, enabling robust low-bit quantization that would otherwise fail on standard Transformers.

\begin{figure}[t]
\centering
\includegraphics[width=1\columnwidth]{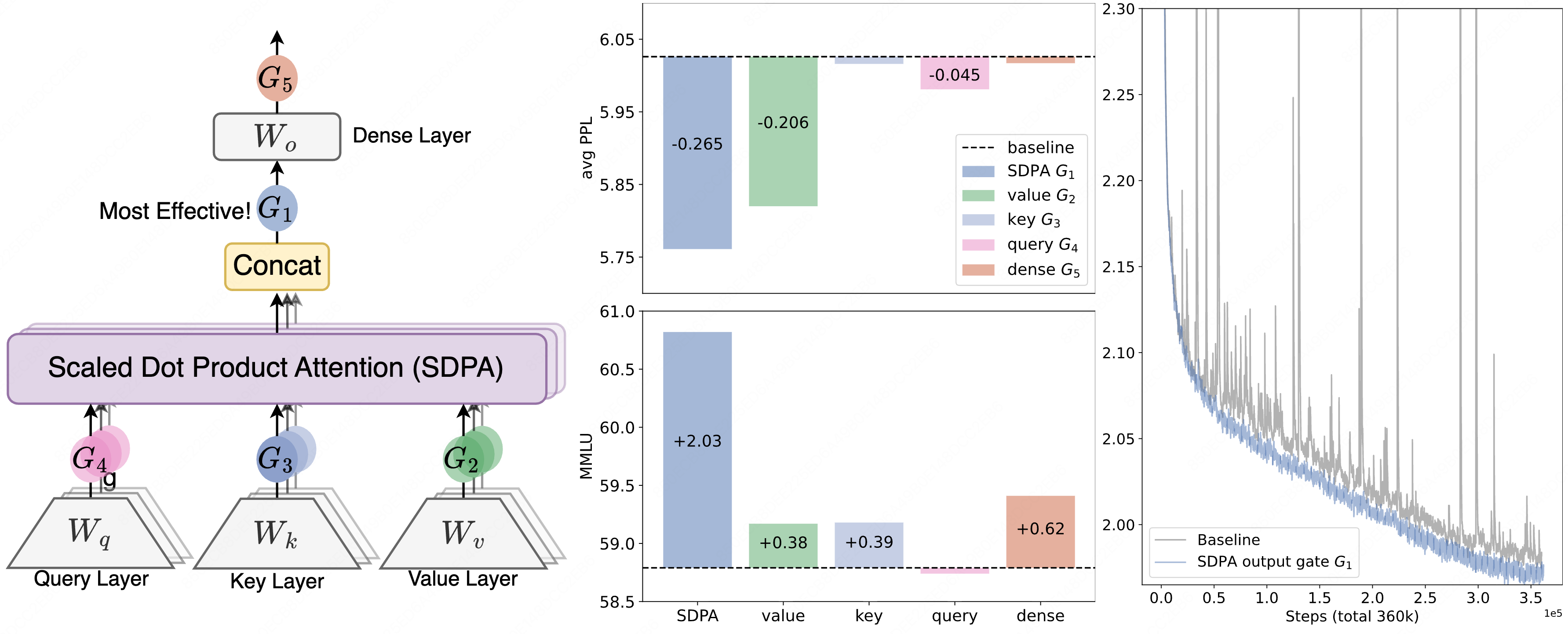}
\caption{Gating position exploration and performance comparison. \textbf{Left:} Investigated positions for applying gating operations. \textbf{Middle:} Performance of 15B MoE models. Gating after SDPA (G1) yields the best overall results; gating after the Value layer (G2) also improves performance, particularly in perplexity. \textbf{Right:} Training loss over 3.5T tokens for baseline vs. SDPA-gated 1.7B dense models. Gating reduces final loss and enhances stability by mitigating loss spikes, enabling higher learning rates and better scaling. The figure is adapted from~\cite{qiu2025gated}.}
\label{fig:qiu_gating_position}
\end{figure}
\begin{figure}[t]
\centering

\includegraphics[width=1\columnwidth]{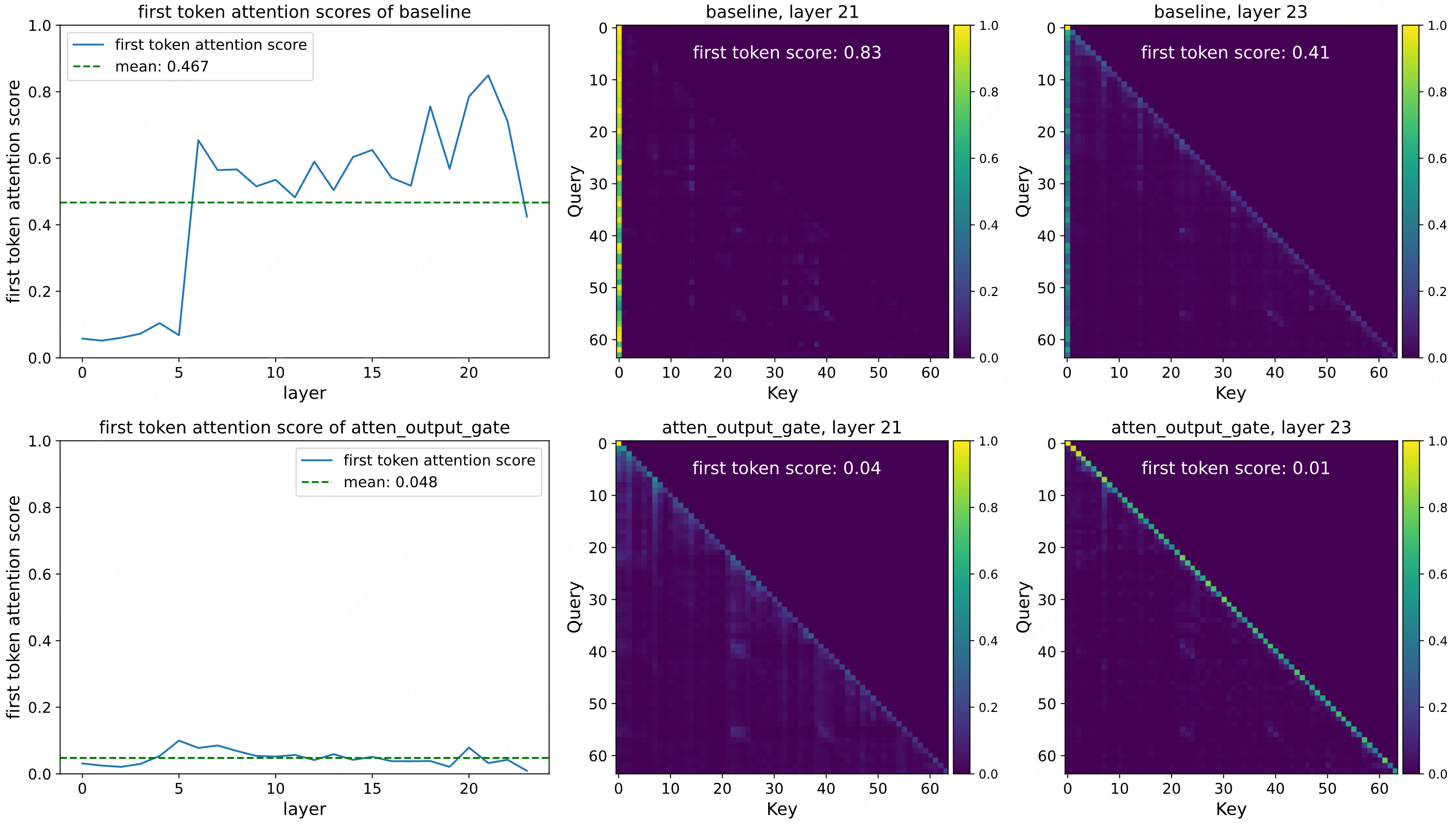}
\caption{AS mitigation with \textit{Gated Attention}. \textbf{Left:} Proportion of attention allocated to the initial token per layer. The baseline model devotes 46.7\% of attention scores (averaged across layers) to the first token; gating reduces this to 4.8\%. \textbf{Right:} Average attention map weights per head. In layer 21, the baseline AS (83\% on the first token) drops to 4\% with gating. 
% In the final output layer, gating amplifies the model's tendency to attend to individual tokens within the sequence. 
The figure is adapted from~\cite{qiu2025gated}.}

\label{fig:qiu_attention_sink}
\end{figure}
\begin{figure}[t]
\centering
\vspace{-4mm}
\includegraphics[width=1\columnwidth]{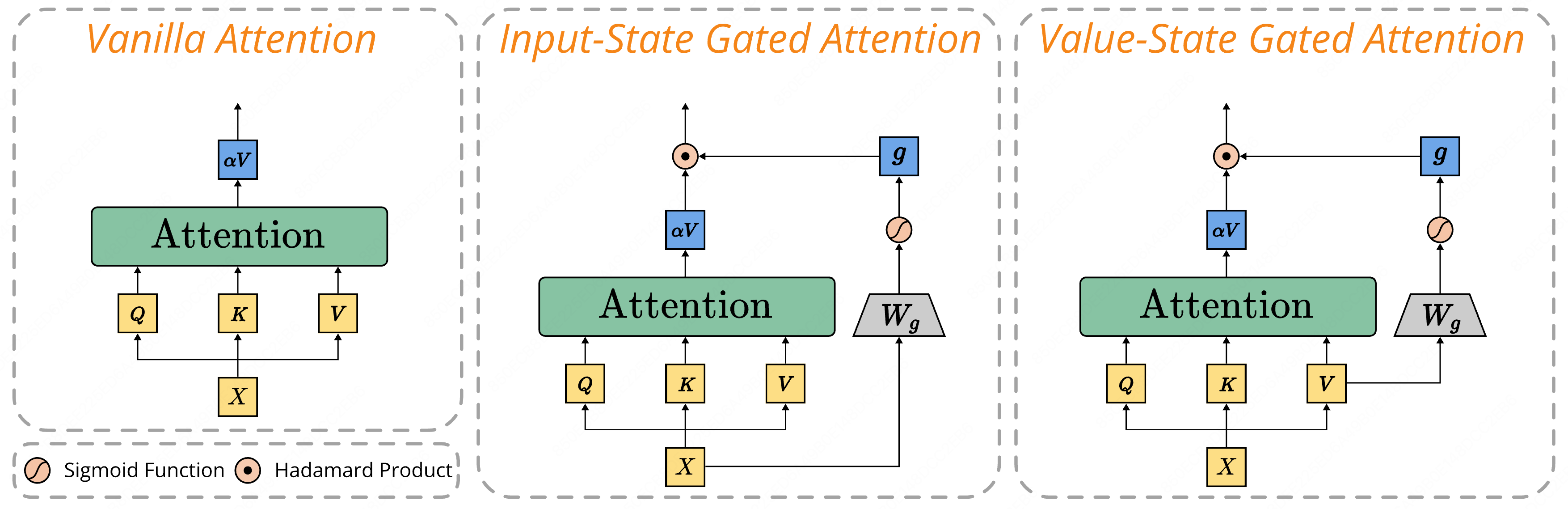}
\caption{Architecture of \textit{Value-State Gated Attention}. Unlike vanilla attention or input-state gated attention, VGA introduces a value-state gating mechanism to modulate the attention output. The figure is adapted from~\cite{bu2025value}.}
\label{fig:vga_architecture}
\end{figure}
\subsubsection{Practical Approaches}

Beyond the \textit{\textbf{Gated Attention Mechanisms}} proposed in \textit{Quantizable Transformers}~\cite{bondarenko2023quantizable}, several subsequent works have extended the paradigm to further suppress AS, improve training efficiency, or adapt the mechanism to large-scale language models. Two representative advances are reviewed below.

\paragraph{Non-linear and Sparse Gated Attention.}  
The work \textit{Gated Attention for Large Language Models}~\cite{qiu2025gated} systematically investigates the design space of gating-augmented softmax attention. Through a comprehensive comparison over 30 variants of 15B MoE models and 1.7B dense models trained on 3.5 trillion tokens, the authors identify that applying a gating mechanism after the Scaled Dot-Product Attention (SDPA) consistently improves performance, enhances training stability, and mitigates AS. The exploration covers two primary gating forms: (i) \textit{head-wise gating}, where a query-dependent scalar gate $\sigma(g_h(Q))$ modulates the entire SDPA output per head; and (ii) \textit{element-wise gating}, where a gate vector produced by $\sigma(G(\mathbf{x}))$ is applied element-wise to the SDPA output, offering finer control but introducing significantly more parameters. The head-wise scalar gate is found to achieve the best trade-off between effectiveness and efficiency.
The core formulation of this variant is:

\begin{equation}
\text{GatedAttention}(Q,K,V) = \sigma(g_h(Q)) \cdot \text{Softmax}\left(\frac{QK^\top}{\sqrt{d}}\right)V,
\label{eq:qiu_gated}
\end{equation}

where $\sigma(\cdot)$ is the sigmoid function, and $g_h(Q)$ is a head-specific, query-dependent scalar gate. This design introduces non-linearity upon the low-rank mapping in softmax attention and yields query-dependent sparse gating scores.
Empirically, the proposed \textit{\textbf{Gated Attention Mechanisms}} reduce training loss, enhance stability, and naturally mitigate massive activation and AS. Figure~\ref{fig:qiu_gating_position} illustrates the gating positions explored and the resulting performance gains. As shown in Figure~\ref{fig:qiu_attention_sink}, \textit{\textbf{Gated Attention Mechanisms}} drastically reduce the AS phenomenon across layers. Notably, this design has been adopted in production-scale models including \textit{Qwen3-Next} and \textit{Qwen3.5}~\cite{qiu2025gated,qwenai2026}.

\paragraph{Value-State Gated Attention.}  
An alternative direction is presented in \textit{Value-State Gated Attention}~\cite{bu2025value}. Instead of gating the attention output, the authors propose to gate the \textit{value representations} before they are weighted by the attention matrix. The core insight is that gating the value-state with a function of itself creates a direct regulatory pathway, decoupling value and attention score updates more effectively than prior methods that gate on input embeddings. Through a theoretical analysis of the underlying gradients, the authors show that this design allows the model to suppress a token's contribution based on its emergent value representation.
The Value-State Gated Attention (VGA) is defined as:

\begin{equation}
\text{VGA}(Q,K,V) = \text{Softmax}\left(\frac{QK^\top}{\sqrt{d}}\right) \bigl( \sigma(G_v(V)) \odot V \bigr),
\label{eq:vga}
\end{equation}

where $\sigma(\cdot)$ is the sigmoid function, $G_v(\cdot)$ is a learnable projection that produces a gating vector of the same dimension as the value vectors, and $\odot$ denotes element-wise multiplication. Unlike output-gating approaches, the gate in VGA is applied directly to the value matrix before the softmax-weighted combination, suppressing the contribution of sink tokens at the value level.

The architecture is illustrated in Figure~\ref{fig:vga_architecture}. Experiments on BERT, RoBERTa, and LLaMA-2-7B demonstrate that VGA significantly mitigates the formation of AS, stabilizes value-state norms, improves downstream task performance and enhances quantization fidelity~\cite{bu2025value}.

\subsubsection{Discussion and Insights}

\textbf{Advantages.}
The core advantage of \textit{\textbf{Gated Attention Mechanisms}} is their ability to decouple the no-op behavior from the softmax attention logits. This decoupling breaks the self-reinforcing cycle that gives rise to AS by providing an alternative pathway for attention heads to produce near-zero updates. Instead of forcing extreme logits onto sink tokens, the learnable gate directly suppresses the attention output. Empirical evidence across multiple architectures consistently shows that gating effectively removes AS, enhances training stability by suppressing loss spikes, and improves long-context extrapolation performance. The mechanism adds minimal computational overhead and critically enables robust low-bit quantization that would otherwise fail on standard Transformers.

\textbf{Limitations.}  
Despite its effectiveness, gated attention exhibits several key limitations. 
First, it requires training from scratch; gate parameters cannot be directly injected into pretrained models without retraining, restricting its applicability for model adaptation or post-hoc enhancement. 
Second, the gating operation introduces non-negligible parameters, particularly for element-wise variants that modulate each dimension independently. 
Third, the training and inference dynamics through which gated attention disrupts the no-op cycle remain poorly understood. 
Open questions include how gate values evolve during optimization, when they converge toward near-zero states, and how this suppression interacts with value norms during inference.
Fourth, the lack of standardized evaluations makes it difficult to quantify AS mitigation effectiveness and compare across different gated attention variants.

\textbf{Future Directions.}
First, developing lightweight post-hoc or parameter-efficient methods, such as adapter-based injection or fine-tuning techniques, to incorporate gate parameters into existing pretrained models without expensive training from scratch. 
Second, designing more parameter-efficient gating variants, such as shared or low-rank gates, to reduce the computational overhead of fine-grained element-wise modulation. 
Third, investigating the training dynamics of gate evolution to better understand how gated attention disrupts the no-op cycle. 
Fourth, establishing standardized evaluation benchmarks with consistent metrics for AS mitigation effectiveness to enable fair comparison across different gated attention variants.

% zunhai

\subsection{Modified Softmax Functions}
\label{sec_5_2_Modified_Softmax_Functions}

\begin{tcolorbox}[takeawaysbox]
{\large \textbf{\textcolor{TikTokPink}{\textit{Key Takeaways:}}}}
  \begin{enumerate}[leftmargin=*, label=\arabic*)]
    \item \textbf{Core Mechanism:}
    \textit{\textbf{Modified Softmax Functions}} directly intervene in Softmax normalization to prevent extreme logits and forced attention allocation, eliminating AS at its mathematical root without introducing additional parameters.

    \item \textbf{Practical Approaches:}
    Three families of modifications have been proposed: output-constrained Softmax, normalization-free attention, and pre-Softmax modulation. These approaches effectively reduce sink token rates and activation outliers, thereby enabling low-bit quantization.

    \item \textbf{Discussion and Insights:}
    Despite their effectiveness, \textit{\textbf{Modified Softmax Functions}} face three key challenges: excessive flattening of attention distributions may degrade performance on tasks requiring sharp attention, they require training from scratch, and they risk incompatibility with existing optimized attention kernels. 
    Future research should focus on striking a better balance between AS elimination and attention sharpness, developing lightweight post-hoc adaptation methods, and ensuring compatibility with efficient attention kernels for practical deployment.
  \end{enumerate}
\end{tcolorbox}

\subsubsection{Core Mechanism}

\textit{\textbf{Modified Softmax Functions}} offer another direct approach to mitigating AS by intervening in the Softmax normalization itself. This line of work is another direct response to the \textit{\textbf{Softmax Limitations and No-Op Theory}} (discussed in \S~\ref{sec_4_1_Softmax_Limitations}). Unlike gated mechanisms that decouple no-op behavior via an additional learnable gate, \textit{\textbf{Modified Softmax Functions}} alter the Softmax computation to prevent the formation of extreme logits and the resulting AS. A representative work in this direction is \textit{Quantizable Transformers}~\cite{bondarenko2023quantizable}, which introduces \textit{clipped softmax} as a lightweight alternative.

As established in the no-op theory, the sum-to-one constraint forces attention heads to concentrate probability mass on sink tokens when they need to produce a near-zero update, leading to extreme pre-Softmax logits and massive activation outliers. To break this cycle, \textit{clipped softmax} \cite{bondarenko2023quantizable} modifies the output of the standard Softmax by stretching and then clipping it into a finite range:

\begin{equation}
\text{ClippedSoftmax}(\mathbf{x}; \zeta, \gamma) = \text{clip}\bigl((\zeta - \gamma) \cdot \text{Softmax}(\mathbf{x}) + \gamma,\ 0,\ 1\bigr),
\label{eq:clipped_softmax}
\end{equation}

where $\zeta \ge 1$ and $\gamma \le 0$ are hyperparameters. This operation maps the original $[0,1]$ probability output to $[\gamma, \zeta]$, then clips it back to $[0,1]$. Consequently, exact zeros or ones can be achieved with a finite input range.

The core insight is that this transformation directly addresses the root cause of AS. By limiting the maximum attention probability and blocking gradient flow for clipped values, the model cannot rely on extreme logits to form a strong sink and is forced to learn an outlier-free strategy for no-op updates. Empirically, models pre-trained with \textit{clipped softmax} learn significantly smaller outliers while maintaining task performance, enabling full INT8 quantization of activations without additional effort.
Other \textit{\textbf{Modified Softmax Functions}} follow a similar philosophy and will be discussed in the following subsection.

\subsubsection{Practical Approaches}

Beyond the \textit{clipped softmax} discussed in the core mechanism, several other \textit{\textbf{Modified Softmax Functions}} have been proposed to mitigate AS. These approaches can be categorized into three families based on their intervention strategy: (i) \textit{Output-Constrained Softmax}, which constrains the output range of Softmax; (ii) \textit{Normalization-free Attention}, which eliminates the sum-to-one normalization constraint; and (iii) \textit{Pre-Softmax Modulation}, which modulates the logits or variance before Softmax.

\begin{figure}[t]
\centering
\includegraphics[width=1\columnwidth]{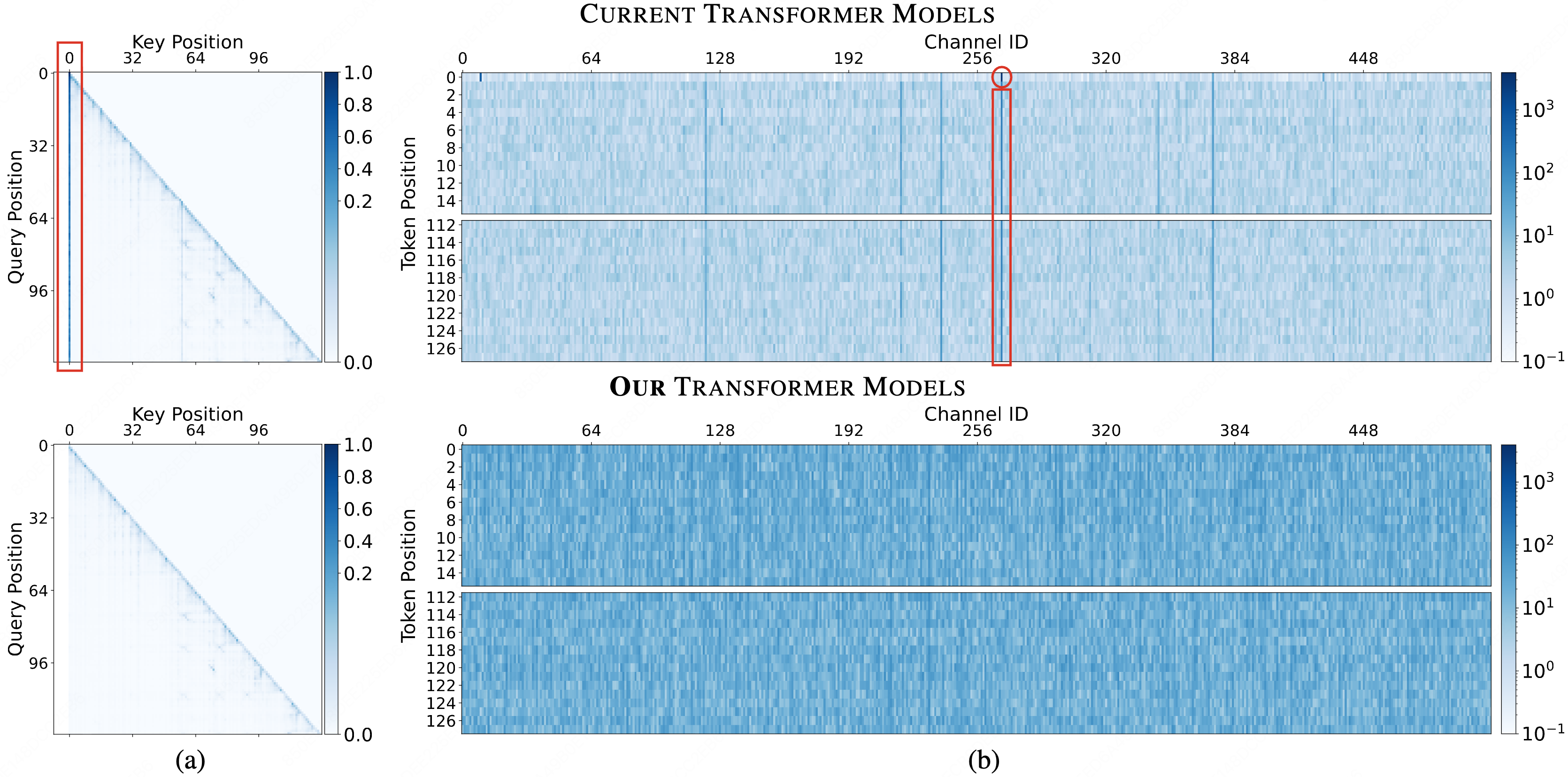}
\caption{\textbf{Top:} Mean attention map across all heads and layers of GPT2-Medium (baseline): the first token dominates attention (red box). Mean hidden state across layers: outlier activations emerge in specific feature dimensions (red box); the first token position exhibits the most extreme outliers (red circle). \textbf{Bottom:} Replacing canonical Softmax with \textit{Softmax-1} eliminates first-token dominance. The figure is adapted from~\cite{kaul2025attention}.}
\label{fig:softmax-1}
\end{figure}

\paragraph{Output-Constrained Softmax.}
This family retains the Softmax framework but directly restricts its output range or rescales its distribution to prevent extreme probabilities that lead to AS.

  \begin{itemize}
  
    % \item \textit{Clipped Softmax.} As detailed in the core mechanism, \textit{Quantizable Transformers}~\cite{bondarenko2023quantizable} proposes this method to constrain the Softmax output range and block gradients for clipped values, enabling robust low-bit quantization.

    \item \textit{Softmax-1.} \textit{From Attention to Activation}~\cite{kaul2025attention} identifies that standard Softmax forces attention onto the first token under causal masking. \textit{Softmax-1} modifies the normalization by adding a constant 1 to the denominator, allowing sub-unit summation:
    \begin{equation}
    \text{Softmax-1}(\mathbf{z})_i = \frac{e^{z_i}}{1 + \sum_j e^{z_j}}.
    \end{equation}
    This modification reduces first-token attention from 65\% to 3.3\% and lowers activation kurtosis from 1657 to 3.1, enabling robust 4-bit quantization. Figure~\ref{fig:softmax-1} illustrates the effect of Softmax-1 on attention maps and activation outliers.

    \item \textit{Elastic-Softmax.} The work \textit{Attention Needs to Focus}~\cite{fu2026attention} introduces \textit{Elastic-Softmax} to mitigate attention underload (manifested as AS). It relaxes the standard Softmax by applying a temperature $T>1$ or a power exponent $\alpha<1$:
    \begin{equation}
    \text{Elastic-Softmax}(\mathbf{z})_i = \frac{e^{z_i / T}}{\sum_j e^{z_j / T}} \quad \text{or} \quad \frac{(e^{z_i})^\alpha}{\sum_j (e^{z_j})^\alpha}.
    \end{equation}
    Flattening the distribution suppresses forced attention on irrelevant tokens. Experiments report 59.58\% attention sparsity and effective AS mitigation.
  
  \end{itemize}

\begin{figure}[t]
\centering
\includegraphics[width=1\columnwidth]{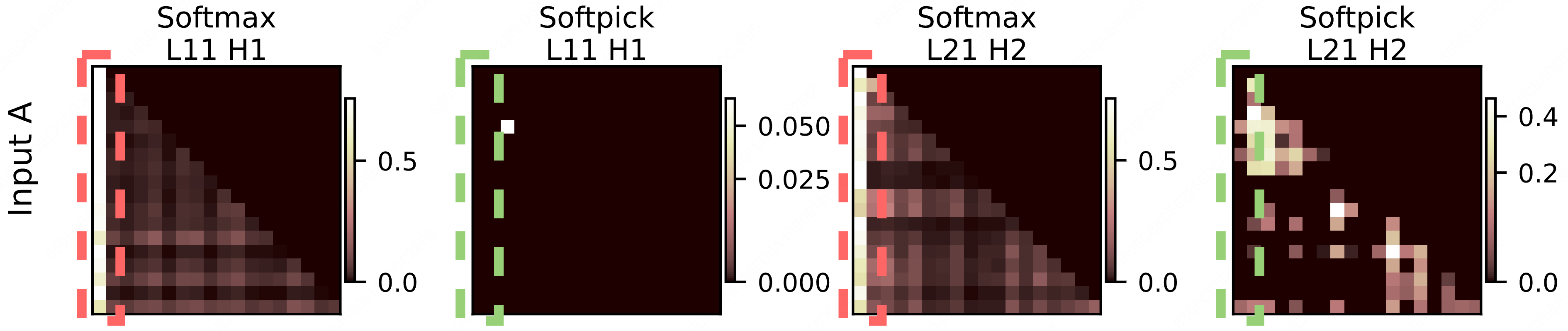}
\caption{Attention maps of Softmax and \textit{Softpick}. Using \textit{Softpick} effectively eliminates AS. The figure is adapted from~\cite{zuhri2025softpick}.}
\label{fig:softpick-2}
\end{figure}
\paragraph{Normalization-Free Attention.}
This family abandons the sum-to-one constraint entirely, replacing Softmax with functions that do not compete across tokens, thereby eliminating the root cause of AS.

  \begin{itemize}
  
  \item \textit{Softpick.} The work \textit{Softpick}~\cite{zuhri2025softpick} proposes a rectified, non-normalized attention function. Starting from a standard Softmax output, it subtracts a threshold $\tau$ and applies ReLU:
  \begin{equation}
  \text{Softpick}(\mathbf{z})_i = \max\left(0, \frac{e^{z_i}}{\sum_j e^{z_j}} - \tau\right).
  \end{equation}
  Because the sum of outputs is no longer 1, the model can assign near-zero weights to all tokens when no update is needed. Empirical results on 340M models show a 0\% sink rate, reduction of activation kurtosis from 33,510 to 340.
  Figure~\ref{fig:softpick-2} visualizes the attention maps of Softmax versus \textit{Softpick}.

  \item \textit{SWAT.} The work \textit{Sliding Window Attention Training}~\cite{fu2025sliding} replaces Softmax with the element-wise sigmoid function:
  \begin{equation}
  \text{SigmoidAttn}(Q,K) = \sigma\left(\frac{QK^\top}{\sqrt{d}}\right), \quad \sigma(x) = \frac{1}{1+e^{-x}}.
  \end{equation}
  There is no normalization across tokens, so each query-key pair is independent, making AS impossible by construction. \textit{SWAT} combines this with sliding window training and achieves competitive performance on long-context benchmarks.

  \end{itemize}

\paragraph{Pre-Softmax Modulation.}
This family retains the Softmax function but modifies its inputs (logits) or controls their statistical properties to shape the resulting attention distribution.

  \begin{itemize}
  
    \item \textit{Integral Attention.} The work \textit{Integral Transformer}~\cite{kobyzev2025integral} denoises attention by integrating signals sampled from the logit distribution. Conceptually, it replaces the deterministic logits with an expected value over a noise distribution:
    \begin{equation}
    \text{IntegralAttn}(Q,K) = \text{Softmax}\left( \mathbb{E}_{\epsilon \sim \mathcal{N}(0,\sigma^2)}[ \text{Logit}(Q,K) + \epsilon ] \right).
    \end{equation}
    This smoothing produces more balanced logits before Softmax, reducing disproportionate weight on sink tokens and mitigating AS as a result. \textit{Integral Transformer} outperforms baselines on knowledge and reasoning benchmarks and reduces rank collapse.

  \end{itemize}

\subsubsection{Discussion and Insights}

\textbf{Advantages.}  
\textit{\textbf{Modified Softmax Functions}} tackle the root cause of AS by directly intervening in Softmax normalization. Unlike gated mechanisms that introduce additional parameters, these methods modify the computation itself, incurring zero parameter overhead while effectively eliminating extreme logits and forced attention allocation. Empirical results across diverse model scales demonstrate that variants such as \textit{Softpick} and \textit{Sigmoid attention} achieve near-zero sink rates and significantly reduce activation outliers, enabling robust low-bit quantization without compromising task performance. Their simplicity and architecture-agnostic design further facilitate adoption in existing Transformer implementations.

\textbf{Limitations.}  
Despite their effectiveness, \textit{\textbf{Modified Softmax Functions}} present several trade-offs. 
First, excessive flattening of attention distributions can diminish the model's capacity to concentrate on genuinely informative tokens, potentially harming performance on tasks that demand sharp attention. 
Second, they require training from scratch, as the modified Softmax cannot be retrofitted into pretrained models without retraining. 
Third, modifying the Softmax function may introduce incompatibility with existing optimized attention kernels, thereby limiting practical deployment in efficient inference pipelines.

\textbf{Future Directions.}
Future research should focus on three directions. First, developing modified Softmax functions that strike a better balance between eliminating AS and preserving sharp attention for genuinely informative tokens, thereby maintaining task performance. 
Second, designing post-hoc or lightweight adaptation methods that can be applied to existing pretrained models without expensive training from scratch. 
Third, ensuring compatibility with existing optimized attention kernels to facilitate practical deployment in efficient inference pipelines and large-scale production systems.

% Yifan

\subsection{Learnable Attention Bias}
\label{sec_5_3_Learnable_Attention_Bias}

\begin{tcolorbox}[takeawaysbox]
{\large \textbf{\textcolor{TikTokPink}{\textit{Key Takeaways:}}}}
\begin{enumerate}[leftmargin=*, label=\arabic*)]
    \item \textbf{Core Mechanism:}  
    \textit{\textbf{Learnable Attention Bias}} explicitly replaces the implicit bias induced by AS with a trainable explicit attention bias mechanism. This allows precise, interpretable modulation of attention in no-update scenarios, directly controlling sink token influence.
    
    \item \textbf{Practical Approaches:} Four families of explicit bias that effectively mitigate AS have been proposed: key-value bias concatenation, key bias, scaling factor bias, and denominator bias.

    \item \textbf{Discussion and Insights:}  
    Despite its effectiveness, it requires training from scratch, lacks standardized evaluations for AS mitigation, and suffers from an incomplete understanding of training dynamics. Future research should focus on developing lightweight post-hoc methods to inject learnable bias into pretrained models, establishing standardized benchmarks for fair comparison, and investigating the interaction between explicit biases and training dynamics.
\end{enumerate}
\end{tcolorbox}

\subsubsection{Core Mechanism}

As detailed in \S~\ref{sec_4_3_Implicit_Attention_Bias}, AS acts as an \textit{\textbf{Implicit Attention Bias}}, contributing almost uniformly to the attention output across different query positions and inputs and effectively functioning as a fixed bias term. 
Based on this insight, \textit{\textbf{Learnable Attention Bias}} introduces dedicated parameters that replicate this bias effect in a controlled and interpretable manner. 
Various implementations of \textit{\textbf{Learnable Attention Bias}} have been proposed in recent studies and are discussed in the following subsection.

\subsubsection{Practical Approaches}

Several concrete instantiations of \textit{\textbf{Learnable Attention Bias}} have been proposed, differing in where the bias is inserted and how it interacts with the attention computation.

\begin{itemize}

\item \textbf{Key-Value Bias Concatenation.}
The earliest explicit instantiation appears in \textit{Massive Activations}~\cite{sun2024massive}, where the authors augment the attention mechanism by concatenating learnable key and value vectors to the existing key and value matrices. The formulation is:

\begin{equation}
\mathrm{Attention}(Q,K,V;\mathbf{k}',\mathbf{v}') = \mathrm{softmax}\left(\frac{Q[K^\top \ \mathbf{k}']}{\sqrt{d}}\right)\left[\begin{array}{c} V \\ \mathbf{v}'^{\top} \end{array}\right],
\end{equation}

where \(\mathbf{k}', \mathbf{v}' \in \mathbb{R}^d\) are learnable parameters per attention head. Training with this explicit bias eliminates \textit{Massive Activations} and AS, confirming that AS is a substitute for an explicit learnable bias.

\item \textbf{Key Bias.}
The empirical study \textit{When Attention Sink Emerges}~\cite{gu2025attention} provides causal evidence by introducing learnable key biases that directly absorb attention. This approach adds a learnable bias matrix \(K_{\text{bias}}\) to the original key matrix:

\begin{equation}
\mathrm{Attention}(Q,K,V) = \mathrm{softmax}\left(\frac{Q(K + K_{\text{bias}})^\top}{\sqrt{d}}\right)V,
\end{equation}

where \(K_{\text{bias}}\) is a head‑specific learnable matrix. With only key biases, the AS disappears and attaches to the bias position, proving that AS can be completely replaced by an explicit key bias.

\item \textbf{Scaling Factor Bias.}
In \textit{Systematic Outliers}~\cite{an2025systematic}, the authors demonstrate that AS function as implicit context‑aware scaling factors. They propose an explicit context‑aware scaling factor \(S_c(x)\) that dynamically adjusts the attention output:

\begin{equation}
\mathrm{Attention}(Q,K,V) = S_c(x)\cdot\mathrm{softmax}\!\left(\frac{QK^\top}{\sqrt{d}}\right)V,
\end{equation}

where \(S_c(x)\) is a learnable scalar that depends on the input context. Structurally eliminating outliers via this scaling factor accelerates convergence and improves model compression~\cite{an2025systematic}.

\item \textbf{Denominator Bias.}
The most parameter‑efficient instantiation modifies the Softmax denominator directly. Both \textit{MiMo-V2-Flash}~\cite{xiao2026mimo} and \textit{GPT-OSS}~\cite{agarwal2025gpt} introduce a learnable scalar per attention head into the denominator of the Softmax normalization:

\begin{equation}
\mathrm{Softmax}_{\text{LAB}}(\mathbf{z})_i = \frac{e^{z_i}}{\sum_{j} e^{z_j} + b},
\end{equation}

where \(b\) is a head‑specific learnable parameter. This term creates a virtual sink that absorbs excess attention probability when no real token is relevant, allowing the model to pay no attention to any token by allocating mass to a dummy position~\cite{agarwal2025gpt}. 

\end{itemize}

\subsubsection{Discussion and Insights}

\textbf{Advantages.}  
\textit{\textbf{Learnable Attention Bias}} replaces the implicit bias induced by AS with an explicit, trainable mechanism. This approach enhances interpretability and provides fine-grained control over attention in no-update scenarios, effectively eliminating the reliance on AS and reducing associated activation outliers. 

\textbf{Limitations.}  
Despite its effectiveness, \textit{\textbf{Learnable Attention Bias}} has notable limitations. It requires training from scratch, as the bias parameters cannot be retrofitted into pretrained models without retraining. Different design choices entail trade-offs; for instance, key-value bias concatenation introduces a significant number of parameters. Furthermore, the lack of standardized evaluations to assess AS mitigation effectiveness makes it difficult to compare different implementations consistently. Finally, there is an incomplete understanding of how explicit biases interact with attention distributions and training dynamics.

\textbf{Future Directions.}
To address the limitations, future research should explore several promising directions. 
First, developing post-hoc or lightweight fine-tuning methods, such as adapter-based or parameter-efficient transfer learning techniques, would enable the injection of learnable bias into existing pretrained models without expensive training from scratch. 
Second, establishing standardized evaluation benchmarks with consistent metrics for AS mitigation effectiveness, computational overhead, and parameter efficiency would facilitate fair comparison across different bias implementations and accelerate progress in this direction.
Third, investigating the interaction between explicit biases and attention distributions as well as training dynamics would deepen theoretical understanding and guide the design of more effective bias mechanisms.

% Yifan

\subsection{Pre-training Interventions}
\label{sec_5_4_Pre-training_Interventions}

\begin{tcolorbox}[takeawaysbox]
{\large \textbf{\textcolor{TikTokPink}{\textit{Key Takeaways:}}}}
\begin{enumerate}[leftmargin=*, label=\arabic*)]
    \item \textbf{Core Mechanism:}
    \textit{\textbf{Pre-training Interventions}} proactively modulate training dynamics via the optimizer, loss function, or normalization scheme to reduce the emergence of AS and activation outliers, without modifying the model architecture.

    \item \textbf{Practical Approaches:}
    These interventions can be grouped into three categories: (i) loss function regularization, (ii)optimizer replacement, and (iii) integrated frameworks combining multiple strategies. 
    They have been validated at production scale and support robust low-bit quantization.

    \item \textbf{Discussion and Insights:}
    While \textit{\textbf{Pre-training Interventions}} provide proactive and architecture-agnostic AS mitigation, they come with key limitations: (i) they require training from scratch, limiting applicability to pre-trained models, and (ii) some methods introduce additional computational overhead. 
    Future research should focus on lightweight post-hoc interventions for pretrained models, as well as more efficient pre-training interventions with minimal computational overhead.
\end{enumerate}
\end{tcolorbox}

\subsubsection{Core Mechanism}

\begin{figure}[t]
\centering
\includegraphics[width=1\columnwidth]{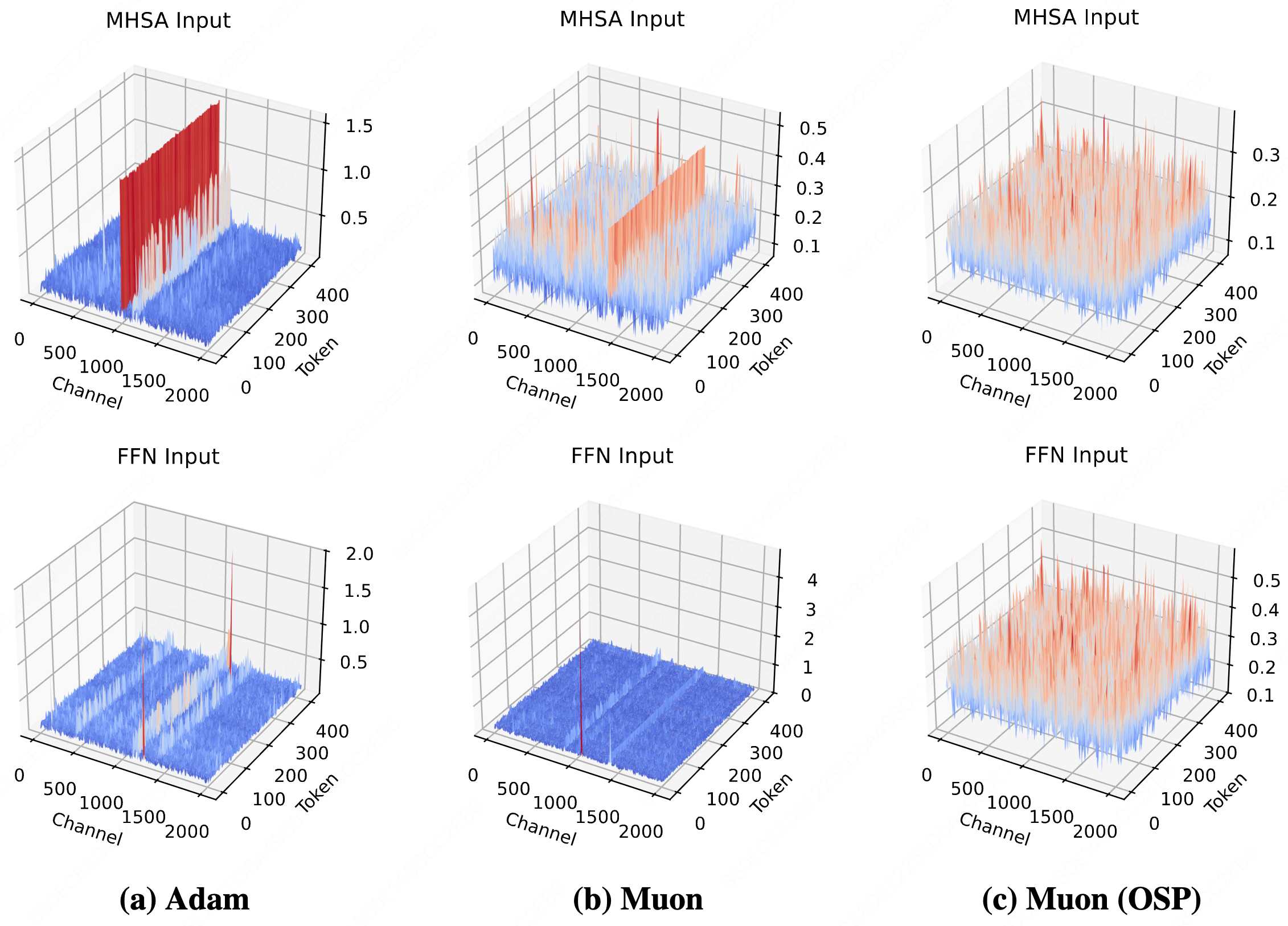}
\caption{Activation distribution in 1.4B models trained on 100B tokens. Three optimization strategies: (a) Adam, (b) Muon, (c) \textit{OSP}. Muon alone provides insufficient outlier mitigation; \textit{OSP} eliminates outliers. Adapted from~\cite{park2025outlier}.}
\label{fig:osp-1}
\end{figure}

Although sink behavior is rooted in architectural factors such as Softmax normalization, accumulating evidence suggests that optimization dynamics can influence the severity and manifestation of AS.
\textit{\textbf{Pre-training Interventions}} target the training process itself, encompassing choices of optimizer, loss function, normalization scheme, and regularization strategy, rather than modifications to the model architecture. 
For example, standard adaptive optimizers like Adam have been shown to favor certain privileged bases in weight matrices, producing activation spikes that closely align with the emergence of AS~\cite{park2025outlier}, as illustrated in Figure~\ref{fig:osp-1}. 
Beyond these effects, training dynamics including gradient noise and parameter updates can further exacerbate outlier formation.
As a result, deliberate adjustments to the training recipe can guide the optimization away from solutions that rely on AS.
These proactive, architecture-agnostic interventions complement the reactive architectural modifications discussed earlier.

% The following subsection reviews concrete instantiations of \textit{\textbf{Pre-training Interventions}}, organized by the component of the training recipe they modify.

\subsubsection{Practical Approaches}

Several concrete instantiations of \textit{\textbf{Pre-training Interventions}} have been proposed. They are organized into three categories based on the aspect of the training recipe they modify.

\paragraph{Loss Function Interventions.}
Adding auxiliary regularization terms to the training objective can directly penalize outlier formation or enforce desirable properties in attention distributions.

\begin{itemize}
  \item \textit{TWEO}. The work \cite{liang2025tweo} demonstrates that extreme outliers are a data‑independent artifact of training, arising from co‑linearity in weight matrices. The proposed loss regularizer penalizes the tails of activation distributions, effectively suppressing outlier growth. While the exact formulation involves a scaling factor that rapidly increases the penalty for large values, the key effect is to reduce activation outliers from over 10000 to below 20. Under standard FP8 training which fails catastrophically, \textit{TWEO} achieves performance comparable to the BF16 baseline while increasing training throughput by 36\%. It also enables, for the first time, hardware‑friendly W8A8 per‑tensor static quantization of LLMs at state‑of‑the‑art quality.
  
  \item \textit{Sink-Aware Training}. The study \cite{fu2026attention2} proves that AS naturally construct a MoE mechanism, explaining head collapse where only a subset of heads contribute. To mitigate this, the authors introduce an auxiliary load balancing loss tailored for attention layers. The loss encourages uniform utilization across heads. Experiments show that this method achieves effective head load balancing and improves performance across vanilla, sink, and gated attention variants.
  
  \item \textit{Decorrelation Loss}. In audio‑visual speech recognition, the work \cite{cappellazzo2025mitigating} observes that intermediate sink tokens exhibit high cosine similarity with the beginning‑of‑sequence token, amplifying activation spikes. The authors propose a decorrelation loss that reduces this similarity:
  \begin{equation}
  \mathcal{L}_{\text{decorr}} = \frac{1}{N} \sum_{i \neq \text{BOS}} \frac{\mathbf{h}_{\text{BOS}} \cdot \mathbf{h}_i}{\|\mathbf{h}_{\text{BOS}}\| \|\mathbf{h}_i\|},
  \end{equation}
  where \(\mathbf{h}\) are hidden states. This intervention mitigates intermediate sinks, improving word error rates under high feature downsampling while maintaining stability at lower rates.
  
\end{itemize}

\paragraph{Optimizer Interventions.}
Standard adaptive optimizers such as Adam have been identified as a primary source of activation outliers. Replacing or modifying the optimizer can suppress these effects.

\begin{itemize}
  \item \textit{OrthoAdam}. The work \cite{kaul2025attention} identifies adaptive optimizers as the main contributor to large outlier activations. The proposed \textit{OrthoAdam} uses orthogonal matrices to transform gradients, storing them in an alternative basis that prevents accumulation in privileged directions. This orthogonal transformation reduces activation kurtosis from 1657 to 3.1 and the perplexity penalty under 4‑bit weight quantization from 3565 to 0.3.
  
  \item \textit{Muon Optimizer}. The \textit{Outlier-Safe Pre-Training (OSP)}  framework \cite{park2025outlier} adopts the \textit{Muon Optimizer}, which eliminates privileged bases in weight matrices while maintaining training efficiency. Privileged bases refer to parameter directions excessively amplified by Adam‑style updates, leading to co‑linearity and activation spikes. By removing these bases, \textit{Muon Optimizer} prevents extreme activation values without sacrificing convergence speed.
\end{itemize}

\paragraph{Outlier-Safe Pre-Training (OSP) Framework.}
Rather than a single intervention, \textit{OSP} \cite{park2025outlier} combines three complementary innovations to proactively prevent outlier formation.

\begin{itemize}
  \item \textit{Muon Optimizer}. As described above, Muon eliminates privileged bases in weight matrices.
  
  \item \textit{Single-Scale RMSNorm}. Standard RMSNorm uses learnable per‑channel scales, which can inadvertently amplify outlier dimensions. \textit{OSP} replaces this with a single scalar scale per layer:
  \begin{equation}
  \text{RMSNorm}(x) = \frac{x}{\sqrt{\mathbb{E}[x^2] + \epsilon}} \cdot \gamma,
  \end{equation}
  where \(\gamma\) is a scalar. This prevents channel‑wise amplification while preserving representational power.
  
  \item \textit{Learnable Embedding Projection}. \textit{OSP} introduces a learnable projection matrix after the embedding layer to redistribute activation magnitudes. 
  This prevents the embedding matrix from directly producing extreme values.
\end{itemize}

The \textit{OSP} framework was validated by training a 1.4B parameter model on 1 trillion tokens, producing the first production‑scale LLM without extreme activation outliers.

These interventions demonstrate that modifying the training recipe can effectively eliminate AS and massive activations at their source. The \textit{OSP} framework provides a comprehensive solution that combines multiple strategies to achieve outlier‑free pre‑training without sacrificing efficiency or performance.

\subsubsection{Discussion and Insights}

\textbf{Advantages.}
\textit{\textbf{Pre-training Interventions}} tackle AS at its origin by shaping optimization dynamics rather than altering model architecture. This proactive and architecture-agnostic strategy complements architectural modifications such as gated attention or modified softmax. By suppressing outliers during training, these interventions render models inherently robust to low-bit quantization.

\textbf{Limitations.}
Despite their effectiveness, \textit{\textbf{Pre-training Interventions}} face three main limitations. First, they require training from scratch; their parameters cannot be directly incorporated into pretrained models, limiting flexibility for adaptation. 
Second, most interventions introduce additional computational overhead during training, such as auxiliary loss computation, thereby imposing extra computational burden on standardized pre-training pipelines.

\textbf{Future Directions.}
To address the aforementioned limitations, future research should explore several promising directions. 
First, developing lightweight post-hoc intervention methods, such as efficient continual training, that can be applied to existing pretrained models would greatly enhance practical flexibility. 
Second, designing more efficient pre-training interventions with minimal computational overhead, such as parameter-free regularization techniques, could reduce the burden on standard pre-training pipelines.

% Wu Wei

\subsection{Other Mitigation Techniques}
\label{sec_5_5_Additional_Techniques_for_Mitigating_Attention_Sink}
Beyond the perspectives discussed above, several additional strategies provide complementary approaches for mitigating AS. 
We briefly summarize these techniques below. 
To conclude this section, we then present a consolidated overview of all AS mitigation methods across two analytical levels.  

\begin{itemize}[leftmargin=*, label=\textbullet]

\item \textbf{Modified Norm Layer.}
\textit{Massive Emergence Layer} shows that RMSNorm and FFN jointly generate massive activations in a specific middle layer. To modify this norm-induced behavior, it selectively suppresses those activations by applying a training-free mask to the affected dimensions within the ME layer and all subsequent layers, thereby alleviating AS and improving LLM performance \cite{shi2026single}.
\textit{Head-wise RMSNorm} modifies standard layer normalization by applying RMSNorm independently to each attention head's value aggregation output instead of over the entire vector. This restores statistical parity across token positions and stabilizes the variance discrepancy originating from self-attention, significantly accelerating pre-training convergence \cite{li2026structural}.
\textit{Outlier-driven rescaling} unifies attention and residual sinks by revealing that outliers function as rescaling factors together with softmax and RMSNorm. To modify this norm-dependent mechanism, it absorbs outliers into learnable parameters or replaces their rescaling effect with explicit gated rescaling, eliminating the need for extreme activation values while preserving training stability and improving quantization robustness \cite{qiu2026unifiedviewattentionresidual}.

\item \textbf{Residual Gated Rescaling.}
\textit{Outlier-driven rescaling} reveals that attention and residual sinks work jointly with softmax and RMSNorm to rescale other components. To preserve this function without extreme outliers, explicit gated rescaling (GatedNorm) absorbs the rescaling effect into learnable parameters, improving training performance and quantization robustness \cite{qiu2026unifiedviewattentionresidual}.

\item \textbf{Value Modification.} 
\textit{V-scale} identifies gradient sinks as the backward-pass counterpart of attention sinks, where massive activations act as adaptive regulators that attenuate localized training pressure. It modifies backward-pass gradients on the Value path to suppress massive activations while preserving sink functionality \cite{chen2026attention}.

\item \textbf{Tuning Auxiliary Loss.} 
\textit{KARMA} reveals that action-only training objectives induce attention sinks, and adds a semantic reconstruction regularizer as a train-only auxiliary loss. This enforces semantic decodability while optimizing the action objective, thereby mitigating semantic collapse \cite{sun2026karma}. 
\textit{Surgery} measures sink divergence per attention head, observes that harmful fine-tuning increases positive-divergence heads, and introduces a regularizer that suppresses positive sink divergence, steering heads toward the negative group \cite{liu2026surgery}. 
\textit{Decorrelated LLM-AVSR} observes that intermediate sink tokens exhibit high cosine similarity to the BOS token, and adds a decorrelation loss that reduces this similarity, effectively mitigating intermediate sinks and massive activations \cite{cappellazzo2025mitigating}. 

\item \textbf{Architectural Isolation.}
To mitigate AS in ViTs, the EDIT has been proposed \cite{feng2026edit}. 
Unlike standard ViTs where the \texttt{[CLS]} token often attracts excessive attention, EDIT adopts a layer-aligned encoder-decoder design: the encoder processes image patches via self-attention, while the decoder uses cross-attention to progressively refine representations from low- to high-level features. 
Evaluations on ImageNet and transfer learning benchmarks demonstrate consistent performance improvements over DeiT3, confirming EDIT's effectiveness in reducing AS and enhancing visual feature extraction.

\end{itemize}

\paragraph{Summary of Mitigation Strategies for Attention Sink.}  
In summary, the AS mitigating methods discussed in this section can be grouped into two overarching principles: (i) providing explicit, controllable alternatives that render AS unnecessary, and (ii) disrupting the causal chain that gives rise to AS. The following overview categorizes the surveyed techniques according to these complementary strategies.

\begin{itemize}
    
    \item \textbf{Providing Explicit Alternatives} (substituting implicit AS with learnable, controllable mechanisms):
    \begin{itemize}
        \item \textit{\textbf{Gated Attention Mechanisms}} (\S~\ref{sec_5_1_Gated_Attention_Mechanisms}): A learnable gate directly modulates attention outputs, enabling no-op updates and eliminating the need for sink tokens.
        \item \textit{\textbf{Learnable Attention Bias}} (\S~\ref{sec_5_3_Learnable_Attention_Bias}): Explicit attention biases absorb excess attention mass, precisely replacing the implicit bias induced by AS.
        \item \textit{\textbf{Residual Gated Rescaling}} (\S~\ref{sec_5_5_Additional_Techniques_for_Mitigating_Attention_Sink}): Explicit gated rescaling absorbs the rescaling effect of outlier-driven sinks into learnable parameters, providing a controllable alternative that eliminates the need for extreme activation values while preserving training stability.
        \item \textit{\textbf{Architectural Isolation}} (\S~\ref{sec_5_5_Additional_Techniques_for_Mitigating_Attention_Sink}): The encoder-decoder architecture redistributes attention away from the \textit{[CLS]} token in ViTs, substituting sink concentration with progressive feature refinement.
    \end{itemize}

    \item \textbf{Cutting the Causal Chain} (eliminating AS by addressing its root causes):
    \begin{itemize}
        \item \textit{\textbf{Modified Softmax Functions}} (\S~\ref{sec_5_2_Modified_Softmax_Functions}): Techniques such as clipping, re-centering, or replacing Softmax remove the sum-to-one constraint that forces attention onto sink tokens.
        \item \textit{\textbf{Pre-training Interventions}} (\S~\ref{sec_5_4_Pre-training_Interventions}): Adjustments to the optimizer, loss function, and normalization scheme suppress the formation of outliers at their source.
        \item \textit{\textbf{Modified Norm Layer}} (\S~\ref{sec_5_5_Additional_Techniques_for_Mitigating_Attention_Sink}): Interventions such as ME layer masking, head-wise RMSNorm, and outlier absorption into normalization parameters directly suppress the generation of massive activations and variance discrepancies at their source.
        \item \textit{\textbf{Value Modification}} (\S~\ref{sec_5_5_Additional_Techniques_for_Mitigating_Attention_Sink}): Modifying the backward-pass gradients on the Value path or reducing the internal values of specific tokens disrupts the forward or backward propagation of AS without eliminating sink functionality.
        \item \textit{\textbf{Tuning Auxiliary Loss}} (\S~\ref{sec_5_5_Additional_Techniques_for_Mitigating_Attention_Sink}): Regularization losses such as semantic reconstruction, sink divergence suppression, and decorrelation loss intervene during training to prevent AS formation without architectural changes.
    \end{itemize}
    
\end{itemize}% Zunhai

%----------------------------------Section 6----------------------------------%
\clearpage
\section{Applications and Practical Guidelines}
\label{sec_7_Applications}

This section categorizes research on AS by application domain and provides practical guidance for managing AS. 
For each domain, we present concrete recommendations for selecting AS-related techniques, aligned with model architecture and task-specific requirements.

\subsection{Model Pre-training}
\label{sec_7_1_Model_Pretraining}

For ViT pre-training, apply \textbf{\textit{Learnable Prefix Tokens}} can stabilize optimization by absorbing sink-related attention artifacts and alleviating attention entropy collapse \cite{zhang2026one, simeoni2025dinov3, wang2025vggt}. 
For LLM pre-training, \textbf{\textit{Sink Token Preservation}} retains early sink tokens as stable attention anchors, which can further support efficient sparse attention patterns \cite{ge2025little}.
To mitigate AS during pre-training, several architectural or functional modifications can be introduced. \textbf{\textit{Learnable Attention Bias}} methods encourage attention toward designated sink positions \cite{xiao2026mimo, agarwal2025gpt, an2025systematic}. 
\textbf{\textit{Gated Attention Mechanisms}} suppress undesirable attention allocation through nonlinear gating \cite{bu2025value, qiu2025gated, qwenai2026, bondarenko2023quantizable}. \textbf{\textit{Modified Softmax Functions}} can substantially reduce AS by relaxing the competitive constraints imposed by standard Softmax \cite{zuhri2025softpick, hongvariance, bondarenko2023quantizable}.
In addition, \textbf{\textit{Pre-training Interventions}} can address AS-related outliers from the optimization perspective. Representative strategies include auxiliary losses that penalize extreme outlier formation \cite{liang2025tweo, team2025longcat}, outlier-safe optimizers that reduce optimizer-induced outlier growth \cite{park2025outlier}, and gated normalization schemes that rescale outlier-dominated activations \cite{qiu2026unifiedviewattentionresidual}.

\subsection{Model Tuning}
\label{sec_7_2_Model_Tuning}

For mitigating harmful fine-tuning effects such as catastrophic forgetting or backdoor injection, apply \textbf{\textit{Sink Token Repurposing}} to detect and preserve AS patterns as indicators of model corruption \cite{liu2026surgery}. 
To understand the theoretical origin of AS dynamics during tuning, refer to the analysis of rotary position embeddings, which reveals inevitable AS convergence in autoregressive transformers \cite{chen2024rotary}. 
% When AS mitigating is desired during model tuning, employ a regularizer to explicitly suppress sink artifacts while preserving task performance \cite{liu2026surgery}.

\subsection{Efficient Inference}
\label{sec_7_3_Model_Inference}

For KV cache compression, apply \textbf{\textit{Sink Token Preservation}} to retain initial AS tokens as fixed anchors, enabling aggressive eviction or quantization without performance collapse \cite{xiao2024efficient, zhang2023h2o, cai2024pyramidkv, liu2024intactkv, hooper2024kvquant, duanmu2024skvq, su2025rotatekv, su2025akvq, shutova2025cache, chen2024prefixquant}; for sparse attention, preserve AS to stabilize block‑wise or streaming patterns \cite{khaki2025sparsevila, fu2025h2eal, ge2025little, acharya2025star, xiao2025duoattention, zhao2024buzz}. 
For other accelerations, inject \textbf{\textit{Learnable Prefix Tokens}} as dedicated AS to absorb outliers and enable low‑bit quantization \cite{zhang2024sinklora, son2024prefixing, dong2025hymba, deng2025unigist, hu2025epic} or repurpose AS as geometric anchors for token selection via \textbf{\textit{Sink Token Repurposing}} \cite{chen2025omnisparse, park2025keydiff, li2024streamingdialogue, yang2025earn, shin2025orthorank}. 
% When AS mitigating is desired for further efficiency, apply \textbf{\textit{Gated Attention Mechanism}} to remove AS \cite{qiu2025gated, bu2025value, bondarenko2023quantizable} or replace softmax with \textbf{\textit{Modified Softmax Functions}} to prevent AS formation \cite{zuhri2025softpick, kaul2025attention, bondarenko2023quantizable}.

\subsection{Mechanism Interpretability}
\label{sec_7_4_Mechanism_Interpretability}

To interpret AS as a consequence of Softmax's inherent limitations, as well as the no-op hypothesis, refer to \textbf{\textit{Softmax Limitations \& No-Op Theory}} \cite{fu2026attention2, fu2026attention, hongvariance, gu2025attention, kaul2025attention, bu2025value, qiu2025gated, chen2024rotary, bondarenko2023quantizable}. 
When analyzing AS through outlier circuits and massive activations that bias attention logits, adopt \textbf{\textit{Outliers Circuits}} \cite{cappellazzo2025mitigating, queipo2026attention, sun2024massive, an2025systematic, yona2025interpreting, kang2025see, su2026unveiling, zuhri2025softpick, guo2024active, puccetti2022outlier, bondarenko2021understanding, kovaleva2021bert, luo2021positional, clark2019does}; for understanding AS as an implicit bias from model parameters or attention dynamics, consider \textbf{\textit{Implicit Attention Bias}} \cite{han2026zerotuning, an2025systematic, sun2024massive}. 
For geometric interpretations where sink tokens serve as stable anchors in representation space, apply \textbf{\textit{Geometric Anchoring}} \cite{chen2025omnisparse, ruscio2025you, shin2025orthorank, park2025keydiff, dong2024exploring, zhang2025anchor}. 
To understand AS from positional bias (e.g., RoPE, causal mask, or structural bias), consider \textbf{\textit{Structural Bias}} \cite{fu2026attention, xiong2025dope, salvatore2025lost, shang2025forgetting, rulli2025attention, wu2025emergence, yan2024unveiling, zhang2026drives, luo2021positional, chen2024rotary}. 
Other theoretical perspectives include \textbf{\textit{Anti-Overmixing}} \cite{barbero2025llms, barbero2024transformers, geshkovski2023emergence}, \textbf{\textit{Catch-Tag-Release Theory}} \cite{zhangattention}, and \textbf{\textit{Active-Dormant Attention Theory}} \cite{guo2024active}.

\subsection{Reducing Hallucination}
\label{sec_7_5_Reducing_Hallucination}

For MLLMs suffering from visual hallucinations, apply \textbf{\textit{Attention Redistribution}} to shift attention mass from AS tokens to informative visual tokens \cite{tu2026attention, jiao2025don, zhuang2025vasparse, zhang2026drives}
Additionally, redistribute attention from AS-mapped fixed vocabulary tokens to enhance factual generation by enabling the model to dynamically reallocate attention weights \cite{chenvocabulary}. 
For preserving beneficial AS, leverage dense visual AS heads in shallow layers to maintain global context and reduce hallucination via \textbf{\textit{Sink Token Repurposing}} \cite{zhang2025shallow, zhang2024seeing}. 
For long-context LLMs applied to multi-modal tasks, preserve initial AS tokens as context anchors to stabilize attention and reduce hallucination via \textbf{\textit{Sink Token Preservation}} \cite{liu2026sinktrack}.

\subsection{Safety \& Robustness}
\label{sec_7_6_Safety_Robustness}

For adversarial attacks, adversaries can exploit initial AS tokens as ideal backdoor gateways to implant triggers with high stealth and effectiveness via \textbf{\textit{Sink Token Repurposing}} \cite{shang2025forgetting}.
For MLLMs, attackers can induce additional AS tokens through adversarial visual inputs to amplify dataset bias and trigger hallucination attacks \cite{wang2025mirage}.
For defense, repurpose register token embeddings as robust features to enhance model robustness against adversarial perturbations or explicitly introduce robustness tokens that function as AS to absorb adversarial noise \cite{pulfer2024robustness}.

\subsection{General Capability Enhancement}
\label{sec_7_7_General_Capability_Enhancement}

For training-free LLM improvement, apply \textbf{\textit{Attention Redistribution}} via attention calibration to harness hidden AS \cite{yu2024unveiling} or treat initial AS as a programmable control knob to systematically optimize attention dynamics \cite{han2026zerotuning}. 
For mitigating position bias, redistribute attention from advantage positions (e.g., sequence start) to disadvantaged positions via inter-position distillation \cite{wang2025position}. 
For domain-specific tasks (e.g., CTR prediction), inject \textbf{\textit{Learnable Prefix Tokens}} as artificial AS to aggregate local context and stabilize attention \cite{li2025ctr}.
For converting decoders to text encoders, mask the first token to surgically eliminate AS interference via \textbf{\textit{Sink Token Repurposing}} \cite{lin2025look}. 
When AS mitigating is expected, replace softmax with \textbf{\textit{Modified Softmax Functions}} to denoise attention on low-semantic tokens \cite{kobyzev2025integral}.

\subsection{Long-Context Enhancement}
\label{sec_7_8_Long-Context_Enhancement}

This section categorizes research on AS by application domain and provides practical guidance for managing AS. For each domain, we present concrete recommendations for selecting AS-related techniques aligned with model architecture and task-specific requirements.
For extending LLMs to unlimited streaming inputs without fine-tuning, preserving initial AS tokens in the KV cache via \textbf{\textit{Sink Token Preservation}} is recommended \cite{xiao2024efficient, liu2026sinktrack, yu2025sliding, chen2025edgeinfinite, zhang2025attention, wang2024greater, yang2025seed, acharya2025star, ge2025little, chen2025magicpig, xiao2025duoattention}. 
In video generation models, retaining deep AS tokens as global anchors helps avoid temporal drift \cite{yi2025deep, shin2026motionstream, liu2026rolling}. 
For training-based approaches, adopting \textbf{\textit{Learnable Prefix Tokens}} to create dedicated AS anchors \cite{zhang2024sinklora, elawady2024relic} or using \textbf{\textit{Learnable Attention Bias}} for sliding window attention \cite{xiao2026mimo} proves effective; alternatively, repurposing end-of-turn tokens as dialogue AS via \textbf{\textit{Sink Token Repurposing}} is a viable strategy \cite{li2024streamingdialogue}. 
When AS mitigation is expected, applying \textbf{\textit{Gated Attention Mechanism}} removes AS while improving length extrapolation \cite{qiu2025gated}, or replacing softmax with \textbf{\textit{Modified Softmax Functions}} prevents AS formation altogether \cite{fu2025sliding}. 
For block-wise sparse attention, preserving initial anchors avoids local AS artifacts \cite{acharya2025star}. 
In efficient retrieval, keeping AS on GPU enables exact computation \cite{chen2025magicpig}. For streaming heads, retaining AS supports aggressive cache compression \cite{xiao2025duoattention}. 
In continual learning, redistributing attention leverages AS in long contexts \cite{bai2025does}.

\subsection{Multi-Modal Enhancement}
\label{sec_7_9_Multi-Modal_Enhancement}

For MLLMs suffering from visual hallucinations, applying \textbf{\textit{Attention Redistribution}} shifts attention mass from visual or OCR sink tokens to informative regions \cite{kang2025see, baek2025large, zhang2026drives}. For ViT-based encoders, adopting \textbf{\textit{Learnable Prefix Tokens}} (e.g., register tokens) absorbs sink artifacts during training \cite{darcet2024vision, chen2025vision}, while test-time register injection serves training-free scenarios \cite{jiang2025vision}. For VLA models, these registers can be repurposed as spatial memory via \textbf{\textit{Sink Token Repurposing}} \cite{koo2025retovla}. For tasks requiring global semantics, \textbf{\textit{Sink Token Preservation}} keeps ViT sink tokens as semantic anchors \cite{luo2026sink}. For robust adaptation, reusing register token embeddings as additional features through \textbf{\textit{Sink Token Repurposing}} proves effective \cite{yellapragada2025leveraging}. For efficient inference, pruning non-essential tokens with visual sinks as stable anchors is enabled by \textbf{\textit{Sink Token Preservation}} \cite{lu2025artifacts}.

%----------------------------------Section 7----------------------------------%
\clearpage
\section{Challenges and Future Directions}
\label{sec_6_Challenges_and_Future_Directions}

Having surveyed the landscape of AS research across \textit{\textbf{Fundamental Utilization}}, \textit{\textbf{Mechanistic Interpretation}}, and \textit{\textbf{Strategic Mitigation}}, it is clear that substantial progress has been made. Nevertheless, several challenges remain, limiting both theoretical understanding and practical deployment. 
In the following, we distill the key open challenges and outline promising future directions that span the field.

\subsection{Challenges}

\textbf{Computational Overhead and Kernel Compatibility.} 
Efficient and accurate detection of dynamic sinks remains an open challenge, as dynamic identification incurs additional computational overhead \cite{kang2025see,yu2024unveiling,su2025rotatekv}. 
Moreover, many techniques operate on attention scores after Softmax, limiting compatibility with high‑performance attention implementations. 
\textit{\textbf{Attention Redistribution}} also incurs additional cost for modifying and reallocating attention scores, which can become a bottleneck in large‑scale models \cite{kang2025see,yu2024unveiling,tu2026attention}.
\textit{\textbf{Gated Attention Mechanisms}} introduce non‑negligible latency, particularly for element‑wise variants that modulate each dimension independently \cite{qiu2025gated,bu2025value}. 
Furthermore, \textit{\textbf{Modified Softmax Functions}} may lack efficient implementations that integrate seamlessly with optimized attention kernels, complicating their deployment in high‑performance settings \cite{zuhri2025softpick,kaul2025attention}.

\textbf{Training from Scratch and Adaptation Cost.} 
Most mitigation methods such as \textit{\textbf{Gated Attention Mechanisms}},  \textit{\textbf{Modified Softmax Functions}}, and \textit{\textbf{Learnable Attention Bias}} require training from scratch \cite{qiu2025gated,bu2025value,sun2024massive,zuhri2025softpick,kaul2025attention}. 
Their parameters cannot be directly injected into pretrained models without retraining. 
\textit{\textbf{Learnable Prefix Tokens}} also demand additional training or fine‑tuning, which can be costly for very large models \cite{zuhri2025softpick,kaul2025attention}. 
These methods severely limit practical adoption for already pretrained large models, as full retraining is often prohibitively expensive in terms of time and computational resources. 
Lightweight adaptation techniques such as adapters or continual pre‑training remain largely unexplored for AS, leaving a critical gap between research insights and real‑world deployment.

\textbf{Incomplete Understanding of Training Dynamics.} 
While the \textit{\textbf{Softmax Limitations and No‑Op Theory}} explains the emergence of AS, it does not capture the complex training dynamics that give rise to no-op behaviors \cite{bondarenko2023quantizable}. 
The evolution of mutual reinforcement between attention scores and value states during optimization remains largely unexplored. 
The training dynamics that lead to systematic alignment of weights, activations, and attention outliers are not completely formalized \cite{an2025systematic,yu2024super}, leaving questions about their emergence, stability, and evolution in pre-training. 
Likewise, the dynamics that produce AS as implicit biases remain unclear. 
This gap affects both \textit{\textbf{Mechanistic Interpretation}} and \textit{\textbf{Strategic Mitigation}}.

% \textbf{Lack of Standardized Benchmark for AS and Outlier Mitigation.}  
% Assessing the effectiveness of AS elimination and outlier suppression remains challenging due to the absence of widely adopted benchmarks. Existing studies employ heterogeneous metrics, including sink ratio, activation kurtosis, attention entropy, low-bit quantization accuracy, perplexity, and task-specific performance, which are often difficult to compare directly. Furthermore, there is no common standard that unifies model architectures, sequence lengths, or evaluation protocols. Establishing a standardized benchmark is therefore critical to enable fair, reproducible cross-method comparisons and to facilitate systematic progress toward robust and reliable AS mitigation.

\subsection{Future Directions}

\textbf{Efficient and Lightweight AS Handling.}  
Ensuring computational efficiency in AS-related operations remains a critical priority. 
This encompasses lightweight detection of dynamic sinks \cite{yu2024unveiling,kang2025see}, efficient implementation of \textit{\textbf{Attention Redistribution}} \cite{tu2026attention}, low-latency execution of \textit{\textbf{Gated Attention Mechanisms}} \cite{qiu2025gated}, rapid geometric measure computation \cite{ruscio2025you}, and the development of \textit{\textbf{Modified Softmax Functions}} compatible with optimized attention kernels \cite{zuhri2025softpick}.
In architectures such as ViTs and MLLMs, where sinks frequently concentrate on uninformative background patches, the need for efficient handling is particularly pronounced \cite{darcet2024vision,kang2025see}. 
Future research should focus on efficient and lightweight AS handling methods, thereby enabling practical deployment of AS-aware strategies in large-scale models without compromising speed or scalability.

\textbf{Lightweight Adaptation for Pre-trained Models.}  
Mitigating AS without relying on full retraining is crucial for practical deployment.
Future research should focus on parameter-efficient adaptation techniques that integrate AS-aware components directly into pretrained models, avoiding the need to train from scratch. 
Promising strategies include the use of adapters, low-rank updates such as LoRA, and continual pre-training with inserted gates, modified softmax, learnable biases, or prefix tokens. 
The overarching objective is to maintain the original model’s functionality while effectively suppressing AS and minimizing activation outliers. 
Advancements in this direction would democratize AS mitigation, enabling widespread adoption across the extensive ecosystem of existing pretrained models.

\textbf{Theoretical Formalization of Training Dynamics.}  
A comprehensive theoretical framework is essential to elucidate the emergence, evolution, and functional role of AS during pre‑training \cite{an2025systematic,bondarenko2023quantizable}. 
Critical open questions include the mutual reinforcement mechanisms between attention scores and value states, the stabilization dynamics of \textbf{\textit{Outlier Circuits}} and others. 
Formalizing the interactions among Softmax constraints, optimization dynamics, and implicit bias formation would offer principled guidance for the design of effective interventions and enable reliable prediction of AS behavior. 
Advancements in this direction would reinforce both \textit{\textbf{Mechanistic Interpretation}} and \textit{\textbf{Strategic Mitigation}}.

\textbf{AS Handling in Emerging Architectures.}  
Beyond the architectures surveyed above, the Transformer landscape continues to evolve rapidly. Emerging paradigms, such as hybrid linear attention architectures \cite{qwenai2026,team2025kimi,yang2025gated} and 3D Transformers for spatial reasoning \cite{wang2025vggt,wang2025continuous,jin2026zipmap}, offer new frontiers for AS research. 
Investigating how AS manifest and interact with these architectural innovations represents a largely unexplored direction, with potential implications for efficiency, interpretability, and task-specific performance.

\textbf{Unified Theoretical Framework.}  
Existing research provides multiple valuable perspectives on AS \cite{bondarenko2023quantizable,an2025systematic,sun2024massive,ruscio2025you}. 
While each interpretation offers important insights, a coherent framework integrating these views remains absent. 
Such a framework would streamline the theoretical landscape, consolidate disparate findings, guide mechanistic interpretation, and enable principled mitigation design, accelerating progress and supporting systematic investigation of AS across diverse Transformer architectures.

\textbf{Standardized Benchmark for AS and Outlier Mitigation.}  
Evaluating the effectiveness of AS elimination and outlier suppression remains challenging due to the lack of widely adopted benchmarks. 
Different mitigation strategies cannot be fairly compared in terms of efficacy, computational overhead, parameter introduction, or other critical factors \cite{qiu2025gated,zuhri2025softpick,sun2024massive,park2025outlier}. 
Establishing a standardized benchmark would facilitate fair and reproducible comparisons across diverse mitigation strategies, accelerate the identification of the most effective approaches, and guide the design of robust and generalizable AS mitigation solutions.

\textbf{Systematic Cross‑Architecture and Cross‑Modal Investigation.}  
Techniques developed for AS in one domain often remain confined to that specific domain. 
For instance, \textit{\textbf{Gated Attention Mechanisms}} have been primarily validated in rapidly evolving LLMs, with limited exploration in vision transformers or multimodal architectures \cite{qiu2025gated,bu2025value}. 
Systematic studies on cross-architecture and cross-modal transfer are needed to determine which methods generalize effectively and which require adaptation. 
Such investigations would accelerate the design of universally robust solutions.

\textbf{Synergistic Integration of Multiple AS Handling Techniques.}  
Current AS handling methods often focus on individual strategies in isolation. 
Exploring the coordinated use of complementary techniques within the same overarching category may enhance efficiency, robustness, and generalizability beyond what each method achieves independently. 
Systematic investigation of such intra-category synergies represents a promising direction for designing hybrid approaches that surpass the capabilities of standalone methods.

\section{Conclusion}
\label{sec_8_Conclusion}

In this work, we present the first comprehensive survey of AS in Transformer architectures, systematically synthesizing over 180 studies across three dimensions: \textit{\textbf{Fundamental Utilization}}, \textit{\textbf{Mechanistic Interpretation}}, and \textit{\textbf{Strategic Mitigation}}. 
Our review reveals that AS profoundly influences training dynamics, model interpretability, and inference efficiency across diverse architectures. 
Empirical utilization strategies demonstrate how AS can be leveraged to improve performance, mechanistic studies elucidate its underlying causes and functional roles, and mitigation techniques provide effective approaches to control or suppress AS for enhanced robustness and low-bit deployment. 
Despite these advances, several challenges remain, including computational efficiency, the necessity of training from scratch, and an incomplete understanding of training dynamics. 
We highlight promising directions for future research, including efficient and lightweight AS handling, AS in emerging architectures, and standardized benchmarks for mitigating AS.
By integrating insights from utilization, interpretation, and mitigation, this survey establishes a foundation for understanding AS and guides the development of more robust and interpretable Transformer models.% Zunhai

%----------------------------------Section 9----------------------------------%
\section{Limitations}
\label{sec_9_Limitations}

Despite the broad scope of this survey, certain limitations should be noted. 
Our analysis primarily focuses on well-established Transformer architectures, including CLMs, LLMs, MLLMs, MoE LLMs, and ViTs, which have been extensively studied in prior literature. 
Emerging or specialized architectures, such as hybrid-linear attention models \cite{qwenai2026,team2025kimi,chen2026hybrid,yang2025gated}, VGGT \cite{wang2025vggt,zhuo2025streaming,jin2026zipmap} and others, are not comprehensively covered due to the limited availability of AS-related studies. 
Nevertheless, we believe that the insights and methodologies presented here are broadly applicable and can inform understanding across other model architectures. 
As research on novel architectures continues to expand, we will incorporate relevant studies to further enhance the comprehensiveness of this survey.% Zunhai

% \section{Acknowledgments}
% \input{Attention_Sink_in_Transformers/Acknowledgments/Acknowledgments}% Zunhai

%%%%%%%%%%%%%%%%%%%%%%%%%%%%%%%%%%%%%%%%%%%%%%%%%%%%%%%%%%%%

\appendix
\clearpage
\section{Comprehensive Overview of Surveyed Papers}
\label{Appendix:A}

%-----------------------------------None-----------------------------------%
\def\None{\textcolor{gray}{-}}
%-----------------------------------Utilization-----------------------------------%
\definecolor{colorUtilize}{RGB}{0, 51, 102} 
\newcommand{\UtilizeStyle}[1]{\textcolor{colorUtilize}{\textit{\footnotesize #1}}}
\def\UtiPreSink{\UtilizeStyle{\begin{tabular}[c]{@{}c@{}} Sink Token\\Preservation \end{tabular}}}
\def\UtiAttnRedis{\UtilizeStyle{\begin{tabular}[c]{@{}c@{}} Attention\\Redistribution\end{tabular}}}
\def\UtiPreixTrainable{\UtilizeStyle{\begin{tabular}[c]{@{}c@{}} Learnable\\Prefix Tokens\end{tabular}}}
\def\UtiSinkUtilize{\UtilizeStyle{\begin{tabular}[c]{@{}c@{}} Sink Tokens\\Repurposing\end{tabular}}}
%-----------------------------------Interpretation-----------------------------------%
\definecolor{colorInterpret}{RGB}{0, 100, 0}
\newcommand{\InterpretStyle}[1]{\textcolor{colorInterpret}{\textit{\footnotesize #1}}}
\def\InterpretSoftNoOp{\InterpretStyle{\begin{tabular}[c]{@{}c@{}} Softmax Limitations\\\& No-Op Theory\end{tabular}}}
\def\InterpretOutlier{\InterpretStyle{Outlier Circuits}}
\def\InterpretAttnBias{\InterpretStyle{\begin{tabular}[c]{@{}c@{}} Implicit\\Attention Bias\end{tabular}}}
\def\InterpretGeoAnchor{\InterpretStyle{Geometric Anchoring}}
\def\InterpretStructural{\InterpretStyle{Structural Bias}}
\def\InterpretAntiOverMix{\InterpretStyle{Anti-Overmixing}}
\def\InterpretSEEnergy{\InterpretStyle{\begin{tabular}[c]{@{}c@{}} Spectral-Energy\\Association\end{tabular}}}
\def\InterpretActDor{\InterpretStyle{\begin{tabular}[c]{@{}c@{}} Active-Dormant\\Attention\end{tabular}}}
\def\InterpretMixComRefine{\InterpretStyle{\begin{tabular}[c]{@{}c@{}} Mix-Compress-Refine\end{tabular}}}
\def\InterpretOutlierResca{\InterpretStyle{\begin{tabular}[c]{@{}c@{}} Outlier-Driven\\Rescaling\end{tabular}}}
\def\Interpretposink{\InterpretStyle{\begin{tabular}[c]{@{}c@{}} P0 Sink Circuit\end{tabular}}}
\def\Interpretnaivemoe{\InterpretStyle{\begin{tabular}[c]{@{}c@{}} Native MoE \end{tabular}}}
\def\Interpretentropic{\InterpretStyle{\begin{tabular}[c]{@{}c@{}} Entropic Optimal\\Transport \end{tabular}}}
\def\Interpretsecondsink{\InterpretStyle{\begin{tabular}[c]{@{}c@{}} Secondary Sink\\Formation \end{tabular}}}
\def\Interpretwaiver{\InterpretStyle{\begin{tabular}[c]{@{}c@{}} Waiver Phenomenon \end{tabular}}}

%-----------------------------------Elimination-----------------------------------%
\definecolor{colorEliminate}{RGB}{139, 0, 0}
\newcommand{\EliminateStyle}[1]{\textcolor{colorEliminate}{\textit{\footnotesize #1}}}
\def\EliminateGatedAttn{\EliminateStyle{\begin{tabular}[c]{@{}c@{}} Gated Attention\end{tabular}}}
\def\EliminateModiSoftmax{\EliminateStyle{\begin{tabular}[c]{@{}c@{}} Modified Softmax\end{tabular}}}
\def\EliminateLearnBias{\EliminateStyle{\begin{tabular}[c]{@{}c@{}} Learnable\\Attention Bias\end{tabular}}}
\def\EliminatePretrainInterv{\EliminateStyle{\begin{tabular}[c]{@{}c@{}} Pre-training\\Interventions\end{tabular}}}
\def\EliminateRisiGate{\EliminateStyle{\begin{tabular}[c]{@{}c@{}} Residual\\Gated Rescaling\end{tabular}}}
\def\EliminateTuningAux{\EliminateStyle{\begin{tabular}[c]{@{}c@{}} Tuning\\Auxiliary Loss\end{tabular}}}
\def\EliminateModiNorm{\EliminateStyle{\begin{tabular}[c]{@{}c@{}} Modified\\Norm Layer\end{tabular}}}
\def\EliminateValueModi{\EliminateStyle{\begin{tabular}[c]{@{}c@{}} Value Modification\end{tabular}}}
\def\EliminateArchitIsolation{\EliminateStyle{\begin{tabular}[c]{@{}c@{}} Architectural\\Isolation\end{tabular}}}

%-----------------------------------Application-----------------------------------%
\newcommand{\AppStyle}[1]{\textit{\footnotesize #1}}

\def\AppModelTrain{\AppStyle{Model Pre-training}}
\def\AppModelTune{\AppStyle{Model Tuning}}
\def\AppModelInfer{\AppStyle{Efficient Inference}}

\def\AppModelInter{\AppStyle{\begin{tabular}[c]{@{}c@{}}Mechanism\\Interpretability\end{tabular}}}
\def\AppHallu{\AppStyle{\begin{tabular}[c]{@{}c@{}}Reducing\\Hallucination\end{tabular}}}
\def\AppSafe{\AppStyle{Safety \& Robustness}}

\def\AppGenEnhan{\AppStyle{\begin{tabular}[c]{@{}c@{}}General Capability\\Enhancement\end{tabular}}}
\def\AppVisEnhan{\AppStyle{\begin{tabular}[c]{@{}c@{}}Multi-Modal\\Enhancement\end{tabular}}}
\def\AppLongEnhan{\AppStyle{\begin{tabular}[c]{@{}c@{}}Long-Context\\Enhancement\end{tabular}}}

%-----------------------------------Application-----------------------------------%

% \begin{landscape}
\begin{longtable}{@{}cccccccc@{}}
\caption{Summary of Surveyed Papers. 
Each paper is annotated with tags corresponding to specific aspects of \textbf{\textit{Fundamental Utilization}} (\S \ref{sec_3_Fundamental_Utilization}), \textbf{\textit{Mechanistic Interpretation}} (\S\ref{sec_4_Mechanistic_Interpretation}), and \textbf{\textit{Strategic Mitigation}} (\S\ref{sec_5_Strategic_Mitigation}) of AS. 
As most studies do not target all three key aspects, the symbol ``-'' denotes the absence of a particular dimension in a given work.
}
\label{tab:my-table}\\
\toprule
% \rowcolor[HTML]{FFCE93}
\textbf{Paper} & \textbf{\begin{tabular}[c]{@{}c@{}} \S\ref{sec_3_Fundamental_Utilization}\\Utilization\end{tabular}} & \textbf{\begin{tabular}[c]{@{}c@{}} \S\ref{sec_4_Mechanistic_Interpretation}\\Interpretation\end{tabular}} & \textbf{\begin{tabular}[c]{@{}c@{}} \S\ref{sec_5_Strategic_Mitigation}\\Mitigation\end{tabular}} & \textbf{\begin{tabular}[c]{@{}c@{}} \S\ref{sec_7_Applications}\\Applications\end{tabular}} & \textbf{Venue} & \textbf{Year} & \textbf{Link} \\* \midrule
\endfirsthead
\endhead
\multicolumn{8}{c}{\cellcolor[HTML]{EFEFEF}\textbf{Classical Language Models}} \\* \midrule
\cite{mutisya2026attention} & \UtiAttnRedis & \None & \None & \AppModelInter & ArXiv & 2026 & \href{https://arxiv.org/abs/2605.01229}{Link} \\ \hline
\cite{ruscio2025you} & \None  & \InterpretGeoAnchor & \None & \AppModelInter  & NeurIPS  & 2025 & \href{https://arxiv.org/abs/2508.02546}{Link} \\ \hline
\cite{li2025ctr} & \UtiPreixTrainable  & \InterpretGeoAnchor & \None & \AppGenEnhan & ArXiv & 2025 & \href{https://arxiv.org/abs/2508.03668}{Link} \\ \hline
\cite{bai2025does} & \UtiAttnRedis & \None & \None & \begin{tabular}[c]{@{}c@{}}\AppModelTune\\ \AppModelInter\end{tabular} & COLM & 2024 & \href{https://arxiv.org/abs/2410.05648}{Link} \\ \hline
\cite{bondarenko2023quantizable} & \None & \begin{tabular}[c]{@{}c@{}}\InterpretSoftNoOp\\ \InterpretOutlier\end{tabular} & \begin{tabular}[c]{@{}c@{}}\EliminateGatedAttn\\ \EliminateModiSoftmax\end{tabular} & \begin{tabular}[c]{@{}c@{}}\AppModelTrain\\ \AppModelInfer\end{tabular} & NeurIPS & 2023 & \href{https://arxiv.org/abs/2306.12929}{Link} \\ \hline
\cite{puccetti2022outlier} & \None & \InterpretOutlier & \None & \AppModelInter & EMNLP & 2022 & \href{https://arxiv.org/abs/2205.11380}{Link} \\ \hline
\cite{luo2021positional} & \None & \begin{tabular}[c]{@{}c@{}} \InterpretOutlier \\ \InterpretStructural\end{tabular} & \None & \AppModelInter & ACL & 2021 & \href{https://arxiv.org/abs/2011.04393}{Link} \\ \hline
\cite{kovaleva2021bert} & \None & \begin{tabular}[c]{@{}c@{}} \InterpretOutlier \\ \InterpretStructural\end{tabular} & \None & \AppModelInter & ACL & 2021 & \href{https://arxiv.org/abs/2105.06990}{Link} \\ \hline
\cite{bondarenko2021understanding} & \UtiPreSink & \InterpretOutlier & \None & \begin{tabular}[c]{@{}c@{}} \AppModelInfer \\ \AppModelInter\end{tabular} & EMNLP & 2021 & \href{https://arxiv.org/abs/2109.12948}{Link} \\ \hline
\cite{clark2019does} & \None & \InterpretStructural & \None & \AppModelInter & ACL & 2019 & \href{https://arxiv.org/abs/1906.04341}{Link} \\ \hline

\multicolumn{8}{c}{\cellcolor[HTML]{EFEFEF}\textbf{Large Language Models}} \\* \midrule
\cite{shi2026single} & \None & \begin{tabular}[c]{@{}c@{}}\InterpretOutlier\\ \InterpretStructural\end{tabular} & \EliminateModiNorm & \AppModelInter & ICML & 2026 & \href{https://arxiv.org/pdf/2605.08504v2}{Link} \\ \hline
\cite{liu2026slash} & \UtiAttnRedis & \InterpretSoftNoOp & \None & \begin{tabular}[c]{@{}c@{}}\AppModelInter\\ \AppGenEnhan\end{tabular} & ICML & 2026 & \href{https://arxiv.org/abs/2605.10503v1}{Link} \\ \hline
\cite{li2026structural} & \None & \begin{tabular}[c]{@{}c@{}}\InterpretSoftNoOp\\ \InterpretOutlier\\ \InterpretStructural\end{tabular} & \EliminateModiNorm & \begin{tabular}[c]{@{}c@{}}\AppModelTrain\\ \AppModelInter\end{tabular} & ICML & 2026 & \href{https://arxiv.org/abs/2605.06611}{Link} \\ \hline
\cite{deepseekai2026deepseekv4} & \None & \None & \EliminateLearnBias & \AppModelTrain & ArXiv & 2026 & \href{https://huggingface.co/deepseek-ai/DeepSeek-V4-Pro/blob/main/DeepSeek_V4.pdf}{Link} \\ \hline
\cite{sukenik2026sink} & \None & \begin{tabular}[c]{@{}c@{}}\InterpretSoftNoOp\\ \InterpretAntiOverMix\end{tabular} & \None & \AppModelInter & ArXiv & 2026 & \href{https://arxiv.org/abs/2605.08453}{Link} \\ \hline
\cite{ran2026mechanistic} & \None & \begin{tabular}[c]{@{}c@{}}\InterpretOutlier\\ \InterpretStructural\end{tabular} & \None & \AppModelInter & ArXiv & 2026 & \href{https://arxiv.org/abs/2604.14722}{Link} \\ \hline
\cite{liu2026sinkrouter} & \UtiSinkUtilize & \None & \None & \AppModelInfer & ArXiv & 2026 & \href{https://arxiv.org/abs/2604.16883}{Link} \\ \hline
\cite{sun2026sepseq} & \UtiPreixTrainable & \InterpretSoftNoOp & \None & \begin{tabular}[c]{@{}c@{}}\AppModelInfer\\ \AppLongEnhan\end{tabular} & ArXiv & 2026 & \href{https://arxiv.org/abs/2604.07737}{Link} \\ \hline
\cite{liu2026depth} & \UtiPreixTrainable & \InterpretStructural & \EliminatePretrainInterv & \AppModelTrain & ArXiv & 2026 & \href{https://arxiv.org/abs/2604.18128}{Link} \\ \hline
\cite{li2026paosc} & \UtiAttnRedis & \None & \None & \AppModelInfer & WWW & 2026 & \href{https://dl.acm.org/doi/abs/10.1145/3774904.3792911}{Link} \\ \hline
\cite{binkowski2026attention} & \UtiSinkUtilize & \None & \None & \AppSafe & ICML & 2026 & \href{https://arxiv.org/abs/2604.10697}{Link} \\ \hline
\cite{sun2026spike} & \None & \InterpretStructural & \None & \AppModelInter & ICML & 2026 & \href{https://arxiv.org/abs/2603.05498}{Link} \\ \hline
\cite{chen2026attention} & \None & \InterpretOutlier & \begin{tabular}[c]{@{}c@{}}\\ \EliminatePretrainInterv\\ \EliminateValueModi\end{tabular} & \AppModelInter & ArXiv & 2026 & \href{https://arxiv.org/abs/2603.17771}{Link} \\ \hline
\cite{ran2026attention} & \None & \InterpretSoftNoOp & \EliminateModiSoftmax & \AppModelInter & ArXiv & 2026 & \href{https://arxiv.org/abs/2603.11487}{Link} \\ \hline
\cite{wang2026summarize} & \UtiAttnRedis & \None & \None & \AppModelInfer & ArXiv & 2026 & \href{https://arxiv.org/abs/2603.11067}{Link} \\ \hline
\cite{peng2026attention} & \None & \begin{tabular}[c]{@{}c@{}}\InterpretStructural\\ \Interpretposink\end{tabular} & \None & \AppModelInter & ArXiv & 2026 & \href{https://arxiv.org/abs/2603.06591}{Link} \\ \hline
\cite{sun2026karma} & \None & \None & \EliminateTuningAux & \AppModelTune & ArXiv & 2026 & \href{https://arxiv.org/abs/2603.22779}{Link} \\ \hline
\cite{liu2026sinktrack} & \UtiPreSink & \None & \None & \begin{tabular}[c]{@{}c@{}} \AppLongEnhan \\ \AppHallu\end{tabular} & ICLR & 2026 & \href{https://openreview.net/forum?id=Gg1aPETCL6}{Link} \\ \hline
\cite{fu2026attention2} & \None & \Interpretnaivemoe & \EliminatePretrainInterv & \begin{tabular}[c]{@{}c@{}}\AppModelInter\\ \AppSafe\end{tabular} & ICML & 2026 & \href{https://arxiv.org/abs/2602.01203}{Link} \\ \hline
\cite{liu2026surgery} & \UtiSinkUtilize & \None & \EliminateTuningAux & \AppModelTune & ICML & 2026 & \href{https://arxiv.org/abs/2602.05228}{Link} \\ \hline
\cite{bae2026affine} & \None & \None & \EliminateModiSoftmax & \AppModelTrain & ICML & 2026 & \href{https://arxiv.org/abs/2602.23057}{Link} \\ \hline
\cite{hosseini2026innerq} & \UtiPreSink & \None & \None & \AppModelInfer & ArXiv & 2026 & \href{https://arxiv.org/abs/2602.23200}{Link} \\ \hline
\cite{liu2026krause} & \None & \None & \EliminateModiSoftmax & \begin{tabular}[c]{@{}c@{}}\AppModelInfer\\ \AppGenEnhan\end{tabular} & ICML & 2026 & \href{https://arxiv.org/abs/2602.11534}{Link} \\ \hline
\cite{li2026transformers} & \None & \InterpretStructural & \None & \AppModelInter & ArXiv & 2026 & \href{https://arxiv.org/abs/2602.05927}{Link} \\ \hline
\cite{fu2026attention} & \None & \InterpretSoftNoOp & \EliminateModiSoftmax & \begin{tabular}[c]{@{}c@{}}\AppModelInfer\\ \AppGenEnhan\end{tabular} & ArXiv & 2026 & \href{https://www.arxiv.org/abs/2601.00919}{Link} \\ \hline
\cite{xiao2026mimo} & \None & \None & \EliminateLearnBias & \begin{tabular}[c]{@{}c@{}}\AppModelTrain\\ \AppLongEnhan\end{tabular} & ArXiv & 2026 & \href{https://arxiv.org/abs/2601.02780}{Link} \\ \hline
\cite{qiu2026unifiedviewattentionresidual} & \None & \InterpretOutlierResca & \begin{tabular}[c]{@{}c@{}}\EliminateModiNorm\\ \EliminateRisiGate\end{tabular} & \AppModelTrain & ArXiv & 2026 & \href{https://arxiv.org/abs/2601.22966}{Link} \\ \hline
\cite{queipo2026attention} & \None & \begin{tabular}[c]{@{}c@{}}\InterpretOutlier\\ \InterpretMixComRefine\end{tabular} & \None & \AppModelInter & ICLR & 2026 & \href{https://arxiv.org/abs/2510.06477}{Link} \\ \hline
\cite{desai2026vattention} & \UtiPreSink & \None & \None & \AppModelInfer & ICLR & 2026 & \href{https://arxiv.org/abs/2510.05688}{Link} \\ \hline
\cite{su2026unveiling} & \None & \InterpretOutlier & \None & \AppModelInter & ICLR & 2026 & \href{https://arxiv.org/abs/2507.23279}{Link} \\ \hline
\cite{han2026zerotuning} & \UtiAttnRedis & \InterpretAttnBias & \None & \begin{tabular}[c]{@{}c@{}}\AppGenEnhan\\ \AppModelInter\end{tabular} & ICLR & 2026 & \href{https://arxiv.org/abs/2505.11739}{Link} \\ \hline
\cite{litman2026you} & \None & \Interpretentropic & \EliminateModiSoftmax & \AppGenEnhan & ArXiv & 2026 & \href{https://arxiv.org/abs/2601.15380}{Link} \\ \hline
\cite{sok2026garbage} & \UtiSinkUtilize & \None & \None & \AppModelInfer & ArXiv & 2026 & \href{https://arxiv.org/abs/2601.06787}{Link} \\ \hline
\cite{huang2026threshold} & \None & \InterpretSoftNoOp & \EliminateModiSoftmax & \AppLongEnhan & ArXiv & 2026 & \href{https://arxiv.org/abs/2601.12145}{Link} \\ \hline
\cite{wong2025existence} & \None & \Interpretsecondsink & \None & \AppModelInter & ArXiv & 2025 & \href{https://arxiv.org/abs/2512.22213}{Link} \\ \hline
\cite{yu2025sliding} & \UtiPreSink & \None & \None & \AppLongEnhan & ArXiv & 2025 & \href{https://arxiv.org/abs/2512.10411}{Link} \\ \hline
\cite{xiong2025dope} & \None & \InterpretStructural & \None & \begin{tabular}[c]{@{}c@{}}\AppLongEnhan\\ \AppModelInter\end{tabular} & ArXiv & 2025 & \href{https://arxiv.org/abs/2511.09146}{Link} \\ \hline
\cite{liang2025tweo} & \None & \InterpretOutlier & \EliminatePretrainInterv & \AppModelTrain & ArXiv & 2025 & \href{https://arxiv.org/abs/2511.23225}{Link} \\ \hline
\cite{salvatore2025lost} & \None & \InterpretStructural & \None & \AppModelInter & ArXiv & 2025 & \href{https://arxiv.org/abs/2510.10276}{Link} \\ \hline
\cite{yang2025cacheclip} & \UtiPreSink & \None & \None & \AppModelInfer & ArXiv & 2025 & \href{https://arxiv.org/abs/2510.10129}{Link} \\ \hline
\cite{shang2025forgetting} & \UtiSinkUtilize & \None & \None & \begin{tabular}[c]{@{}c@{}}\AppSafe\\ \AppModelInter\end{tabular} & ArXiv & 2025 & \href{https://arxiv.org/abs/2510.17021}{Link} \\ \hline
\cite{fang2025artificial} & \UtiPreSink & \None & \None & \AppModelInfer & ArXiv & 2025 & \href{https://arxiv.org/abs/2510.07318}{Link} \\ \hline
\cite{gu2025obcache} & \UtiPreSink & \None & \None & \AppModelInfer & ArXiv & 2025 & \href{https://arxiv.org/abs/2510.07651}{Link} \\ \hline
\cite{mu2025sals} & \UtiPreSink & \None & \None & \AppModelInfer & NeurIPS & 2025 & \href{https://arxiv.org/abs/2510.24273}{Link} \\ \hline
\cite{bu2025value} & \None & \begin{tabular}[c]{@{}c@{}}\InterpretSoftNoOp\\ \InterpretOutlier\end{tabular} & \EliminateGatedAttn & \begin{tabular}[c]{@{}c@{}}\AppModelInfer\\ \AppModelTrain\\ \AppModelInter\end{tabular} & ArXiv & 2025 & \href{https://arxiv.org/abs/2510.09017}{Link} \\ \hline
\cite{zhu2025ojakv} & \UtiPreSink & \None & \None & \AppModelInfer & ArXiv & 2025 & \href{https://arxiv.org/abs/2509.21623}{Link} \\ \hline
\cite{su2025kvsink} & \UtiPreSink & \begin{tabular}[c]{@{}c@{}}\InterpretSoftNoOp\\ \InterpretOutlier\end{tabular} & \None & \begin{tabular}[c]{@{}c@{}}\AppModelInfer\\ \AppModelInter\end{tabular} & COLM & 2025 & \href{https://arxiv.org/abs/2508.04257}{Link} \\ \hline
\cite{agarwal2025gpt} & \None & \None & \EliminateLearnBias & \AppModelTrain & ArXiv & 2025 & \href{https://arxiv.org/abs/2508.10925}{Link} \\ \hline
\cite{ruscio2025you} & \None & \InterpretGeoAnchor & \None & \AppModelInter & NeurIPS & 2025 & \href{https://arxiv.org/abs/2508.02546}{Link} \\ \hline
\cite{li2025ctr} & \UtiPreixTrainable & \InterpretGeoAnchor & \None & \AppGenEnhan & ArXiv & 2025 & \href{https://arxiv.org/abs/2508.03668}{Link} \\ \hline
\cite{kobyzev2025integral} & \None & \None & \EliminateModiSoftmax & \AppGenEnhan & EMNLP & 2025 & \href{https://arxiv.org/abs/2508.18387}{Link} \\ \hline
\cite{fu2025h2eal} & \UtiPreSink & \None & \None & \AppModelInfer & ICCAD & 2025 & \href{https://arxiv.org/abs/2508.16653}{Link} \\ \hline
\cite{wang2025position} & \UtiAttnRedis & \InterpretStructural & \None & \AppGenEnhan & EMNLP & 2025 & \href{https://arxiv.org/abs/2508.15709}{Link} \\ \hline
\cite{he2025trianglemix} & \UtiPreSink & \None & \None & \AppModelInfer & ArXiv & 2025 & \href{https://arxiv.org/abs/2507.21526}{Link} \\ \hline
\cite{shin2025orthorank} & \UtiPreSink & \InterpretGeoAnchor & \None & \begin{tabular}[c]{@{}c@{}}\AppModelInfer\\ \AppModelInter\end{tabular} & ICML & 2025 & \href{https://arxiv.org/abs/2507.03865}{Link} \\ \hline
\cite{yang2025earn} & \UtiPreixTrainable & \None & \None & \begin{tabular}[c]{@{}c@{}}\AppModelInfer\\ \AppModelInter\end{tabular} & KDD & 2025 & \href{https://arxiv.org/abs/2507.00715}{Link} \\ \hline
\cite{qi2025deltallm} & \UtiPreSink & \None & \None & \AppModelInfer & ArXiv & 2025 & \href{https://arxiv.org/abs/2507.19608}{Link} \\ \hline
\cite{park2025outlier} & \None & \InterpretOutlier & \EliminatePretrainInterv & \AppModelTrain & ACL & 2025 & \href{https://arxiv.org/abs/2506.19697}{Link} \\ \hline
\cite{yu2025two} & \UtiPreSink & \None & \None & \begin{tabular}[c]{@{}c@{}}\AppModelInfer\\ \AppModelInter\end{tabular} & ArXiv & 2025 & \href{https://arxiv.org/abs/2506.12220}{Link} \\ \hline
\cite{yao2025learn} & \UtiPreSink & \None & \None & \AppModelInfer & ArXiv & 2025 & \href{https://arxiv.org/abs/2506.15704}{Link} \\ \hline
\cite{qiu2025gated} & \None & \InterpretSoftNoOp & \EliminateGatedAttn & \begin{tabular}[c]{@{}c@{}}\AppModelInfer\\ \AppModelTrain\\ \AppLongEnhan\\ \AppModelInter\end{tabular} & NeurIPS & 2025 & \href{https://arxiv.org/abs/2505.06708}{Link} \\ \hline
\cite{willette2025delta} & \UtiPreSink & \None & \None & \AppModelInfer & ArXiv & 2025 & \href{https://arxiv.org/abs/2505.11254}{Link} \\ \hline
\cite{zuhri2025softpick} & \None & \begin{tabular}[c]{@{}c@{}}\InterpretSoftNoOp\\ \InterpretOutlier\end{tabular} & \EliminateModiSoftmax & \begin{tabular}[c]{@{}c@{}}\AppModelInfer\\ \AppModelTrain\\ \AppModelInter\end{tabular} & ArXiv & 2025 & \href{https://arxiv.org/abs/2504.20966}{Link} \\ \hline
\cite{barbero2025llms} & \None & \InterpretAntiOverMix & \None & \AppModelInter & COLM & 2025 & \href{https://arxiv.org/abs/2504.02732}{Link} \\ \hline
\cite{park2025keydiff} & \UtiSinkUtilize & \InterpretGeoAnchor & \None & \begin{tabular}[c]{@{}c@{}}\AppModelInfer\\ \AppModelInter\end{tabular} & NeurIPS & 2025 & \href{https://arxiv.org/abs/2504.15364}{Link} \\ \hline
\cite{chen2025edgeinfinite} & \UtiPreSink & \None & \None & \AppLongEnhan & ACL & 2025 & \href{https://arxiv.org/abs/2503.22196}{Link} \\ \hline
\cite{yona2025interpreting} & \None & \InterpretOutlier & \None & \begin{tabular}[c]{@{}c@{}}\AppModelInter\\ \AppSafe\end{tabular} & ICML & 2025 & \href{https://arxiv.org/abs/2503.08908}{Link} \\ \hline
\cite{xiao2025efficient} & \UtiPreSink & \None & \None & \AppModelInfer & ACL & 2025 & \href{https://arxiv.org/abs/2503.08640}{Link} \\ \hline
\cite{wu2025emergence} & \None & \InterpretStructural & \None & \AppModelInter & ICML & 2025 & \href{https://arxiv.org/abs/2502.01951}{Link} \\ \hline
\cite{an2025systematic} & \None & \begin{tabular}[c]{@{}c@{}}\InterpretOutlier\\ \InterpretAttnBias\end{tabular} & \EliminateLearnBias & \begin{tabular}[c]{@{}c@{}}\AppModelTrain\\ \AppModelInter\end{tabular} & ICLR & 2025 & \href{https://arxiv.org/abs/2502.06415}{Link} \\ \hline
\cite{fu2025sliding} & \None & \InterpretSoftNoOp & \EliminateModiSoftmax & \AppLongEnhan & ArXiv & 2025 & \href{https://arxiv.org/abs/2502.18845}{Link} \\ \hline
\cite{jin2025massive} & \None & \InterpretOutlier & \None & \AppModelInter & ArXiv & 2025 & \href{https://arxiv.org/abs/2502.01563}{Link} \\ \hline
\cite{zhangattention} & \None & \InterpretOutlier & \None & \AppModelInter & NeurIPS & 2025 & \href{https://arxiv.org/abs/2502.00919}{Link} \\ \hline
\cite{kamoda2025weight} & \None & \InterpretStructural & \None & \AppModelInter & NAACL & 2025 & \href{https://arxiv.org/abs/2501.15754}{Link} \\ \hline
\cite{su2025rotatekv} & \UtiPreSink & \InterpretOutlier & \None & \AppModelInfer & IJCAI & 2025 & \href{https://arxiv.org/abs/2501.16383}{Link} \\ \hline
\cite{he2025task} & \UtiPreSink & \None & \None & \AppModelInfer & ArXiv & 2025 & \href{https://arxiv.org/abs/2501.15113}{Link} \\ \hline
\cite{wang2025llms} & \UtiPreSink & \None & \None & \AppModelInfer & ArXiv & 2025 & \href{https://arxiv.org/abs/2503.08879}{Link} \\ \hline
\cite{deng2025unigist} & \UtiPreixTrainable & \None & \None & \AppModelInfer & NeurIPS & 2025 & \href{https://arxiv.org/abs/2509.15763}{Link} \\ \hline
\cite{shutova2025cache} & \UtiPreSink & \None & \None & \AppModelInfer & ICML & 2025 & \href{https://arxiv.org/abs/2501.19392}{Link} \\ \hline
\cite{gao2025softplus} & \None & \None & \EliminateModiSoftmax & \begin{tabular}[c]{@{}c@{}}\AppGenEnhan\\ \AppLongEnhan\end{tabular} & ICML & 2026 & \href{https://arxiv.org/abs/2501.13428}{Link} \\ \hline
\cite{qwenai2026} & \None & \None & \EliminateGatedAttn & \AppModelTrain & ArXiv & 2025 & \href{https://qwen.ai/blog?id=4074cca80393150c248e508aa62983f9cb7d27cd&from=research.latest-advancements-list}{Link} \\ \hline
\cite{khalil2025singular} & \None & \InterpretGeoAnchor & \None & \begin{tabular}[c]{@{}c@{}}\AppModelInfer\\ \AppModelInter\end{tabular} & NeurIPS & 2025 & \href{https://openreview.net/forum?id=AJgzo9JhtM}{Link} \\ \hline
\cite{zhang2025attention} & \UtiPreSink & \InterpretSoftNoOp & \None & \begin{tabular}[c]{@{}c@{}}\AppLongEnhan\\ \AppModelInter\end{tabular} & ACL & 2025 & \href{https://arxiv.org/abs/2412.16545}{Link} \\ \hline
\cite{liu2025sgd} & \UtiPreSink & \None & \None & \AppModelInfer & NeurIPS & 2025 & \href{https://openreview.net/pdf?id=XM31M3uSUU}{Link} \\ \hline
\cite{zeng2025subkv} & \UtiPreSink & \None & \None & \AppModelInfer & ArXiv & 2025 & \href{https://onlinelibrary.wiley.com/doi/abs/10.1002/spe.3422}{Link} \\ \hline
\cite{hanevolving} & \UtiPreSink & \None & \None & \AppModelInfer & ArXiv & 2025 & \href{https://openreview.net/forum?id=YPjcyMzhE2}{Link} \\ \hline
\cite{zhang2025anchor} & \UtiPreSink & \InterpretGeoAnchor & \None & \begin{tabular}[c]{@{}c@{}}\AppModelInfer\\ \AppModelInter\end{tabular} & ArXiv & 2025 & \href{https://arxiv.org/abs/2411.06680}{Link} \\ \hline
\cite{lin2025look} & \UtiSinkUtilize & \None & \None & \AppGenEnhan & ACL & 2025 & \href{https://aclanthology.org/2025.acl-long.1113.pdf}{Link} \\ \hline
\cite{zhang2025leank} & \UtiPreSink & \None & \None & \AppModelInfer & EMNLP & 2025 & \href{https://arxiv.org/abs/2508.02215}{Link} \\ \hline
\cite{hongvariance} & \None & \InterpretSoftNoOp & \EliminateModiSoftmax & \begin{tabular}[c]{@{}c@{}}\AppModelTrain\\ \AppModelInter\end{tabular} & EMNLP & 2025 & \href{https://aclanthology.org/2025.emnlp-main.421v2.pdf}{Link} \\ \hline
\cite{xiang2025dfrot} & \None & \InterpretOutlier & \EliminatePretrainInterv & \AppModelInfer & COLM & 2025 & \href{https://arxiv.org/abs/2412.00648}{Link} \\ \hline
\cite{wu2025scope} & \UtiPreSink & \None & \None & \AppModelInfer & ArXiv & 2025 & \href{https://arxiv.org/abs/2412.13649}{Link} \\ \hline
\cite{acharya2025star} & \UtiPreSink & \None & \None & \begin{tabular}[c]{@{}c@{}}\AppModelInfer\\ \AppLongEnhan\end{tabular} & ICML & 2025 & \href{https://arxiv.org/abs/2411.17116}{Link} \\ \hline
\cite{gu2025attention} & \None & \begin{tabular}[c]{@{}c@{}}\InterpretSoftNoOp\\ \InterpretAttnBias\end{tabular} & \begin{tabular}[c]{@{}c@{}}\EliminateModiSoftmax\\ \EliminateLearnBias\end{tabular} & \AppModelInter & ICLR & 2025 & \href{https://arxiv.org/abs/2410.10781}{Link} \\ \hline
\cite{hu2025epic} & \UtiPreixTrainable & \InterpretOutlier & \None & \AppModelInfer & ICML & 2025 & \href{https://arxiv.org/abs/2410.15332}{Link} \\ \hline
\cite{guo2024active} & \None & \begin{tabular}[c]{@{}c@{}}\InterpretSoftNoOp\\ \InterpretOutlier\\ \InterpretActDor\end{tabular} & \None & \AppModelInter & ArXiv & 2024 & \href{https://arxiv.org/abs/2410.13835}{Link} \\ \hline
\cite{zhao2024buzz} & \UtiPreSink & \None & \None & \AppModelInfer & ArXiv & 2024 & \href{https://arxiv.org/abs/2410.23079}{Link} \\ \hline
\cite{xiao2025duoattention} & \UtiPreSink & \None & \None & \begin{tabular}[c]{@{}c@{}}\AppModelInfer\\ \AppLongEnhan\end{tabular} & ICLR & 2025 & \href{https://arxiv.org/abs/2410.10819}{Link} \\ \hline
\cite{chen2024prefixquant} & \UtiPreixTrainable & \None & \None & \AppModelInfer & ArXiv & 2024 & \href{https://arxiv.org/abs/2410.05265}{Link} \\ \hline
\cite{ge2025little} & \UtiPreSink & \None & \None & \begin{tabular}[c]{@{}c@{}}\AppModelInfer\\ \AppModelTrain\\ \AppLongEnhan\end{tabular} & ICLR & 2025 & \href{https://arxiv.org/abs/2410.01485}{Link} \\ \hline
\cite{chen2025magicpig} & \UtiPreSink & \InterpretGeoAnchor & \None & \begin{tabular}[c]{@{}c@{}}\AppModelInfer\\ \AppLongEnhan\\ \AppModelInter\end{tabular} & ICLR & 2025 & \href{https://arxiv.org/abs/2410.16179}{Link} \\ \hline
\cite{elawady2024relic} & \UtiPreSink & \None & \None & \AppLongEnhan & ArXiv & 2024 & \href{https://arxiv.org/abs/2410.02751}{Link} \\ \hline
\cite{kaul2025attention} & \None & \begin{tabular}[c]{@{}c@{}}\InterpretSoftNoOp\\ \InterpretOutlier\end{tabular} & \begin{tabular}[c]{@{}c@{}}\EliminateModiSoftmax\\ \EliminatePretrainInterv\end{tabular} & \begin{tabular}[c]{@{}c@{}}\AppModelInfer\\ \AppModelInter\end{tabular} & ICLR & 2025 & \href{https://arxiv.org/abs/2410.17174}{Link} \\ \hline
\cite{jo2024a2sf} & \UtiPreSink & \None & \None & \AppModelInfer & ArXiv & 2024 & \href{https://arxiv.org/abs/2407.20485}{Link} \\ \hline
\cite{yan2024unveiling} & \None & \Interpretwaiver & \EliminateValueModi & \AppModelInter & ArXiv & 2024 & \href{https://arxiv.org/abs/2407.01601}{Link} \\ \hline
\cite{jiang2024minference} & \UtiPreSink & \None & \None & \AppModelInfer & NeurIPS & 2024 & \href{https://arxiv.org/abs/2407.02490}{Link} \\ \hline
\cite{barbero2024transformers} & \None & \InterpretAntiOverMix & \None & \AppModelInter & NeurIPS & 2024 & \href{https://arxiv.org/abs/2406.04267}{Link} \\ \hline
\cite{son2024prefixing} & \UtiPreixTrainable & \InterpretOutlier & \None & \AppModelInfer & EMNLP & 2024 & \href{https://arxiv.org/abs/2406.12016}{Link} \\ \hline
\cite{zhang2024sinklora} & \UtiPreSink & \None & \None & \begin{tabular}[c]{@{}c@{}}\AppModelInfer\\ \AppLongEnhan\end{tabular} & ArXiv & 2024 & \href{https://arxiv.org/abs/2406.05678}{Link} \\ \hline
\cite{cai2024pyramidkv} & \UtiPreSink & \None & \None & \AppModelInfer & ArXiv & 2024 & \href{https://arxiv.org/abs/2406.02069}{Link} \\ \hline
\cite{guo2024attention} & \UtiPreSink & \InterpretOutlier & \None & \AppModelInfer & EMNLP & 2024 & \href{https://arxiv.org/abs/2406.12335}{Link} \\ \hline
\cite{yu2024unveiling} & \UtiAttnRedis & \InterpretStructural & \None & \AppGenEnhan & ICML & 2025 & \href{https://arxiv.org/abs/2406.15765}{Link} \\ \hline
\cite{dong2024exploring} & \None & \InterpretGeoAnchor & \None & \AppLongEnhan & NeurIPS & 2024 & \href{https://arxiv.org/abs/2405.18009}{Link} \\ \hline
\cite{duanmu2024skvq} & \UtiPreSink & \None & \None & \AppModelInfer & COLM & 2024 & \href{https://arxiv.org/abs/2405.06219}{Link} \\ \hline
\cite{liu2024intactkv} & \UtiPreSink & \InterpretOutlier & \None & \AppModelInfer & ACL & 2024 & \href{https://arxiv.org/abs/2403.01241}{Link} \\ \hline
\cite{cancedda2024spectral} & \None & \InterpretSEEnergy & \None & \AppModelInter & ACL & 2024 & \href{https://arxiv.org/abs/2402.09221}{Link} \\ \hline
\cite{sun2024massive} & \None & \begin{tabular}[c]{@{}c@{}}\InterpretOutlier\\ \InterpretAttnBias\end{tabular} & \EliminateLearnBias & \AppModelInter & COLM & 2024 & \href{https://arxiv.org/abs/2402.17762}{Link} \\ \hline
\cite{liao2024free} & \UtiPreSink & \InterpretOutlier & \EliminateModiSoftmax & \begin{tabular}[c]{@{}c@{}}\AppModelInfer\\ \AppModelInter\end{tabular} & ArXiv & 2024 & \href{https://arxiv.org/abs/2402.12102}{Link} \\ \hline
\cite{xiao2024infllm} & \UtiPreSink & \None & \None & \AppModelInfer & NeurIPS & 2024 & \href{https://arxiv.org/abs/2402.04617}{Link} \\ \hline
\cite{wang2024greater} & \UtiPreSink & \None & \None & \begin{tabular}[c]{@{}c@{}}\AppModelInfer\\ \AppLongEnhan\end{tabular} & COLM & 2024 & \href{https://arxiv.org/abs/2401.11504}{Link} \\ \hline
\cite{hooper2024kvquant} & \UtiPreSink & \None & \None & \AppModelInfer & NeurIPS & 2024 & \href{https://arxiv.org/abs/2401.18079}{Link} \\ \hline
\cite{sandal2024zero} & \UtiPreSink & \None & \None & \AppGenEnhan & ArXiv & 2024 & \href{https://arxiv.org/abs/2401.08683}{Link} \\ \hline
\cite{gurnee2024universal} & \None & \InterpretOutlier & \None & \AppModelInter & TMLR & 2024 & \href{https://arxiv.org/abs/2401.12181}{Link} \\ \hline
\cite{zhang2024q} & \UtiPreSink & \None & \None & \AppModelInfer & MLSys & 2024 & \href{https://proceedings.mlsys.org/paper_files/paper/2024/hash/bbb7506579431a85861a05fff048d3e1-Abstract-Conference.html}{Link} \\ \hline
\cite{li2024streamingdialogue} & \UtiSinkUtilize & \None & \None & \begin{tabular}[c]{@{}c@{}}\AppModelInfer\\ \AppLongEnhan\end{tabular} & NeurIPS & 2024 & \href{https://arxiv.org/abs/2403.08312}{Link} \\ \hline
\cite{chen2024rotary} & \None & \begin{tabular}[c]{@{}c@{}}\InterpretSoftNoOp\\ \InterpretStructural\end{tabular} & \None & \begin{tabular}[c]{@{}c@{}}\AppModelInter\\ \AppModelTune\end{tabular} & NeurIPS & 2024 & \href{https://proceedings.neurips.cc/paper_files/paper/2024/hash/61f425da6e0a201b8fe1454601abfba5-Abstract-Conference.html}{Link} \\ \hline
\cite{ge2023model} & \UtiPreSink & \None & \None & \AppModelInfer & ICLR & 2024 & \href{https://arxiv.org/abs/2310.01801}{Link} \\ \hline
\cite{xiao2024efficient} & \begin{tabular}[c]{@{}c@{}}\UtiPreSink\\ \UtiPreixTrainable\end{tabular} & \InterpretSoftNoOp & \None & \begin{tabular}[c]{@{}c@{}}\AppModelInfer\\ \AppLongEnhan\\ \AppModelInter\end{tabular} & ICLR & 2024 & \href{https://arxiv.org/abs/2309.17453}{Link} \\ \hline
\cite{han2024lm} & \UtiPreSink & \InterpretSoftNoOp & \None & \AppModelInfer & NAACL & 2024 & \href{https://arxiv.org/abs/2308.16137}{Link} \\ \hline
\cite{zhang2023h2o} & \UtiPreSink & \None & \None & \AppModelInfer & NeurIPS & 2023 & \href{https://arxiv.org/abs/2306.14048}{Link} \\ \hline
\cite{geshkovski2023emergence} & \None & \InterpretAntiOverMix & \None & \AppModelInter & NeurIPS & 2023 & \href{https://arxiv.org/abs/2305.05465}{Link} \\ \hline

\multicolumn{8}{c}{\cellcolor[HTML]{EFEFEF}\textbf{Mixture-of-Experts Large Language Models}} \\* \midrule
\cite{deepseekai2026deepseekv4} & \None & \None & \EliminateLearnBias & \AppModelTrain & ArXiv & 2026 & \href{https://huggingface.co/deepseek-ai/DeepSeek-V4-Pro/blob/main/DeepSeek_V4.pdf}{Link} \\ \hline
\cite{xiao2026mimo} & \None & \None & \EliminateLearnBias & \begin{tabular}[c]{@{}c@{}}\AppModelTrain\\ \AppLongEnhan\end{tabular} & ArXiv & 2026 & \href{https://arxiv.org/abs/2601.02780}{Link} \\ \hline
\cite{su2026unveiling} & \None & \InterpretOutlier & \None & \AppModelInter & ICLR & 2026 & \href{https://arxiv.org/abs/2507.23279}{Link} \\ \hline
\cite{team2025longcat} & \None & \None & \EliminatePretrainInterv & \AppModelTrain & ArXiv & 2025 & \href{https://arxiv.org/abs/2509.01322}{Link} \\ \hline
\cite{agarwal2025gpt} & \None & \None & \EliminateLearnBias & \AppModelTrain & ArXiv & 2025 & \href{https://arxiv.org/abs/2508.10925}{Link} \\ \hline
\cite{fu2025h2eal} & \UtiPreSink & \None & \None & \AppModelInfer & ICCAD & 2025 & \href{https://arxiv.org/abs/2508.16653}{Link} \\ \hline
\cite{qiu2025gated} & \None & \InterpretSoftNoOp & \EliminateGatedAttn & \begin{tabular}[c]{@{}c@{}}\AppModelInfer\\ \AppModelTrain\\ \AppLongEnhan\\ \AppModelInter\end{tabular} & NeurIPS & 2025 & \href{https://arxiv.org/abs/2505.06708}{Link} \\ \hline
\cite{qwenai2026} & \None & \None & \EliminateGatedAttn & \AppModelTrain & ArXiv & 2026 & \href{https://qwen.ai/blog?id=4074cca80393150c248e508aa62983f9cb7d27cd&from=research.latest-advancements-list}{Link} \\ \hline
\cite{sun2024massive} & \None & \begin{tabular}[c]{@{}c@{}}\InterpretOutlier\\ \InterpretAttnBias\end{tabular} & \EliminateLearnBias & \AppModelInter & COLM & 2024 & \href{https://arxiv.org/abs/2402.17762}{Link} \\ \hline

\multicolumn{8}{c}{\cellcolor[HTML]{EFEFEF}\textbf{Multi-Modal Large Language Models}} \\* \midrule
\cite{xiao2026not} & \UtiAttnRedis & \None & \None & \AppHallu & ArXiv & 2026 & \href{https://arxiv.org/abs/2605.10676v1}{Link} \\ \hline
\cite{chen2026vocabulary} & \UtiSinkUtilize & \None & \None & \AppHallu & ArXiv & 2026 & \href{https://arxiv.org/abs/2605.10622v1}{Link} \\ \hline
\cite{jung2026probing} & \UtiSinkUtilize & \None & \None & \AppHallu & ArXiv & 2026 & \href{https://arxiv.org/abs/2605.10815v2}{Link} \\ \hline
\cite{choi2026sinks} & \UtiAttnRedis & \None & \None & \AppHallu & ArXiv & 2026 & \href{https://arxiv.org/abs/2604.03316}{Link} \\ \hline
\cite{kim2026sink} & \UtiAttnRedis & \None & \None & \begin{tabular}[c]{@{}c@{}}\AppModelInfer\\ \AppVisEnhan\end{tabular} & ArXiv & 2026 & \href{https://arxiv.org/abs/2604.20937}{Link} \\ \hline
\cite{shukla2026sage} & \UtiSinkUtilize & \None & \None & \AppHallu & ArXiv & 2026 & \href{https://arxiv.org/abs/2603.27898}{Link} \\ \hline
\cite{liu2026sinktrack} & \UtiPreSink & \None & \None & \begin{tabular}[c]{@{}c@{}}\AppLongEnhan\\ \AppHallu\end{tabular} & ICLR & 2026 & \href{https://arxiv.org/abs/2604.10027}{Link} \\ \hline
\cite{lyu2026revealing} & \UtiAttnRedis & \None & \None & \begin{tabular}[c]{@{}c@{}}\AppVisEnhan\\ \AppHallu\end{tabular} & ArXiv & 2026 & \href{https://arxiv.org/abs/2602.15556}{Link} \\ \hline
\cite{jiang2026kvsmooth} & \UtiAttnRedis & \None & \None & \AppHallu & ArXiv & 2026 & \href{https://arxiv.org/abs/2602.04268}{Link} \\ \hline
\cite{tu2026attention} & \UtiAttnRedis & \None & \None & \begin{tabular}[c]{@{}c@{}}\AppHallu\\ \AppVisEnhan\end{tabular} & IJCV & 2026 & \href{https://arxiv.org/abs/2503.08342}{Link} \\ \hline
\cite{luo2026sink} & \UtiPreSink & \None & \None & \begin{tabular}[c]{@{}c@{}}\AppVisEnhan\\ \AppModelInter\end{tabular} & ICLR & 2026 & \href{https://arxiv.org/abs/2510.08510}{Link} \\ \hline
\cite{chen2025omnisparse} & \UtiSinkUtilize & \InterpretGeoAnchor & \None & \begin{tabular}[c]{@{}c@{}}\AppModelInfer\\ \AppModelInter\end{tabular} & ArXiv & 2025 & \href{https://arxiv.org/abs/2511.12201}{Link} \\ \hline
\cite{cappellazzo2025mitigating} & \None & \InterpretOutlier & \EliminateTuningAux & \begin{tabular}[c]{@{}c@{}}\AppGenEnhan\\ \AppModelInter\end{tabular} & ArXiv & 2025 & \href{https://arxiv.org/abs/2510.22603}{Link} \\ \hline
\cite{aman2025bitmar} & \UtiPreSink & \None & \None & \AppModelInfer & EMNLP & 2025 & \href{https://arxiv.org/abs/2510.10560}{Link} \\ \hline
\cite{khaki2025sparsevila} & \UtiPreSink & \None & \None & \begin{tabular}[c]{@{}c@{}}\AppModelInfer\\ \AppVisEnhan\end{tabular} & CVPR & 2025 & \href{https://arxiv.org/abs/2510.17777}{Link} \\ \hline
\cite{qi2025capturing} & \UtiAttnRedis & \None & \None & \AppHallu & ICML & 2026 & \href{https://arxiv.org/abs/2510.22067}{Link} \\ \hline
\cite{fan2025visipruner} & \UtiPreSink & \None & \None & \begin{tabular}[c]{@{}c@{}}\AppModelInfer\\ \AppModelInter\end{tabular} & EMNLP & 2025 & \href{https://arxiv.org/abs/2510.17205}{Link} \\ \hline
\cite{kang2025pevlm} & \UtiPreSink & \None & \None & \AppModelInfer & ArXiv & 2025 & \href{https://arxiv.org/abs/2506.19651}{Link} \\ \hline
\cite{baek2025large} & \UtiAttnRedis & \None & \None & \AppVisEnhan & EMNLP & 2025 & \href{https://arxiv.org/abs/2505.15865}{Link} \\ \hline
\cite{jiao2025don} & \UtiAttnRedis & \None & \None & \AppHallu & ArXiv & 2025 & \href{https://arxiv.org/abs/2504.09456}{Link} \\ \hline
\cite{kang2025see} & \UtiAttnRedis & \InterpretOutlier & \None & \begin{tabular}[c]{@{}c@{}}\AppVisEnhan\\ \AppModelInter\end{tabular} & ICLR & 2025 & \href{https://arxiv.org/abs/2503.03321}{Link} \\ \hline
\cite{su2025akvq} & \UtiPreSink & \InterpretOutlier & \None & \AppModelInfer & ICME & 2025 & \href{https://arxiv.org/abs/2501.15021}{Link} \\ \hline
\cite{zhuang2025vasparse} & \UtiAttnRedis & \None & \None & \AppHallu & CVPR & 2025 & \href{https://arxiv.org/abs/2501.06553}{Link} \\ \hline
\cite{wang2025mirage} & \UtiSinkUtilize & \None & \None & \begin{tabular}[c]{@{}c@{}}\AppSafe\\ \AppHallu\end{tabular} & USENIX Security & 2025 & \href{https://arxiv.org/abs/2501.15269}{Link} \\ \hline
\cite{zhang2026drives} & \None & \begin{tabular}[c]{@{}c@{}}\InterpretOutlier\\ \InterpretStructural\end{tabular} & \EliminateModiSoftmax & \begin{tabular}[c]{@{}c@{}}\AppVisEnhan\\ \AppHallu\\ \AppModelInter\end{tabular} & ArXiv & 2025 & \href{https://github.com/zhangbaijin/Massive-activations-VLMs}{Link} \\ \hline
\cite{zhang2025shallow} & \begin{tabular}[c]{@{}c@{}}\UtiAttnRedis\\ \UtiSinkUtilize\end{tabular} & \None & \None & \AppHallu & EMNLP & 2025 & \href{https://aclanthology.org/2025.emnlp-main.174/}{Link} \\ \hline
\cite{chenvocabulary} & \UtiAttnRedis & \None & \None & \AppHallu & ArXiv & 2025 & \href{https://openreview.net/forum?id=8o7iV9l27T}{Link} \\ \hline
\cite{lee2025tale} & \UtiPreSink & \None & \None & \AppModelInfer & ArXiv & 2025 & \href{https://direct.mit.edu/tacl/article/doi/10.1162/TACL.a.39/133612}{Link} \\ \hline
\cite{zhang2024seeing} & \begin{tabular}[c]{@{}c@{}}\UtiAttnRedis\\ \UtiSinkUtilize\end{tabular} & \None & \None & \AppHallu & ArXiv & 2024 & \href{https://arxiv.org/abs/2411.09968}{Link} \\ \hline
\cite{yang2025seed} & \UtiPreSink & \None & \None & \begin{tabular}[c]{@{}c@{}}\AppLongEnhan\\ \AppVisEnhan\end{tabular} & CVPR & 2025 & \href{https://arxiv.org/abs/2407.08683}{Link} \\ \hline

\multicolumn{8}{c}{\cellcolor[HTML]{EFEFEF}\textbf{Vision Transformers}} \\* \midrule
\cite{wang2026vit} & \UtiPreixTrainable & \None & \None & \AppVisEnhan & ArXiv & 2026 & \href{https://www.arxiv.org/abs/2602.08071}{Link} \\ \hline
\cite{liu2026krause} & \None & \None & \EliminateModiSoftmax & \begin{tabular}[c]{@{}c@{}}\AppModelInfer\\ \AppGenEnhan\end{tabular} & ICML & 2026 & \href{https://arxiv.org/abs/2602.11534}{Link} \\ \hline
\cite{simeoni2025dinov3} & \UtiPreixTrainable & \None & \None & \begin{tabular}[c]{@{}c@{}}\AppModelTrain\\ \AppVisEnhan\end{tabular} & ArXiv & 2025 & \href{https://arxiv.org/abs/2508.10104}{Link} \\ \hline
\cite{lu2025artifacts} & \UtiSinkUtilize & \None & \None & \begin{tabular}[c]{@{}c@{}}\AppModelInfer\\ \AppVisEnhan\end{tabular} & ArXiv & 2025 & \href{https://arxiv.org/abs/2507.16018}{Link} \\ \hline
\cite{xiao2025focus} & \UtiPreixTrainable & \None & \None & \AppVisEnhan & ArXiv & 2025 & \href{https://arxiv.org/abs/2507.14787}{Link} \\ \hline
\cite{jiang2025vision} & \UtiAttnRedis & \None & \None & \begin{tabular}[c]{@{}c@{}}\AppModelInter\\ \AppVisEnhan\end{tabular} & CVPR & 2025 & \href{https://arxiv.org/abs/2506.08010}{Link} \\ \hline
\cite{lappe2025register} & \UtiPreixTrainable & \None & \None & \AppModelInter & NeurIPS & 2025 & \href{https://arxiv.org/abs/2505.05892}{Link} \\ \hline
\cite{chen2025vision} & \UtiPreixTrainable & \None & \None & \AppVisEnhan & NeurIPS & 2025 & \href{https://arxiv.org/abs/2505.21501}{Link} \\ \hline
\cite{feng2026edit} & \None & \None & \EliminateArchitIsolation & \begin{tabular}[c]{@{}c@{}}\AppVisEnhan\\ \AppModelInter\end{tabular} & ArXiv & 2025 & \href{https://arxiv.org/abs/2504.06738}{Link} \\ \hline
\cite{yellapragada2025leveraging} & \UtiSinkUtilize & \None & \None & \begin{tabular}[c]{@{}c@{}}\AppVisEnhan\\ \AppSafe\end{tabular} & ICASSP & 2025 & \href{https://arxiv.org/abs/2501.04784}{Link} \\ \hline
\cite{sun2024massive} & \None & \begin{tabular}[c]{@{}c@{}}\InterpretOutlier\\ \InterpretAttnBias\end{tabular} & \EliminateLearnBias & \AppModelInter & COLM & 2024 & \href{https://arxiv.org/abs/2402.17762}{Link} \\ \hline
\cite{pulfer2024robustness} & \UtiPreixTrainable & \None & \None & \begin{tabular}[c]{@{}c@{}}\AppVisEnhan\\ \AppSafe\end{tabular} & ECCV & 2024 & \href{https://arxiv.org/abs/2503.10191}{Link} \\ \hline
\cite{darcet2024vision} & \UtiPreixTrainable & \None & \None & \AppVisEnhan & ICLR & 2024 & \href{https://arxiv.org/abs/2309.16588}{Link} \\ \hline
\cite{bondarenko2023quantizable} & \None & \begin{tabular}[c]{@{}c@{}}\InterpretSoftNoOp\\ \InterpretOutlier\end{tabular} & \begin{tabular}[c]{@{}c@{}}\EliminateGatedAttn\\ \EliminateModiSoftmax\end{tabular} & \begin{tabular}[c]{@{}c@{}}\AppModelTrain\\ \AppModelInfer\end{tabular} & NeurIPS & 2023 & \href{https://arxiv.org/abs/2306.12929}{Link} \\ \hline

\multicolumn{8}{c}{\cellcolor[HTML]{EFEFEF}\textbf{Diffusion Transformers}} \\* \midrule
\cite{starodubcev2026registers} & \UtiPreixTrainable & \None & \None & \AppVisEnhan & ArXiv & 2026 & \href{https://arxiv.org/abs/2605.16147}{Link} \\ \hline
\cite{wu2026attention} & \UtiAttnRedis & \None & \None & \AppVisEnhan & ICML & 2026 & \href{https://arxiv.org/abs/2605.09313v2}{Link} \\ \hline
\cite{wu2026taming} & \UtiPreixTrainable & \None & \None & \begin{tabular}[c]{@{}c@{}}\AppModelTrain\\ \AppSafe\end{tabular} & ArXiv & 2026 & \href{https://arxiv.org/abs/2605.05206}{Link} \\ \hline
\cite{mao2026packforcing} & \UtiPreSink & \None & \None & \AppLongEnhan & ArXiv & 2026 & \href{https://arxiv.org/abs/2603.25730}{Link} \\ \hline
\cite{su2026omniforcing} & \UtiSinkUtilize & \None & \None & \begin{tabular}[c]{@{}c@{}}\AppModelInfer\\ \AppVisEnhan\end{tabular} & ArXiv & 2026 & \href{https://arxiv.org/abs/2603.11647}{Link} \\ \hline
\cite{cao2026mlv} & \UtiPreSink & \None & \None & \AppVisEnhan & ArXiv & 2026 & \href{https://arxiv.org/abs/2602.02123}{Link} \\ \hline
\cite{shin2026motionstream} & \UtiPreSink & \None & \None & \begin{tabular}[c]{@{}c@{}}\AppModelInfer\\ \AppLongEnhan\end{tabular} & ICLR & 2026 & \href{https://arxiv.org/abs/2511.01266}{Link} \\ \hline
\cite{liu2026rolling} & \UtiPreSink & \None & \None & \AppLongEnhan & ICLR & 2026 & \href{https://arxiv.org/abs/2509.25161}{Link} \\ \hline
\cite{zhang2026freetext} & \UtiSinkUtilize & \None & \None & \AppVisEnhan & ICML & 2026 & \href{https://arxiv.org/abs/2601.00535}{Link} \\ \hline
\cite{yi2025deep} & \UtiPreSink & \None & \None & \AppLongEnhan & ICML & 2026 & \href{https://arxiv.org/abs/2512.05081}{Link} \\ \hline
\cite{lu2025reward} & \UtiPreSink & \None & \None & \AppModelInfer & ArXiv & 2025 & \href{https://arxiv.org/abs/2512.04678}{Link} \\ \hline
\cite{bandyopadhyay2025block} & \UtiPreSink & \None & \None & \AppModelInfer & ArXiv & 2025 & \href{https://arxiv.org/abs/2511.20426}{Link} \\ \hline
\cite{yang2025longlive} & \UtiPreSink & \None & \None & \AppModelInfer & ArXiv & 2025 & \href{https://arxiv.org/abs/2509.22622}{Link} \\ \hline
\cite{jamal2026diffusion} & \UtiPreixTrainable & \None & \None & \AppModelInter & AAAI & 2026 & \href{https://openreview.net/forum?id=Fc7s3UrQr1}{Link} \\ \hline
\cite{kim2025text} & \UtiAttnRedis & \None & \EliminateTuningAux & \begin{tabular}[c]{@{}c@{}}\AppVisEnhan\\ \AppHallu\end{tabular} & CVPR & 2025 & \href{https://arxiv.org/abs/2411.15236}{Link} \\ \hline

\multicolumn{8}{c}{\cellcolor[HTML]{EFEFEF}\textbf{Diffusion Language Models}} \\* \midrule
\cite{myrzakhan2026sink} & \UtiSinkUtilize & \None & \None & \AppModelInfer & ArXiv & 2026 & \href{https://arxiv.org/abs/2602.17664}{Link} \\ \hline
\cite{long2026focus} & \UtiPreSink & \None & \None & \AppModelInfer & ArXiv & 2026 & \href{https://arxiv.org/abs/2602.02159}{Link} \\ \hline
\cite{zhang2026one} & \UtiPreixTrainable & \InterpretSoftNoOp & \None & \AppGenEnhan & ArXiv & 2026 & \href{https://arxiv.org/abs/2601.19657}{Link} \\ \hline
\cite{dai2026revealing} & \None & \None & \None & \AppModelInter & ArXiv & 2026 & \href{https://arxiv.org/abs/2601.07894}{Link} \\ \hline
\cite{rulli2025attention} & \None & \None & \None & \AppModelInter & ArXiv & 2025 & \href{https://arxiv.org/abs/2510.15731}{Link} \\ \hline

\multicolumn{8}{c}{\cellcolor[HTML]{EFEFEF}\textbf{Linear Attention and Hybrid Linear Attention Models}} \\* \midrule
\cite{dong2025hymba} & \UtiPreixTrainable & \None & \None & \begin{tabular}[c]{@{}c@{}}\AppModelInfer\\ \AppGenEnhan\end{tabular} & ICLR & 2025 & \href{https://arxiv.org/abs/2411.13676}{Link} \\ \hline
\cite{meng2024enhanced} & \UtiPreSink & \None & \None & \begin{tabular}[c]{@{}c@{}}\AppModelTrain\\ \AppLongEnhan\end{tabular} & ArXiv & 2024 & \href{https://arxiv.org/abs/2408.00244}{Link} \\ \hline
\cite{wang2025mamba} & \UtiPreixTrainable & \None & \None & \begin{tabular}[c]{@{}c@{}}\AppModelTrain\\ \AppVisEnhan\end{tabular} & CVPR & 2025 & \href{https://arxiv.org/abs/2405.14858}{Link} \\ \hline

\multicolumn{8}{c}{\cellcolor[HTML]{EFEFEF}\textbf{Vision-Language-Action Models}} \\* \midrule
\cite{koo2025retovla} & \UtiSinkUtilize & \None & \None & \begin{tabular}[c]{@{}c@{}} \AppVisEnhan\end{tabular}  & ArXiv & 2025 & \href{https://arxiv.org/abs/2509.21243}{Link} \\* \midrule

\multicolumn{8}{c}{\cellcolor[HTML]{EFEFEF}\textbf{3D Transformers}} \\* \midrule
\cite{cheng2026longstream} & \UtiPreSink & \None & \None & \AppLongEnhan & ArXiv & 2026 & \href{https://arxiv.org/abs/2602.13172}{Link} \\ \hline
\cite{wang2025vggt} & \UtiPreixTrainable & \None & \None & \begin{tabular}[c]{@{}c@{}}\AppModelTrain\\ \AppVisEnhan\end{tabular} & CVPR & 2025 & \href{https://arxiv.org/abs/2503.11651}{Link} \\ \hline

\multicolumn{8}{c}{\cellcolor[HTML]{EFEFEF}\textbf{Autoregressive Video Diffusion Models}} \\* \midrule
\cite{li2026train} & \UtiPreSink & \None & \None & \AppLongEnhan & ArXiv & 2026 & \href{https://arxiv.org/abs/2602.14027}{Link} \\ \hline

\multicolumn{8}{c}{\cellcolor[HTML]{EFEFEF}\textbf{ Omni-modal Large Language Models}} \\* \midrule
\cite{yoo2026nature} & \UtiSinkUtilize & \InterpretAttnBias & \None & \AppVisEnhan & ArXiv & 2026 & \href{https://arxiv.org/abs/2603.14337}{Link} \\ \hline

\end{longtable}
% \end{landscape}% Yaxiu

%%%%%%%%%%%%%%%%%%%%%%%%%%%%%%%%%%%%%%%%%%%%%%%%%%%%%%%%%%%%

\clearpage
\bibliographystyle{unsrt}
\bibliography{reference}

\end{document}